\documentclass[10pt,twocolumn,letterpaper]{article}
\usepackage[pagenumbers]{cvpr} 

\definecolor{cvprblue}{rgb}{0.21,0.49,0.74}
\usepackage[breaklinks,colorlinks,allcolors=cvprblue]{hyperref}
\usepackage{enumitem} 

\usepackage[shortlabels]{enumitem}
\usepackage{amsmath}
\usepackage{epsfig}
\usepackage{graphicx}
\usepackage{color,soul}
\usepackage{placeins}

\usepackage{caption}
\usepackage{subcaption}

\usepackage{amsmath, amssymb, amsthm}
\usepackage{mathrsfs}
\usepackage{mathtools}
\usepackage[draft]{microtype}  
\usepackage{amsfonts}
\usepackage{bbm}
\usepackage{float}
\usepackage{multirow}
\usepackage{makecell}

\usepackage{tikz}
\usetikzlibrary{positioning,fit,calc,shapes}  
\tikzset{set/.style={draw,circle,inner sep=0pt,align=center}}
\usepackage{tikzscale}
\usepackage{pgfplots}
\usepgfplotslibrary{groupplots,fillbetween, statistics}
\pgfplotsset{compat=newest}
\usetikzlibrary{math}
\usepackage{tabularx} 
\newcolumntype{P}[1]{>{\centering\arraybackslash}p{#1}}

\usepackage{float}

\usepackage{pdflscape}
\usepackage{subcaption}
\usepackage{graphicx}
\usepackage{enumitem}
\usepackage{xcolor}
\usepackage{placeins}

\definecolor{darkgreen}{RGB}{0,120,0}      
\definecolor{golden}{RGB}{218,165,32}

\defcitealias{kazhdan2013screened}{SPSR}
\defcitealias{huo2022iterative}{iPSR}
\defcitealias{hanocka2020point2mesh}{Point2Mesh}
\defcitealias{peng2021sap}{SAP}

\newcounter{cases}
\newcounter{subcases}[cases]

\allowdisplaybreaks[1]
\usepackage{sistyle} 

\numberwithin{equation}{section}

\newcommand{\interior}[1]{%
  {\kern0pt#1}^{\mathrm{o}}}

\definecolor{darkcandyapplered}{rgb}{0.64, 0.0, 0.0}

\def\ours{DIAMOND}

\def\paperID{}
\def\confName{CVPR}
\def\confYear{2026}

\title{DIAMOND-SSS: Diffusion-Augmented Multi-View Optimization for Data-efficient SubSurface Scattering}

\author{Guillermo Figueroa-Araneda \thanks{Equal contribution.}\\
Technical University of Munich\\
{\tt\small guillermo.figueroaa@usm.cl}
\and
Iris Diana Jimenez\footnotemark[1]\\
Ludwig-Maximilians-Universit\"at Munich\\
{\tt\small jimenez@lmu.de}
\and
Florian Hofherr\\
Technical University of Munich\\
{\tt\small florian.hofherr@tum.de}
\and
Manny Ko\\
Independent Researcher\\
{\tt\small man961@yahoo.com}
\and
Hector Andrade-Loarca\\
Technical University of Munich\\
{\tt\small hector.andrade@tum.de}
\and
Daniel Cremers\\  
Technical University of Munich\\
{\tt\small cremers@tum.de}
}

\begin{document}
\maketitle

\begin{abstract}
Subsurface scattering (SSS) gives translucent materials—such as wax, jade, marble, and skin—their characteristic soft shadows, color bleeding, and diffuse glow. Modeling these effects in neural rendering remains challenging due to complex light transport and the need for densely captured multi-view, multi-light datasets (often $>100$ views and 112 OLATs).

We present \textbf{DIAMOND-SSS}, a data-efficient framework for high-fidelity translucent reconstruction from extremely sparse supervision—even as few as ten images. We fine-tune diffusion models for novel-view synthesis and relighting, conditioned on estimated geometry and trained on less than $7\%$ of the dataset, producing photorealistic augmentations that can replace up to $95\%$ of missing captures. To stabilize reconstruction under sparse or synthetic supervision, we introduce illumination-independent geometric priors: a multi-view silhouette consistency loss and a multi-view depth consistency loss. 
Across all sparsity regimes, \textbf{DIAMOND-SSS} achieves state-of-the-art quality in relightable Gaussian rendering, reducing real capture requirements by up to $90\%$ compared to SSS-3DGS.
\end{abstract}

\begin{figure}
\label{fig:teaser}
\centering
\includegraphics[width=1.0\columnwidth]{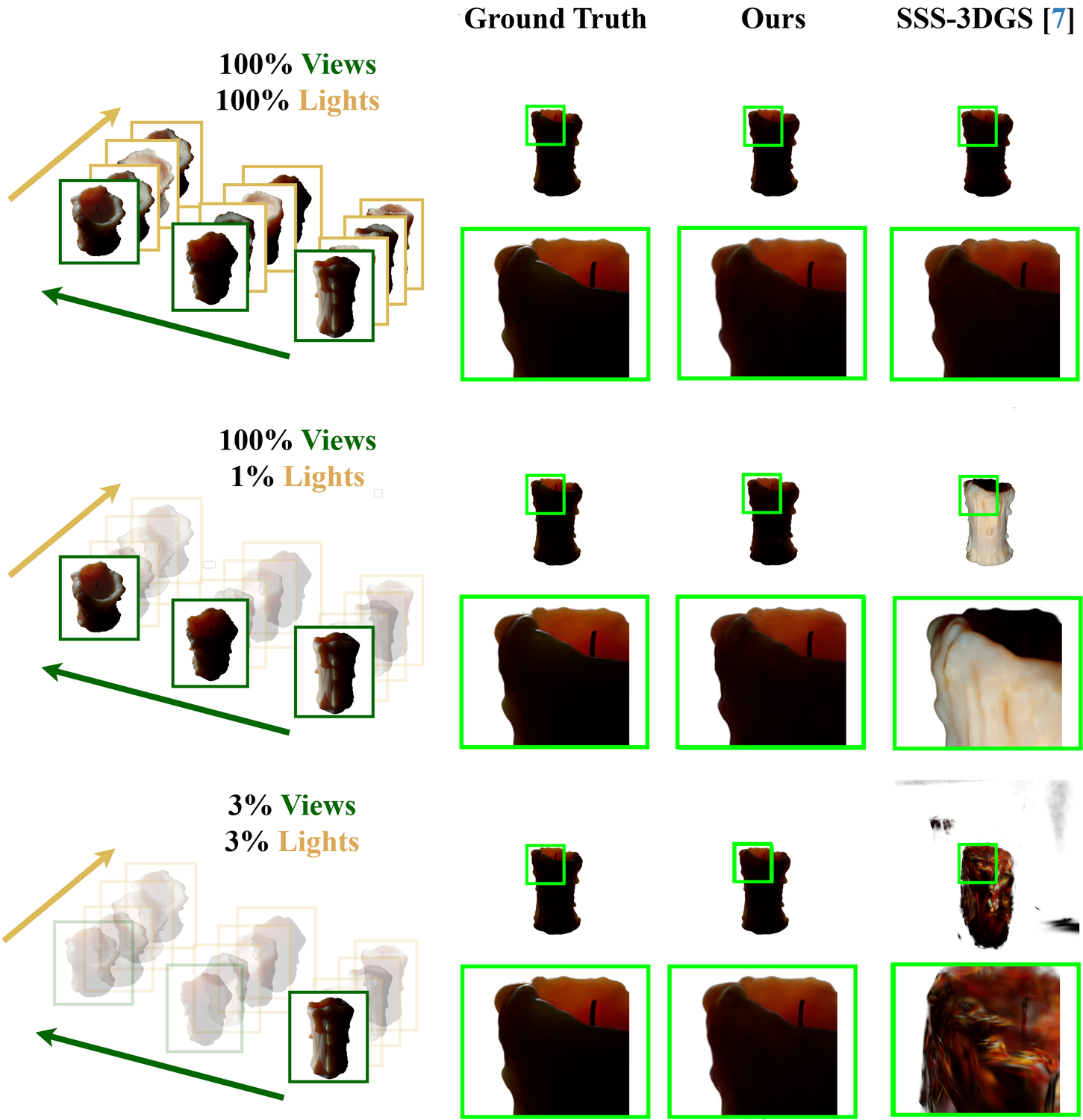}
\caption{
   We propose a data-efficient framework for re-lightable 3D with subsurface scattering. Unlike prior work \cite{dihlmann2024subsurfacescattering3dgaussian} requiring \textcolor{darkgreen}{many viewpoints} under 
\textcolor{golden}{multiple illuminations}, our method uses diffusion models to synthesize novel views and lighting from sparse observations. The baseline \cite{dihlmann2024subsurfacescattering3dgaussian} massively overfits in the second row, for the only given light condition.
}
\end{figure}    
\section{Introduction}
\label{sec:intro}

Rendering translucent materials such as skin, wax, or jade remains challenging in graphics and vision. Their appearance is governed by \emph{subsurface scattering} (SSS)—light entering a surface, scattering within the medium, and exiting elsewhere—producing soft glows, color bleeding, and blurred shadows. Accurately capturing these effects requires jointly modeling geometry, illumination, and volumetric transport, making data acquisition costly and dependent on dense, controlled multi-view and multi-light setups.

Neural rendering methods such as NeRF~\cite{mildenhall2020nerf} and 3D Gaussian Splatting (3DGS)~\cite{kerbl20233dgaussiansplattingrealtime} enable high-quality novel-view synthesis, but entangle lighting and appearance, limiting relighting. Physically based extensions incorporate BRDF models~\cite{phong1975illumination,cook1982reflectance,burley2012physically} under known~\cite{srinivasan2021nerv,Zhang22IRON,Brahimi24SuperVol,brahimi24SparseViewsNearLight} or unknown~\cite{Zhang21PhySG,nerd,neilfpp,gao2024relightable,kaleta2025lumigauss} illumination, or account for indirect lighting~\cite{Zhang2022ModellingIndirIlluminationInvRendering,Zhang2021NeRFactor,Sarkar23LitNerf}. While effective for opaque materials, these methods cannot reproduce the spatially decoupled transport characteristic of translucent media.

Modeling SSS requires volumetric light transport. Classical BSSRDF (Bidirectional Scattering Surface Reflectance Distribution Function) models
~\cite{jensen2001practical,jacques2013optical,donner2006spectral,d2011quantized}
offer analytic approximations but struggle with real-world complexity. Learning-based SSS methods~\cite{Thomson_2024,li2023inverserenderingtranslucentobjects,zhu2023neuralrelightingsubsurfacescattering,TG23NeuralBSSRDF} provide more flexibility but still require dense, calibrated data.

KiloOSF~\cite{thomas2024kiloosf} learns object-centric scattering functions yet remains surface-bound. The current state of the art, SSS-3DGS~\cite{dihlmann2024subsurfacescattering3dgaussian}, augments Relightable 3DGS~\cite{gao2024relightable} with a residual MLP to capture volumetric scattering, achieving photorealistic translucency—but at the cost of dense supervision: over 100 calibrated views and 112 OLAT illuminations per object.

\medskip\noindent\textbf{Our approach.}  
We present \textbf{DIAMOND-SSS}, a data-efficient extension of SSS-3DGS~\cite{dihlmann2024subsurfacescattering3dgaussian} for relightable reconstruction of translucent materials from sparse supervision. While SSS-3DGS requires tens of thousands of images across $\sim$100 views and 112 OLAT illuminations, DIAMOND-SSS attains comparable or superior quality using as little as ten views and one light.

Our key component is a \emph{diffusion-based augmentation pipeline} that synthesizes missing viewpoints and lighting via two geometry-conditioned diffusion models—one for novel-view generation and one for relighting—fine-tuned on less than $7\%$ of the dataset yet producing photorealistic, geometrically aligned augmentations. To stabilize reconstruction under sparse or synthetic data, we introduce two illumination-invariant geometric priors: multi-view silhouette consistency and multi-view depth consistency, which regularize structure across viewpoints and complement diffusion-based augmentation.

Across all supervision regimes, DIAMOND-SSS surpasses prior work. Geometric priors alone improve fidelity, and combined with diffusion-based relighting enable high-quality reconstructions with up to $95\%$ fewer real captures. Even under extreme sparsity (3\% views, 3\% lights), our method faithfully reproduces subsurface scattering effects—unattainable with previous approaches.

\medskip\noindent\textbf{Contributions.}
\begin{itemize}
    \item We introduce a \emph{diffusion-based augmentation pipeline} for subsurface scattering reconstruction, with fine-tuned models for novel-view synthesis and relighting synthesis of translucent objects. This enables photometrically and geometrically consistent supervision even under extremely sparse captures (\cref{subsec:diffusion_augmentation,subsec:diamond_pipeline}).
    \item We design two \emph{multi-view geometric consistency losses}—silhouette and depth—that complement diffusion-based data augmentation by enforcing structural alignment across viewpoints and improving stability under limited or synthetic supervision (\cref{subsec:geometric_consistency}).
    \item We demonstrate that by combining our diffusion-based data augmentation with geometric consistency losses, DIAMOND-SSS achieves high-quality translucent reconstructions with up to \textbf{90\% less real data} than prior SSS-3DGS approaches, while maintaining fidelity across varying sparsity regimes (\cref{sec:experiments}).
\end{itemize}

\section{Related Work}
\label{sec:related}

\subsection{Neural Rendering and Physically-Based Relighting}
\label{subsec:neuralrendering}

Neural Radiance Fields (NeRF) \cite{mildenhall2020nerf} and 3D Gaussian Splatting (3DGS) \cite{kerbl20233dgaussiansplattingrealtime} have established high-quality novel-view synthesis by combining scene representation as radiance fields with volumetric rendering. While they show impressive results they entangle appearance and illumination and effectively bake a static lighting into the scene representation, which does not allow for relighting.

To address this limitation large body of work extends these methods with physically based reflectance modeling. Approaches integrate microfacet \cite{cook1982reflectance} or Disney BRDFs \cite{burley2012physically} under known \cite{srinivasan2021nerv,Zhang22IRON,Brahimi24SuperVol,brahimi24SparseViewsNearLight} or unknown \cite{Zhang21PhySG,nerd,neilfpp,gao2024relightable,kaleta2025lumigauss} lighting conditions, or incorporate indirect illumination \cite{Zhang2022ModellingIndirIlluminationInvRendering} and more general appearance decompositions \cite{Zhang2021NeRFactor,Sarkar23LitNerf}. These methods achieve relightable reconstructions but remain focused on opaque materials where light transport is surface-bounded.

\subsection{Subsurface Scattering Models}
\label{subsec:sss}

Subsurface scattering (SSS) describes light entering a surface, scattering within a medium, and exiting elsewhere. Classical BSSRDF models—dipole~\cite{jensen2001practical}, multipole~\cite{jacques2013optical}, spectral/multilayer skin models~\cite{donner2006spectral}, and quantized diffusion~\cite{d2011quantized}—are efficient but rely on strong assumptions that limit realism.

Learning-based alternatives, including Neural SSS~\cite{Thomson_2024}, Li~\etal~\cite{li2023inverserenderingtranslucentobjects}, Zhu~\etal~\cite{zhu2023neuralrelightingsubsurfacescattering}, and TG~\etal~\cite{TG23NeuralBSSRDF}, relax these assumptions but still require dense multi-view, multi-light supervision. KiloOSF~\cite{thomas2024kiloosf} introduces object-centric scattering functions but remains surface-bound, while SSS-3DGS~\cite{dihlmann2024subsurfacescattering3dgaussian} models volumetric SSS via a residual MLP atop Gaussian splats, still demanding dense captures. Our method leverages diffusion-based synthetic augmentation and illumination-invariant geometric priors to achieve comparable or superior quality with up to \textbf{90\% less real data}.

\subsection{Diffusion Models for Novel View Synthesis and Relighting}
\label{subsec:diffusion}

Diffusion models have recently advanced novel view synthesis (NVS). Single-view methods such as Zero-1-to-3~\cite{liu2023zero} generate pose-controlled views from one RGB input, while multi-view–consistent models including SyncDreamer~\cite{liusyncdreamer}, MVDream~\cite{shimvdream}, Free3D~\cite{zheng24}, and Cat3D~\cite{gao2024cat3d} produce coherent view sets and can augment sparse observations for 3D reconstruction. Controllable diffusion frameworks like ControlNet~\cite{zhang2023controlnet} enable conditioning on geometric cues, and Poirier-Ginter~\etal~\cite{poirier2024radiancefield} use such conditioning for OLAT-based relighting via lighting-direction controls.

Building on these advances, our augmentation pipeline (\cref{subsec:diffusion_augmentation}) employs two geometry-conditioned diffusion models—one for novel-view synthesis and one for relighting—fine-tuned on a small subset of translucent objects. Integrated into our reconstruction framework (\cref{subsec:diamond_pipeline}), these models produce photometrically and geometrically aligned augmentations. To our knowledge, \ours-SSS is the first to combine diffusion-based augmentation with Gaussian Splatting for data-efficient translucent reconstruction.

\subsection{Geometric Supervision under Sparse and Synthetic Settings}
\label{subsec:geometry}

Cross-view geometric constraints provide strong cues for 3D reconstruction, particularly under sparse or inconsistent photometric supervision.  
Silhouette consistency, used since early visual hull methods, has been applied to neural fields~\cite{li2023inverserenderingtranslucentobjects} to improve shape regularity, while multi-view depth consistency enforces alignment across illuminations.  

In our framework (\cref{subsec:geometric_consistency}), we extend these ideas with illumination-invariant multi-view silhouette and depth consistency losses that stabilize optimization when real data are limited or synthetic supervision is used.  
Together with our diffusion-based augmentation (\cref{subsec:diffusion_augmentation}), these geometric priors form the foundation of DIAMOND-SSS’s data efficiency and reconstruction fidelity.

\section{Method}
\label{sec:method}

We build on Subsurface Scattering Gaussian Splatting (SSS-3DGS)~\cite{dihlmann2024subsurfacescattering3dgaussian}, which extends 3D Gaussian Splatting (3DGS)~\cite{kerbl20233dgaussiansplattingrealtime} to represent translucent materials. 
Our method improves data efficiency through two main contributions:
(i) a diffusion-based data augmentation strategy for generating novel views and relighting, and 
(ii) multi-view geometric consistency losses that enhance reconstruction fidelity under sparse captures. 

\begin{figure*}[htb!]
\centering
\includegraphics[width=\textwidth]{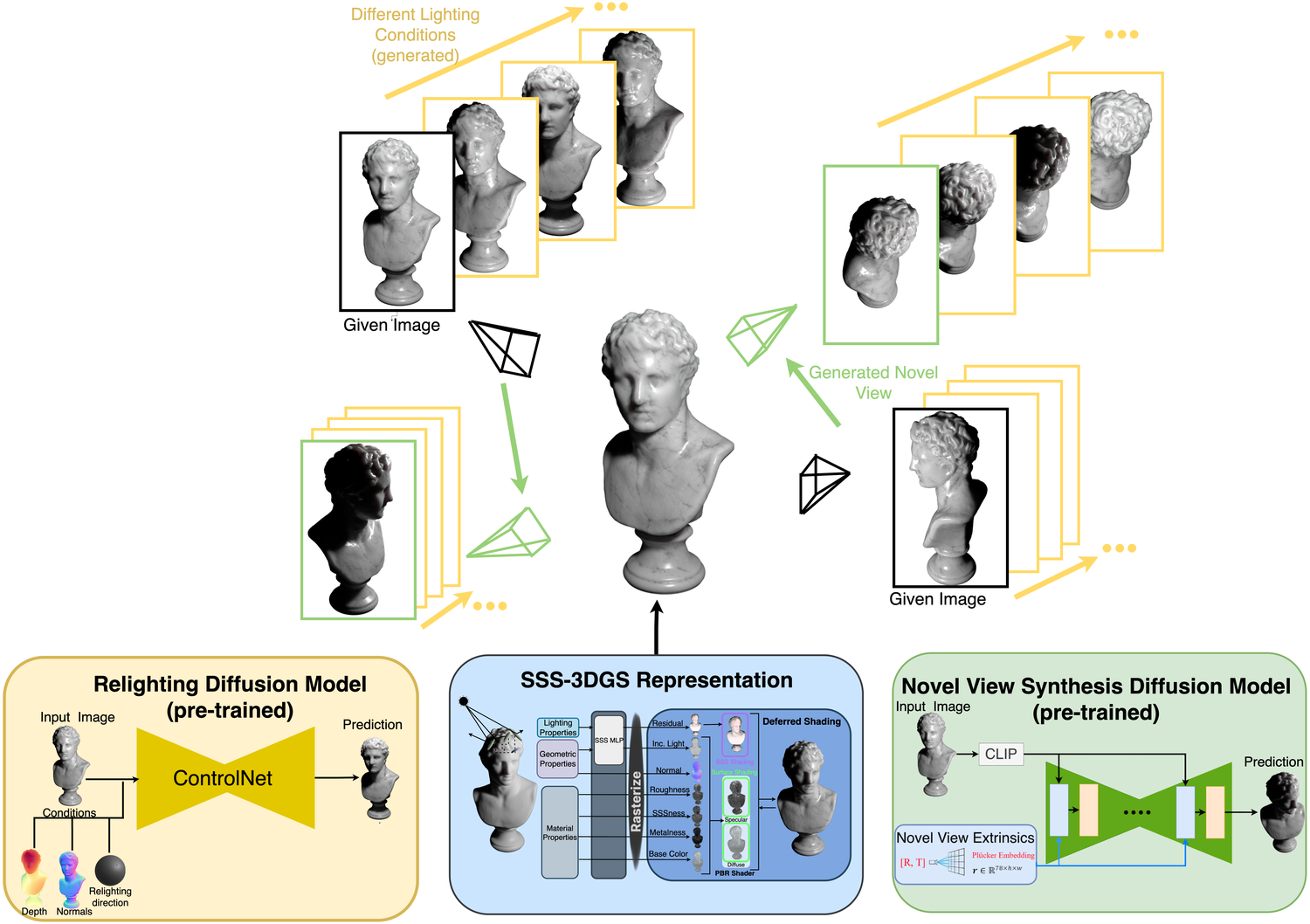}
\caption{
Overview of our data augmentation pipeline. From a few posed input images, we generate additional viewpoints using a novel-view diffusion model (green) and synthetic OLAT variants using a relighting model (yellow). The combined real and synthetic data are used to train SSS-3DGS~\cite{dihlmann2024subsurfacescattering3dgaussian} with an MLP for subsurface scattering, stabilized by our geometric consistency losses (\cref{subsec:geometric_consistency}).
}
\label{fig:approach}
\end{figure*}

\subsection{Recap: SSS-3DGS}
\label{subsec:sss3dgs}

\paragraph{Vanilla 3DGS.}
3D Gaussian Splatting represents a scene as a set of $N$ anisotropic Gaussians:
\[
G = \{ (\mu_i, \Sigma_i, c_i, o_i) \}_{i=1}^N,
\]
where $\mu_i \in \mathbb{R}^3$ is the mean position, $\Sigma_i \in \mathbb{R}^{3 \times 3}$ is the covariance matrix defining scale and orientation, $c_i$ is the color, and $o_i \in [0, 1]$ is the opacity. 
The covariance matrix is parameterized via Cholesky decomposition to ensure positive semi-definiteness, and eigenvalue clamping constrains the spatial extent. 

\paragraph{Extended Shading with BRDF and SSS.}
To model standard surface appearance in Gaussian splatting, each Gaussian is typically augmented with physically motivated material attributes 
$c_i \in [0,1]^3$ (albedo), $r_i \in [0,1]$ (roughness), and $m_i \in [0,1]$ (metalness)~\cite{gao2024relightable}. 
Shading is then computed in a deferred pass using a microfacet BRDF:
\begin{equation}
C_{\text{surf}}(x) = C_{\text{surf}}(x; c, r, m, \mathbf{n}, \ell),
\end{equation}
where $\mathbf{n}$ is the surface normal, and $\ell$ the light direction.
Dihlmann et al.~\cite{dihlmann2024subsurfacescattering3dgaussian} extend this formulation to handle translucent materials by introducing an additive neural residual term that models subsurface scattering. 
The final pixel intensity becomes
\begin{equation}
I(x) = C_{\text{surf}}(x) \;+\; 
C_{\text{sss}}(x; \ell, \mathbf{v}, \phi),
\end{equation}
where $\mathbf{v}$ is the view direction and $\phi$ denotes learned latent features capturing material-dependent scattering behavior.

\paragraph{Surface Shading.}
Under OLAT lighting with directional source $\ell$ and radiance $L(\ell)$, surface shading is given by:
\begin{equation}
C_{\text{surf}} = L(\ell) \cdot f_r(\mathbf{n}, \mathbf{v}, \ell; c, r, m) \cdot \max(0, \mathbf{n} \cdot \ell),
\end{equation}
where the BRDF is composed of diffuse and specular components:
\[
f_r = f_{\text{diff}} + f_{\text{spec}},
\]
\[
f_{\text{diff}} = (1 - m) \frac{c}{\pi}, \quad
f_{\text{spec}} = \frac{D_{\text{GGX}} \cdot G_{\text{Smith}} \cdot F_{\text{Schlick}}}{4 (\mathbf{n} \cdot \mathbf{v})_+ (\mathbf{n} \cdot \ell)_+}.
\]
Here $\alpha = r^2$, $\mathbf{h} = \frac{\mathbf{v} + \ell}{\|\mathbf{v} + \ell\|}$ is the half-vector, and $F_0 = (1 - m) \cdot 0.04 + m \cdot c$. For complete expressions of $D_{\text{GGX}}$, $G_{\text{Smith}}$, and $F_{\text{Schlick}}$, we refer the reader to~\cite{Walter2007GGX, Schlick1994, Smith1967}.

\paragraph{Neural Subsurface Residual.}
To model volumetric scattering, a learned residual term adds view- and light-dependent translucency:
\[
C_{\text{sss}}(x) = sss(x) \cdot f_\phi(\mu(x), \Sigma(x), \mathbf{n}(x), \ell, \mathbf{v}),
\]
where $sss(x) \in [0,1]$ is a learned scalar controlling subsurface strength, and $f_\phi$ is a small MLP. 
This residual effectively encodes the radiance contribution of translucent transport for a single Gaussian, allowing for soft shadows, color bleeding, and glow effects beyond the surface BRDF.

\subsection{Diffusion-Based Data Augmentation}
\label{subsec:diffusion_augmentation}

To enable high-quality translucent reconstruction from sparse captures, we introduce a data-driven augmentation strategy based on pretrained diffusion models. 
We use these models synthesize photometrically plausible novel views and lighting conditions that supplement the limited input data during SSS-3DGS optimization (\cref{subsec:sss3dgs}). 
Both diffusion models are fine-tuned once on a small subset of translucent objects and subsequently applied to all unseen objects, ensuring broad generalization without per-object retraining (\cref{fig:approach}). 
An extended discussion of the architecture and training procedure is provided in the Supplementary Material.

\paragraph{Fine-Tuning Strategy.}
We adapt two publicly available latent diffusion models for our purpose: one for novel-view synthesis (Free3D~\cite{zheng24}) and one for relighting (inspired by ControlNet~\cite{zhang2023controlnet} and Poirier~\etal~\cite{poirier2024radiancefield}). 
Both were originally trained on large-scale datasets of predominantly opaque objects. 
To adapt them for translucent appearance effects, we fine-tune each model on \emph{less than 7\% of the total OLAT dataset} (see~\cref{subsec:datasets}), using only a subset of images from four distinct translucent objects captured under controlled directional lighting and multiple viewpoints. 
All images used for fine-tuning are drawn exclusively from the training split, ensuring that no evaluation views or lights are seen by the diffusion models. 
Fine-tuning uses the original diffusion objectives with perceptual supervision (L1, SSIM, LPIPS) under fixed illumination and deterministic sampling, which encourages geometric consistency and reduces view-to-view drift. We discuss our finetuning trategy with more detail in the supplement, see~\cref{sec:supp_diffusion}.

\paragraph{Novel-View Diffusion.}
For view synthesis, we employ fine-tune Free3D~\cite{zheng24} architecture conditioned on the input image and camera pose through ray-conditioning normalization layers, without architectural modifications. 
The model learns to preserve object identity while improving parallax and silhouette coherence across novel viewpoints. 
During inference, it takes as input the source image and relative camera transformation and generates a geometrically consistent view from the target pose. 
The resulting synthetic views are used for photometric and geometric supervision during SSS-3DGS optimization.

\paragraph{Relighting Diffusion.}
For illumination augmentation, we extend the relighting model of Poirier-Ginter~\etal~\cite{poirier2024radiancefield} by enriching its conditioning. 
In addition to the source image, depth map, and target lighting code (encoded as 9D spherical harmonics), we also concatenate predicted surface normals~\cite{marigold} and depth~\cite{depth_anything_v2}. 
Additional details are provided in \cref{sec:supp_diffusion}. 
Training follows the $v$-parameterization objective combined with perceptual (L1, SSIM, LPIPS), cycle-consistency, and blur-aware losses to encourage smooth irradiance transitions and stable light-transport behavior. 
This enables the synthesis of relit images that reproduce soft shadows, diffuse scattering, and low-frequency volumetric effects characteristic of translucent materials.

We emphasize that all conditioning maps are estimated automatically—no ground-truth normals or depths are required—making the pipeline applicable to casual or legacy captures.

\subsection{Multi-View Geometric Consistency}
\label{subsec:geometric_consistency}

\begin{figure}[htb!]
\centering
\includegraphics[width=0.5\textwidth]{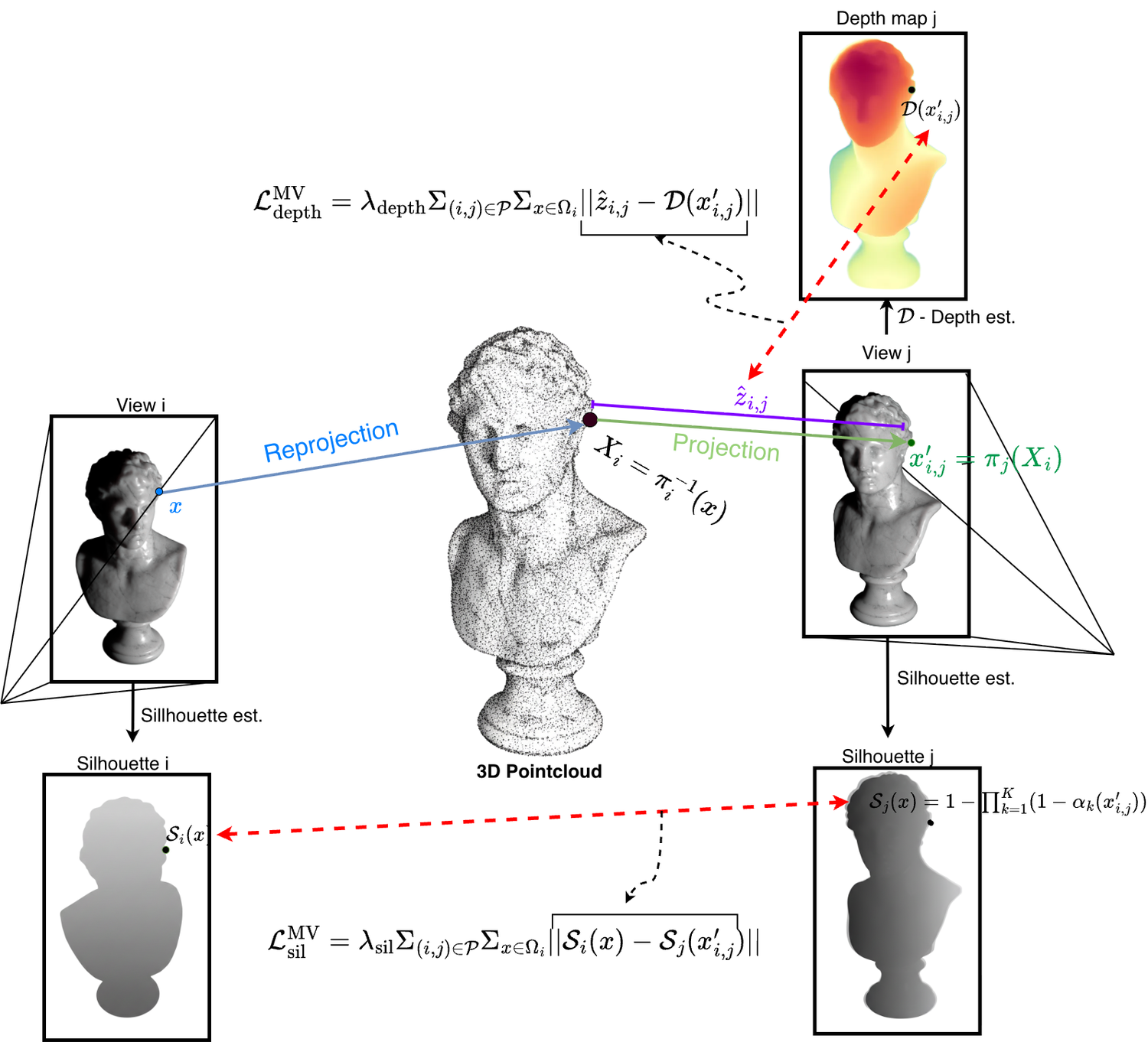}
\caption{
   Visualization of the proposed geometric consistency losses. 
   Silhouette consistency enforces cross-view contour alignment, while depth consistency stabilizes global geometry under varying lighting.
}
\label{fig:depth_cons_loss}
\end{figure}

To further improve Gaussian splatting reconstruction from diffusion-generated images—which may exhibit minor spatial or photometric inconsistencies inherent to generative synthesis (e.g., known geometric failures of diffusion models~\cite{Sarkar_2024_CVPR})—we introduce two illumination-invariant geometric losses. These consist of a silhouette consistency loss and a depth consistency loss that enforce cross-view structural alignment, thereby stabilizing optimization under both real and synthetic supervision. See \cref{fig:depth_cons_loss} for an illustration.

\paragraph{Silhouette Consistency.}
For a pixel $x$ in view $i$, we compute its 3D location $X_i$ using depth $D_i(x)$  (which is obtained by using DepthAnythingv2 \cite{depth_anything_v2}) as well as its reprojection $x_{i,j}'$ to view $j$:
\begin{equation}
\label{eq:backproj}
X_i = \pi_i^{-1}(x, D_i(x)),
\quad x_{i,j}' = \pi_j(X_i),
\end{equation}
where $\pi_i$ is the projection function to view $i$. Using the soft silhouette in view $i$, computed from the Gaussian opacities:
\[
S_i(x) = 1 - \prod_{k=1}^K (1 - \alpha_k(x)),
\]
where $\alpha_k(x)$ is the opacity contribution at pixel $x$ from the $k$-th 3D Gaussian, we define the multi-view silhouette consistency loss:
\begin{equation}
\label{eq:sil_loss}
\mathcal{L}_{\text{sil}}^{\text{MV}} = 
\lambda_{\text{sil}} \sum_{(i,j)} \sum_{x \in \Omega_i} 
\left\lVert S_i(x) - S_j(x_{i,j}') \right\rVert_1.
\end{equation}
This loss encourages consistent contours across views and improves boundary fidelity. See \cref{fig:depth_cons_loss} for an illustration. 

\paragraph{Depth Consistency.}
We additionally enforce cross-view agreement in depth. 
For back-projected point $X_i$ (see \cref{eq:backproj}) and target camera center $C_j$, we compute:
\[
\hat{z}_{i,j} = \|X_i - C_j\|_2, \quad x_{i,j}' = \pi_j(X_i),
\]
and compare it with the rendered depth $D_j(x_{i,j}')$, computed with~\cite{depth_anything_v2}, defining the loss:
\begin{equation}
\label{eq:depth_loss}
\mathcal{L}_{\text{depth}}^{\text{MV}} = 
\lambda_{\text{depth}}\sum_{(i,j)} \sum_{x \in \Omega_i} 
\left\lVert \hat{z}_{i,j} - D_j(x_{i,j}') \right\rVert_1.
\end{equation}
This loss reduces floating geometry and improves global coherence.

\paragraph{Total Loss.}
Our final training objective extends Relightable 3D Gaussians (R3DGS)~\cite{gao2024relightable} by integrating diffusion-augmented supervision and our illumination-invariant geometric priors (\cref{subsec:geometric_consistency}). 
The total loss combines photometric, regularization, and geometric consistency terms:
\begin{equation}
\begin{split}
\mathcal{L}_\text{total} =
&(1 - \lambda_{\text{dssim}})\mathcal{L}_1 +
\lambda_{\text{dssim}}(1 - \text{SSIM}) +
\lambda_{\text{lpips}}\mathcal{L}_{\text{LPIPS}} \\
&+
\lambda_{\text{mask}}\mathcal{L}_{\text{mask}} +
\lambda_{\text{smooth}}\mathcal{L}_{\text{smooth}} +
\lambda_{\text{enh}}\mathcal{L}_{\text{enh}} \\
&+
\lambda_{\text{ray}}\mathcal{L}_{\text{ray}} +
\lambda_{\text{sil}}\mathcal{L}_{\text{sil}}^{\text{MV}} +
\lambda_{\text{depth}}\mathcal{L}_{\text{depth}}^{\text{MV}} .
\end{split}
\end{equation}

Here, $\mathcal{L}_{\text{photo}}$ is a combination of pixel-wise and perceptual losses (L1, SSIM, LPIPS) balancing structural and appearance fidelity.  
Regularization terms $\mathcal{L}_{\text{mask}}$, $\mathcal{L}_{\text{smooth}}$, $\mathcal{L}_{\text{enh}}$, and $\mathcal{L}_{\text{ray}}$ follow prior work~\cite{gao2024relightable}, enforcing spatial coherence, contrast preservation, and physical consistency in transmittance and shading.

\subsection{DIAMOND-SSS Reconstruction Procedure}
\label{subsec:diamond_pipeline}

After fine-tuning, we freeze the diffusion models and use them to augment each scene before 3DGS optimization. For every real view, we synthesize \mbox{2--4} novel viewpoints, and for each real \emph{and} synthetic view we generate \mbox{4--8} relit variants under novel OLAT conditions. This yields an expanded set of real images, diffusion-generated views, and synthetic relightings.

We then optimize a 3DGS model with an SSS residual on this augmented dataset. Photometric losses ($L_1$, SSIM~\cite{wang2004ssim}, LPIPS~\cite{zhang2018lpips}) are applied to all images, with synthetic ones down-weighted by $\alpha = 0.5$ to mitigate diffusion artifacts. Geometric losses (multi-view silhouette and depth consistency) are applied uniformly to real and synthetic images to enforce cross-view structural alignment.

This joint supervision enables robust learning of both shape and appearance from as few as ten real captures per object, significantly reducing acquisition cost. 
Once adapted, the diffusion models can be reused for many reconstructions, enabling scalable capture of translucent materials in unconstrained settings.

\section{Experiments}
\label{sec:experiments}

\begin{figure*}[htb!]
\centering
\resizebox{\textwidth}{!}{%
\renewcommand{\arraystretch}{0.9}
\setlength{\tabcolsep}{2pt}

\begin{tabular}{ccccccccc}
\toprule
\multicolumn{3}{c}{\textbf{Statue}} 
& \multicolumn{3}{c}{\textbf{Candle}} 
& \multicolumn{3}{c}{\textbf{Soap}} \\
\cmidrule(lr){1-3}\cmidrule(lr){4-6}\cmidrule(lr){7-9}
\textbf{GT} & \textbf{Ours} & \textbf{SSS-3DGS \cite{dihlmann2024subsurfacescattering3dgaussian}} &
\textbf{GT} & \textbf{Ours} & \textbf{SSS-3DGS \cite{dihlmann2024subsurfacescattering3dgaussian}} &
\textbf{GT} & \textbf{Ours} & \textbf{SSS-3DGS \cite{dihlmann2024subsurfacescattering3dgaussian}} \\
\midrule

\raisebox{-0.5\height}{\includegraphics[width=0.10\linewidth]{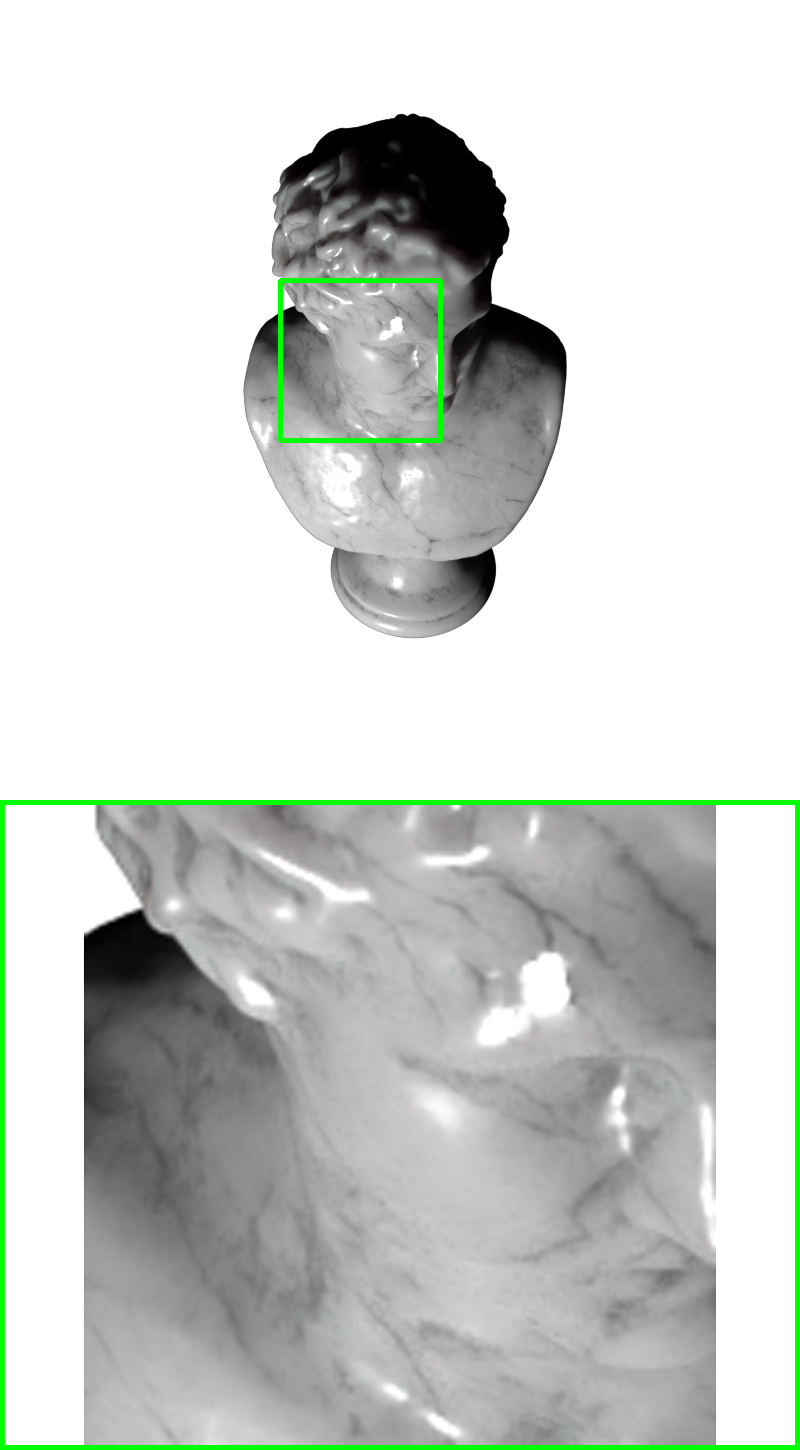}} &
\raisebox{-0.5\height}{\includegraphics[width=0.10\linewidth]{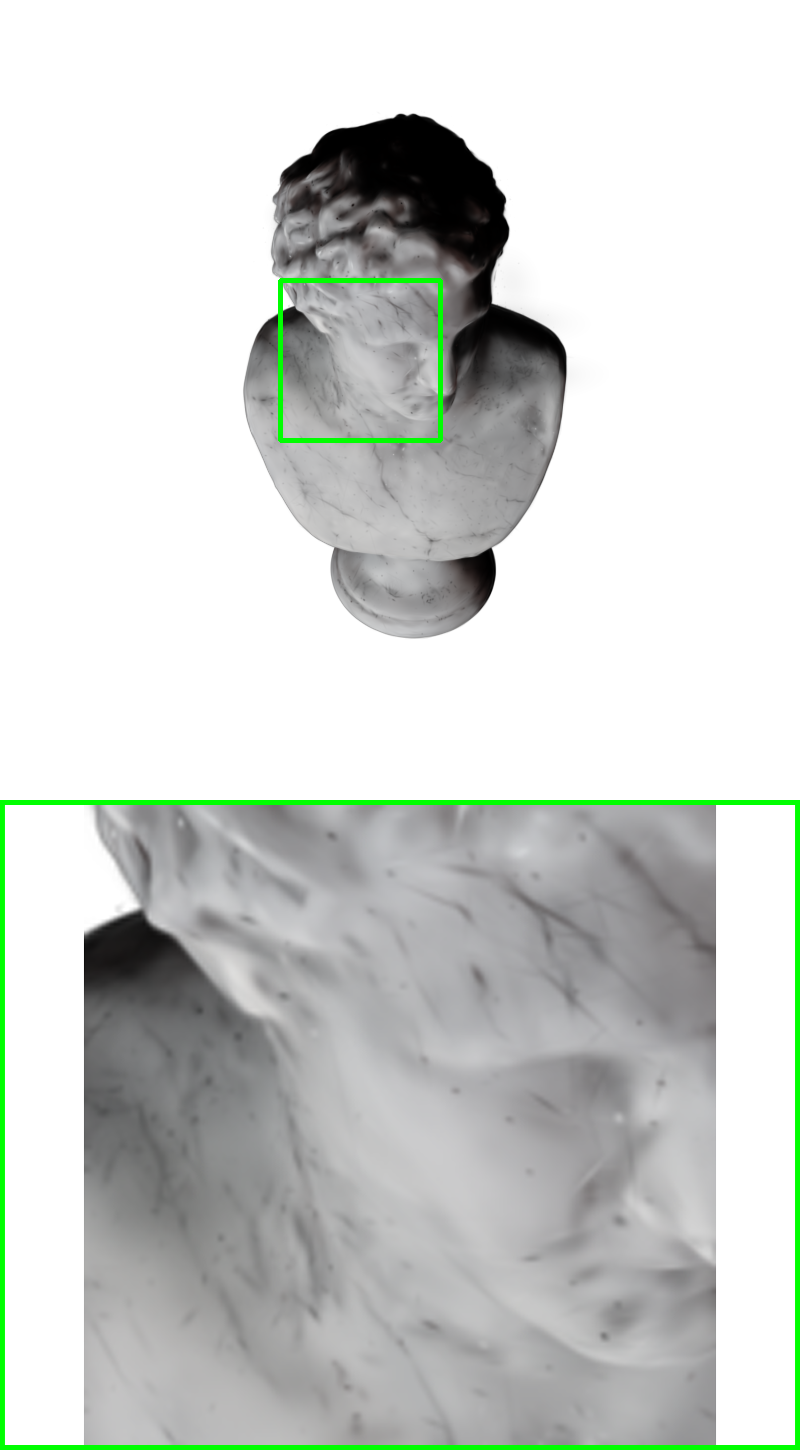}} &
\raisebox{-0.5\height}{\includegraphics[width=0.10\linewidth]{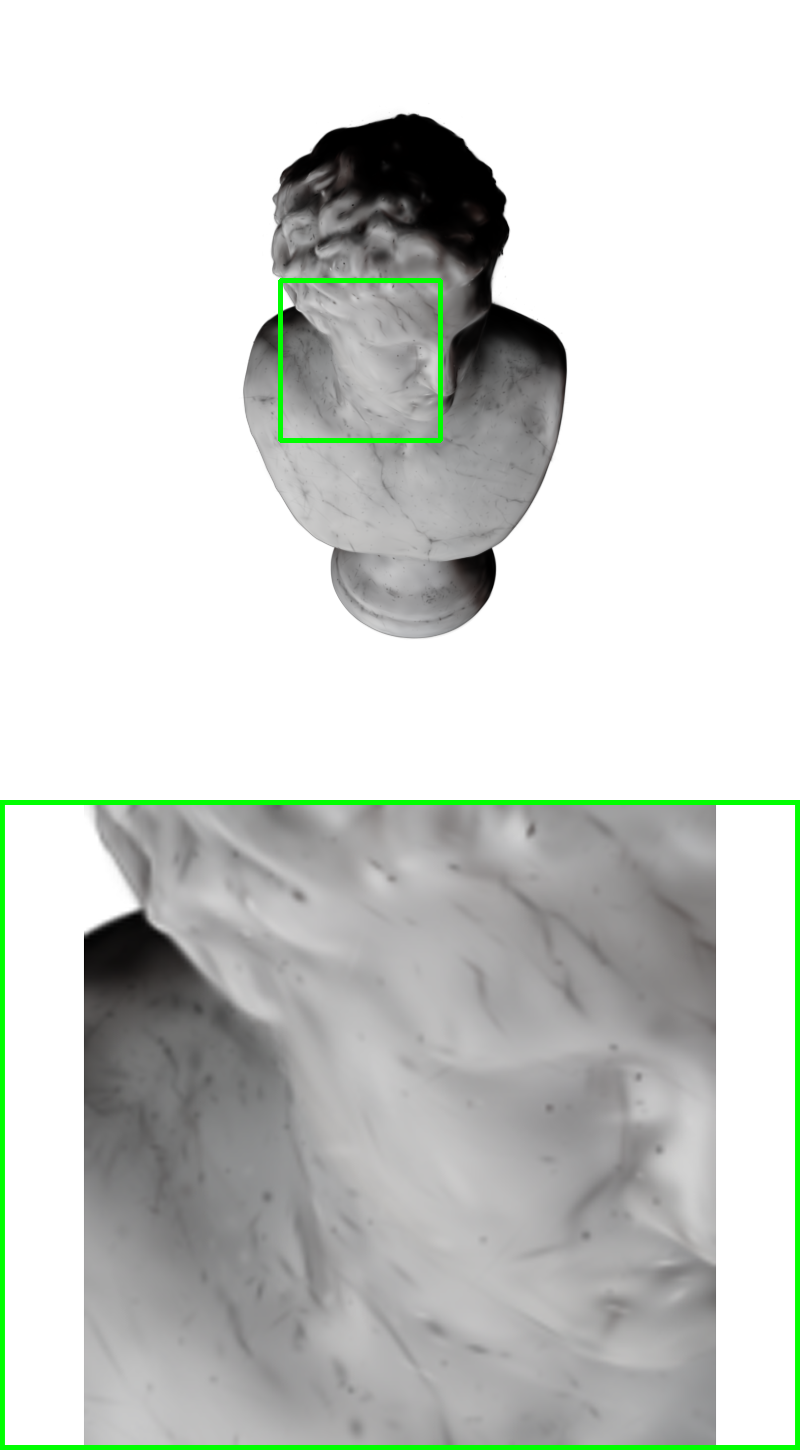}} &
\raisebox{-0.5\height}{\includegraphics[width=0.10\linewidth]{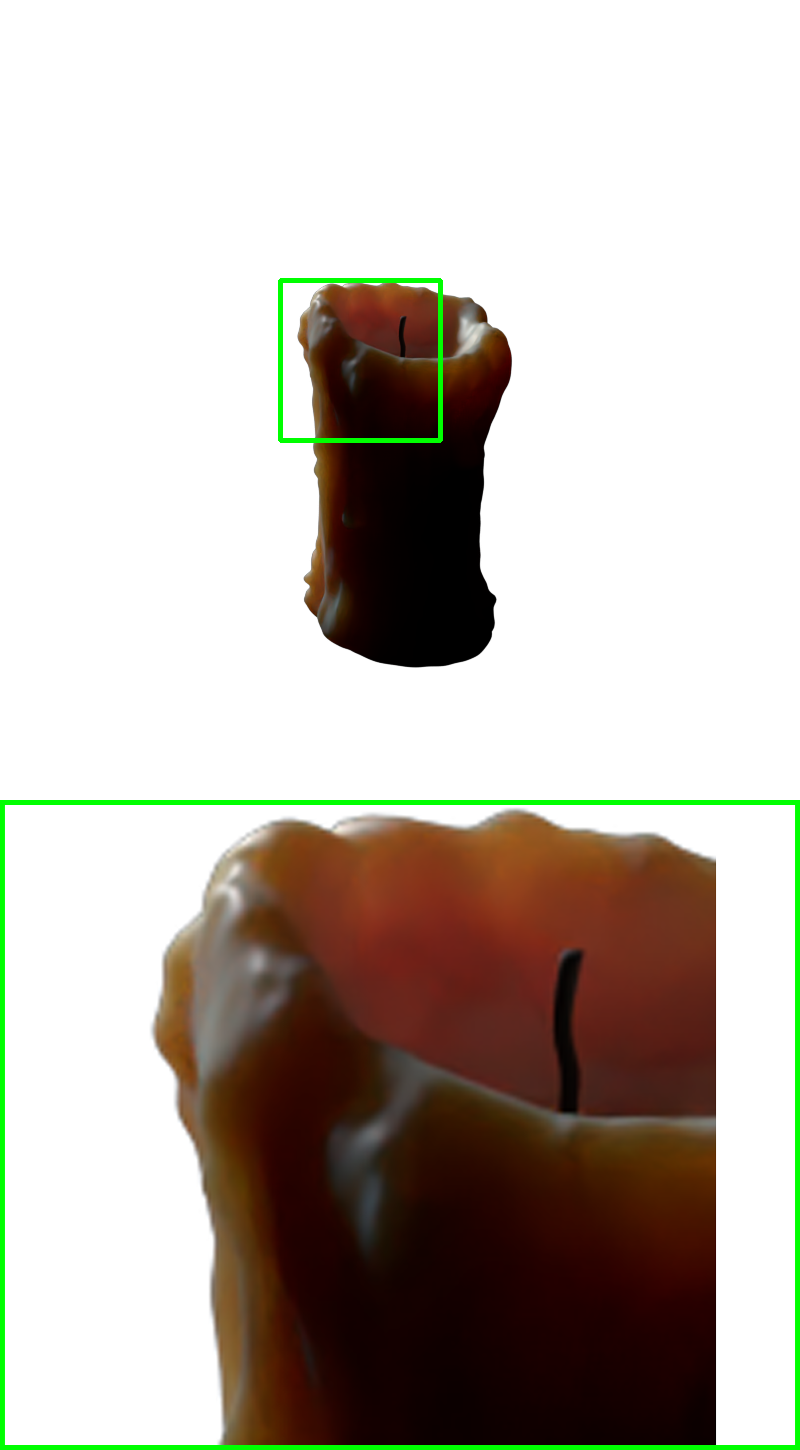}} &
\raisebox{-0.5\height}{\includegraphics[width=0.10\linewidth]{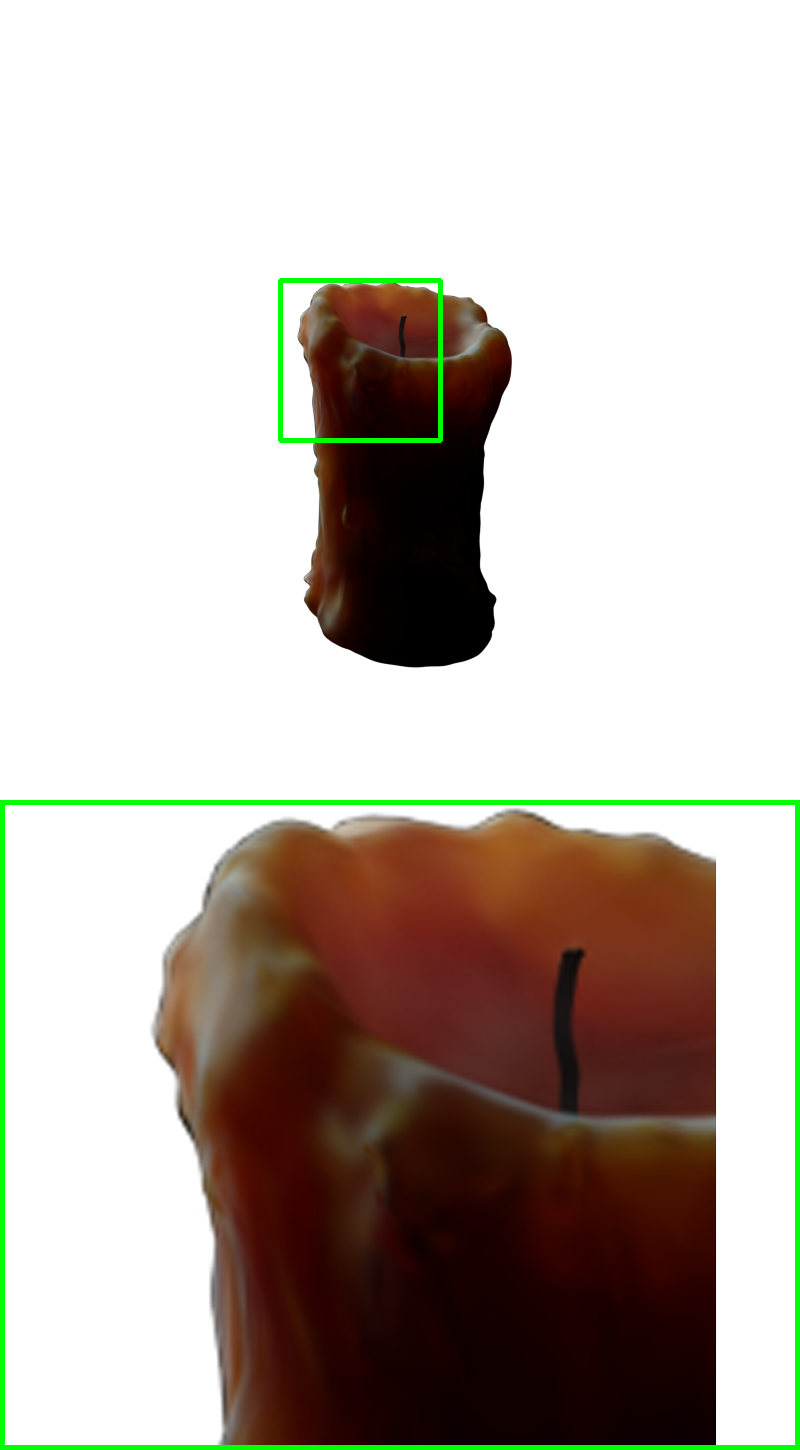}} &
\raisebox{-0.5\height}{\includegraphics[width=0.10\linewidth]{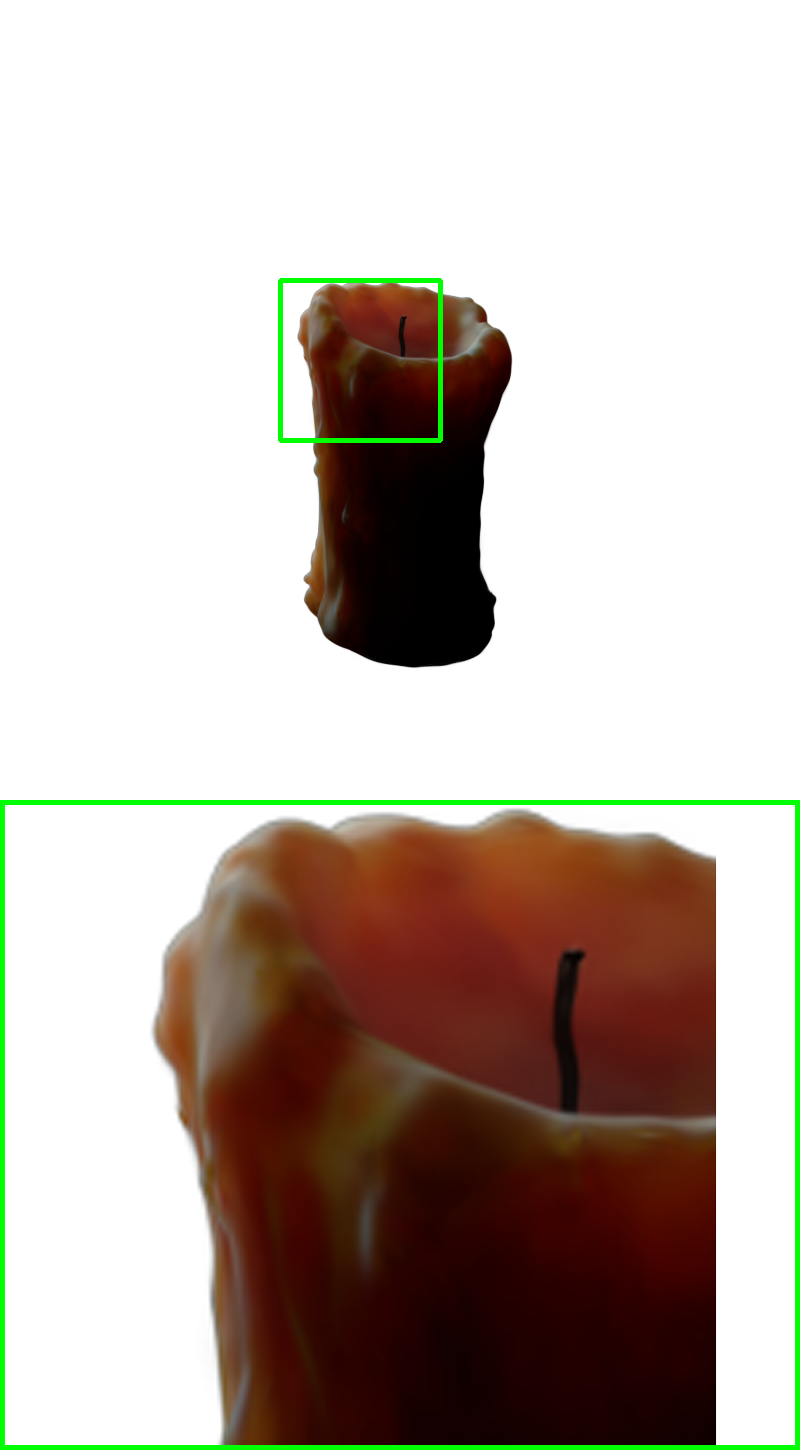}} &
\raisebox{-0.5\height}{\includegraphics[width=0.10\linewidth]{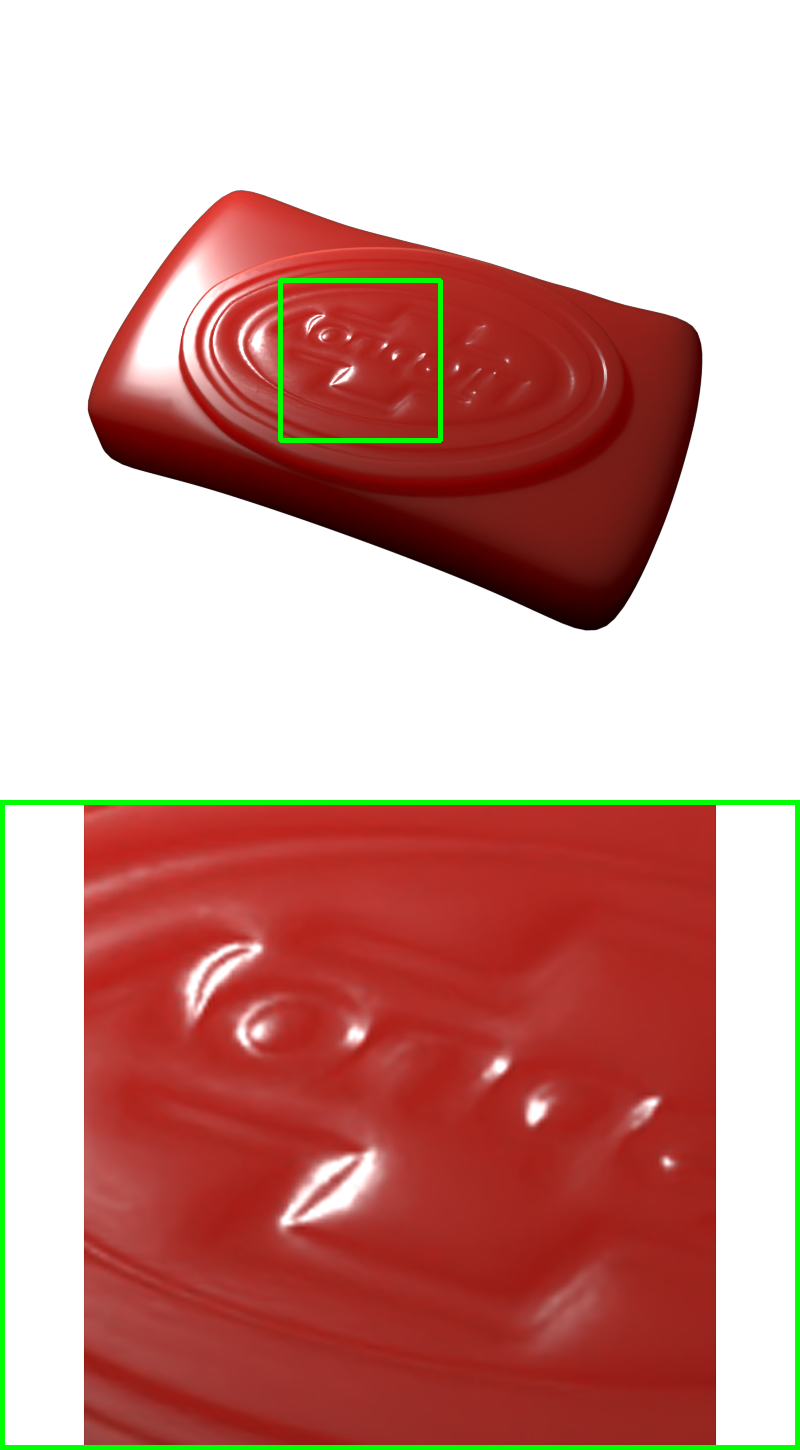}} &
\raisebox{-0.5\height}{\includegraphics[width=0.10\linewidth]{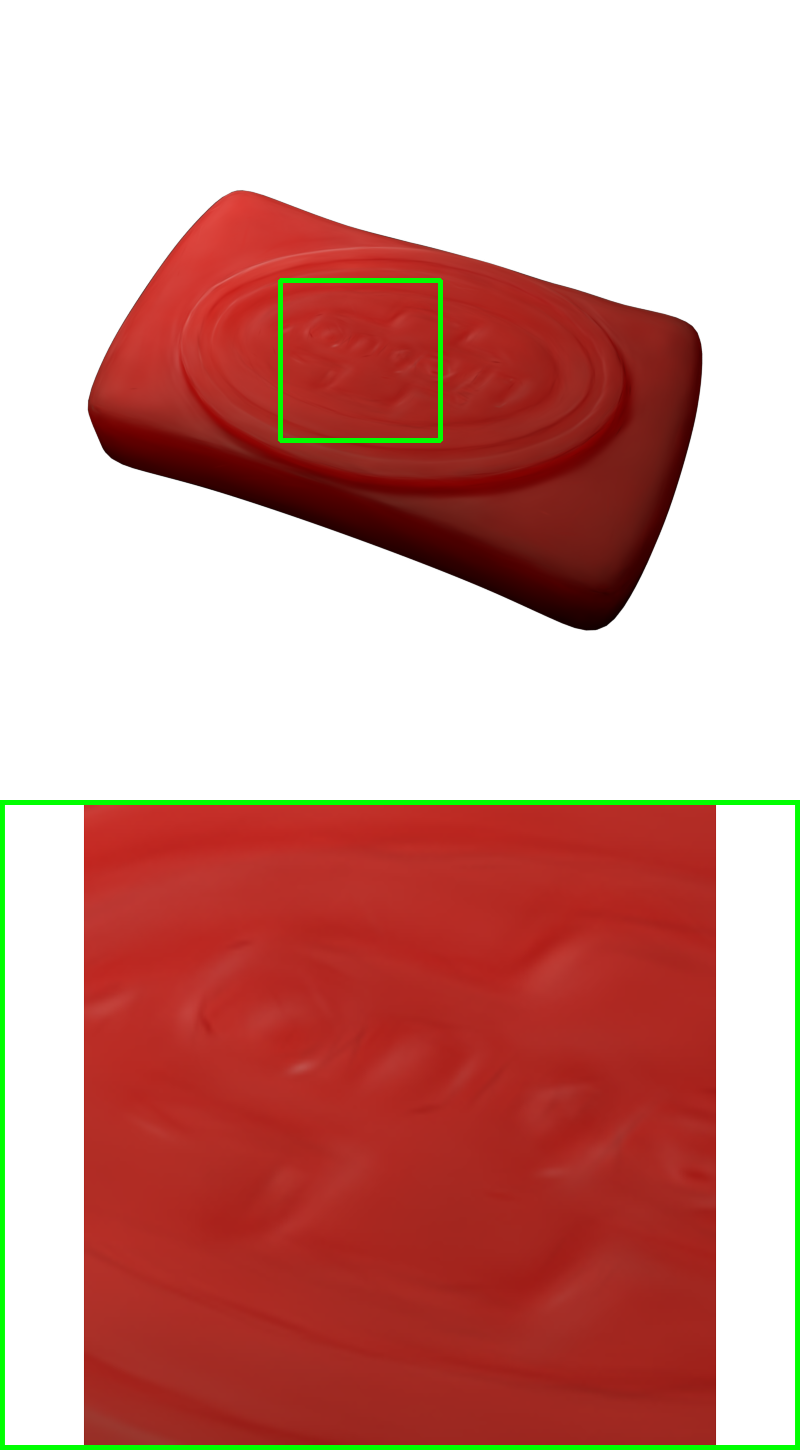}} &
\raisebox{-0.5\height}{\includegraphics[width=0.10\linewidth]{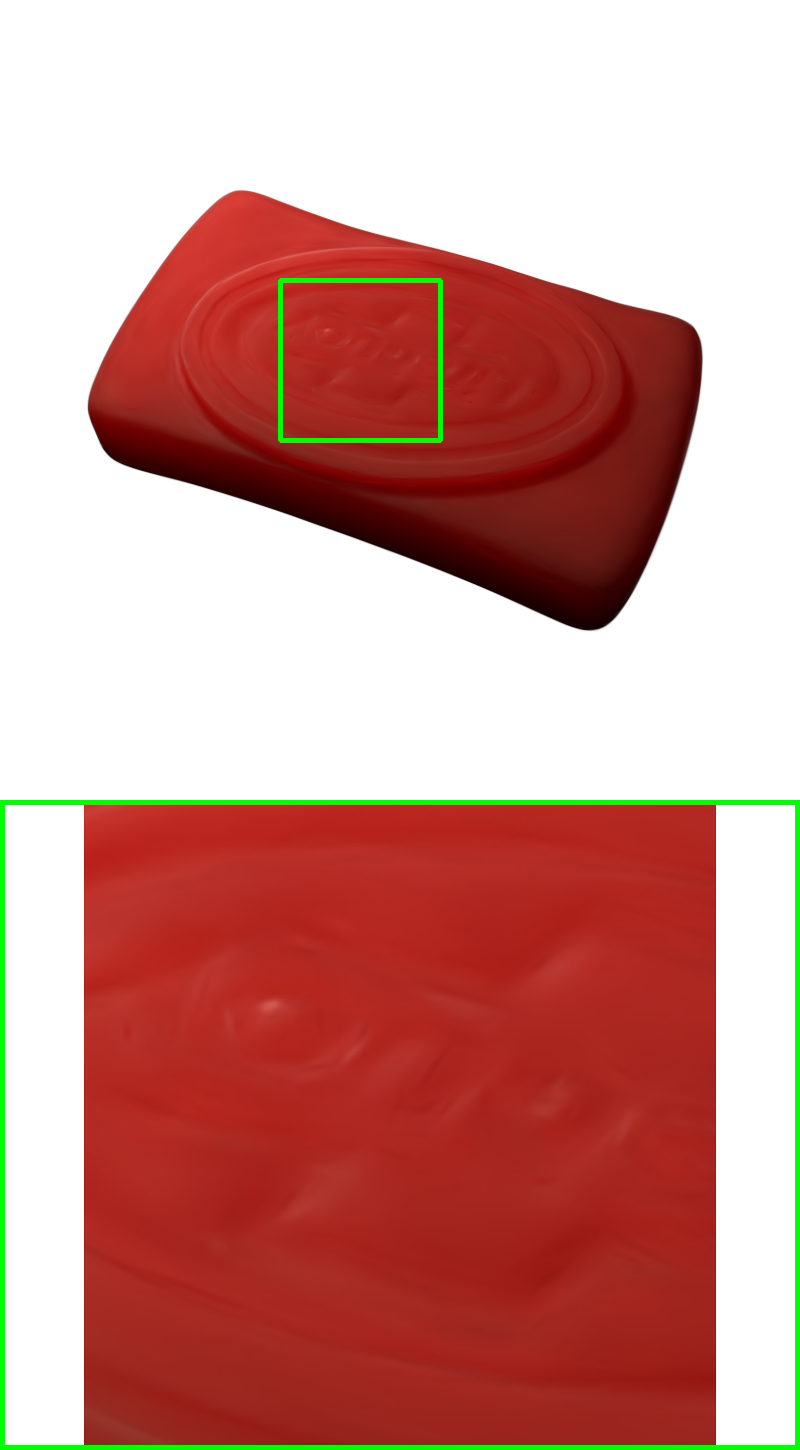}} \\
\midrule

\scriptsize all views,\;single light &
\includegraphics[width=0.10\linewidth]{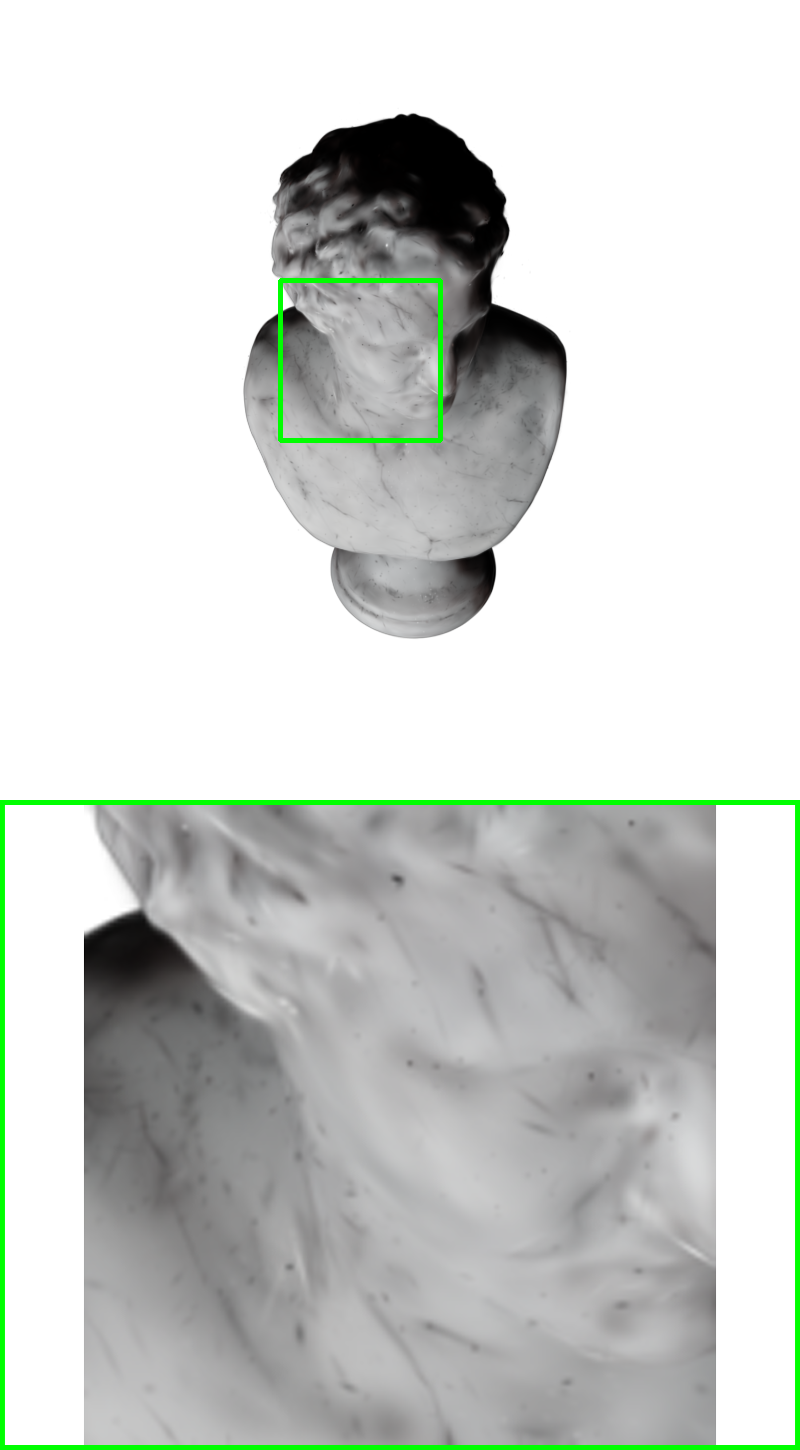} &
\includegraphics[width=0.10\linewidth]{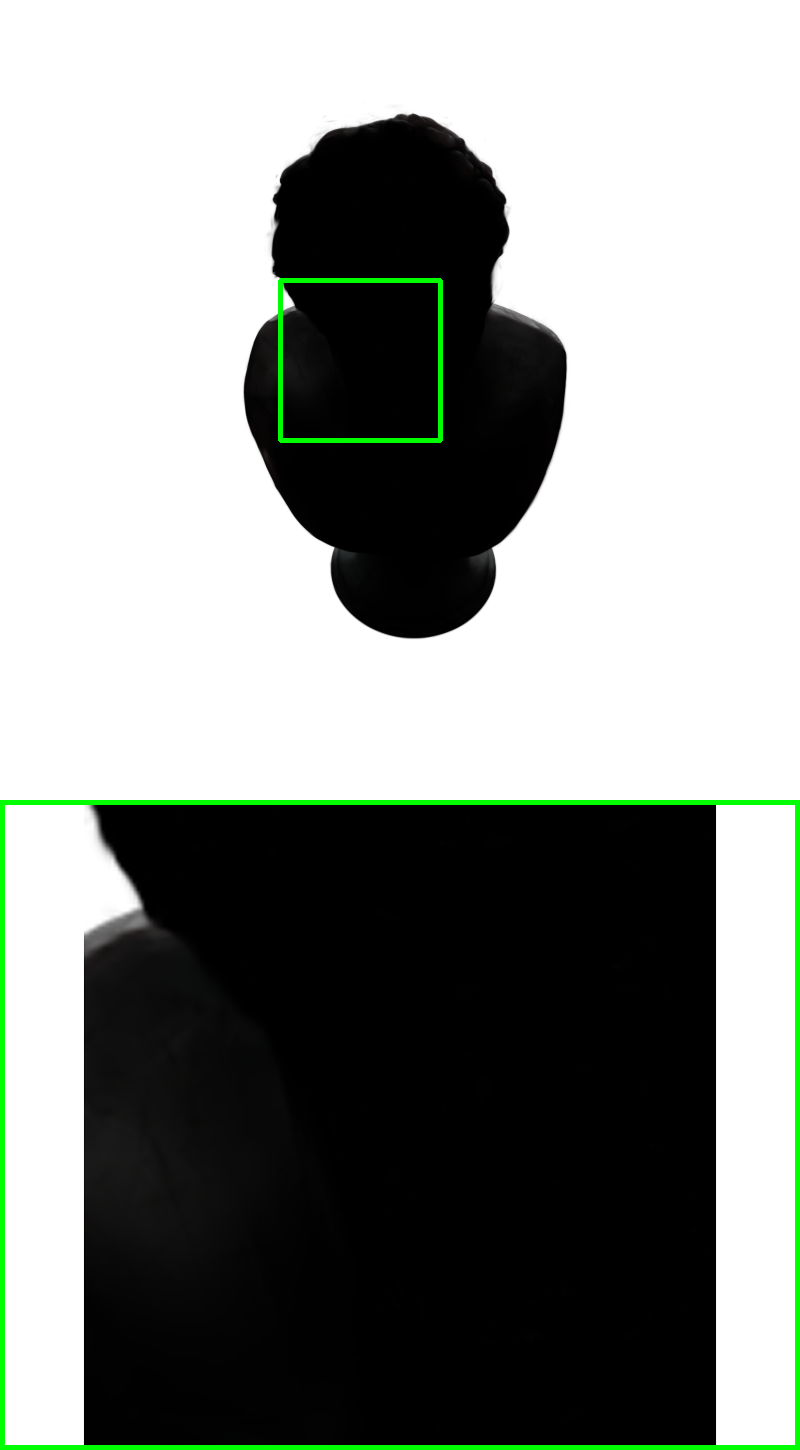} &
\scriptsize all views,\;single light &
\includegraphics[width=0.10\linewidth]{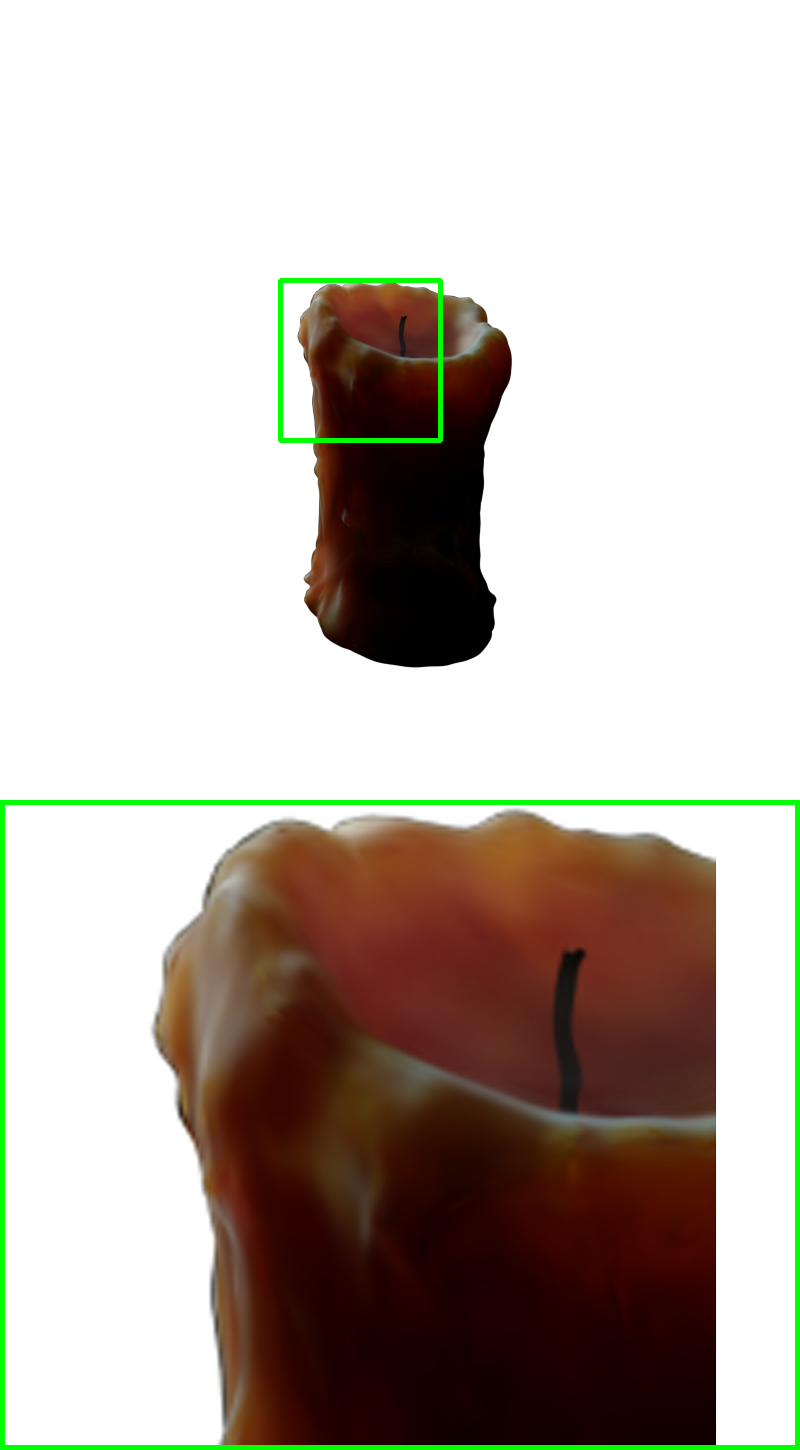} &
\includegraphics[width=0.10\linewidth]{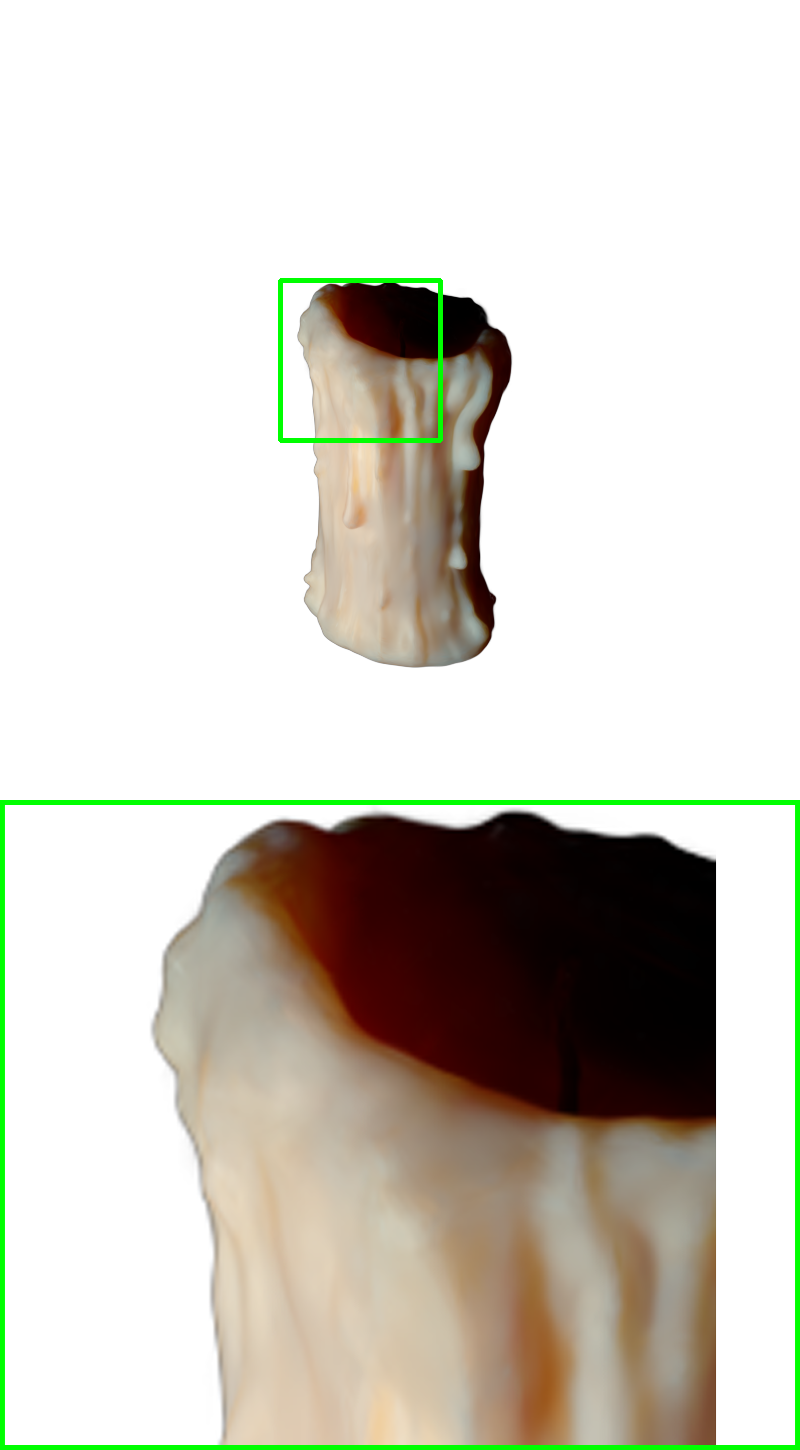} &
\scriptsize all views,\;single light &
\includegraphics[width=0.10\linewidth]{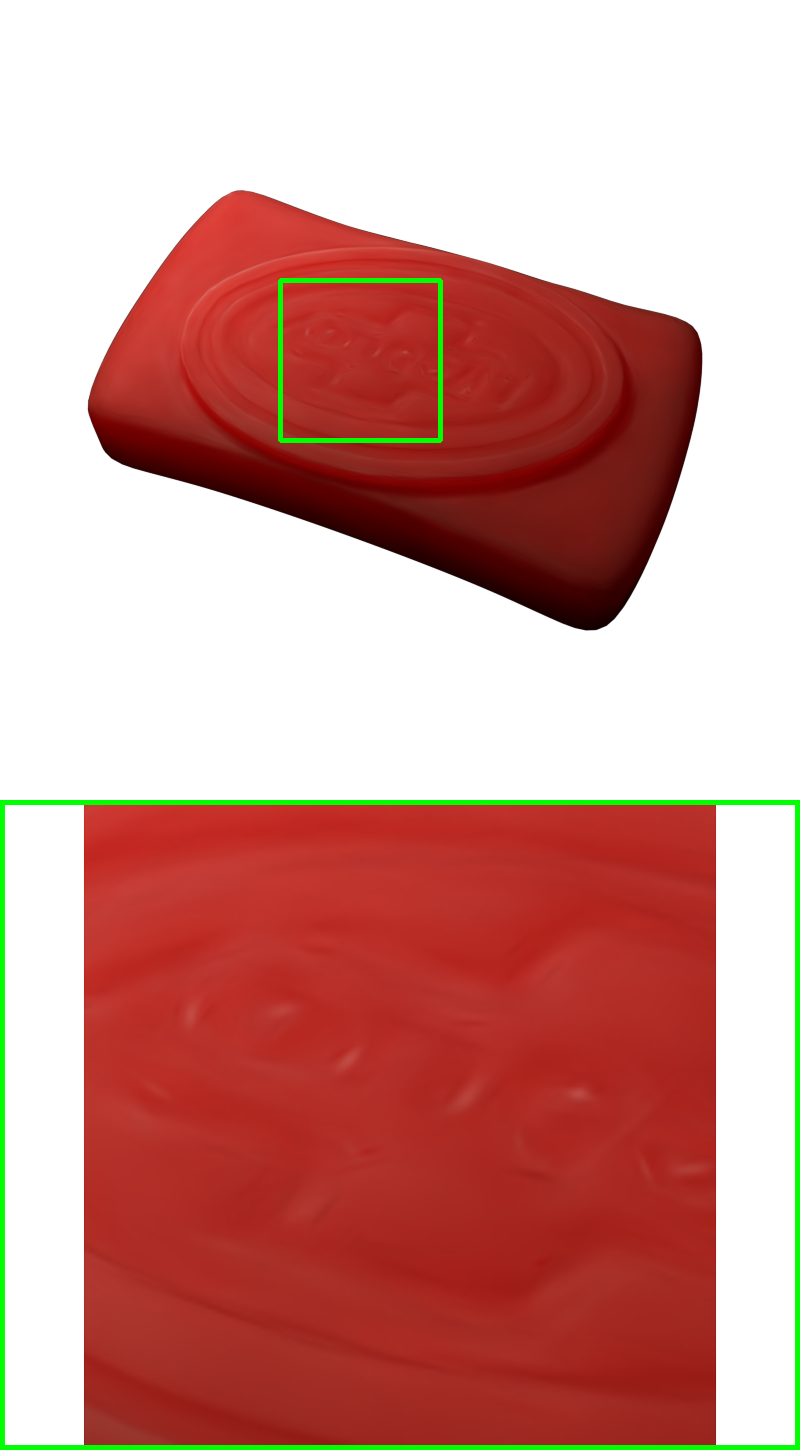} &
\includegraphics[width=0.10\linewidth]{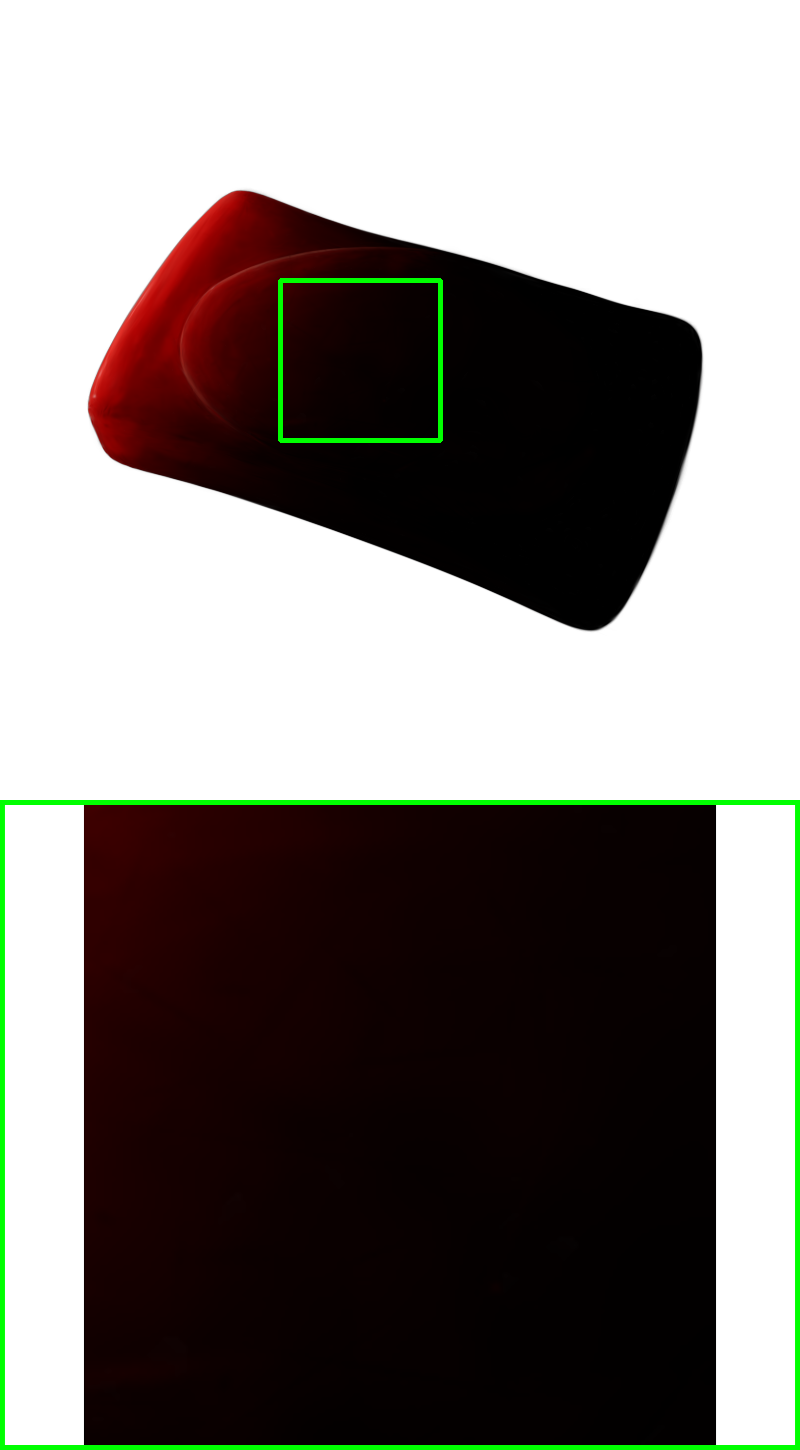} \\
\midrule

\scriptsize 5\% views,\;5\% lights &
\includegraphics[width=0.10\linewidth]{figures/reconstructions/statue/sil_depth_10000_dataset_50pcameras_50plights_00056.png} &
\includegraphics[width=0.10\linewidth]{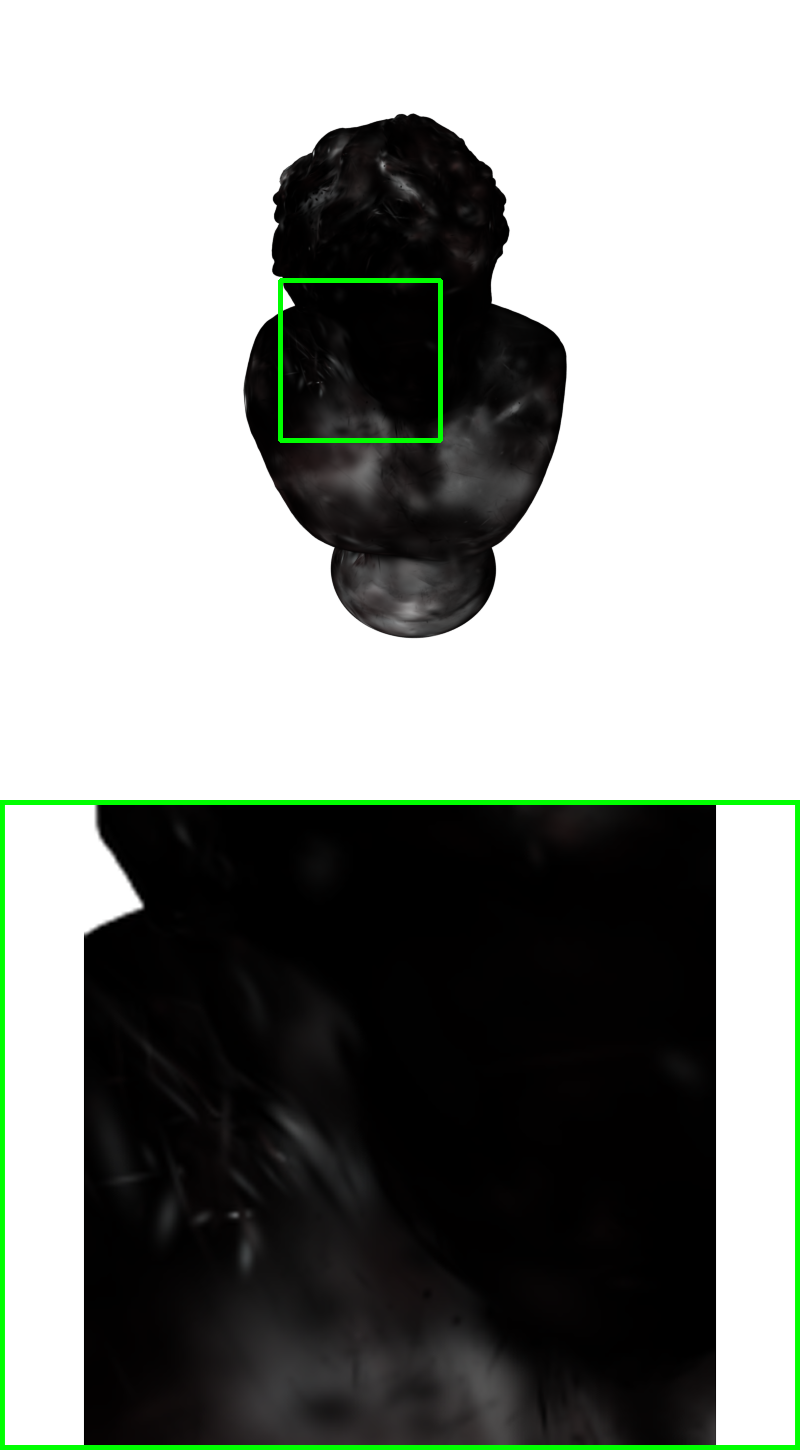} &
\scriptsize 5\% views,\;5\% lights &
\includegraphics[width=0.10\linewidth]{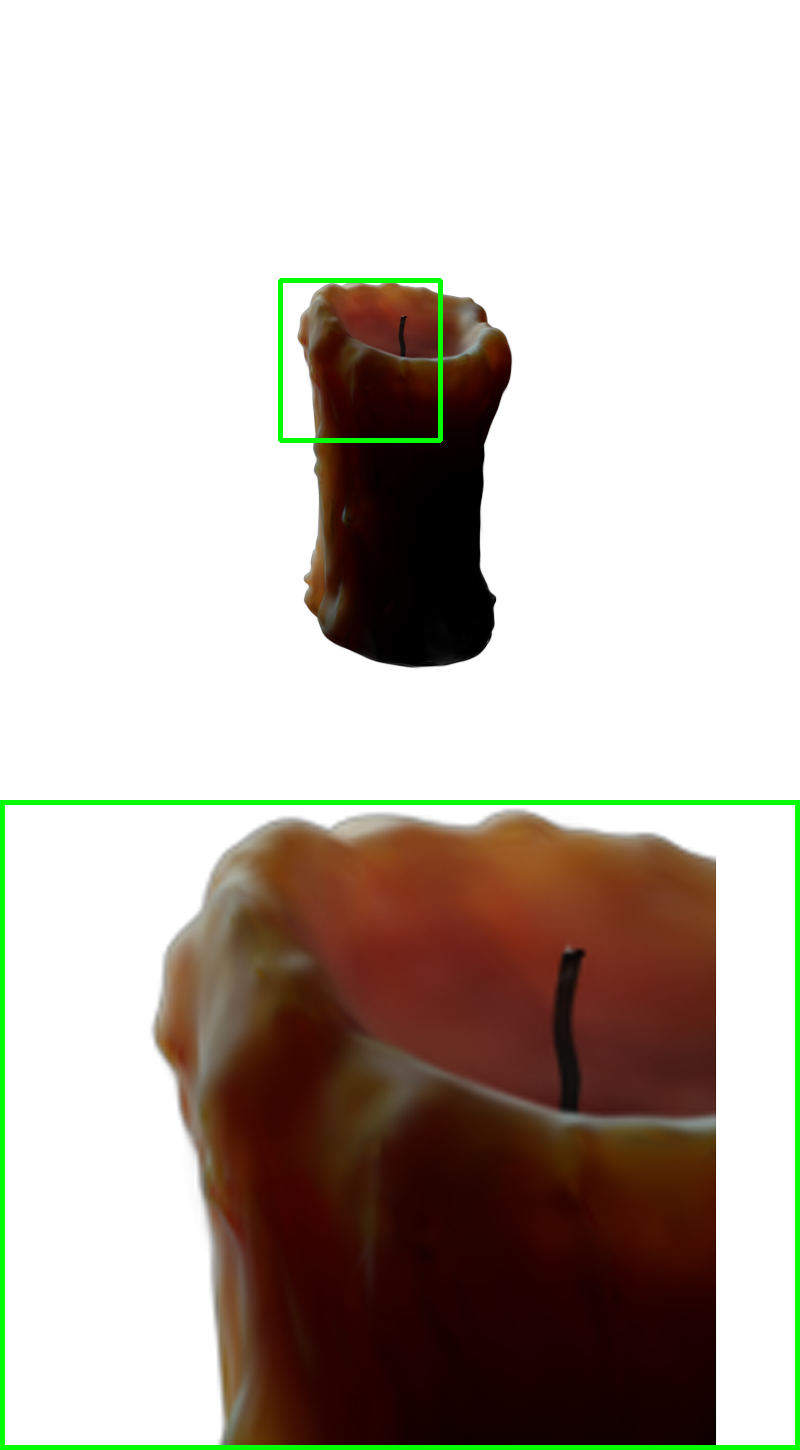} &
\includegraphics[width=0.10\linewidth]{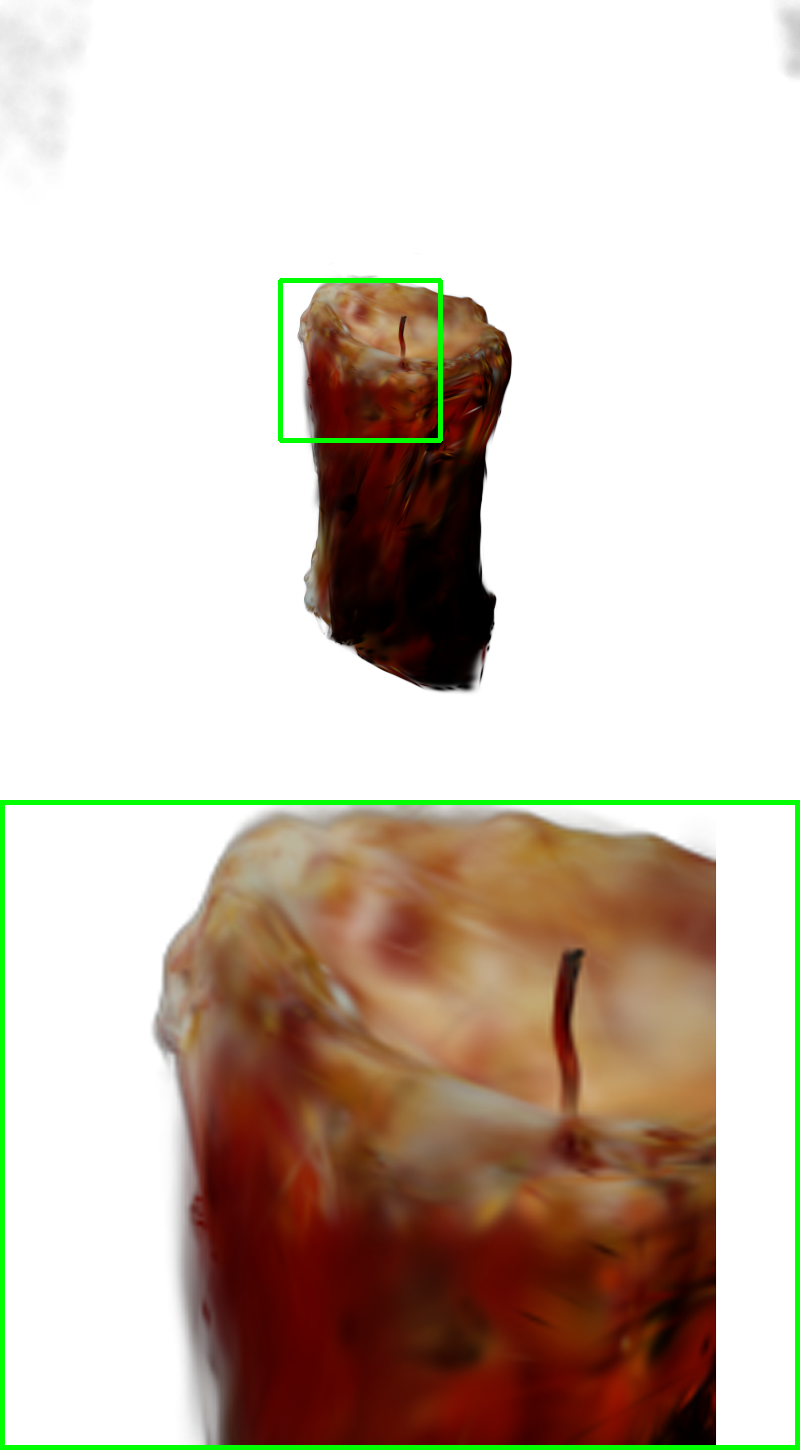} &
\scriptsize 5\% views,\;5\% lights &
\includegraphics[width=0.10\linewidth]{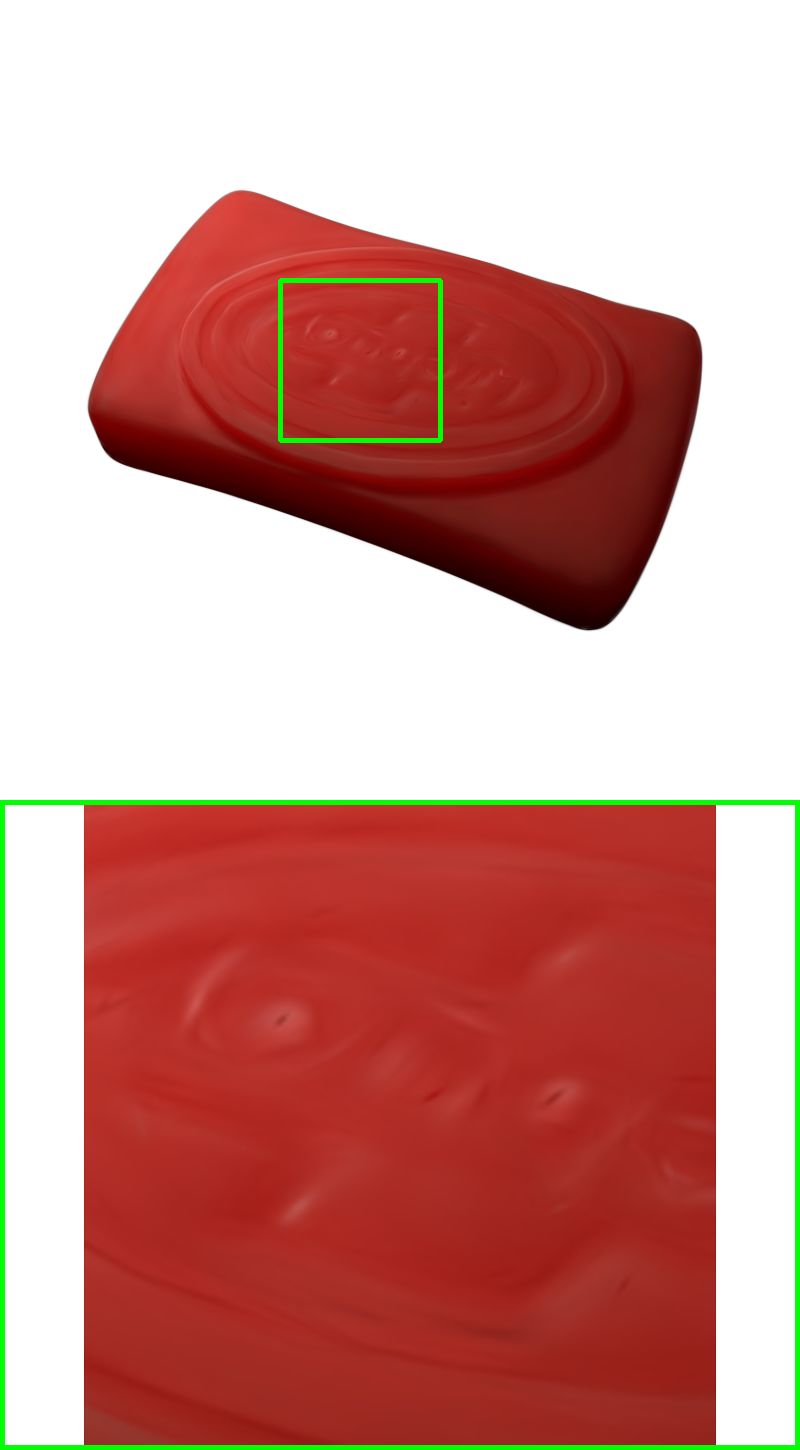} &
\includegraphics[width=0.10\linewidth]{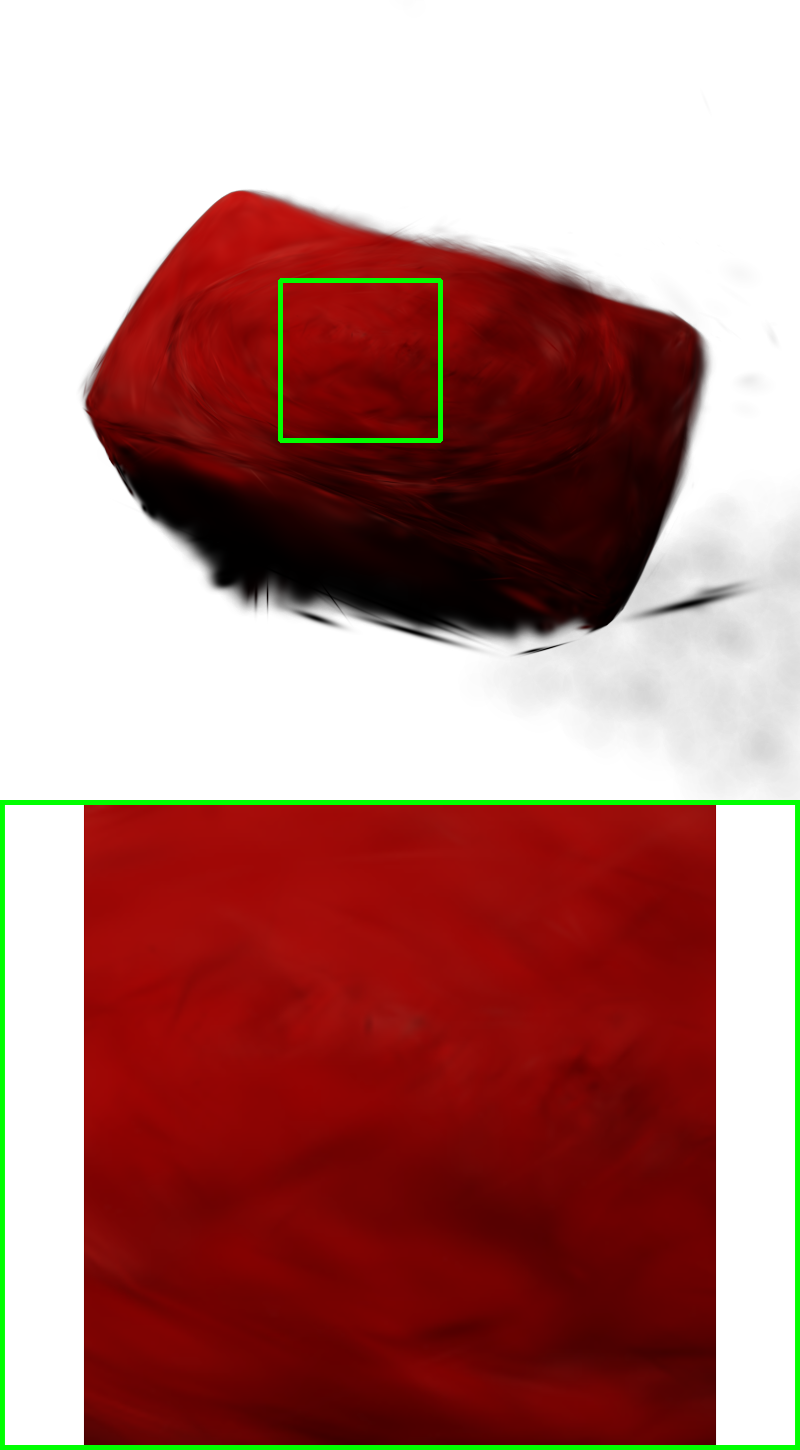} \\
\midrule

\scriptsize 3\% views,\;3\% lights &
\includegraphics[width=0.10\linewidth]{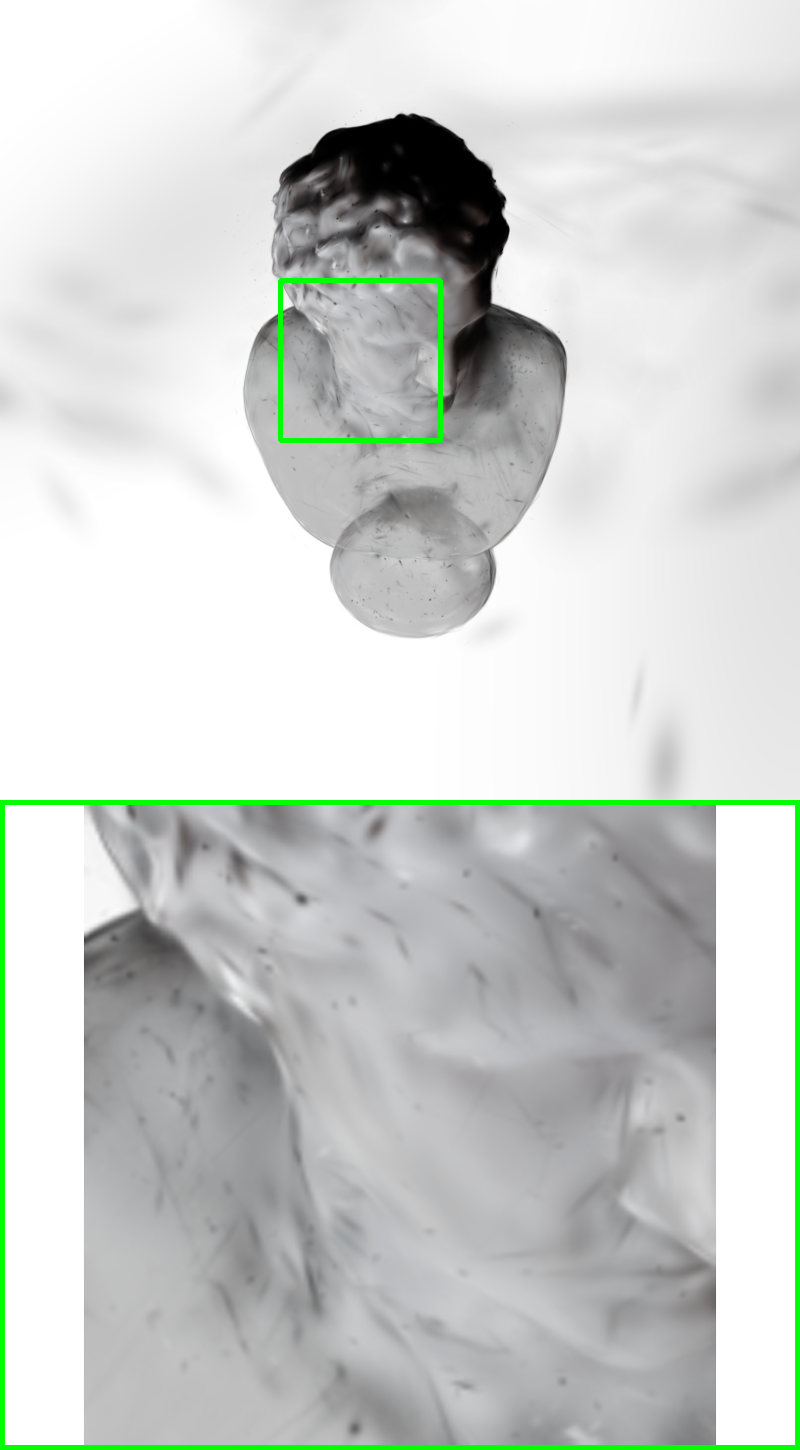} &
\includegraphics[width=0.10\linewidth]{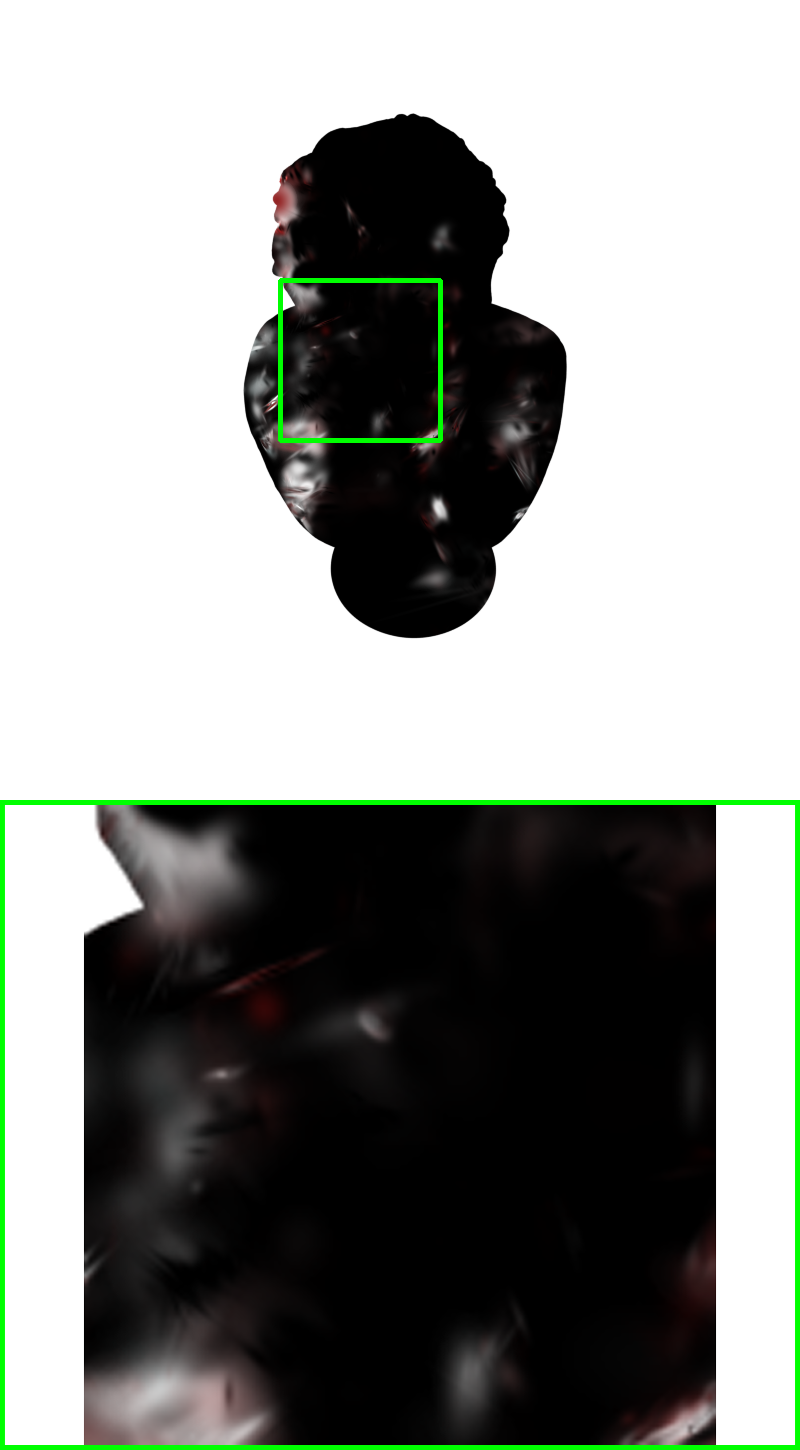} &
\scriptsize 3\% views,\;3\% lights &
\includegraphics[width=0.10\linewidth]{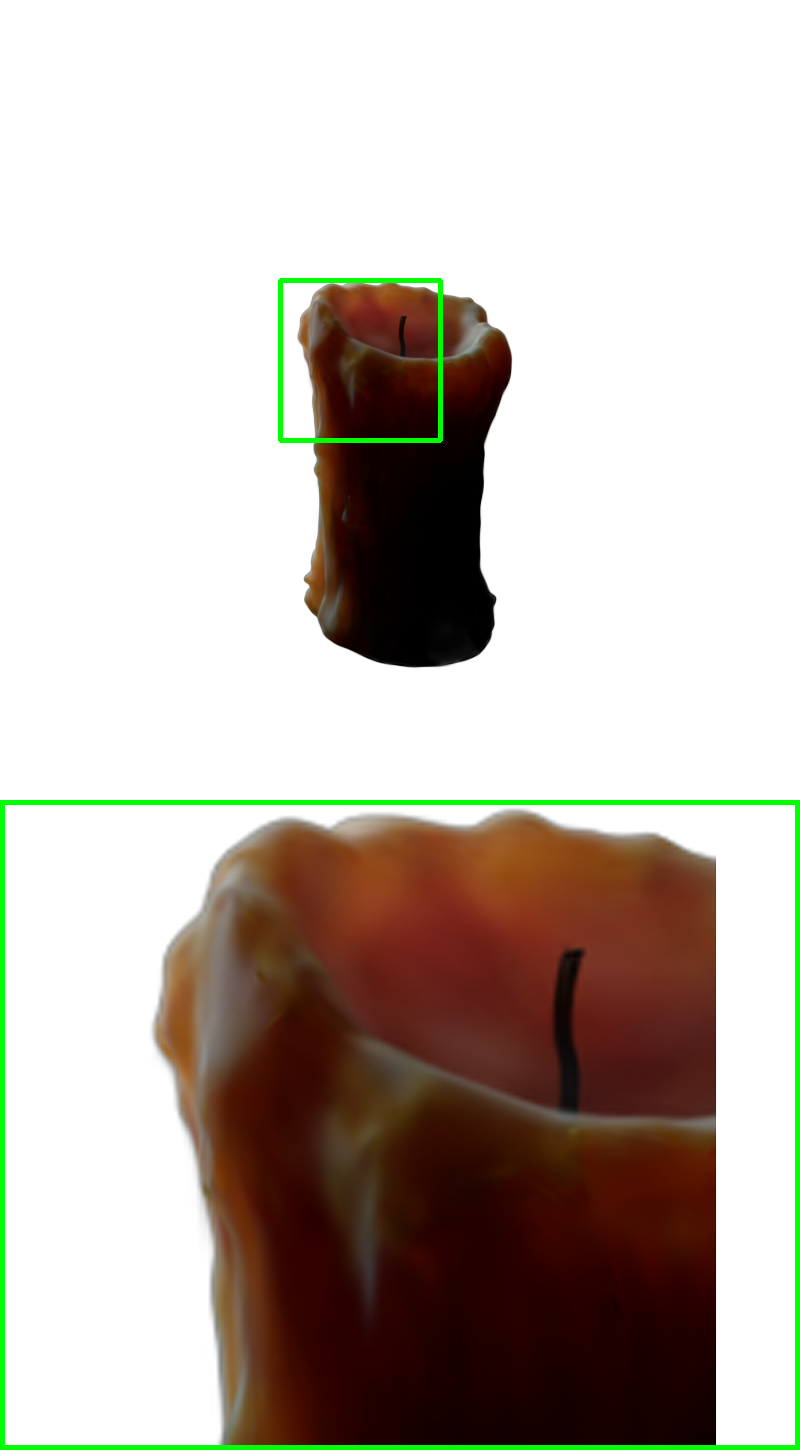} &
\includegraphics[width=0.10\linewidth]{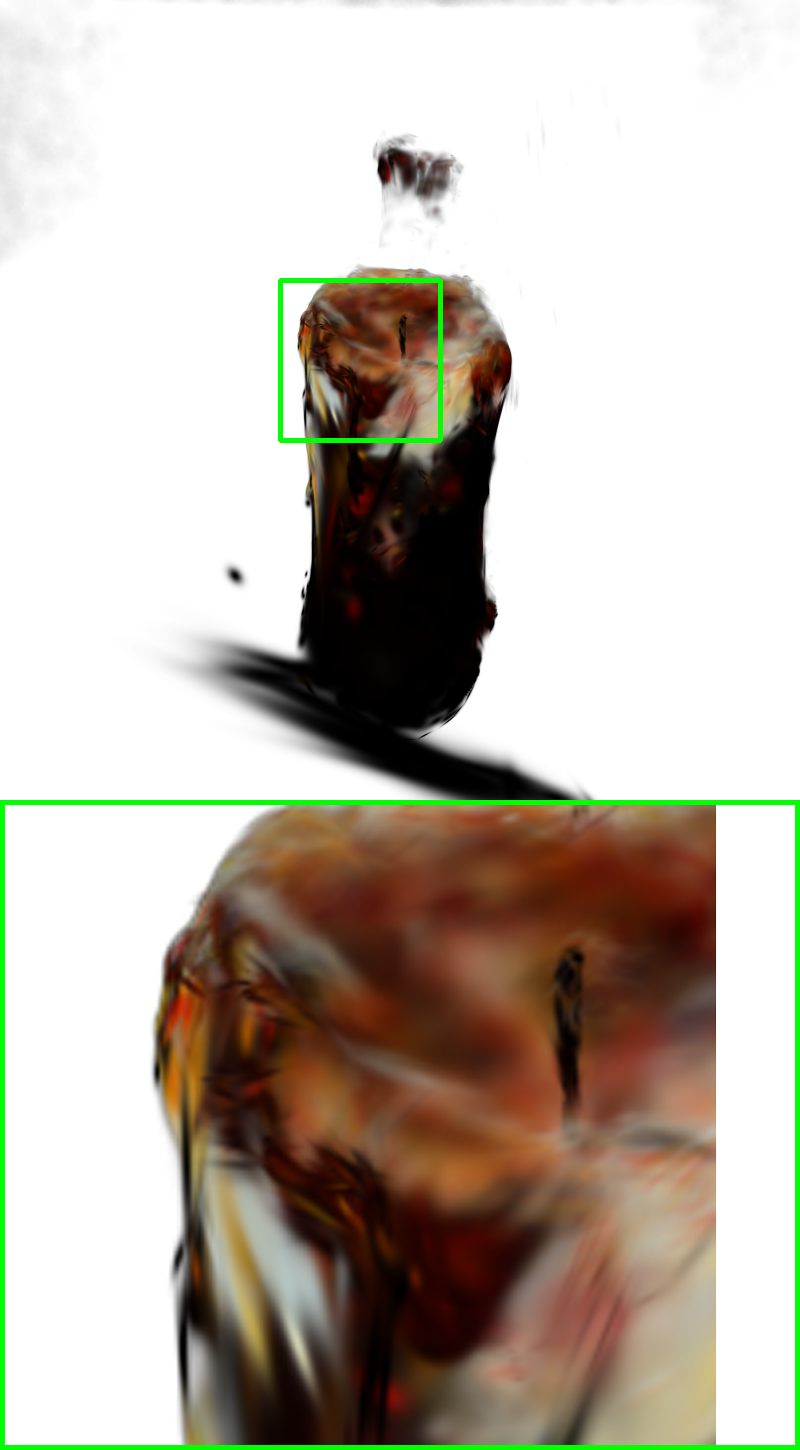} &
\scriptsize 3\% views,\;3\% lights &
\includegraphics[width=0.10\linewidth]{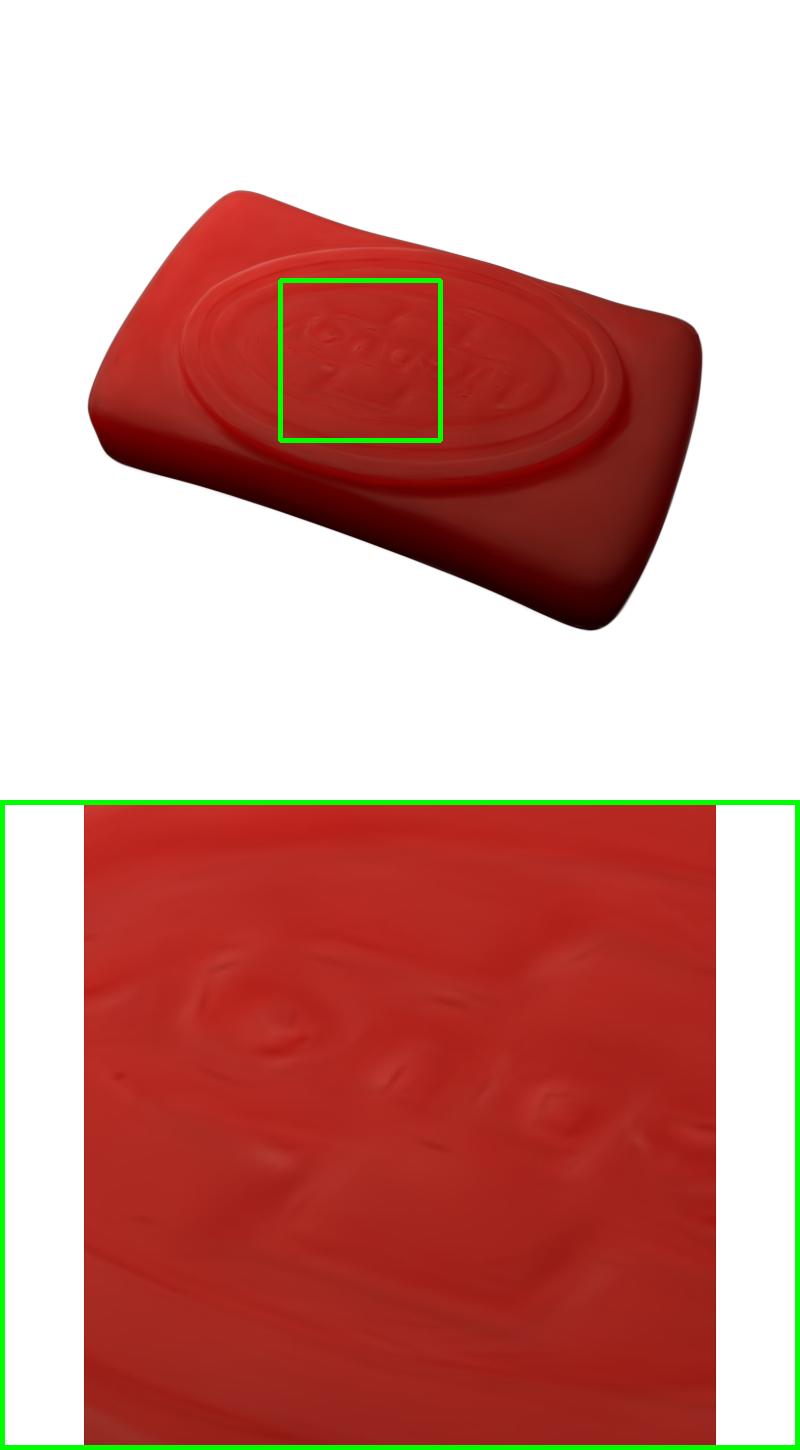} &
\includegraphics[width=0.10\linewidth]{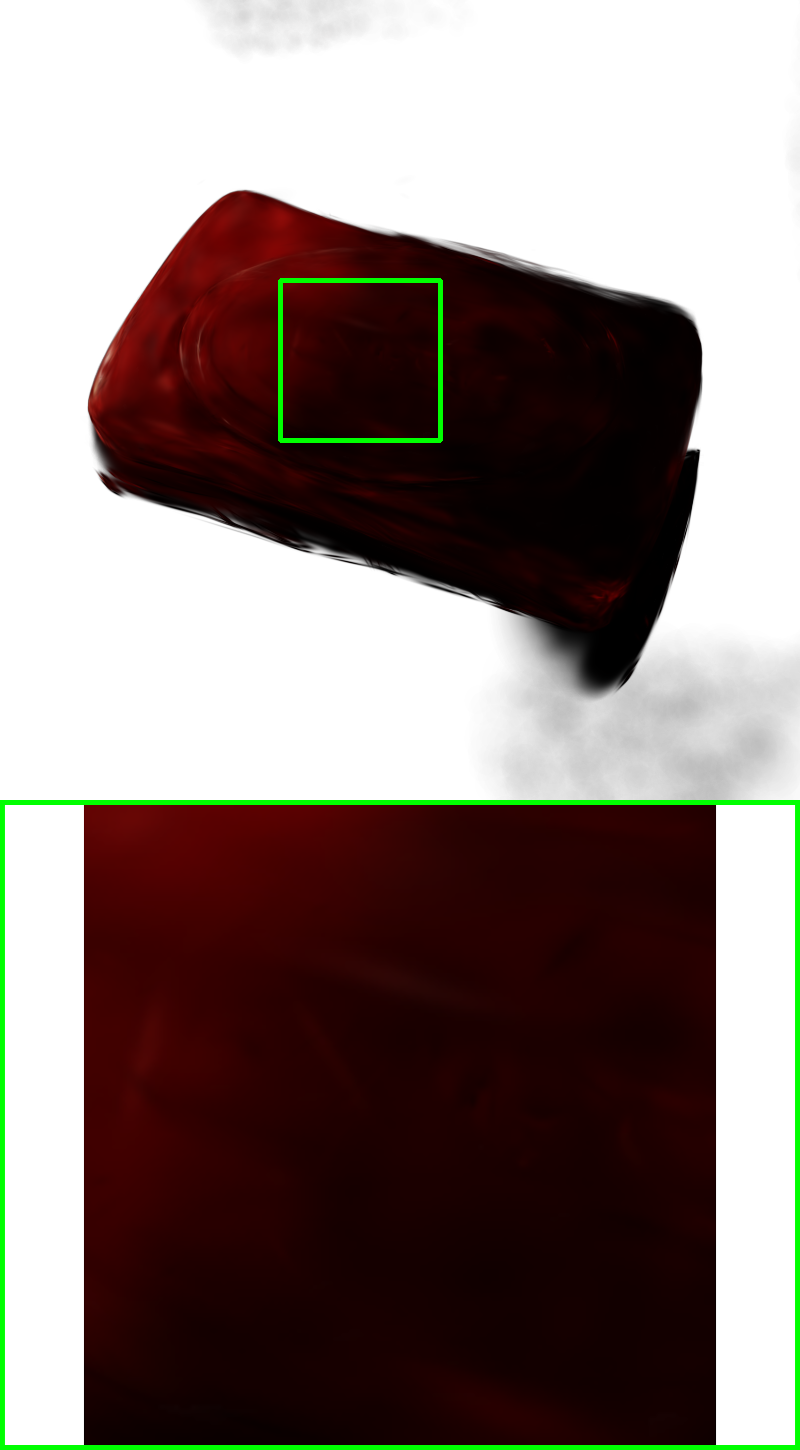} \\
\bottomrule
\end{tabular}
}
\caption{Qualitative comparison of reconstructed translucent appearance under different supervision conditions, the first row shows the setting: all views, all lights. }
\label{fig:qualitative_comparison_all}

\end{figure*}

\begin{table*}[htb!]
\small
\centering
\setlength{\tabcolsep}{4.5pt}
\begin{tabular}{lcccccc}
\toprule
& \multicolumn{3}{c}{\textbf{SSS-3DGS}} & \multicolumn{3}{c}{\textbf{\ours-SSS (ours)}} \\
\cmidrule(lr){2-4} \cmidrule(lr){5-7}
\textbf{Regime} & PSNR↑ & SSIM↑ & LPIPS↓ & PSNR↑ & SSIM↑ & LPIPS↓ \\
\midrule
Full views, full lights      & 35.01 & 0.97 & 0.040 & \textbf{36.79} & \textbf{0.99} & \textbf{0.022} \\
Full views, 1 light / view   & 21.94 & 0.88 & 0.177 & \textbf{35.14} & \textbf{0.98} & \textbf{0.030} \\
5\% views, 5\% lights        & 20.10 & 0.64 & 0.215 & \textbf{24.56} & \textbf{0.89} & \textbf{0.035} \\
3\% views, 3\% lights        & 18.15 & 0.58 & 0.319 & \textbf{21.34} & \textbf{0.86} & \textbf{0.039} \\
5\% views, full lights       & 25.32 & 0.86 & 0.060 & \textbf{32.15} & \textbf{0.96} & \textbf{0.020} \\
3\% views, full lights       & 25.32 & 0.86 & 0.060 & \textbf{31.31} & \textbf{0.96} & \textbf{0.020} \\
\bottomrule
\end{tabular}
\caption{\textbf{All capture regimes (compact).} Metrics averaged over held-out test views/lights.}
\label{tab:all_regimes_compact}
\end{table*}

\begin{table*}[t]
\small
\centering
\setlength{\tabcolsep}{6pt}
\begin{tabular}{llccc}
\toprule
\textbf{Category} & \textbf{Variant} & \textbf{PSNR↑} & \textbf{SSIM↑} & \textbf{LPIPS↓} \\
\midrule
\multirow{4}{*}{\textbf{Geometric  (10\%, 3\% lights)}}
 & w/o silhouette loss (Depth only)     & 20.697 & 0.621 & 0.331 \\
 & w/o depth loss (Silhouette only)     & 20.765 & 0.844 & \textbf{0.036} \\
 & w/o both (geom) \;(Vanilla)          & 20.222 & 0.591 & 0.319 \\
 & \textbf{Full \ours-SSS (Depth+Sil.)} & \textbf{21.342} & \textbf{0.853} & 0.039 \\
\midrule
\multirow{4}{*}{\textbf{Diffusion Augmentation (100\% views, 1 light p/v)}}
 & w/o relighting diffusion    & 23.20 & 0.90 & 0.150 \\
 & w/o novel-view diffusion    & 35.05 & 0.98 & 0.031 \\
 & w/o both (diffusion)        & 23.10 & 0.90 & 0.152 \\
 & \textbf{Full \ours-SSS (Depth+Sil.)} & \textbf{35.14} & \textbf{0.98} & \textbf{0.030} \\
\bottomrule
\end{tabular}
\caption{\textbf{Unified ablation of geometric consistency and diffusion augmentation.} 
Geometric losses are evaluated on \emph{real data} (10\% views, 3\% lights), 
and diffusion variants on the \emph{full views, 1 light per view} regime. }
\label{tab:ablation_combined}
\end{table*}

We evaluate \ours-SSS on translucent object benchmarks following the SSS-3DGS dataset setup~\cite{dihlmann2024subsurfacescattering3dgaussian}. 
Our experiments are designed to validate three key aspects of our approach: 
(1) the effectiveness of diffusion-based augmentation for replacing real multi-light captures (\cref{subsec:diffusion_augmentation}), 
(2) the stabilizing effect of multi-view geometric consistency losses under sparse or synthetic supervision (\cref{subsec:geometric_consistency}), and 
(3) the overall reconstruction quality and data efficiency compared to state-of-the-art SSS reconstruction methods. 
A high-level overview of our pipeline is provided in \cref{fig:approach} and detailed in \cref{sec:method}.

\paragraph{Experimental Protocol.}
To analyze robustness under limited supervision, we adopt a \textbf{view--light pruning strategy} inspired by prior work. 
Beginning with the full OLAT capture ($\sim$100 viewpoints, 112 lights per viewpoint), we progressively subsample camera poses and their associated illuminations while ensuring uniform coverage in pose--light space. 
This controlled reduction yields capture regimes ranging from full data to \textbf{3\%} of views and lights, enabling us to quantify how reconstruction quality and geometric stability degrade across sparsity levels and to identify the practical breakpoints where competing methods fail.

We focus on six capture regimes that isolate each component of \ours-SSS:
\begin{enumerate}
    \item \textbf{All views, all lights}: $\sim$100 views and 112 OLAT illuminations per viewpoint (probes the effect of geometric losses, \cref{subsec:geometric_consistency}).
    \item \textbf{All views, 1 light per view}: $\sim$100 views and 1 OLAT per view (tests diffusion-based relighting, \cref{subsec:diffusion_augmentation}).
    \item \textbf{5\% views, 5\% lights}: $\sim$5\% of views and 5\% of lights (joint sparse, tests both diffusion modules and geometric losses).
    \item \textbf{3\% views, 3\% lights}: $\sim$3\% of views and 3\% of lights (extreme sparse).
    \item \textbf{5\% views, full lights}: $\sim$5\% of views and all lights (tests diffusion novel-view synthesis, \cref{subsec:diffusion_augmentation}).
    \item \textbf{3\% views, full lights}: $\sim$3\% of views and all lights (extreme sparse views; isolates novel-view diffusion).
\end{enumerate}
Following~\cite{dihlmann2024subsurfacescattering3dgaussian}, all comparisons are made against SSS-3DGS, the strongest prior baseline.  
We omit KiloOSF~\cite{thomas2024kiloosf} from our quantitative tables since it performs significantly below SSS-3DGS and our approach already surpasses this state-of-the-art method across all regimes.

\subsection{Datasets}
\label{subsec:datasets}

We use synthetic and real translucent objects from the SSS-3DGS dataset~\cite{dihlmann2024subsurfacescattering3dgaussian}, captured under $\sim$100 viewpoints and 112 one-light-at-a-time (OLAT) illuminations. o our knowledge, this is the only publicly available dataset with these characteristics. Objects include wax candles, jade figurines, plastic toys, and marble statues. For our six evaluation regimes, we subsample the original capture following the previously presented overview. Silhouettes are obtained from provided masks or automatic background subtraction. Depth and normal maps are estimated using off-the-shelf predictors: Depth Anything~\cite{depth_anything_v2} and Marigold~\cite{marigold}, respectively. These cues supervise our geometric consistency terms and condition the diffusion models (\cref{subsec:diffusion_augmentation,subsec:geometric_consistency}). In the supplementary material we present a visual overview of the dataset. As described in \cref{subsec:diffusion_augmentation}, both diffusion models are fine-tuned once using \textbf{$\leq$ 7\% of the total dataset}, drawn proportionally from four representative translucent objects (\emph{car, jam jar, red candle, head}). 
This ensures robustness across material/transport variations while avoiding per-object retraining.

\subsection{Implementation Details}
\label{subsec:implementation}

We initialize 3D Gaussians from COLMAP reconstructions following~\cite{dihlmann2024subsurfacescattering3dgaussian}. 
Unless otherwise stated, each object is optimized with Adam ($5{\times}10^{-3}$ for Gaussian parameters, $1{\times}10^{-3}$ for the MLP, decayed by $0.1$ after 50k steps). 
The SSS residual MLP has four layers with 64 hidden units and ReLU activations (see \cref{subsec:sss3dgs}). A more detailed implementation road-map can be found in the Supplementary section, as well as a table with all the hyperparemters used.

\paragraph{Diffusion-Based Augmentation.}

We fine-tune the official Free3D~\cite{zheng24}
checkpoint on our OLAT multi-view data using the original architecture, loss, and linear noise schedule ($0.00085\!\rightarrow\!0.0120$ over 1000 steps). We train for $\sim$500 epochs with lr $1{\times}10^{-4}$, 100 warm-up steps, batches of 16 views from 4 objects, and $32{\times}32$ latents (scale $0.18215$), using gradient checkpointing. Synthetic views are used during reconstruction with photometric losses down-weighted by $\alpha = 0.5$. For relighting, we adopt the implementation
of Poirier-Ginter\,\etal~\cite{poirier2024radiancefield}, fine-tuning a ControlNet-style diffusion model conditioned on light direction to generate synthetic OLAT data from single-illumination captures. This augmented data trains a relightable 3DGS model with an MLP for light/view conditioning and per-image latent vectors for consistency.

\paragraph{Geometric Consistency.}
We implement the multi-view silhouette and depth consistency losses from \cref{subsec:geometric_consistency} using depth-based reprojection: each pixel in a source view is back-projected into 3D, transformed into the target camera, and bilinearly sampled to obtain the corresponding silhouette and depth values. Consistency is enforced only on geometrically valid correspondences, filtered by visibility checks and per-pixel validity masks derived from depth reprojection. Silhouette alignment is supervised with a binary cross-entropy loss, and depth agreement with an $\ell_1$ loss, both normalized by the number of valid pixels. Unless stated otherwise, we use weights $\lambda_{\text{sil}} = 0.5$ and $\lambda_{\text{depth}} $$= 0.3$.

\subsection{Qualitative Results}
\label{subsec:qualitative}

Figure~\ref{fig:qualitative_comparison_all} compares reconstructions across the regimes. 
Under full supervision, SSS-3DGS already produces high-quality translucency, but suffers from slight geometric inconsistencies and blurry boundaries. 
Adding our geometric consistency losses (\cref{subsec:geometric_consistency}) sharpens silhouettes and improves depth coherence. 
As views and lights become sparse, the baselines exhibit strong degradation, whereas \ours-SSS maintains realistic subsurface scattering, color bleeding, and soft shadows, particularly when diffusion-based relighting (\cref{subsec:diffusion_augmentation}) is used to compensate for missing OLAT captures. More results are depicted in the Supplementary section.

\subsection{Quantitative Results Across Capture Regimes}
\label{subsec:quantitative}

We report PSNR, SSIM, and LPIPS averaged over held-out views and illuminations. 
\cref{tab:all_regimes_compact} consolidates all six regimes, comparing SSS-3DGS and \ours-SSS.  
Across all data regimes, \ours-SSS consistently improves fidelity—especially under sparse supervision—demonstrating that diffusion-based augmentation and geometric priors complement each other.

\subsection{Data Efficiency and Ablations}
\label{subsec:dataefficiency}

We assess each component via ablations disabling geometric losses and/or diffusion augmentation (Table~\ref{tab:ablation_combined}). Removing geometric consistency causes depth and silhouette drift, while omitting diffusion reduces photometric diversity and harms relighting. The full \ours-SSS model offers the best stability–fidelity trade-off and consistently matches or surpasses SSS-3DGS, with the largest gains under sparse supervision (\cref{tab:all_regimes_compact}). Novel-view diffusion is particularly important when views are limited. Together with \cref{fig:qualitative_comparison_all}, these results show that diffusion-based relighting and geometric regularization enable high-quality translucent reconstruction with up to \textbf{90\% less real data}. Additional qualitative ablations appear in the Supplementary section.

\section{Limitations and conclusion}
\label{sec:limitations}

While \ours-SSS achieves high-fidelity translucent reconstruction under sparse supervision, several limitations remain.

\medskip
\noindent\textit{Diffusion realism.}  
Diffusion-augmented views and relit samples, though visually consistent, are not strictly physically accurate and may exhibit slight color or scattering biases. Incorporating physics-based priors could enhance their realism.

\medskip
\noindent\textit{Efficiency and scalability.}  
Diffusion-based augmentation adds computational cost, partly offset by reduced capture needs. Lighter architectures or shared scene-level finetuning may alleviate this overhead.

\medskip
\noindent
\textit{Conclusion.}  
In summary, \ours-SSS couples diffusion-based augmentation with geometric priors to achieve photometrically consistent, data-efficient reconstruction of translucent materials. 
It reduces real capture requirements by orders of magnitude while preserving quality, paving the way toward scalable acquisition of complex appearance in unconstrained settings.

\section{Acknowledgments}

This work was supported by the ERC Advanced Grant SIMULACRON and the Munich Center for Machine Learning (MCML). In addition, the work of G. Figueroa-Araneda was also supported by Universidad T\'ecinca Federico Santa Mar\'ia and Centro Nacional de Inteligencia Artificial in Chile. 

{\small
\bibliographystyle{ieeenat_fullname}
\bibliography{main}
}

\clearpage
\appendix
\section*{Supplementary Material}
\label{sec:supplement}

This Supplementary Material provides additional technical details, ablation studies, and qualitative evaluations that complement the main paper. 
Section~\ref{sec:hyperparams} reports the full implementation details and hyperparameters for all modules in DIAMOND-SSS. 
Section~\ref{sec:supp_qualitative} provides additional qualitative reconstructions under a wide range of capture regimes, demonstrating the robustness of DIAMOND-SSS, particularly in sparse-view and sparse-light settings where SSS-3DGS fails severely. 
Section~\ref{sec:supp_diffusion} details the fine-tuning procedures for the diffusion-based components—multi-view and relighting diffusion—and presents extended qualitative comparisons against off-the-shelf NVS and relighting baselines. 
Section~\ref{sec:supp_geom} expands on the formulation and practical considerations of the multi-view geometric losses, including visibility handling, depth normalization, sampling strategy, and targeted ablation studies on both synthetic and real objects. 
Finally, Section~\ref{sec:supp_repro} summarizes compute requirements, reproducibility notes, and implementation considerations to facilitate future research and adoption.

\section{Implementation and Hyperparameters}
\label{sec:hyperparams}

In this appendix we provide implementation and optimization details that complement the main paper (see \cref{subsec:implementation}).

\subsection{Optimization and Training Schedule}
\label{sec:supp_optim}

We jointly optimize the 3D Gaussian parameters and the SSS residual MLP (see \cref{subsec:sss3dgs}) using the Adam optimizer, with separate learning rates for geometry ($5\times10^{-3}$) and appearance ($1\times10^{-3}$). Both learning rates are reduced by a factor of $0.1$ after $50$k iterations. Training runs for $80$k--$100$k steps depending on the object and capture regime.

Gaussian initialization, densification, and pruning follow the SSS-3DGS pipeline~\cite{dihlmann2024subsurfacescattering3dgaussian}. To ensure fair comparison across ablations, we keep the number of Gaussians fixed throughout optimization.

All experiments are implemented in PyTorch~2.2 with mixed precision and executed either on a single NVIDIA RTX A6000 (48\,GB) or on a workstation equipped with a GPU with at least 24\,GB VRAM, 32\,GB RAM, and roughly 100\,GB of disk space. The environment uses CUDA~11.6 for compatibility with our training framework.

\subsection{Loss Weights}
\label{sec:supp_loss_weights}

The main paper defines the full objective $\mathcal{L}_\text{total}$ in \cref{subsec:geometric_consistency}, combining photometric, regularization, and geometric consistency terms.
For completeness, we summarize the scalar weights used in all experiments.

We employ a mixed photometric objective
\begin{equation}
\label{eq:avg_loss}
\begin{split}
\mathcal{L}_\text{total} =
&(1 - \lambda_{\text{dssim}})\mathcal{L}_1 +
\lambda_{\text{dssim}}(1 - \text{SSIM}) +
\lambda_{\text{lpips}}\mathcal{L}_{\text{LPIPS}} \\
&+
\lambda_{\text{mask}}\mathcal{L}_{\text{mask}} +
\lambda_{\text{smooth}}\mathcal{L}_{\text{smooth}} +
\lambda_{\text{enh}}\mathcal{L}_{\text{enh}} \\
&+
\lambda_{\text{ray}}\mathcal{L}_{\text{ray}} +
\lambda_{\text{sil}}\mathcal{L}_{\text{sil}}^{\text{MV}} +
\lambda_{\text{depth}}\mathcal{L}_{\text{depth}}^{\text{MV}}, 
\end{split}
\end{equation}

The regularization objective proposed in \cite{dihlmann2024subsurfacescattering3dgaussian} combines several auxiliary losses that improve stability, material fidelity, and geometric consistency. First, a normal-consistency term enforces agreement between the predicted normals $\mathbf{N}$ and pseudo-normals $\tilde{\mathbf{N}}$ derived from the rendered depth map $D$ under a local planarity assumption:
\begin{equation}
    \mathcal{L}_{\text{normals}}
    = \lambda_{\text{normals}} \lVert \mathbf{N} - \tilde{\mathbf{N}} \rVert_2 .
\end{equation}
To supervise incident radiance, we constrain the clamped predicted illumination $\overline{L}_{\text{in}}(x,\boldsymbol{\omega_i})$ to match the learned spherical-harmonics visibility $V_{SH}(x,\boldsymbol{\omega_i})$ via an L1 loss
\begin{equation}
    \mathcal{L}_{\text{incident}}
    = \lambda_{\text{incident}}
    \big\|\, \overline{L}_{\text{in}}(x,\boldsymbol{\omega_i})
           - V_{SH}(x,\boldsymbol{\omega_i}) \big\|_1 .
\end{equation}
Foreground consistency is enforced with a spatial masking loss, penalizing the contribution of Gaussians outside the image mask $I_{\text{mask}}$:
\begin{equation}
    \mathcal{L}_{\text{mask}}
    = -\lambda_{\text{mask}}
      \left[
        I_{\text{mask}}\log(\alpha)
        + (1-I_{\text{mask}})\log(1-\alpha)
      \right].
\end{equation}
Material smoothness is encouraged using bilateral smoothing on metalness $m$, roughness $r$, subsurfaceness $\text{sss}$, and base color $\mathbf{b}$, each with its own weight $\lambda_{\text{smooth}}^{(q)}$; for a generic attribute $PI_q$ the loss is
\begin{equation}
    \mathcal{L}_{\text{smooth}}^{(q)}
    = \frac{1}{N}\!
      \sum_{x,y}
      M(x,y)\,
      \|\nabla PI_q(x,y)\|\,
      \exp\big(-\|\nabla I(x,y)\|\big),
\end{equation}
where $I(x,y)$ is the input image and $M(x,y)$ denotes the valid mask. To improve highlight and shadow reconstruction, the authors in \cite{dihlmann2024subsurfacescattering3dgaussian} adopt the R3DGS enhancement loss~\cite{gao2024relightable}, comparing the predicted base color $\mathbf{b}$ to a pseudo-target $T$ constructed from the input RGB image $I_{rgb}$ using a sigmoid-based blending weight $sw$:
\begin{equation}
\begin{aligned}
    \mathcal{L}_{\text{enhance}}
    &= \lambda_{\text{enhance}} \|T - \mathbf{b}\|_1, \\
    T &= sw\cdot I_{rgb}^2
        + (1-sw)\cdot \big[1 - (1-I_{rgb})^2\big].
\end{aligned}
\end{equation}

Finally, we supervise learned visibility using ray-traced ground-truth visibility $V_{RT}$:
\begin{equation}
    \mathcal{L}_{\text{raytrace}}
    = \lambda_{\text{raytrace}}
      \big\| V_{SH}(x,\boldsymbol{\omega_i})
          - V_{RT}(x,\boldsymbol{\omega_i}) \big\|_1 .
\end{equation}
Together, these terms form the complete regularization objective used during optimization. All hyperparameters used in our experiments are presented in~\cref{supp:tab:sss3dgs_hyperparams}. 

\subsection{Hyperparameter Tuning}
\label{sec:hyp_res}

The goal in this section is to understand how different configurations affect the custom objective $\mathcal{L}_\text{total}$ (see Eq.~\ref{eq:avg_loss}), which aggregates photometric and regularization terms into a single performance metric for translucent scenes.

\subsubsection{Parallel Coordinate Analysis}
\label{sec:hyp_parallel}

\begin{figure*}[htb]
    \centering
    \includegraphics[width=\linewidth]{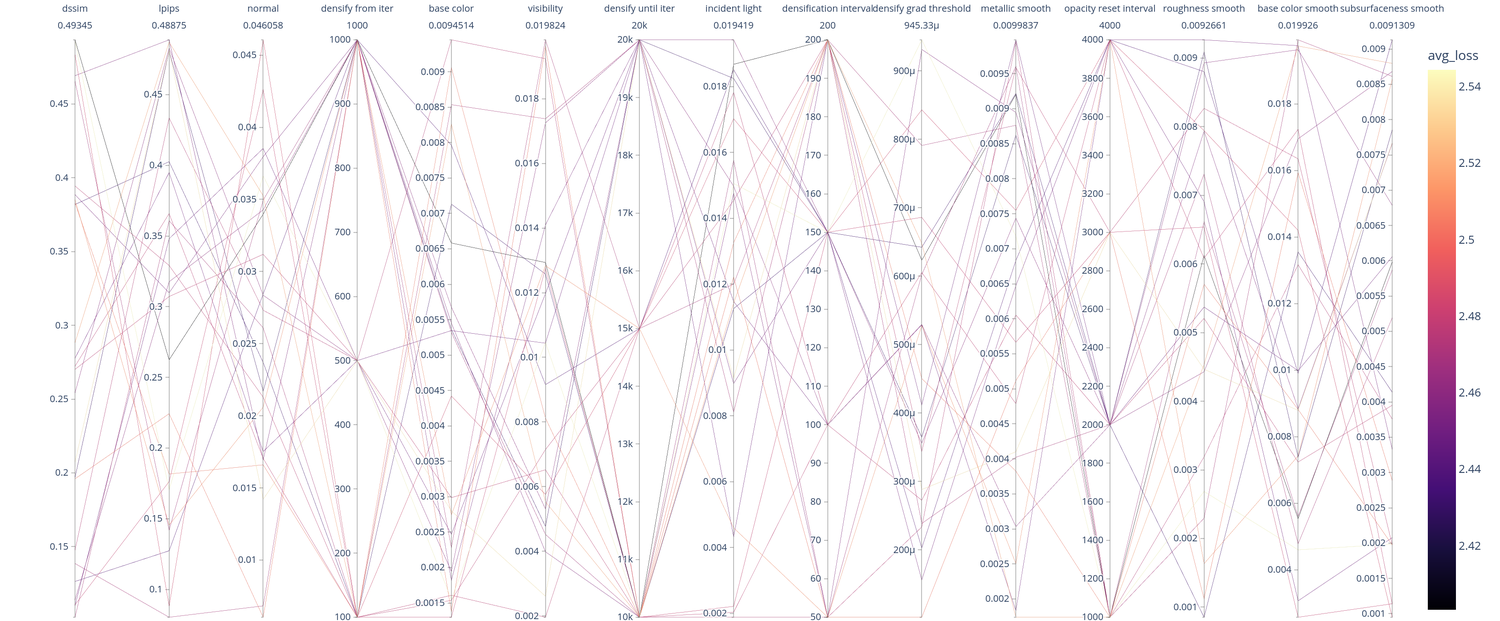}
    \caption{\textbf{Parallel coordinate plot of the hyperparameter search.}
    Each line corresponds to one configuration; lighter colors indicate higher values of the aggregated objective $\mathcal{L}_\text{total}$, as defined in Eq.~\ref{eq:avg_loss}.}
    \label{supp:fig:parallel_plot}
\end{figure*}

\Cref{supp:fig:parallel_plot} shows a parallel coordinate plot where each polyline corresponds to one hyperparameter configuration, and each vertical axis represents a different hyperparameter. The color encodes performance: lighter lines denote higher $\mathcal{L}_\text{total}$ (better configurations), while darker lines correspond to lower-performing settings.

Several trends emerge:

\begin{itemize}[leftmargin=*]
    \item Mid-range values of $\lambda_{\text{dssim}}$ (0.25,0.3)  and $\lambda_{\text{incident\_light}}$ (0.01-0.012) tend to perform best, suggesting that moderate weighting of structural similarity and incident-light regularization provides a good trade-off for reconstruction quality.
    \item Higher values of $\lambda_{\text{normal}}$ (0.045$>$) and $\lambda_{\text{base\_color\_smooth}}$ (0.018$>$) correlate with better performance, indicating that emphasizing surface consistency and base-color smoothness can enhance overall fidelity.
    \item For $\lambda_{\text{roughness\_smooth}}$ (0.002$<$) and \texttt{densify\_until\_iter} (10K$<$), the best configurations cluster towards lower values, suggesting that overly strong roughness smoothing or excessively long densification phases can harm fine detail and lead to suboptimal reconstructions.
\end{itemize}

Overall, these patterns provide guidance on how to balance detail preservation, smoothness, and physical plausibility in SSS-3DGS-style reconstructions.

\begin{table*}[t]
\centering
\small
\begin{tabular}{lclc}
\hline
\multicolumn{2}{c}{\textbf{Training and learning rate parameters}} & \multicolumn{2}{c}{\textbf{Densification and loss parameters}} \\
\hline
\textbf{Configuration} & \textbf{Value} \cite{dihlmann2024subsurfacescattering3dgaussian} & \textbf{Configuration} & S\textbf{Value} \\
\hline
iterations & 10{,}000 & percent\_dense & 0.01 \\
batch size & 4 & densification\_interval & 150 \\
Gaussian learning rate & $5\times 10^{-3}$ & opacity\_reset\_interval & 3000 \\
MLP learning rate & $1\times 10^{-3}$ & densify\_from\_iter & 500 \\
position\_lr\_init & 0.00016 & densify\_until\_iter & 10000 \\
position\_lr\_final & 0.0000016 & densify\_grad\_threshold & 0.0010 \\
position\_lr\_delay\_mult & 0.01 & densify\_grad\_normal\_threshold & -- \\
position\_lr\_max\_steps & 30{,}000 & normal\_densify\_from\_iter & 0 \\
normal\_lr & 0.01 & random\_background & False \\
normals\_lr & 0.01 & sss\_width & 32 \\
sh\_lr & -- & $\lambda_{\text{dssim}}$ & 0.492 \\
feature\_lr & 0.0025 & $\lambda_{\text{lpips}}$ & 0.171 \\
color\_lr & 0.0025 & $\lambda_{\text{mask}}$ & 0.1 \\
opacity\_lr & 0.05 & $\lambda_{\text{smooth}}$ & {[}0.002, 0.002, 0.002, 0.006{]} \\
scaling\_lr & 0.005 & $\lambda_{\text{enh}}$ & 0.005 \\
rotation\_lr & 0.001 & $\lambda_{\text{ray}}$ & 0.01 \\
env\_lr & 0.1 & $\lambda_{\text{sil}}$ & 0.5 \\
env\_rest\_lr & 0.001 & $\lambda_{\text{depth}}$ & 0.3 \\
base\_color\_lr & 0.01 & $\alpha$ (synthetic weight) & 0.5 \\
roughness\_lr & 0.01 & $\lambda_{\text{normal}}$ & 0.037 \\
metallic\_lr & 0.01 & $\lambda_{\text{visibility}}$ & 0.01 \\
subsurfaceness\_lr & 0.01 & $\lambda_{\text{incident\_light}}$ & 0.015 \\
light\_lr & 0.001 & $\lambda_{\text{mask\_entropy}}$ & 0.1 \\
light\_rest\_lr & 0.0001 & $\lambda_{\text{base\_color}}$ & 0.002 \\
light\_init & -1.0 & $\lambda_{\text{base\_color\_smooth}}$ & 0.005 \\
visibility\_lr & 0.0025 & $\lambda_{\text{metallic\_smooth}}$ & 0.007 \\
visibility\_rest\_lr & 0.0025 & $\lambda_{\text{roughness\_smooth}}$ & 0.003 \\
sss\_lr & 0.001 & $\lambda_{\text{subsurfaceness\_smooth}}$ & 0.002 \\
\hline
\end{tabular}
\caption{\textbf{DIAMOND-SSS configuration}
Training and densification hyperparameters for DIAMOND-SSS.}
\label{supp:tab:sss3dgs_hyperparams}
\end{table*}

\subsubsection{Best Configuration vs. Original}
\label{sec:hyp_best_vs_original}

\Cref{supp:tab:hyperparam_comparison} compares the original hyperparameter configuration from~\cite{dihlmann2024subsurfacescattering3dgaussian} with the best-performing combination identified by our tuning procedure.

\begin{table}[t]
\small
\centering
\begin{tabular}{lcc}
\hline
\textbf{Hyperparameter} & \textbf{Original} & \textbf{Best} \\
\hline
$\lambda_{\text{dssim}}$                & 0.200 & 0.492 \\
$\lambda_{\text{lpips}}$               & 0.200 & 0.171 \\
$\lambda_{\text{normal}}$              & 0.020 & 0.037 \\
$\lambda_{\text{visibility}}$          & 0.010 & 0.010 \\
$\lambda_{\text{incident\_light}}$     & 0.020 & 0.015 \\
$\lambda_{\text{base\_color}}$         & 0.005 & 0.002 \\
$\lambda_{\text{base\_color\_smooth}}$ & 0.006 & 0.005 \\
$\lambda_{\text{metallic\_smooth}}$    & 0.002 & 0.007 \\
$\lambda_{\text{roughness\_smooth}}$   & 0.002 & 0.003 \\
$\lambda_{\text{subsurfaceness\_smooth}}$ & 0.002 & 0.002 \\
\texttt{densification\_interval}       & 100   & 150   \\
\texttt{opacity\_reset\_interval}      & 3000  & 1000  \\
\texttt{densify\_from\_iter}           & 500   & 500   \\
\texttt{densify\_until\_iter}          & 15000 & 10000 \\
\texttt{densify\_grad\_threshold}      & 0.0002 & 0.0010 \\
\hline
\end{tabular}
\caption{\textbf{Original vs.\ tuned hyperparameter configuration.}
The ``Best'' setting corresponds to the highest $\mathcal{L}_\text{total}$ found in our search, as illustrated in \cref{supp:fig:parallel_plot}.}
\label{supp:tab:hyperparam_comparison}
\end{table}

A few observations:

\begin{itemize}[leftmargin=*]
    \item The perceptual loss weight $\lambda_{\text{lpips}}$ decreases from $0.2$ to $0.171$, suggesting that a slightly reduced emphasis on perceptual similarity can improve the aggregate objective when combined with other terms.
    \item In contrast, the structural loss weight $\lambda_{\text{dssim}}$ increases substantially from $0.2$ to $0.492$, indicating that stronger structural supervision plays a key role in achieving better reconstructions.
    \item The normal-consistency term $\lambda_{\text{normal}}$ is tuned upward from $0.02$ to $0.037$, reinforcing the benefit of encouraging local surface coherence.
    \item Among regularization terms, $\lambda_{\text{metallic\_smooth}}$ and $\lambda_{\text{roughness\_smooth}}$ increase, whereas $\lambda_{\text{base\_color}}$ and $\lambda_{\text{base\_color\_smooth}}$ are slightly reduced. This suggests that more flexible modeling of specular behavior combined with milder regularization on base color yields better fidelity.
    \item The densification strategy favors a shorter and more aggressive phase: \texttt{densify\_until\_iter} is reduced from 15000 to 10000, \texttt{densification\_interval} is slightly increased, and \texttt{densify\_grad\_threshold} is relaxed. The opacity reset interval is also reduced from 3000 to 1000. Together, these changes promote faster adaptation of the Gaussian set and help avoid early overfitting.
\end{itemize}

Overall, the tuned configuration reflects a nuanced rebalancing of perceptual, structural, and regularization terms, leading to improved performance under the aggregated objective $\mathcal{L}_\text{total}$ while remaining close in spirit to the original SSS-3DGS design.

\section{Additional Qualitative Reconstructions}
\label{sec:supp_qualitative}

To complement the quantitative evaluations presented in the main paper, we provide
additional qualitative reconstructions across a variety of translucent objects and
capture regimes. These visual comparisons illustrate how DIAMOND-SSS behaves under
different levels of supervision, ranging from full-view OLAT captures to extremely
sparse pose–light subsets.

\begin{figure*}[htb!]
\centering
\resizebox{\textwidth}{!}{%
\renewcommand{\arraystretch}{0.9}
\setlength{\tabcolsep}{2pt}

\begin{tabular}{ccccccccc}
\toprule
\multicolumn{3}{c}{\textbf{Plastic Bottle}} 
& \multicolumn{3}{c}{\textbf{Crystal}} 
& \multicolumn{3}{c}{\textbf{Massage Ball}} \\
\cmidrule(lr){1-3}\cmidrule(lr){4-6}\cmidrule(lr){7-9}
\textbf{GT} & \textbf{Ours} & \textbf{SSS-3DGS \cite{dihlmann2024subsurfacescattering3dgaussian}} &
\textbf{GT} & \textbf{Ours} & \textbf{SSS-3DGS \cite{dihlmann2024subsurfacescattering3dgaussian}} &
\textbf{GT} & \textbf{Ours} & \textbf{SSS-3DGS \cite{dihlmann2024subsurfacescattering3dgaussian}} \\
\midrule

\raisebox{-0.5\height}{\includegraphics[width=0.10\linewidth]{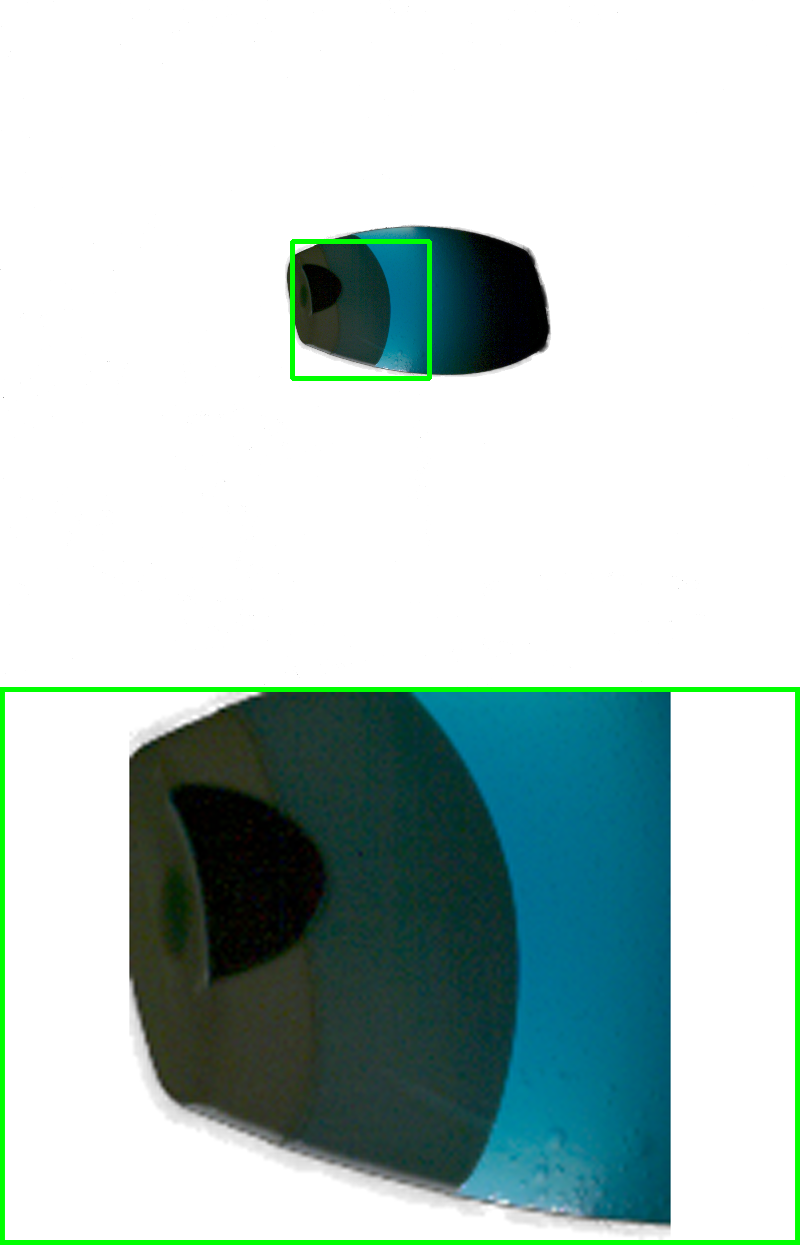}} &
\raisebox{-0.5\height}{\includegraphics[width=0.10\linewidth]{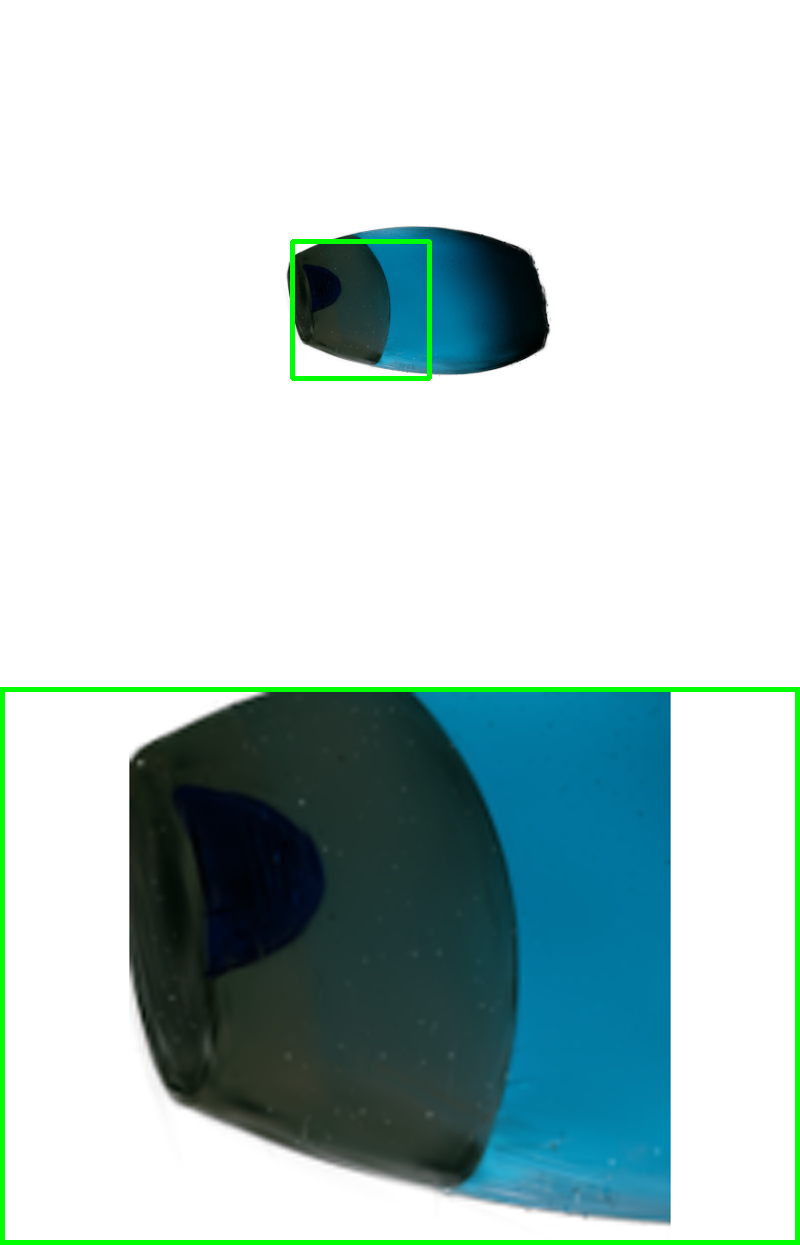}} &
\raisebox{-0.5\height}{\includegraphics[width=0.10\linewidth]{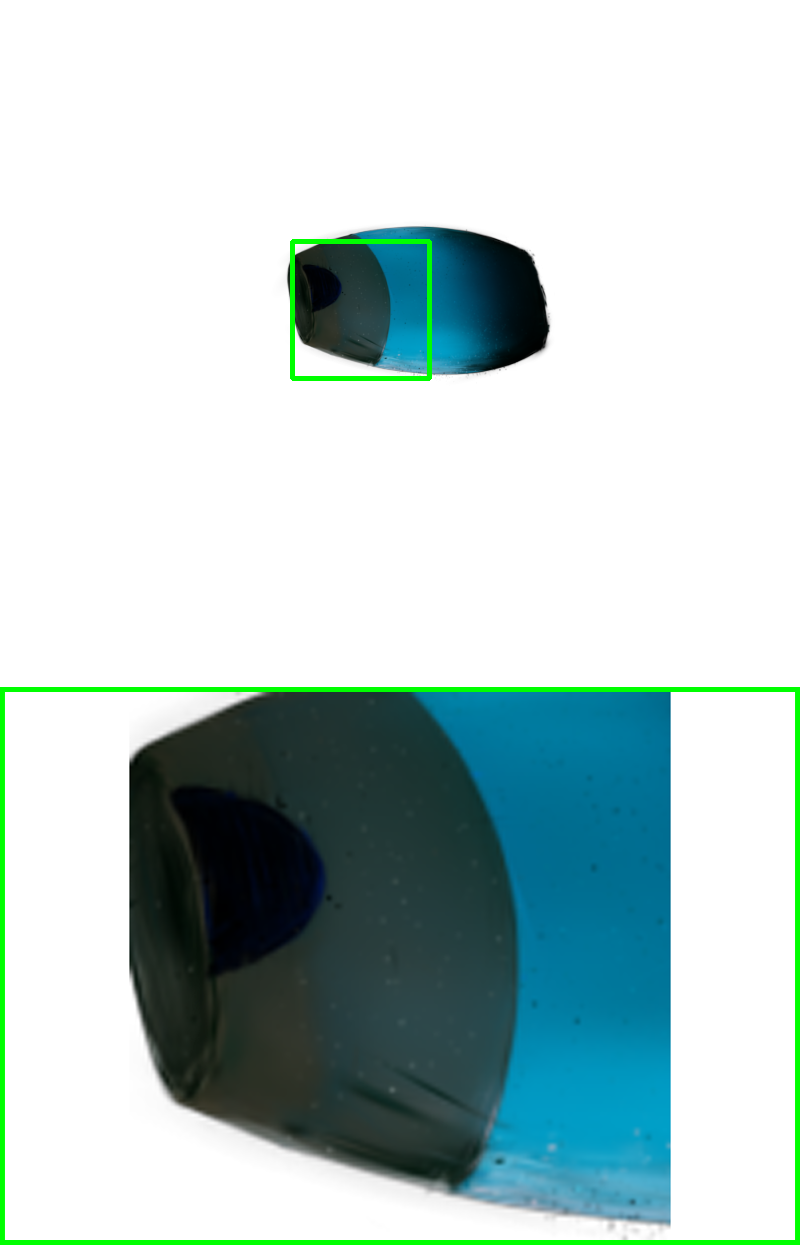}} &
\raisebox{-0.5\height}{\includegraphics[width=0.10\linewidth]{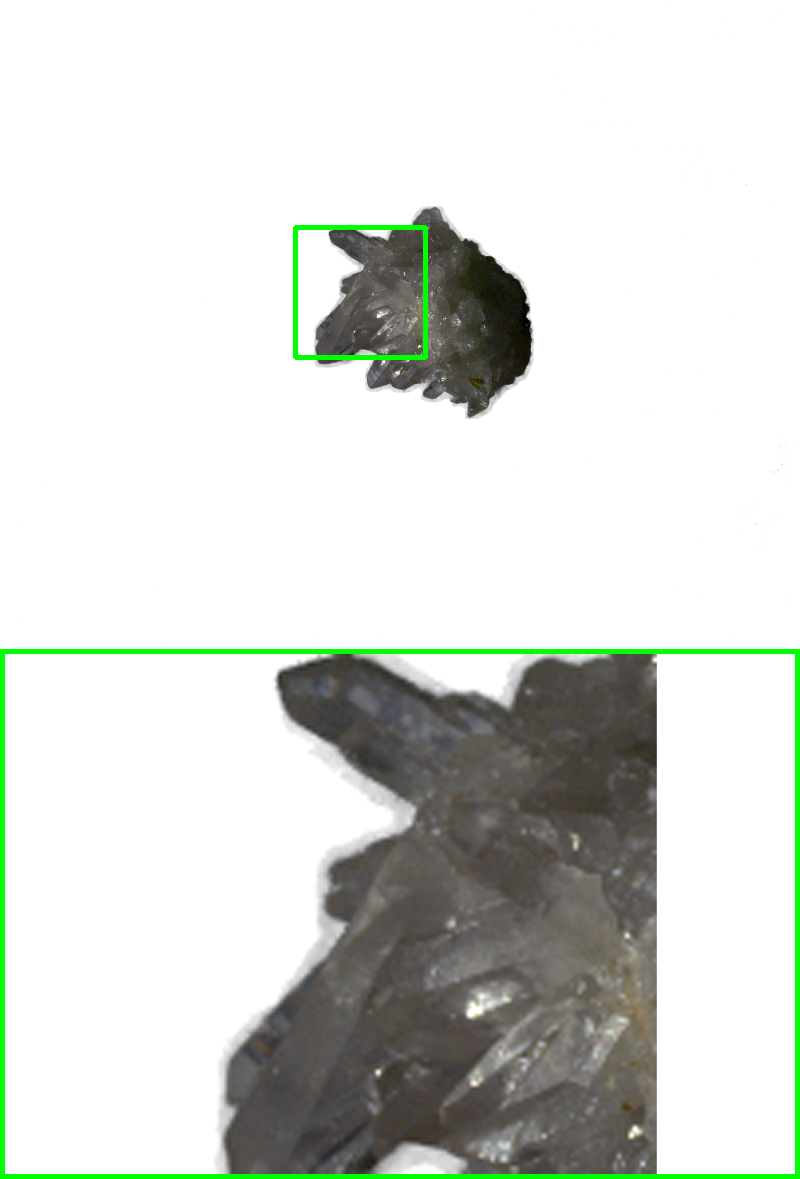}} &
\raisebox{-0.5\height}{\includegraphics[width=0.10\linewidth]{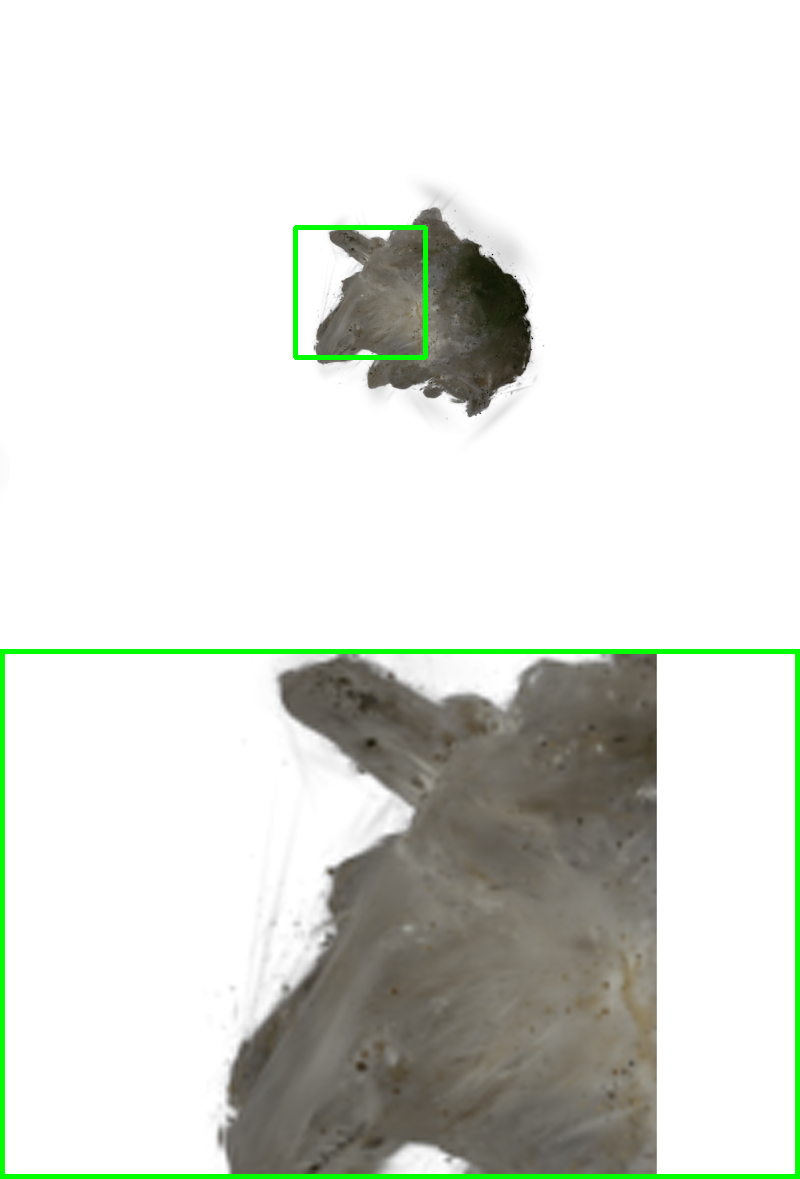}} &
\raisebox{-0.5\height}{\includegraphics[width=0.10\linewidth]{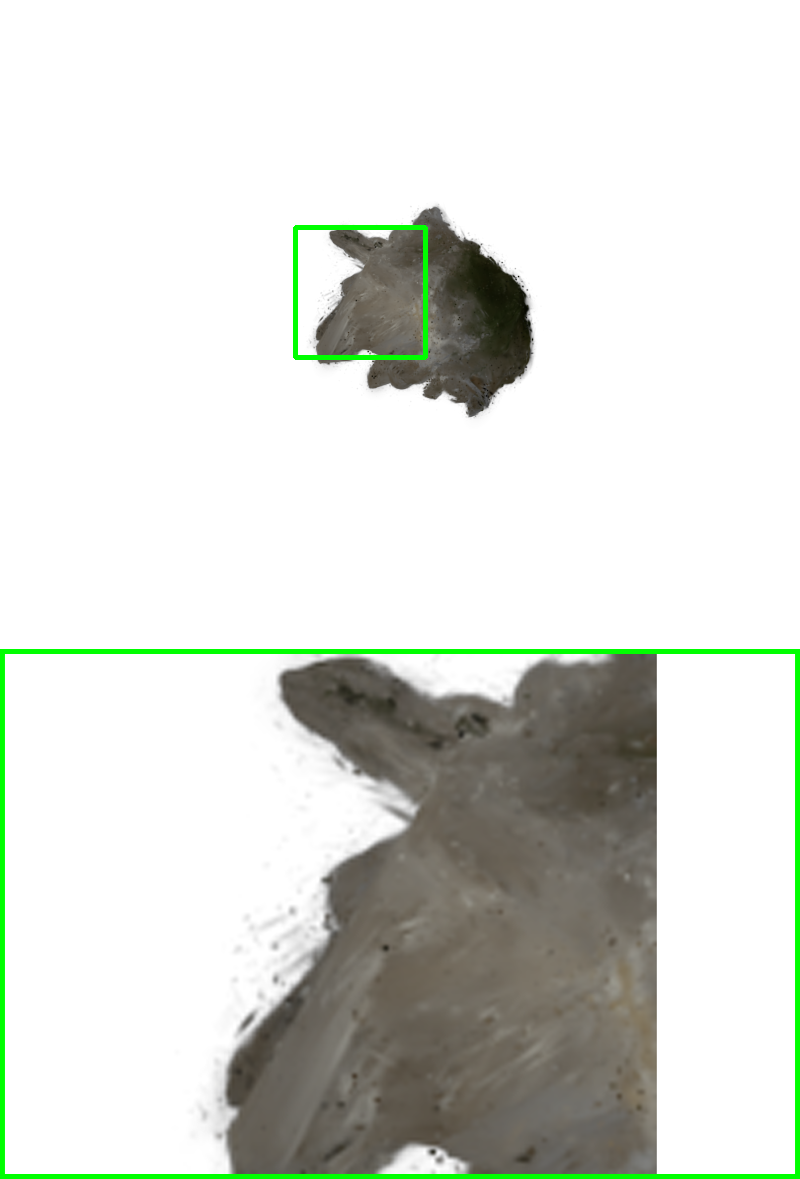}} &
\raisebox{-0.5\height}{\includegraphics[width=0.10\linewidth]{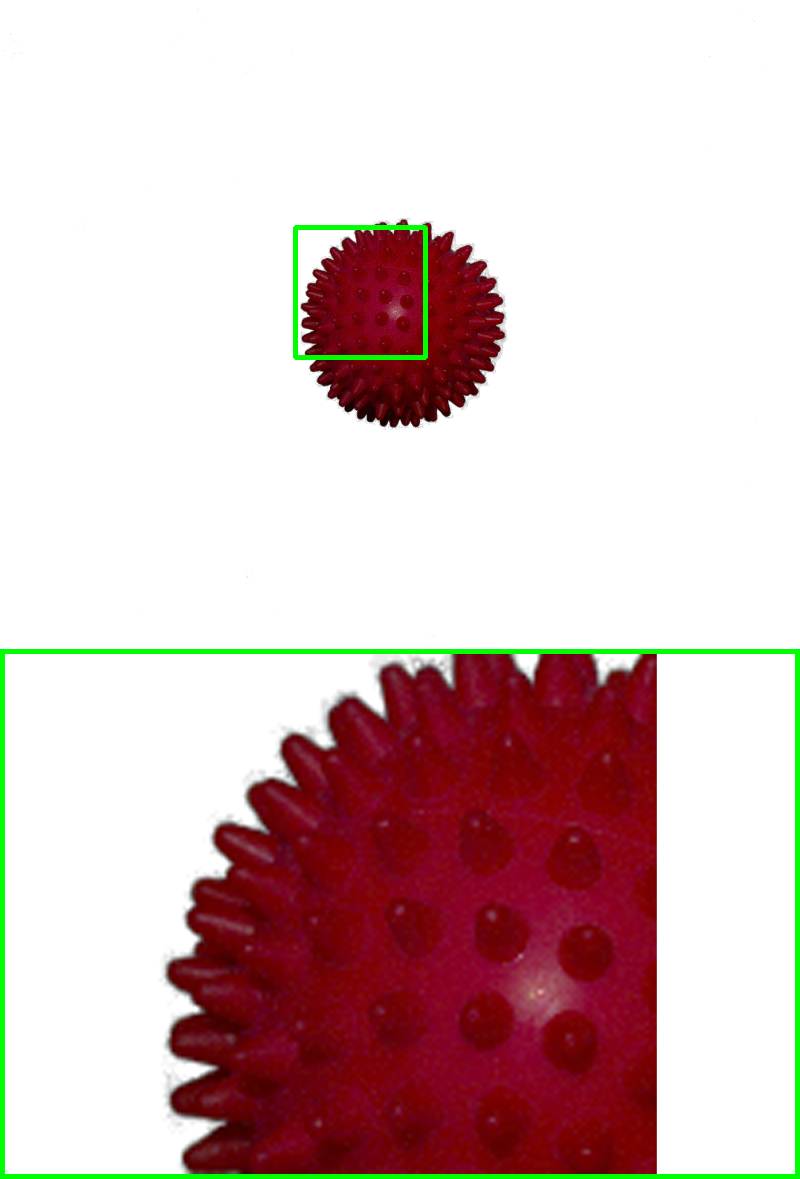}} &
\raisebox{-0.5\height}{\includegraphics[width=0.10\linewidth]{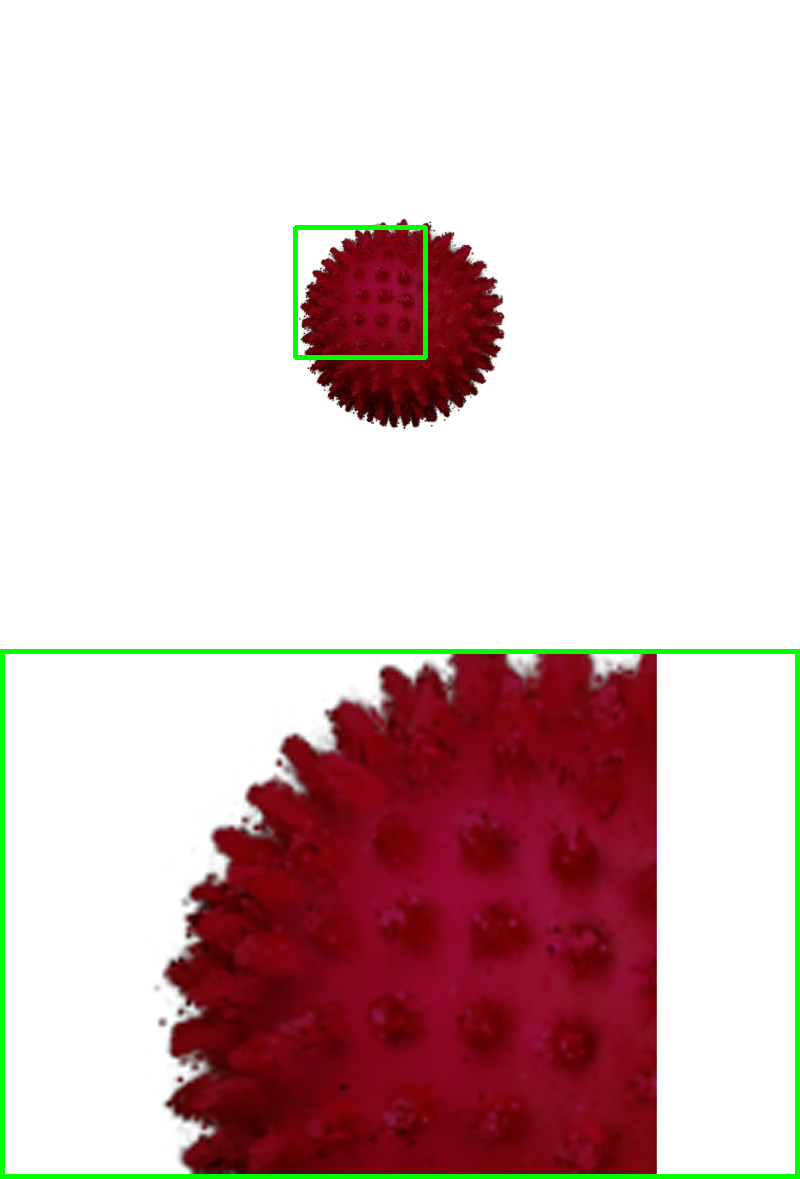}} &
\raisebox{-0.5\height}{\includegraphics[width=0.10\linewidth]{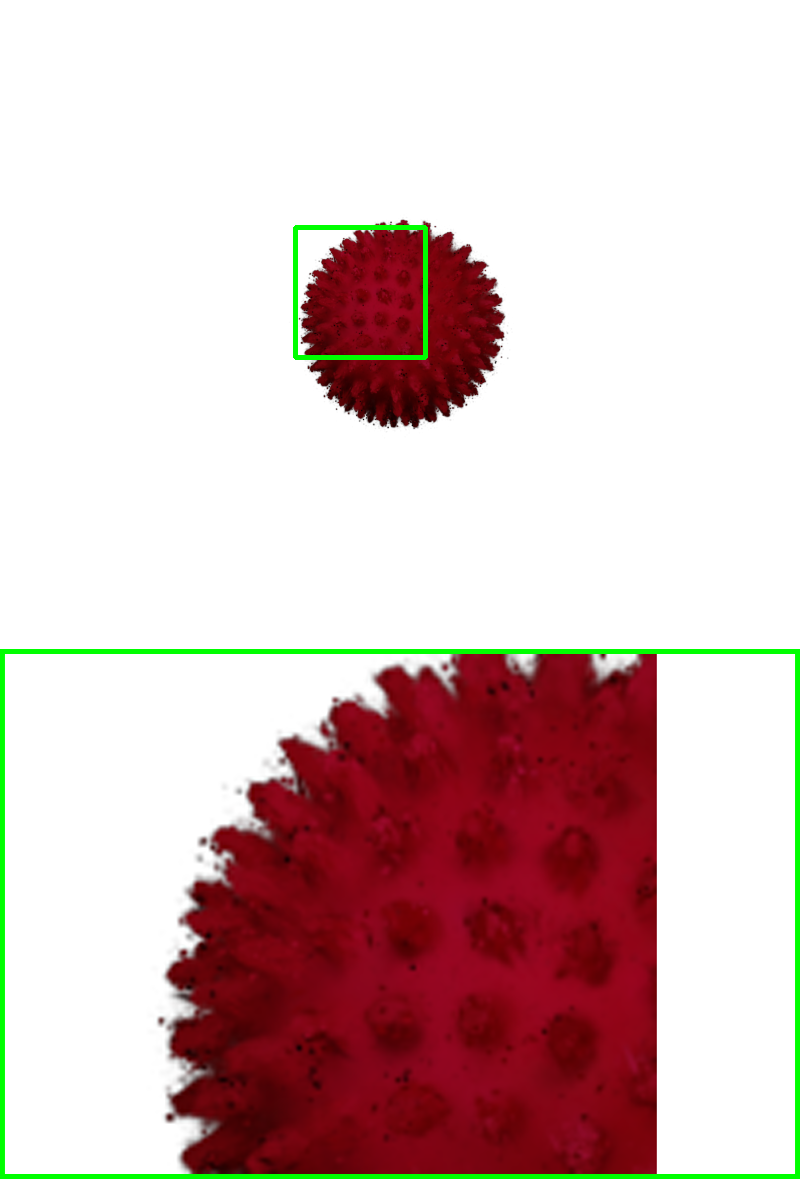}} \\
\midrule

\scriptsize all views,\;single light &
\includegraphics[width=0.10\linewidth]{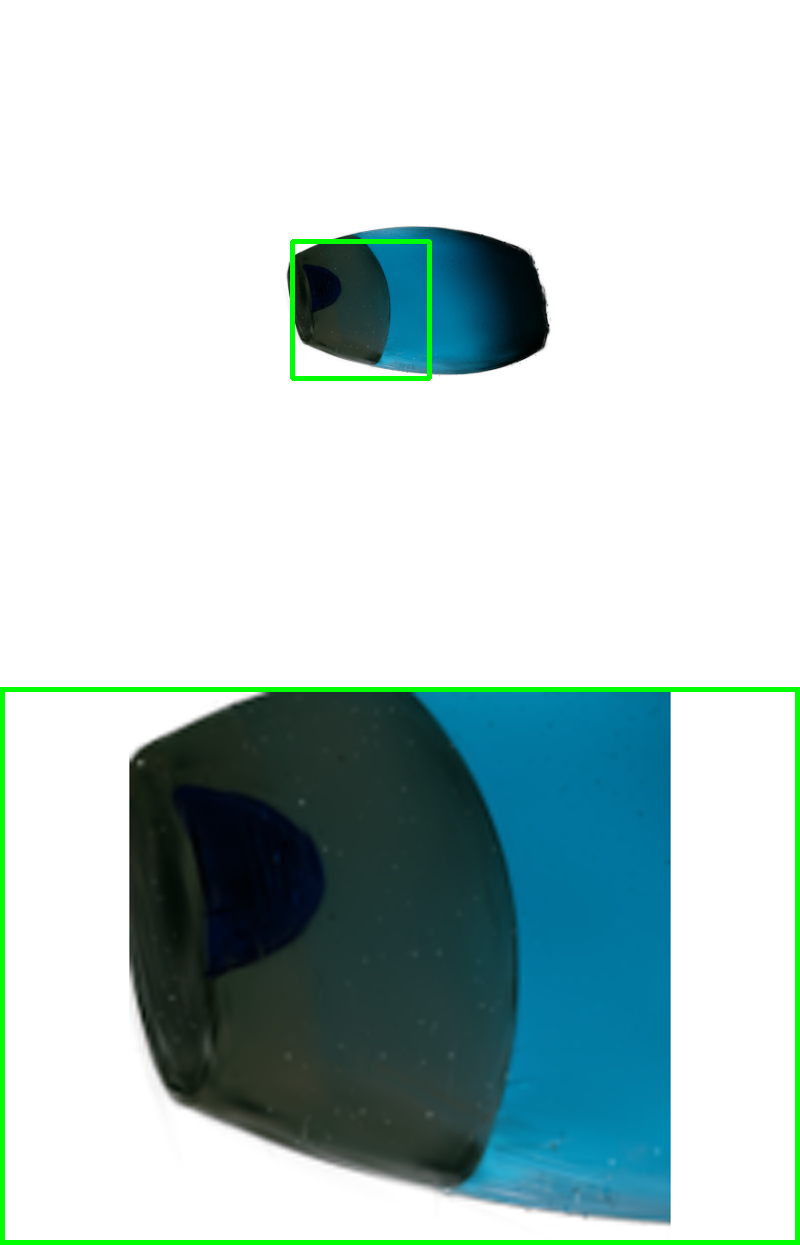} &
\includegraphics[width=0.10\linewidth]{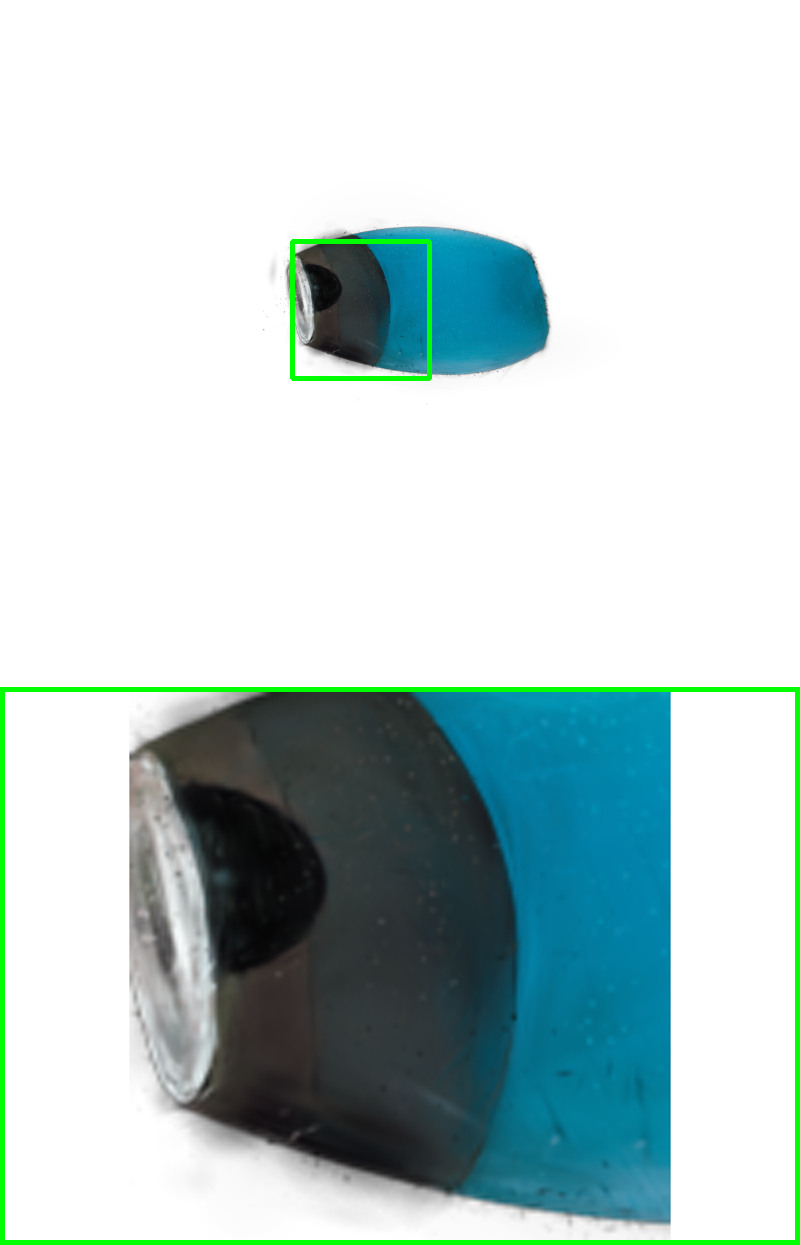} &
\scriptsize all views,\;single light &
\includegraphics[width=0.10\linewidth]{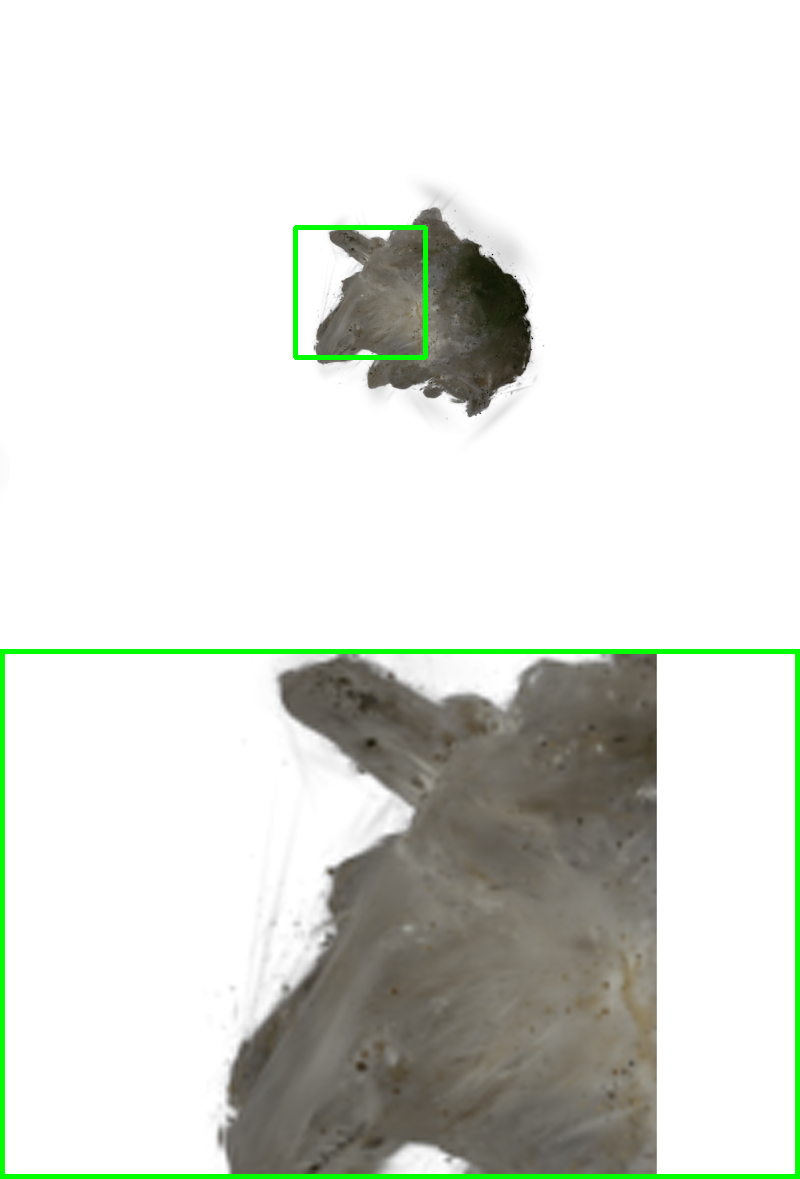} &
\includegraphics[width=0.10\linewidth]{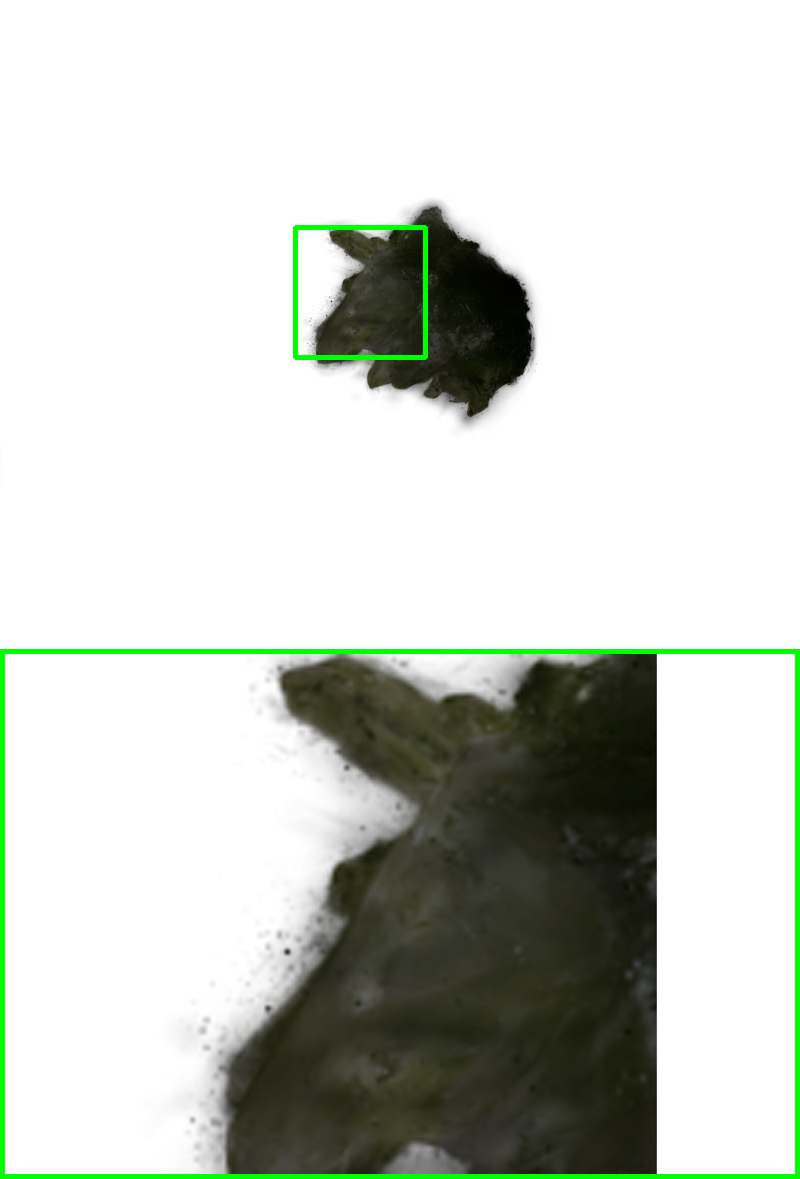} &
\scriptsize all views,\;single light &
\includegraphics[width=0.10\linewidth]{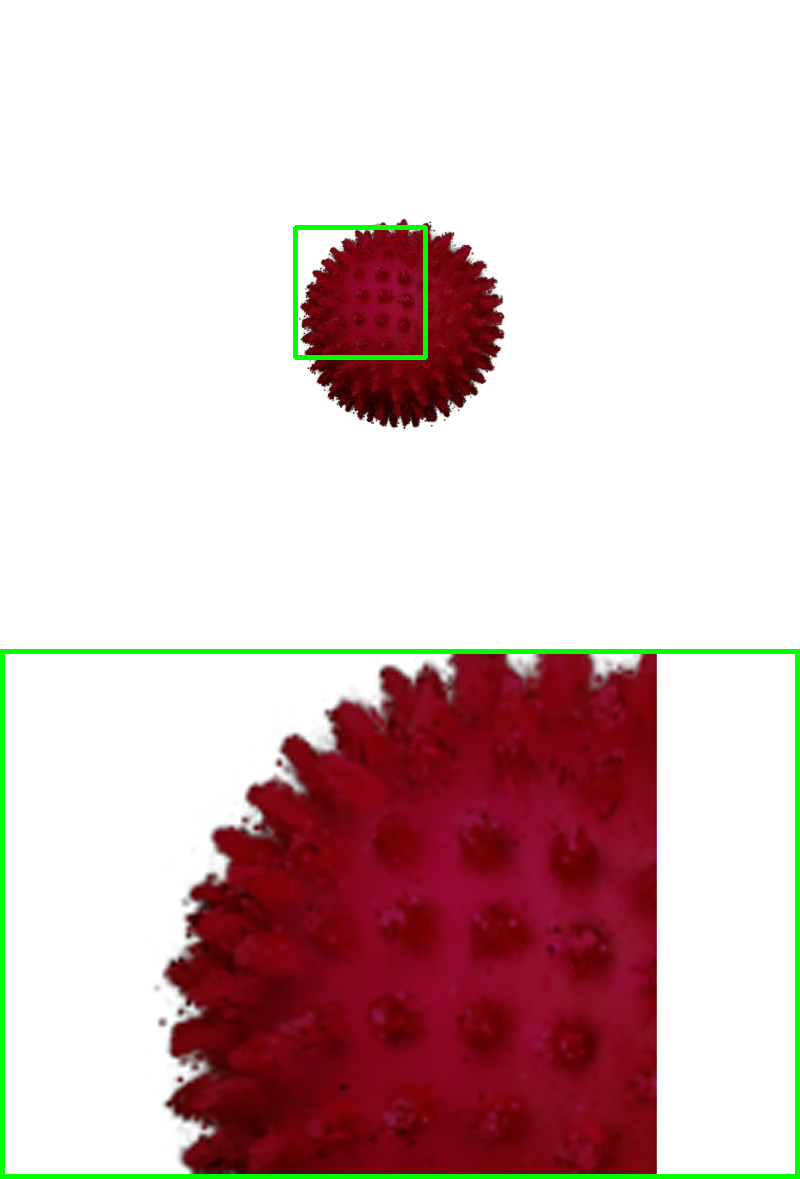} &
\includegraphics[width=0.10\linewidth]{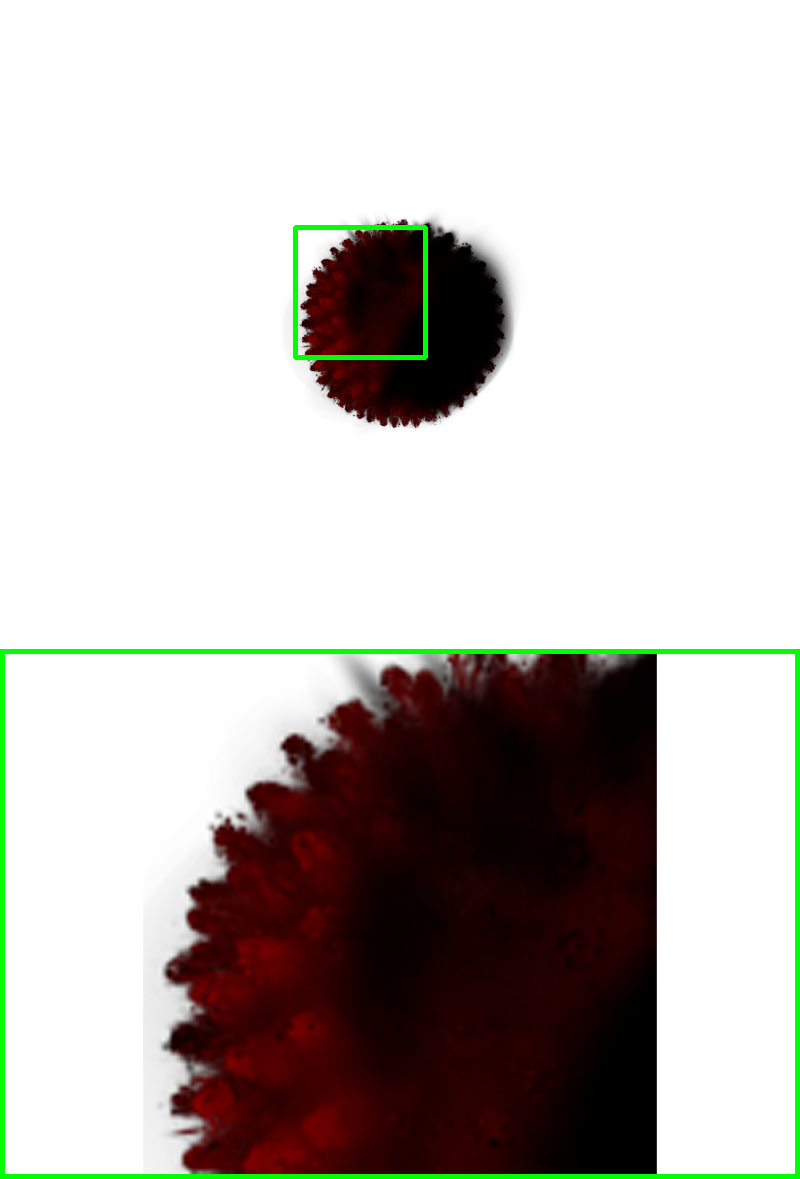} \\
\midrule

\scriptsize 5\% views,\;5\% lights &
\includegraphics[width=0.10\linewidth]{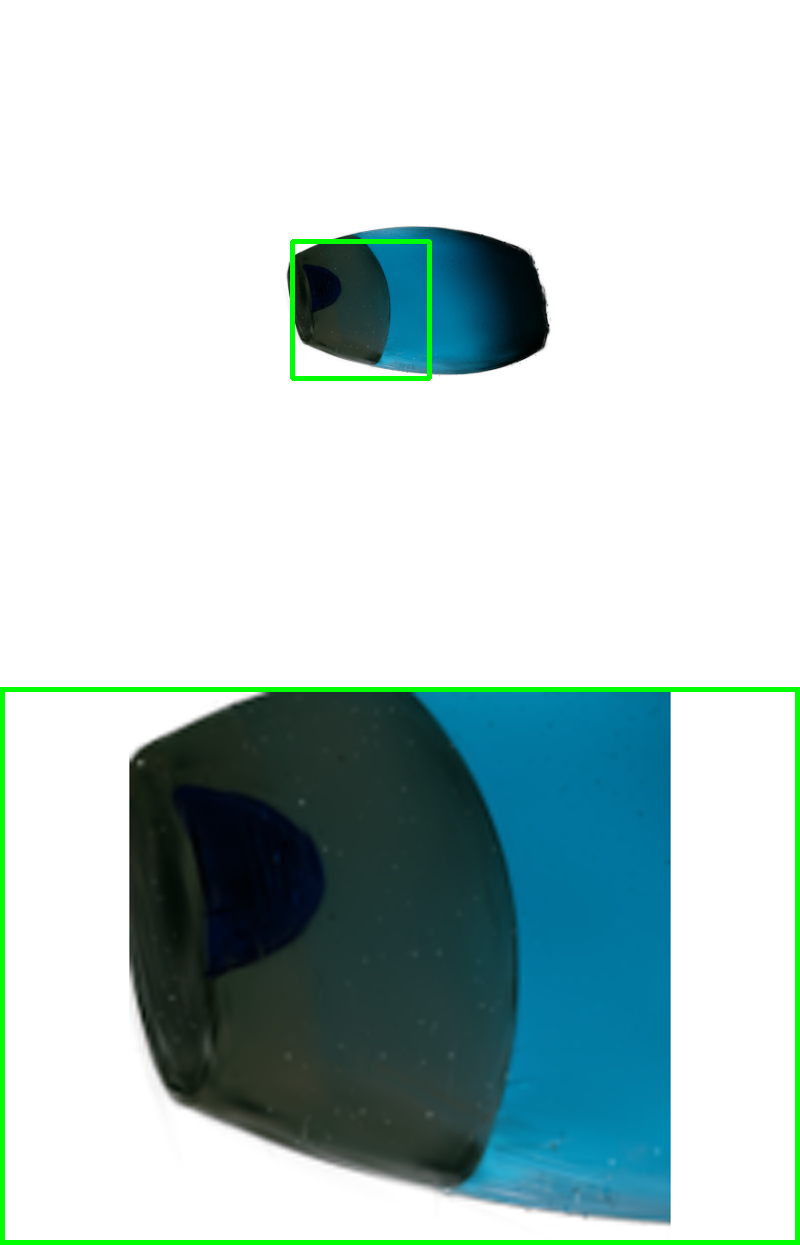} &
\includegraphics[width=0.10\linewidth]{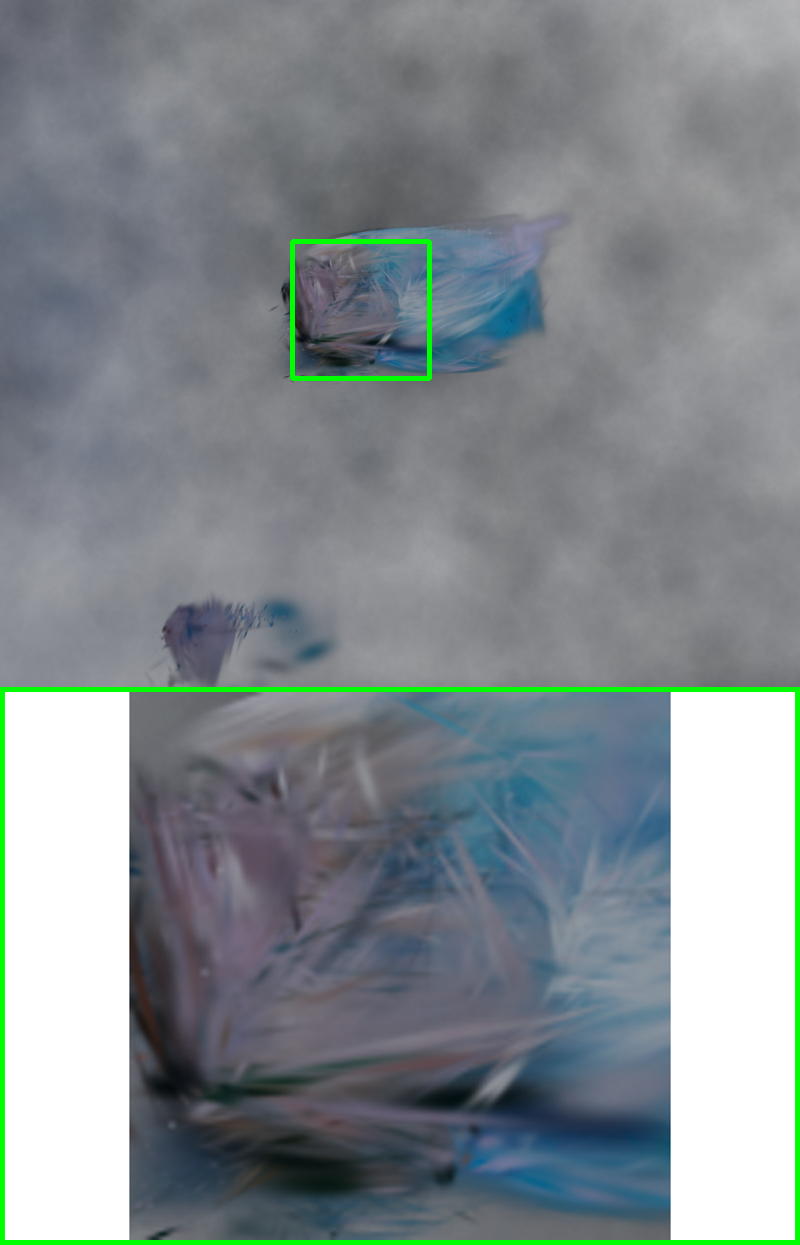} &
\scriptsize 5\% views,\;5\% lights &
\includegraphics[width=0.10\linewidth]{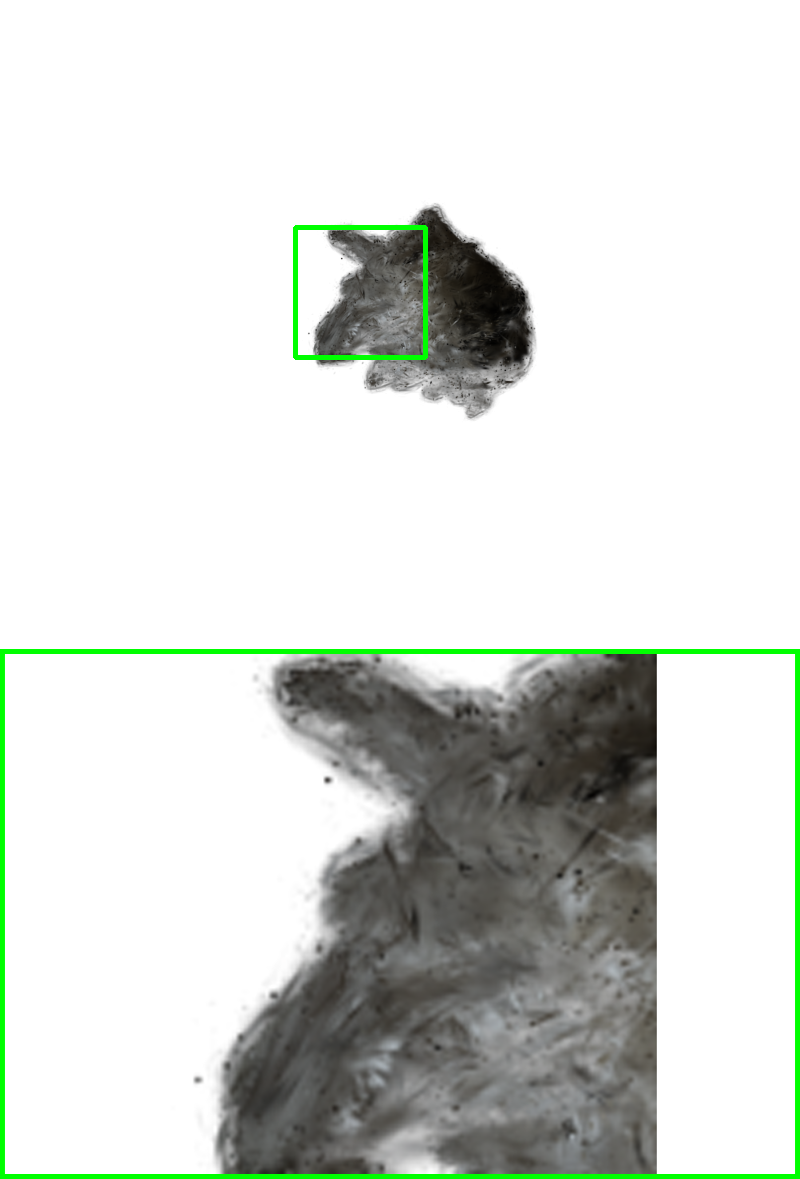} &
\includegraphics[width=0.10\linewidth]{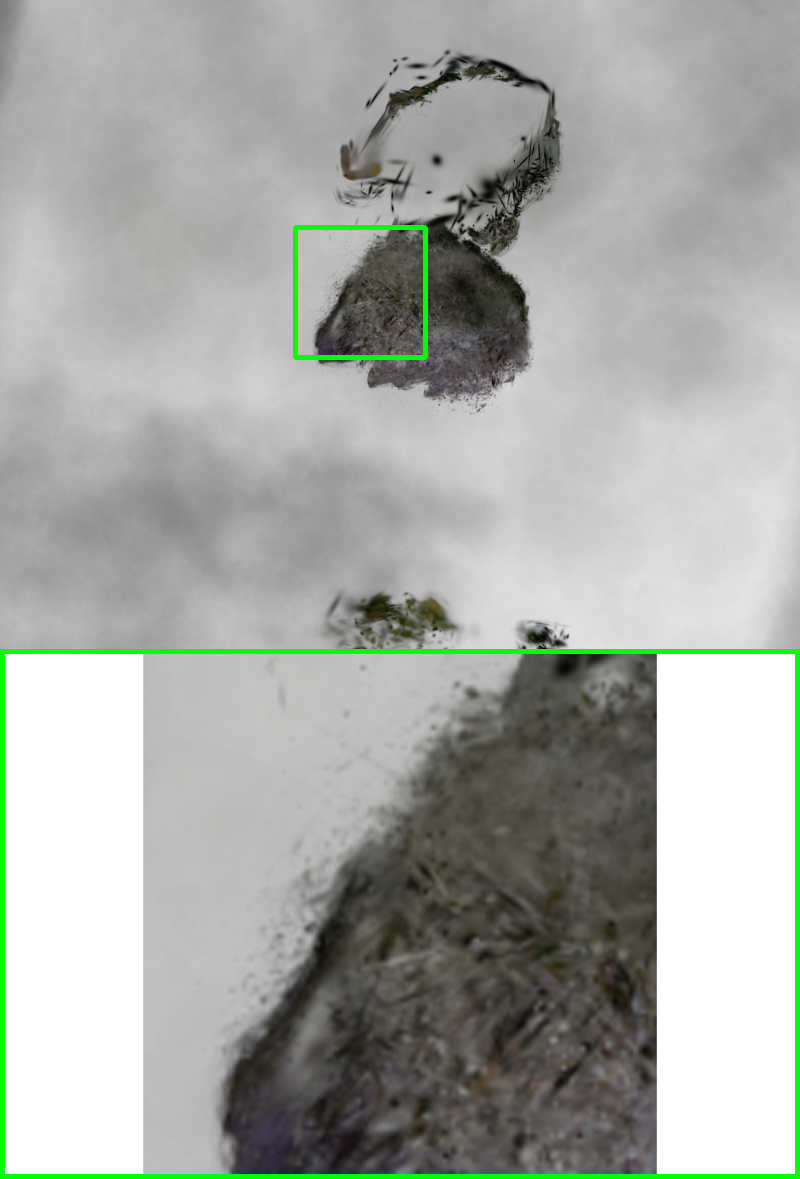} &
\scriptsize 5\% views,\;5\% lights &
\includegraphics[width=0.10\linewidth]{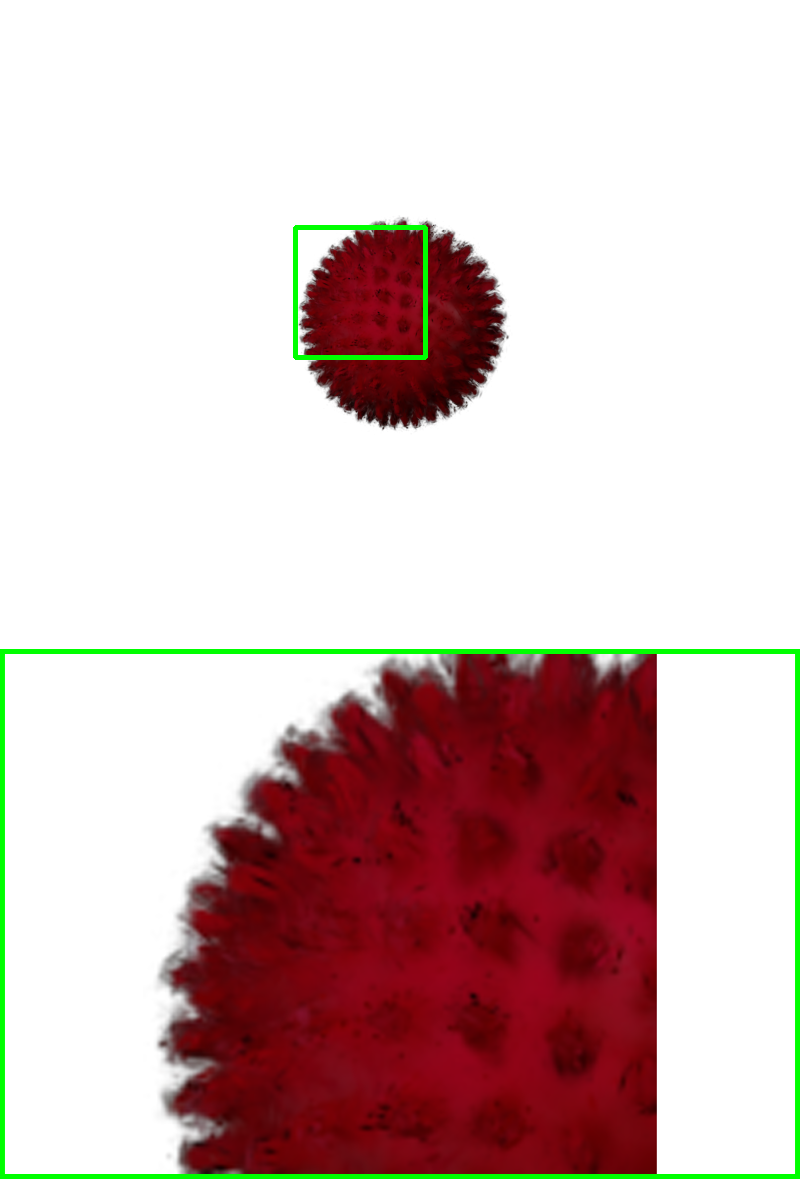} &
\includegraphics[width=0.10\linewidth]{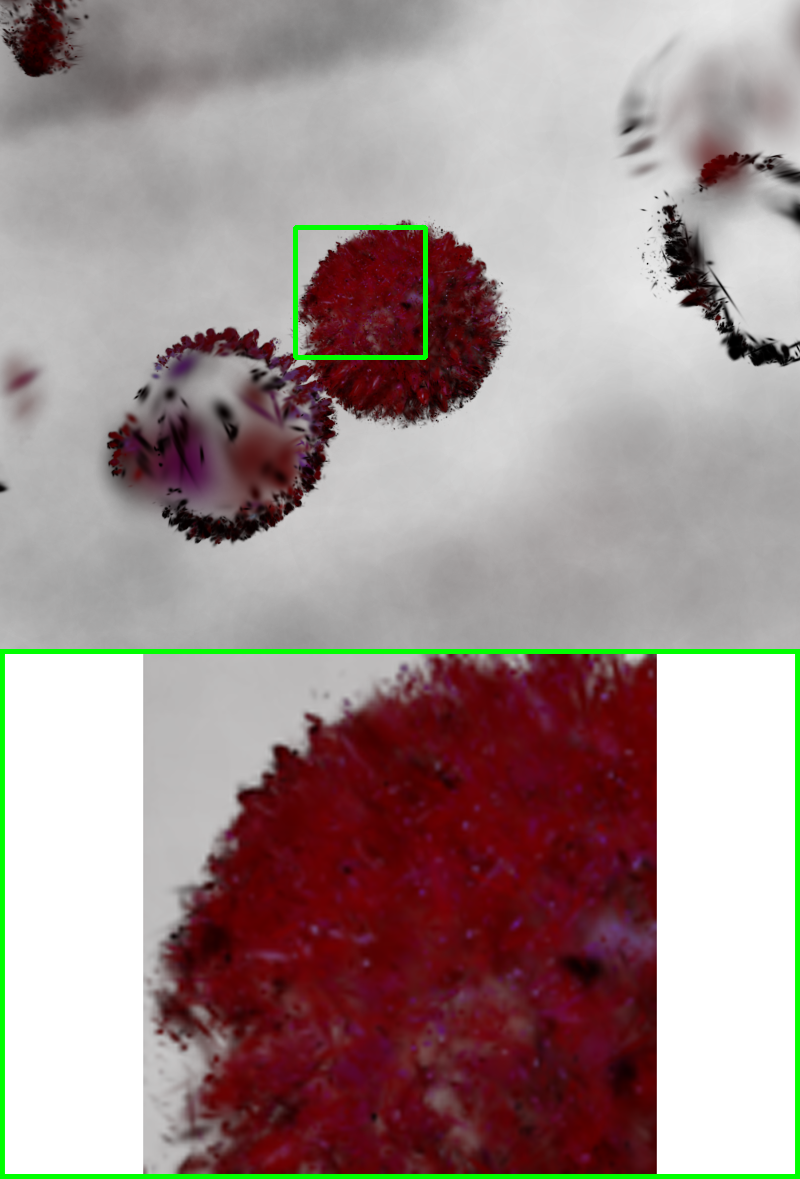} \\
\midrule

\scriptsize 3\% views,\;3\% lights &
\includegraphics[width=0.10\linewidth]{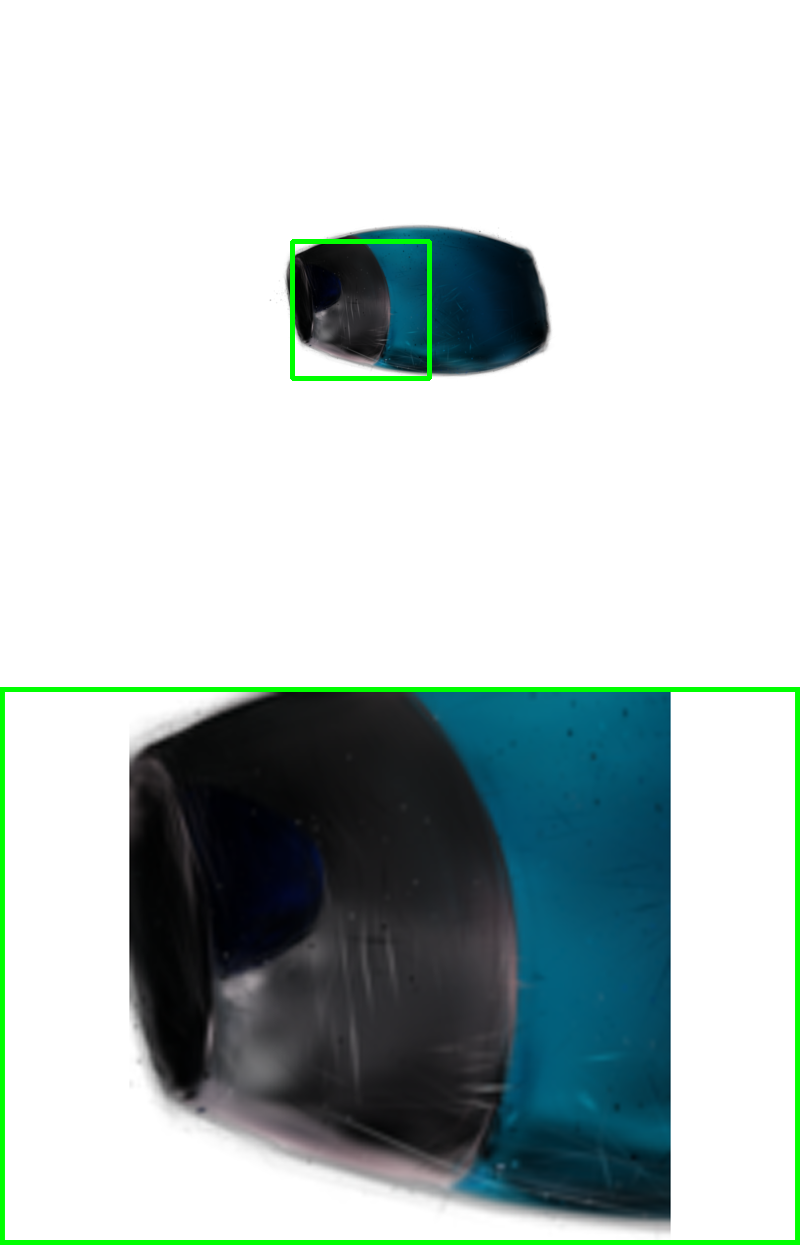} &
\includegraphics[width=0.10\linewidth]{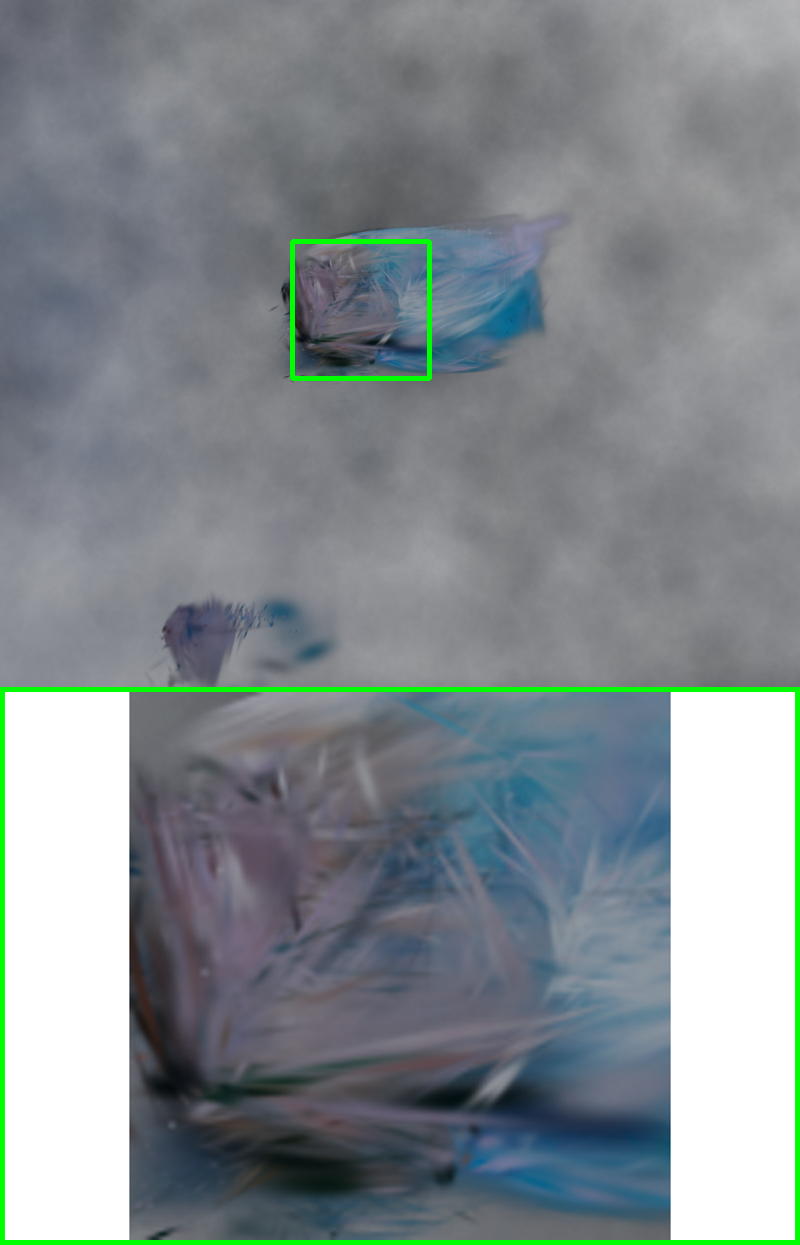} &
\scriptsize 3\% views,\;3\% lights &
\includegraphics[width=0.10\linewidth]{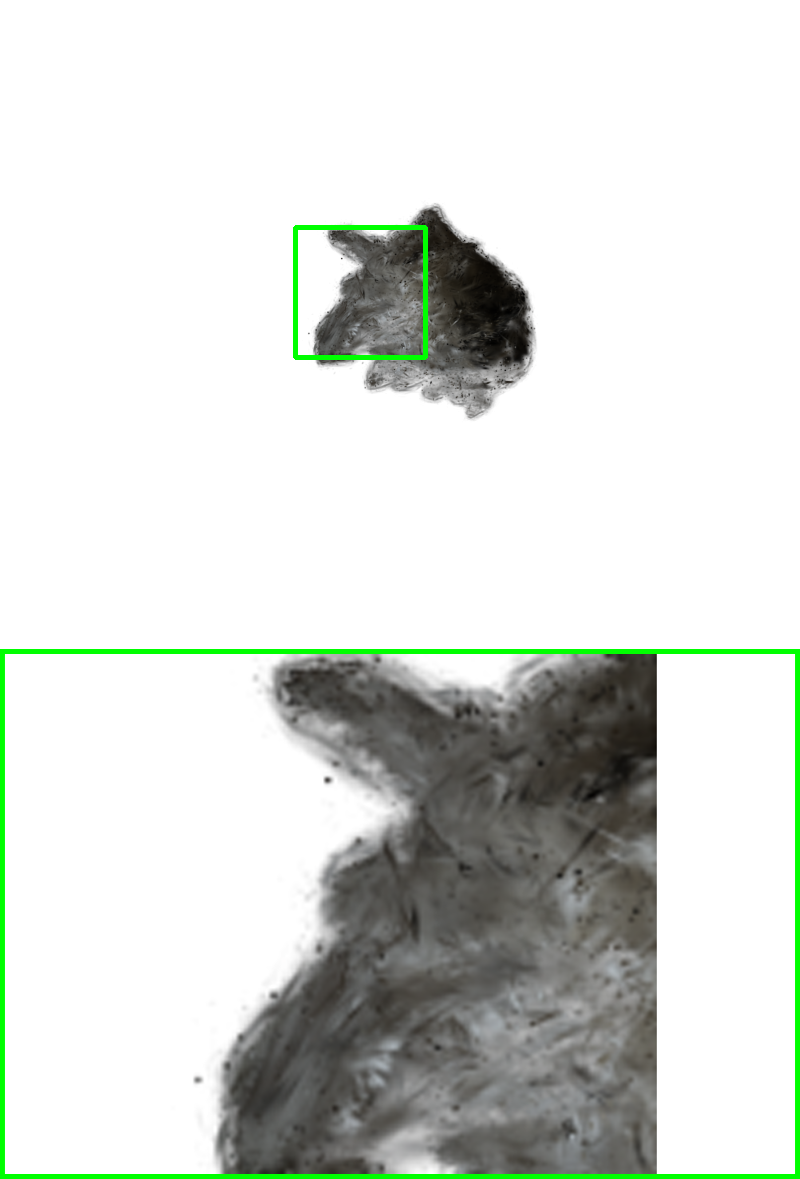} &
\includegraphics[width=0.10\linewidth]{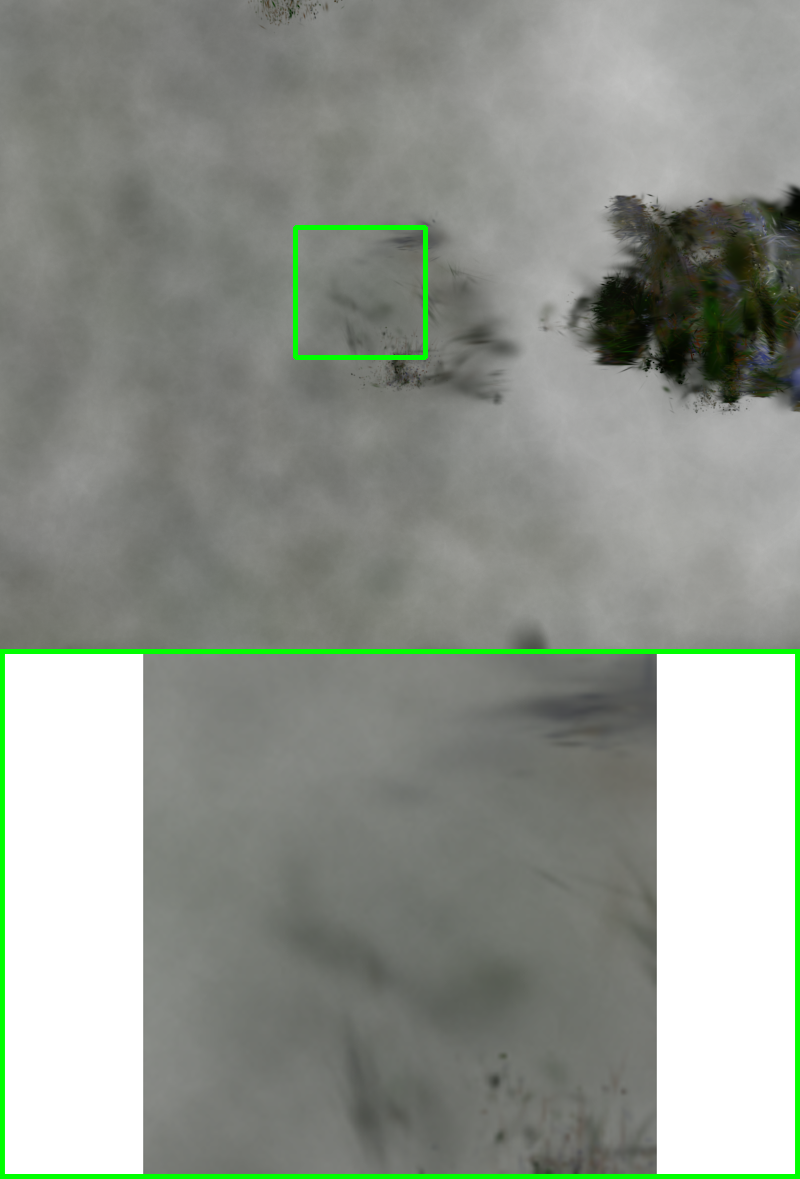} &
\scriptsize 3\% views,\;3\% lights &
\includegraphics[width=0.10\linewidth]{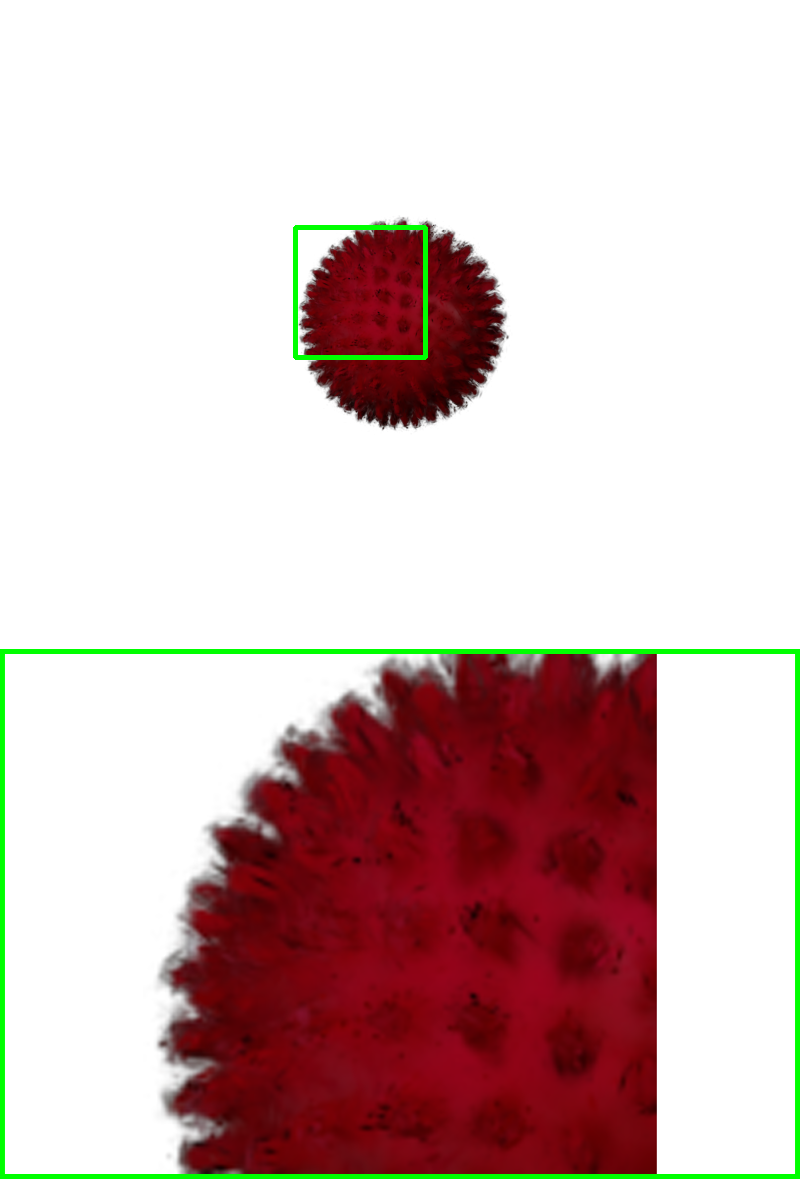} &
\includegraphics[width=0.10\linewidth]{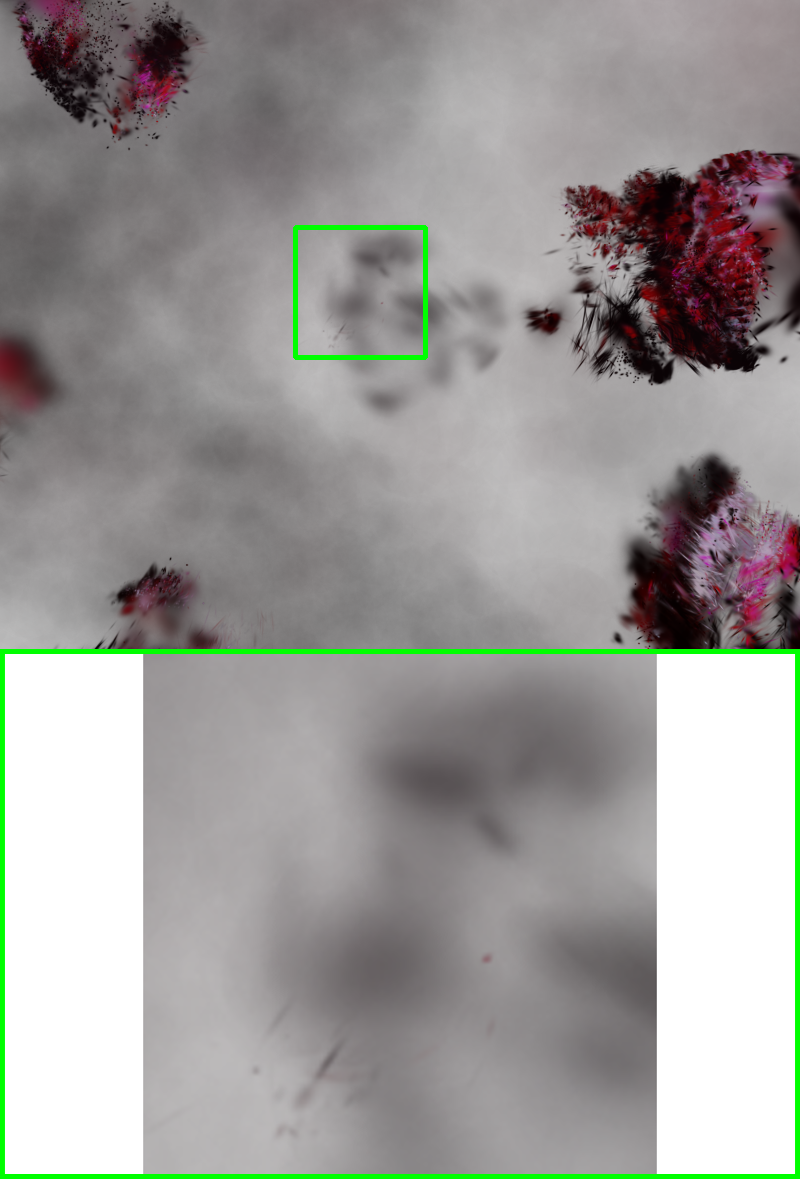} \\
\bottomrule
\end{tabular}
}
\caption{Qualitative comparison of reconstructed translucent appearance under different supervision conditions, the first row shows the setting: all views, all lights. }
\label{fig:qualitative_comparison_all_1}

\end{figure*}

\paragraph{Comparison across objects and capture regimes.}
Figure~\ref{fig:qualitative_comparison_all_1} presents reconstructions for three
representative objects—\emph{Plastic Bottle}, \emph{Crystal}, and
\emph{Massage Ball}—under four supervision levels:  
(i) all views and all lights,  
(ii) all views with a single light per view,  
(iii) 5\% of views and 5\% of lights, and  
(iv) 3\% of views and 3\% of lights.  

In all settings, DIAMOND-SSS produces sharper boundaries, more stable silhouettes,
and more faithful subsurface-scattering cues than SSS-3DGS~\cite{dihlmann2024subsurfacescattering3dgaussian}.  
In the sparse regimes (5\%/5\% and 3\%/3\%), SSS-3DGS fails severely—exhibiting strong boundary leakage, collapsed geometry, and inconsistent opacity—while DIAMOND-SSS remains stable and reconstructs the correct translucent appearance.

\begin{figure*}[htb!]
\centering
\resizebox{\textwidth}{!}{%
\renewcommand{\arraystretch}{0.9}
\setlength{\tabcolsep}{2pt}

\begin{tabular}{ccccccccc}
\toprule
\multicolumn{3}{c}{\textbf{Marmalade}} 
& \multicolumn{3}{c}{\textbf{Tupperware}} 
& \multicolumn{3}{c}{\textbf{Soap Bottle}} \\
\cmidrule(lr){1-3}\cmidrule(lr){4-6}\cmidrule(lr){7-9}
\textbf{GT} & \textbf{Ours} & \textbf{SSS-3DGS \cite{dihlmann2024subsurfacescattering3dgaussian}} &
\textbf{GT} & \textbf{Ours} & \textbf{SSS-3DGS \cite{dihlmann2024subsurfacescattering3dgaussian}} &
\textbf{GT} & \textbf{Ours} & \textbf{SSS-3DGS \cite{dihlmann2024subsurfacescattering3dgaussian}} \\
\midrule

\raisebox{-0.5\height}{\includegraphics[width=0.10\linewidth]{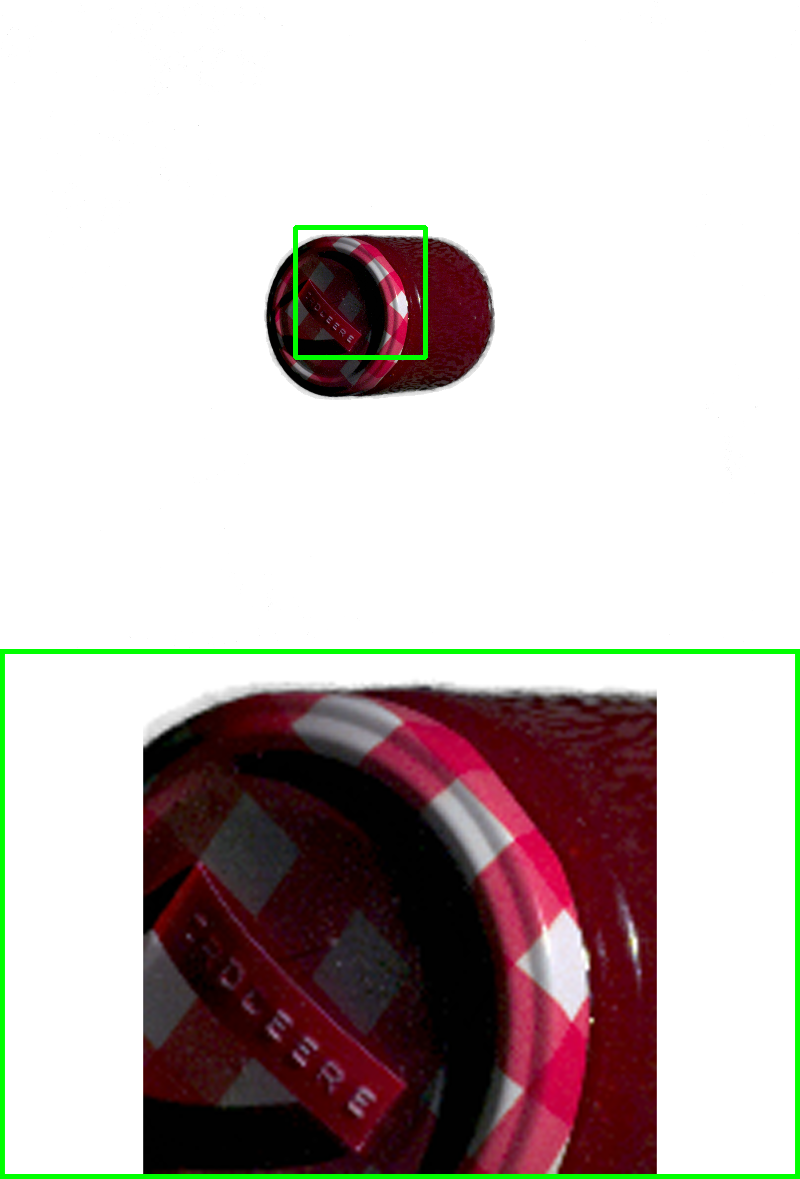}} &
\raisebox{-0.5\height}{\includegraphics[width=0.10\linewidth]{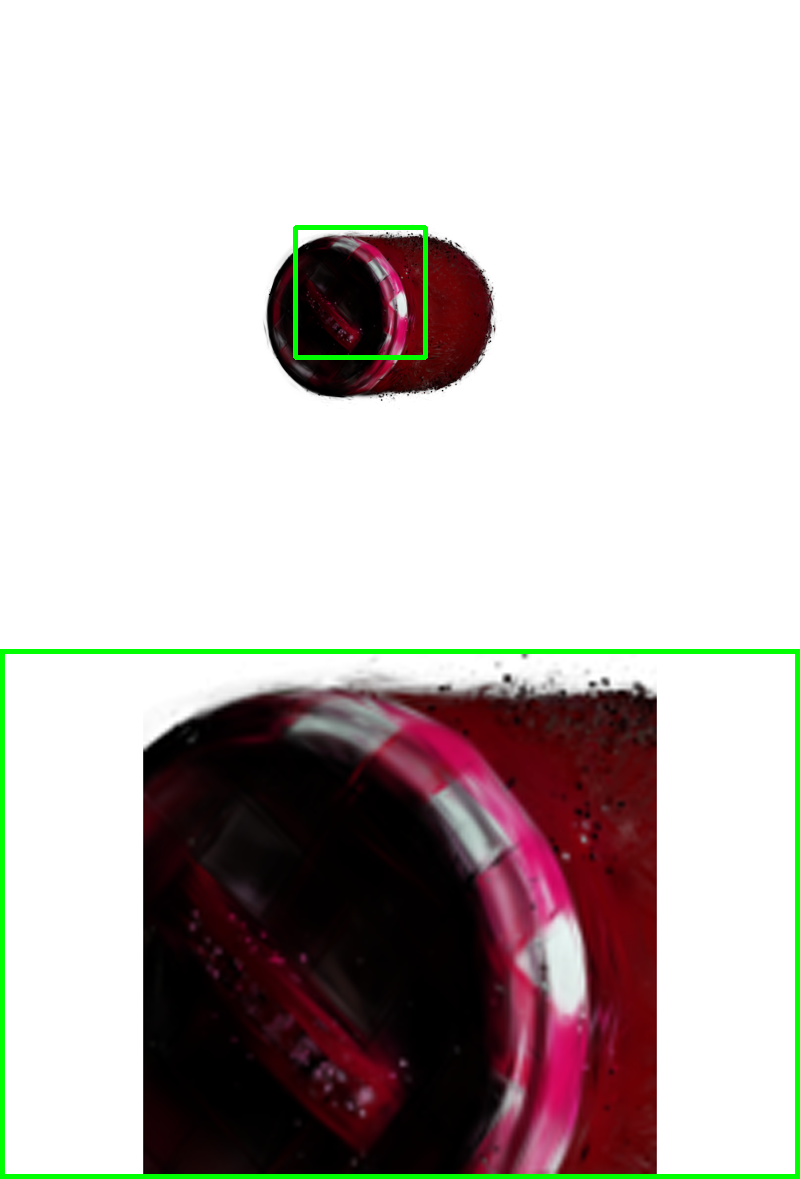}} &
\raisebox{-0.5\height}{\includegraphics[width=0.10\linewidth]{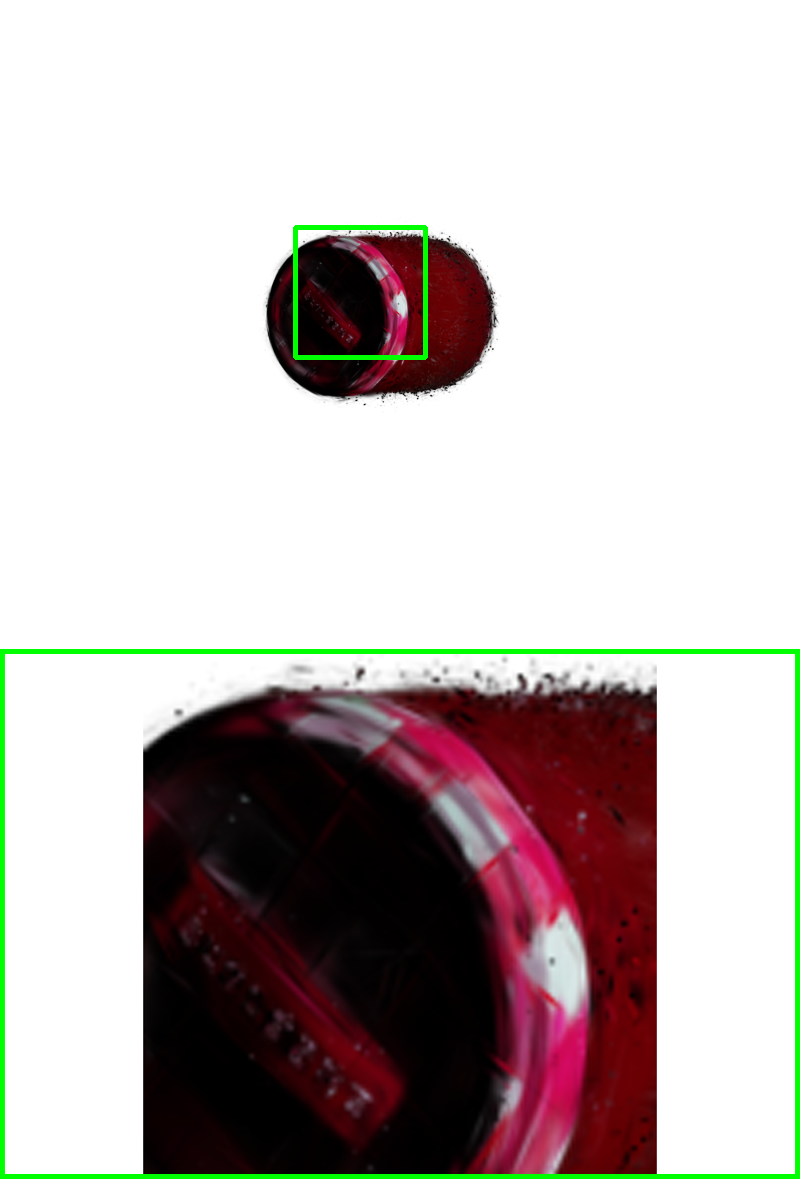}} &
\raisebox{-0.5\height}{\includegraphics[width=0.10\linewidth]{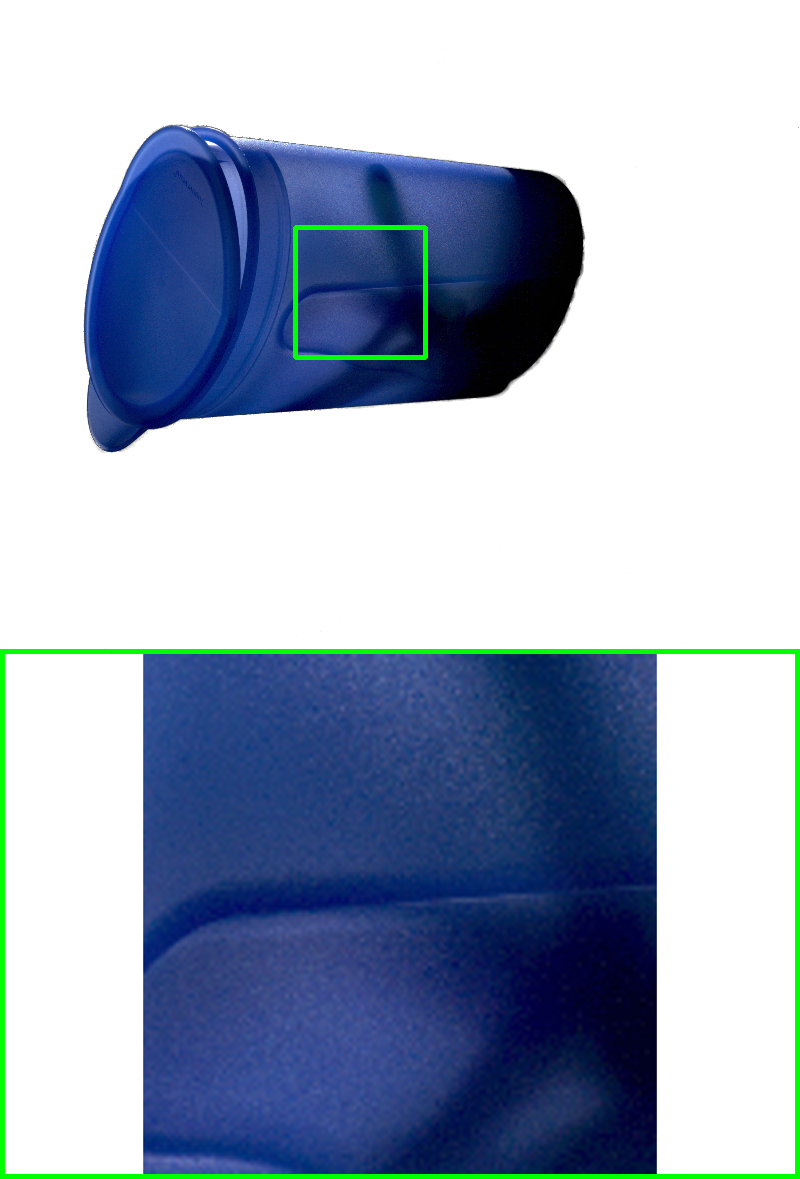}} &
\raisebox{-0.5\height}{\includegraphics[width=0.10\linewidth]{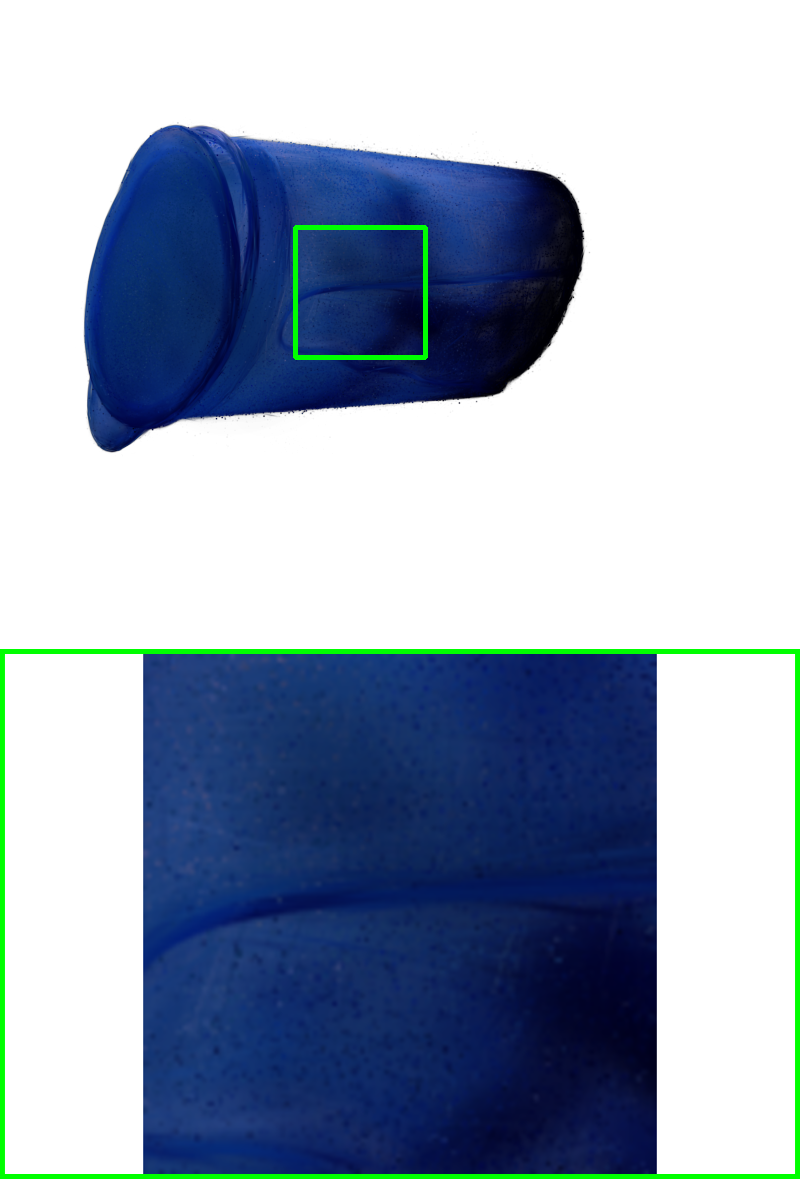}} &
\raisebox{-0.5\height}{\includegraphics[width=0.10\linewidth]{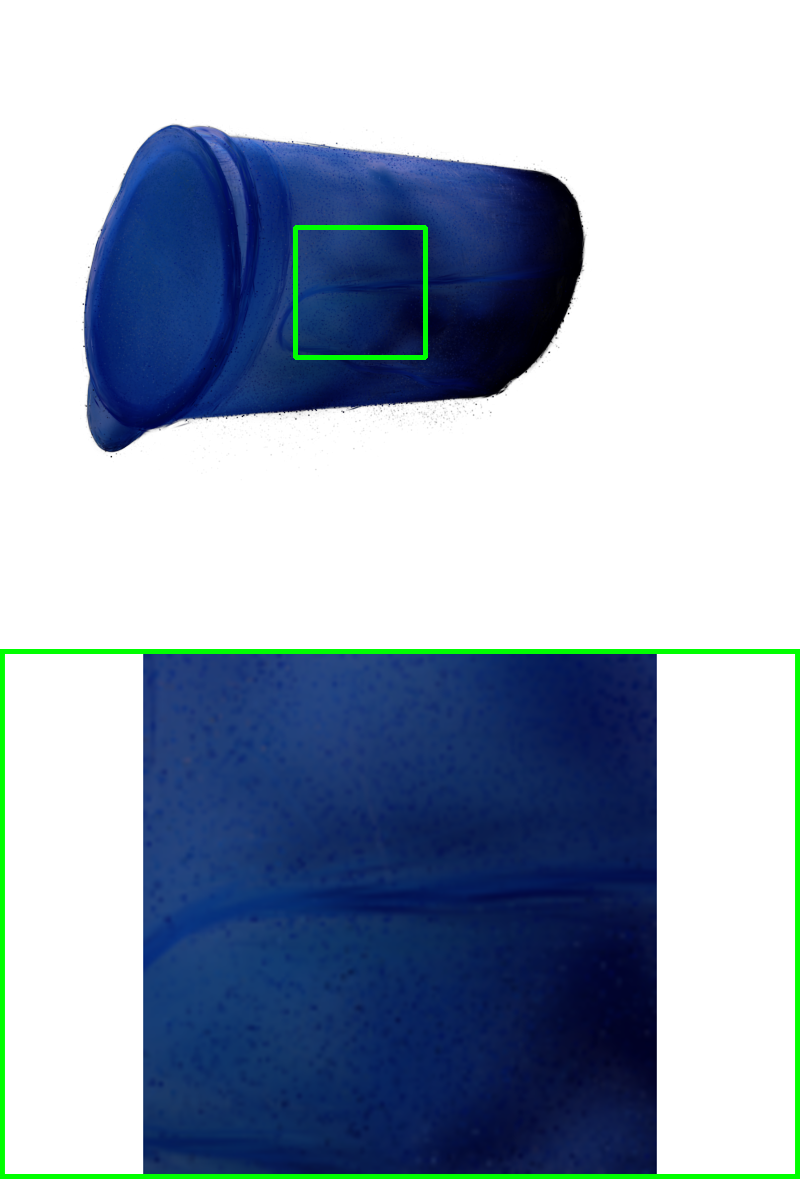}} &
\raisebox{-0.5\height}{\includegraphics[width=0.10\linewidth]{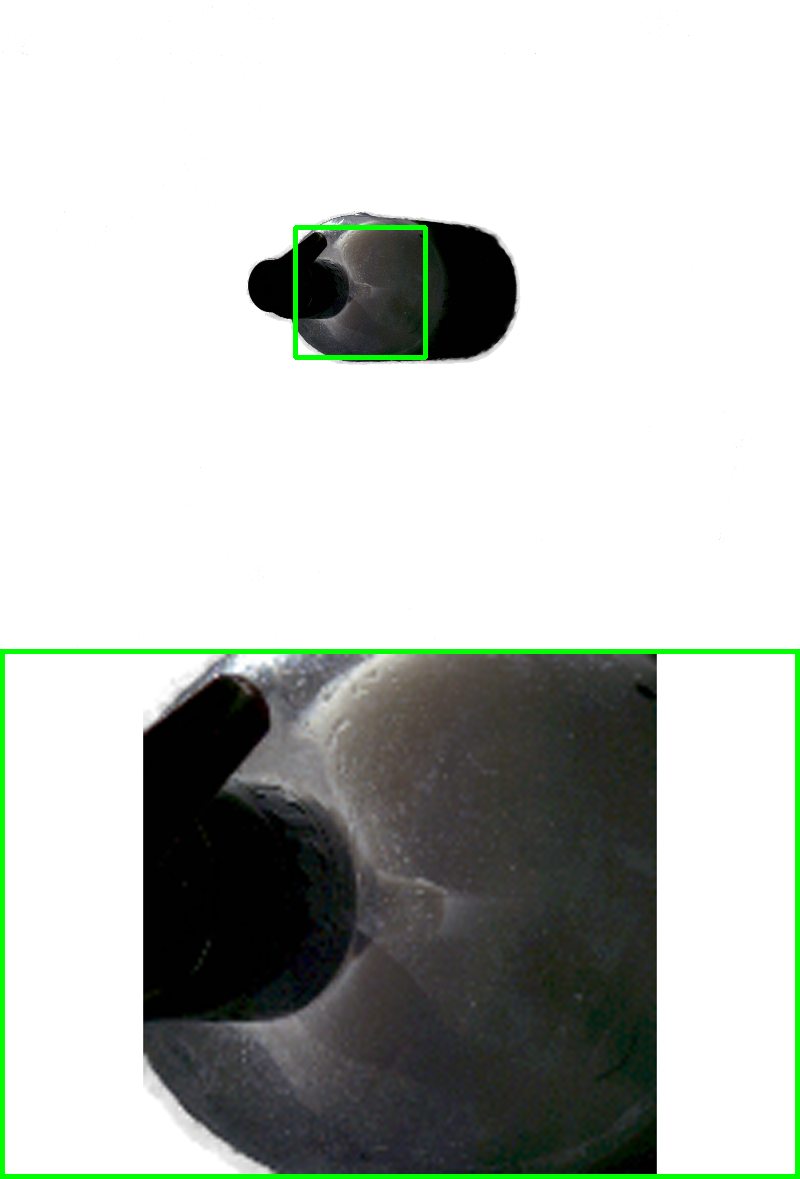}} &
\raisebox{-0.5\height}{\includegraphics[width=0.10\linewidth]{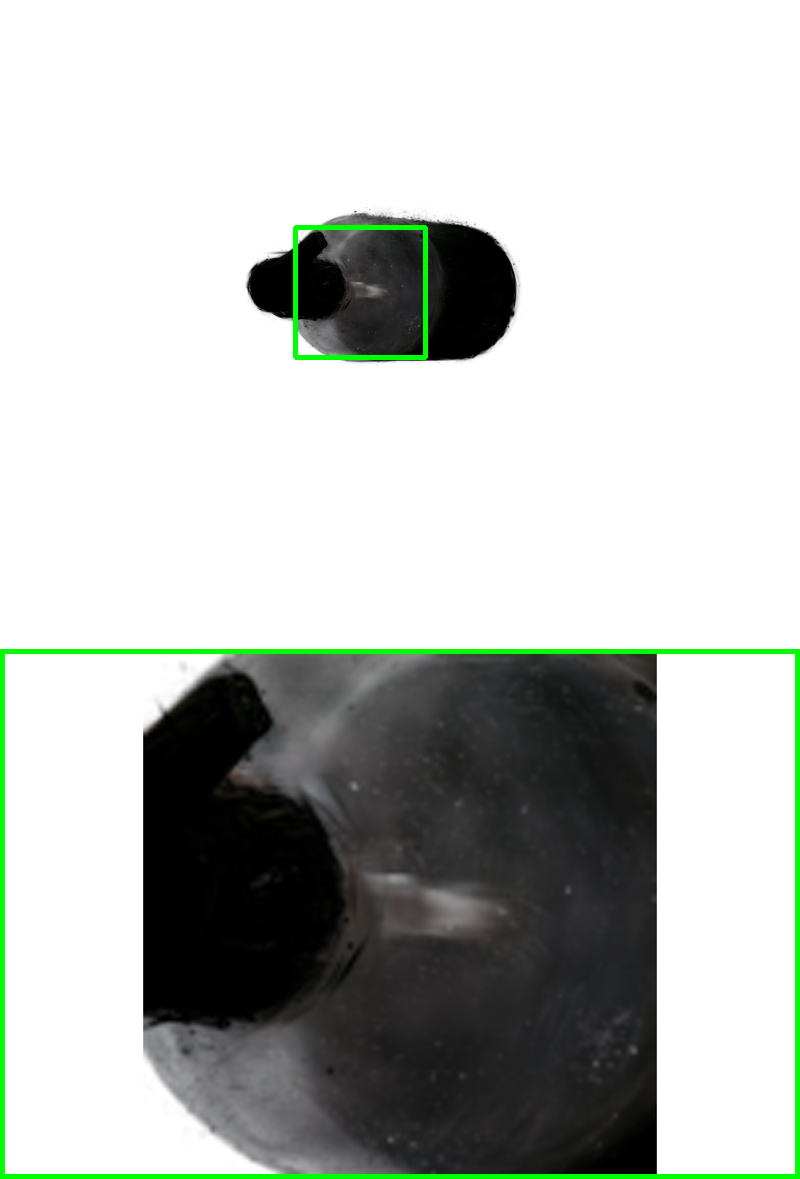}} &
\raisebox{-0.5\height}{\includegraphics[width=0.10\linewidth]{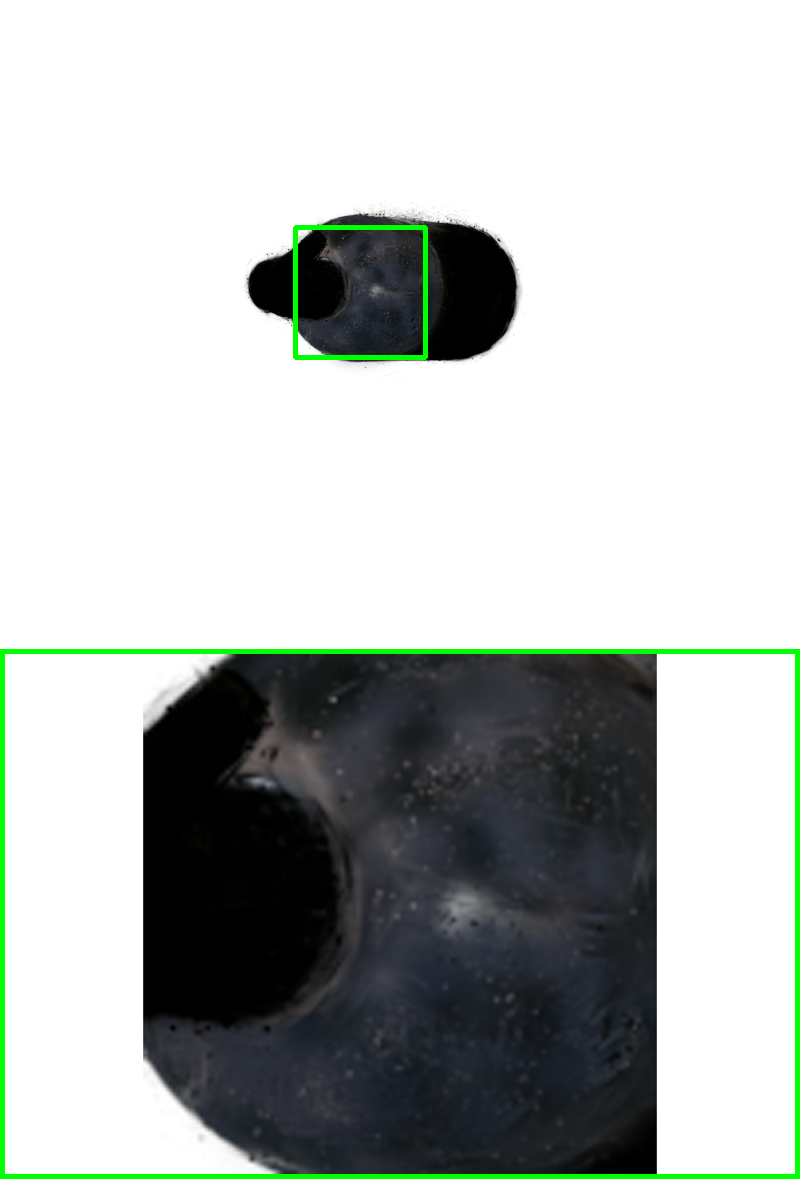}} \\
\midrule

\scriptsize all views,\;single light &
\includegraphics[width=0.10\linewidth]{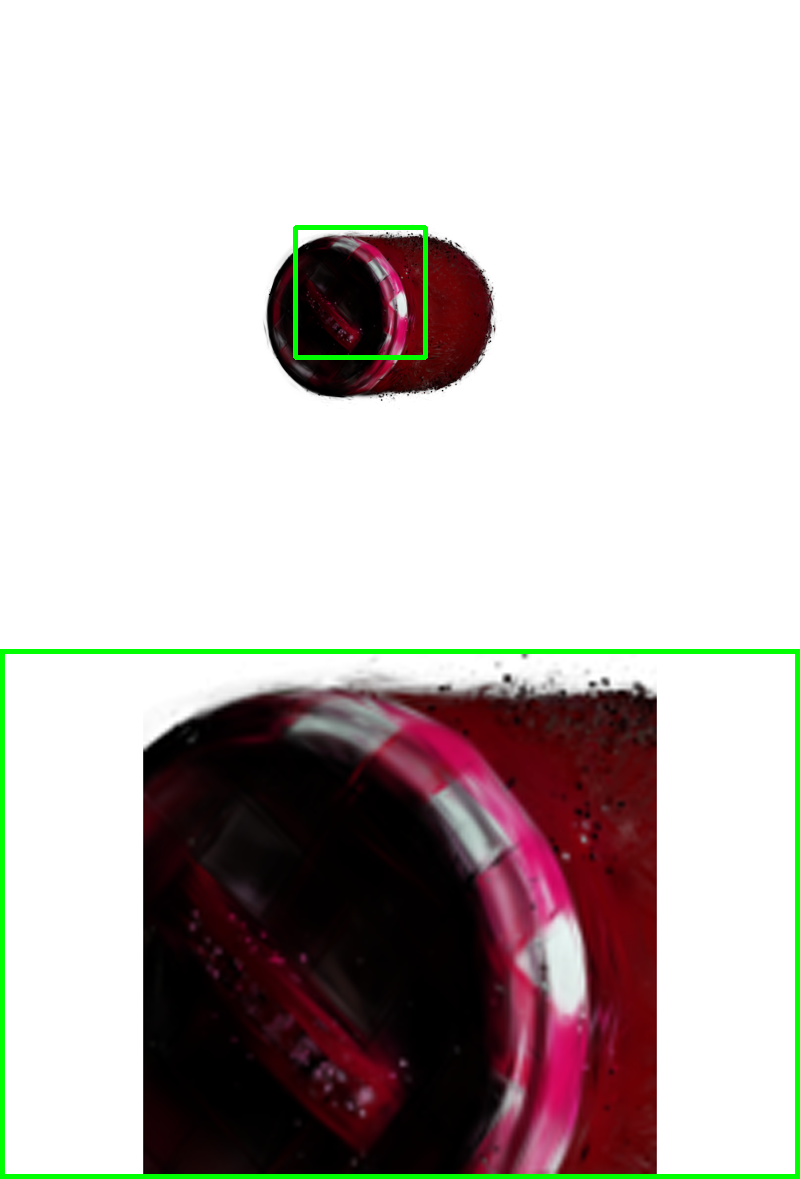} &
\includegraphics[width=0.10\linewidth]{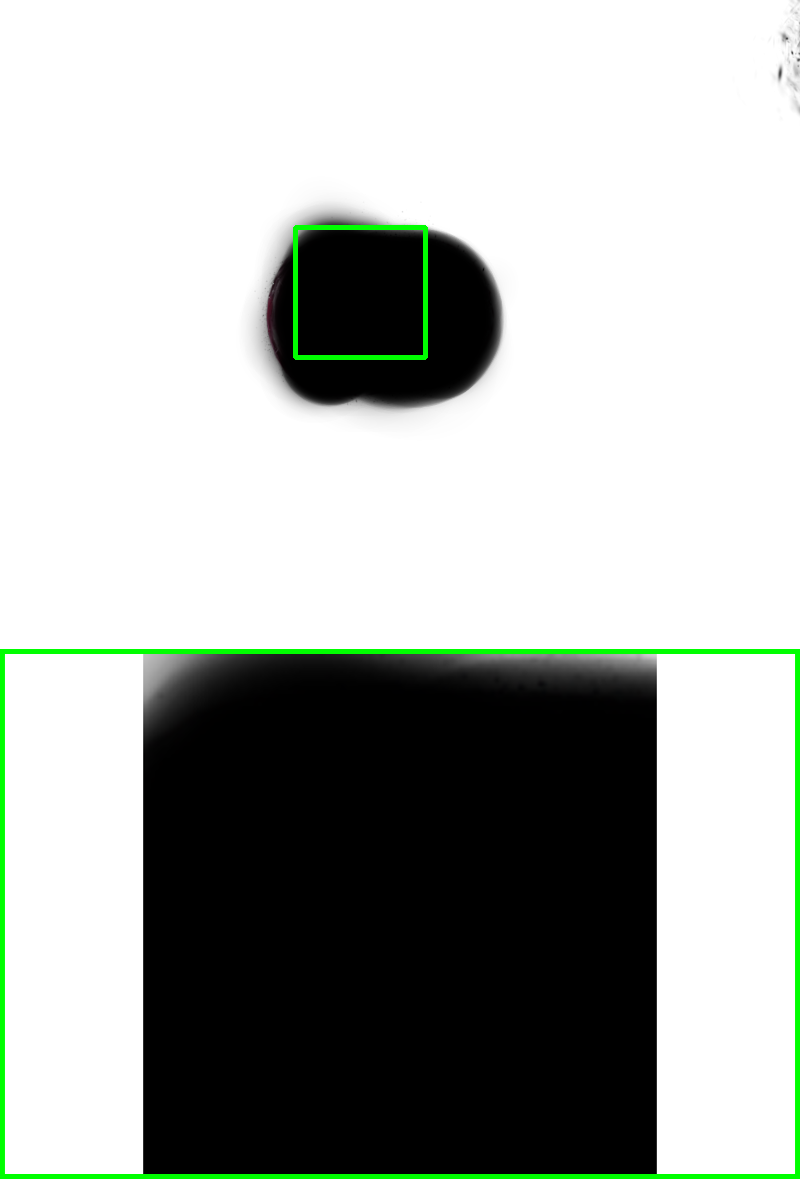} &
\scriptsize all views,\;single light &
\includegraphics[width=0.10\linewidth]{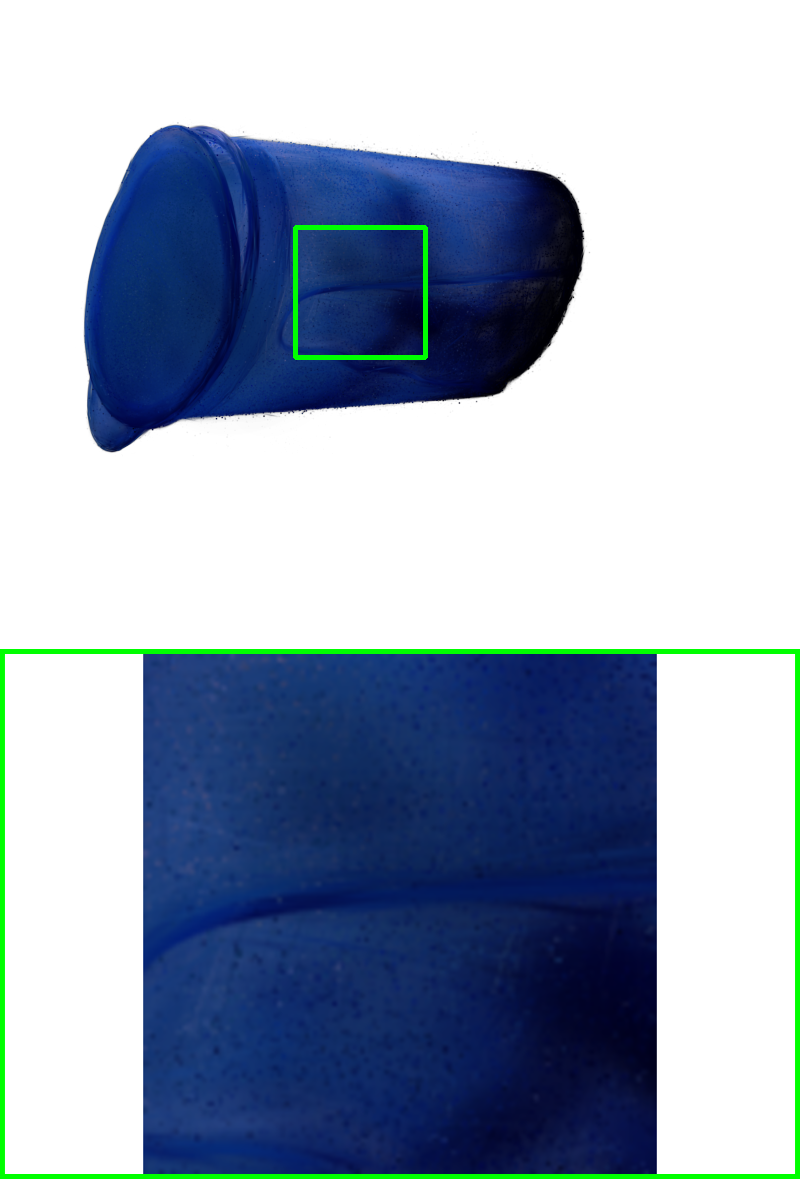} &
\includegraphics[width=0.10\linewidth]{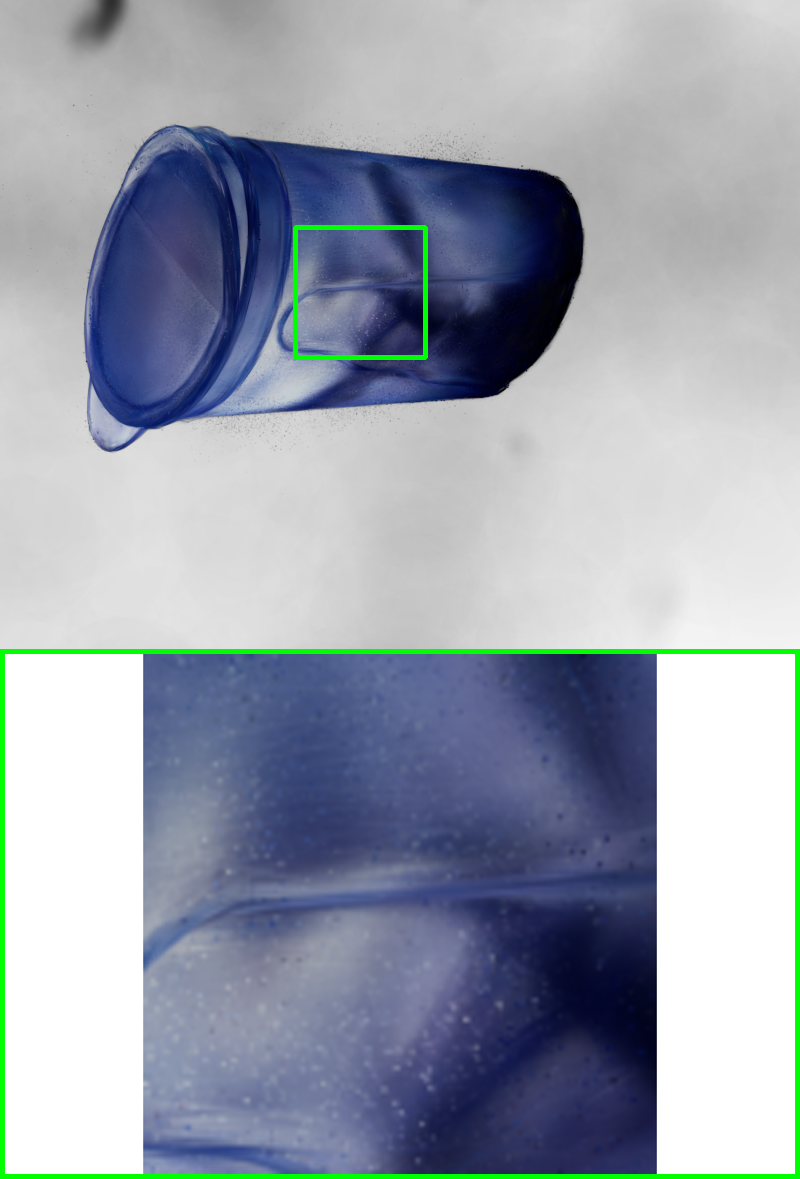} &
\scriptsize all views,\;single light &
\includegraphics[width=0.10\linewidth]{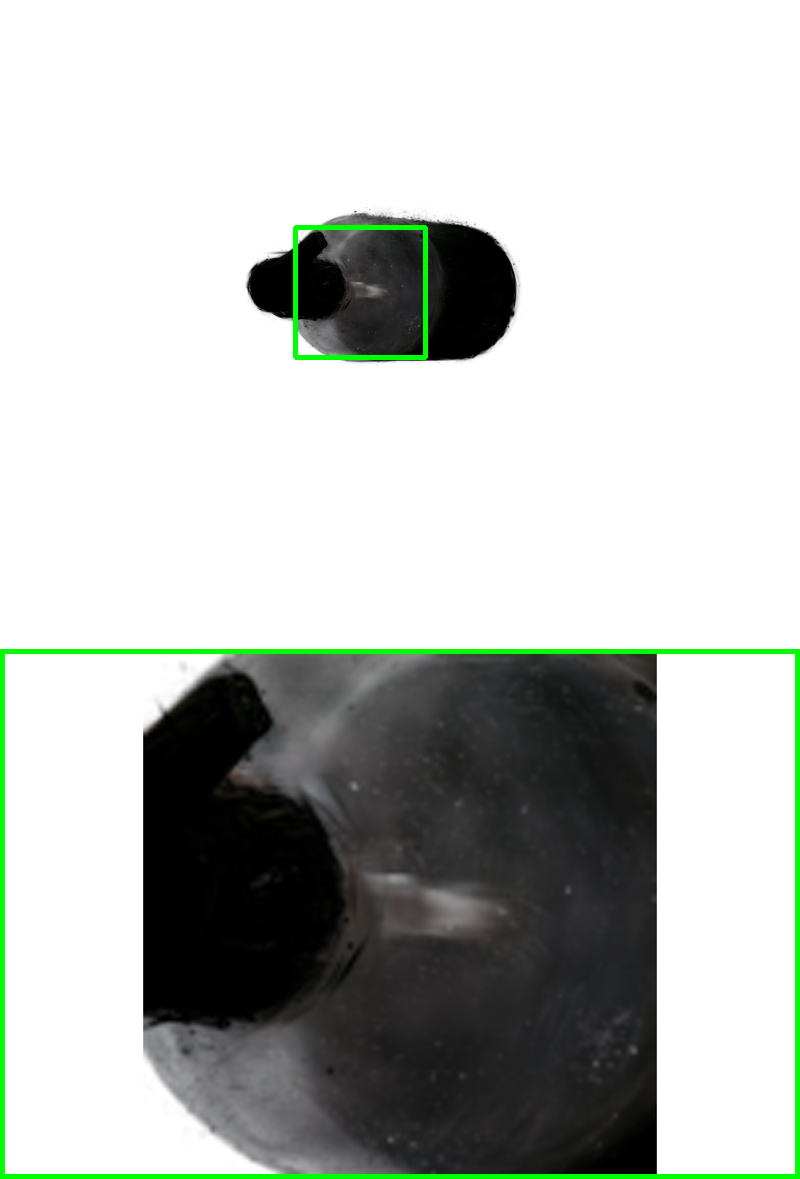} &
\includegraphics[width=0.10\linewidth]{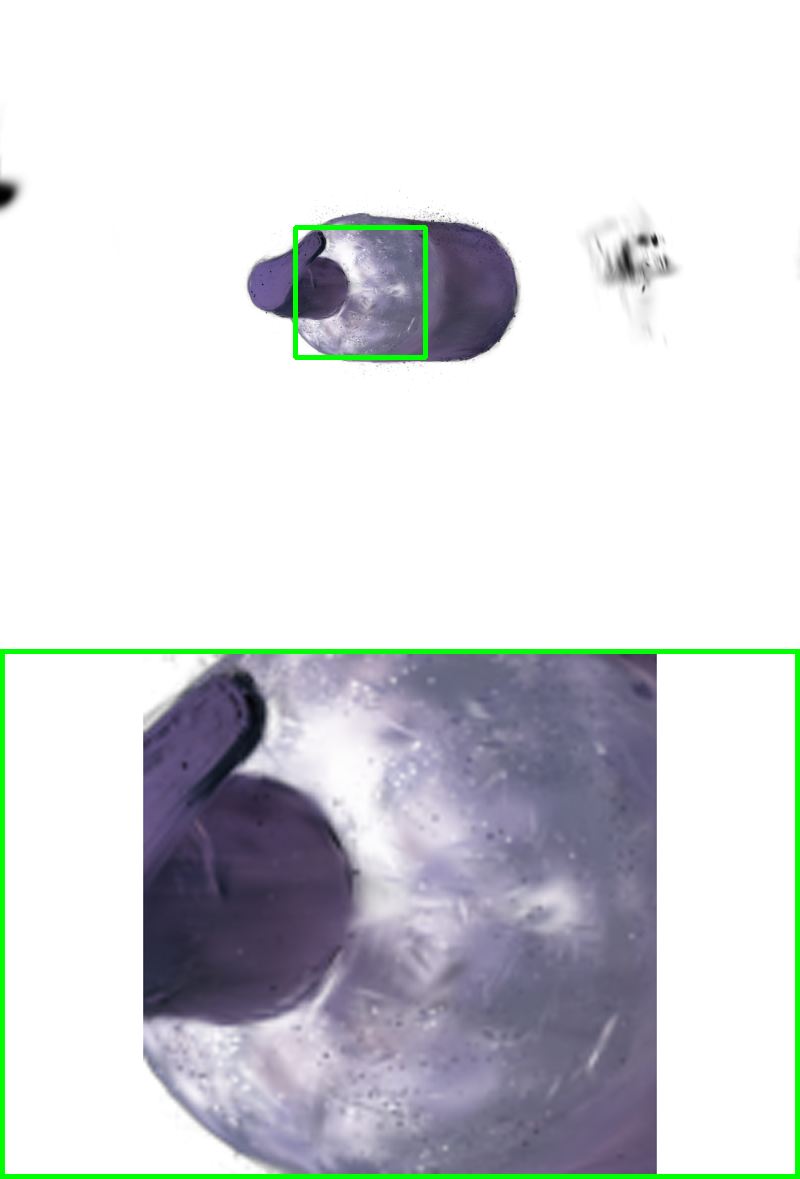} \\
\midrule

\scriptsize 5\% views,\;5\% lights &
\includegraphics[width=0.10\linewidth]{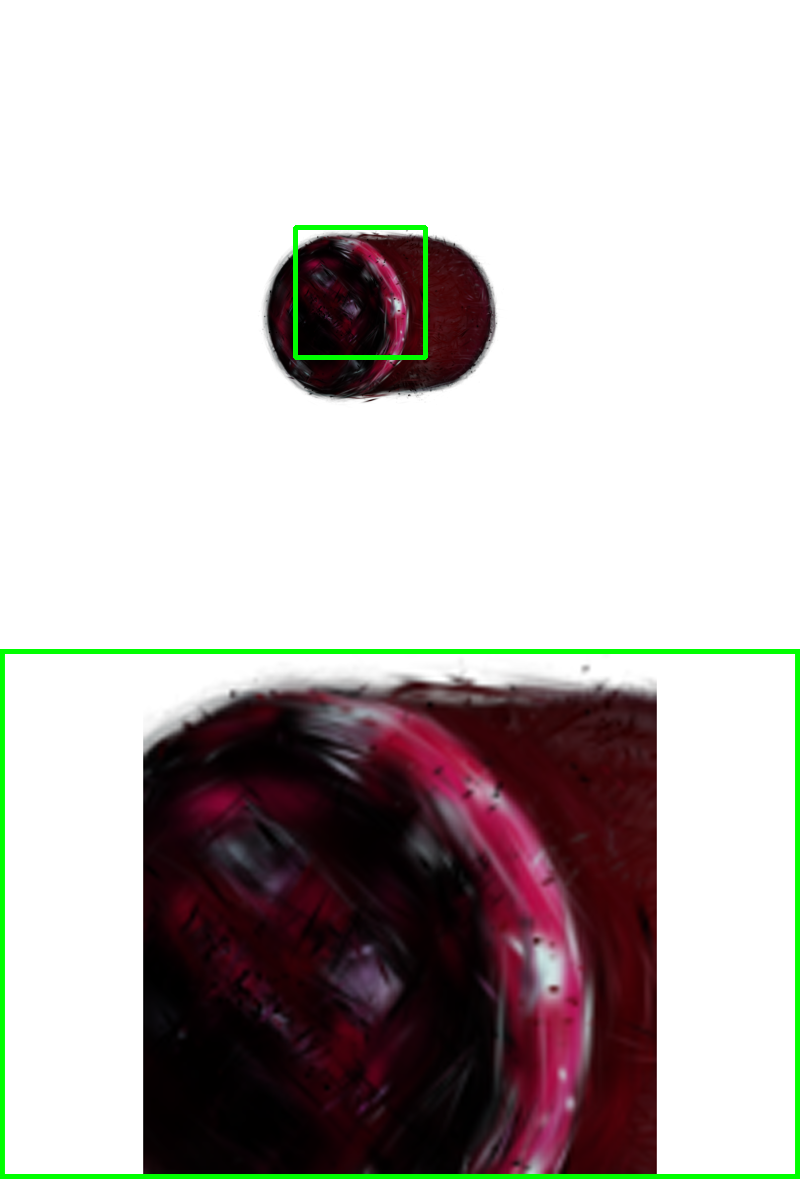} &
\includegraphics[width=0.10\linewidth]{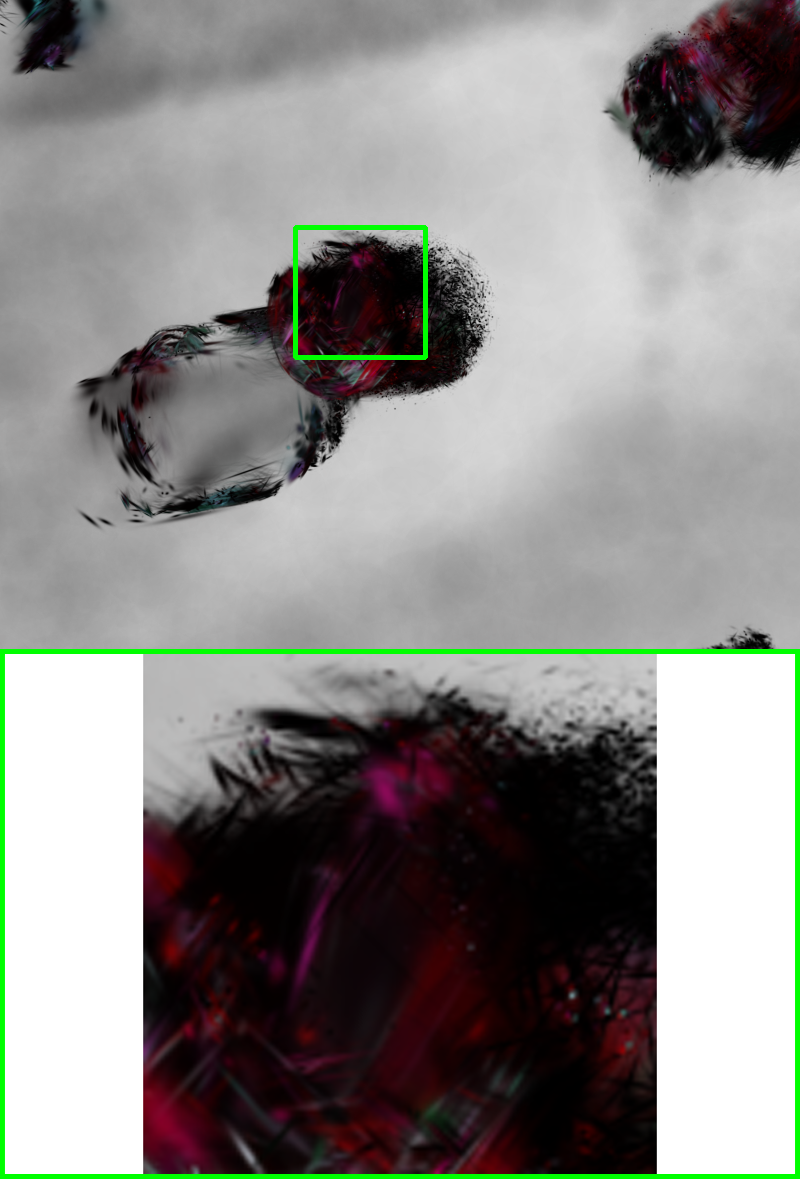} &
\scriptsize 5\% views,\;5\% lights &
\includegraphics[width=0.10\linewidth]{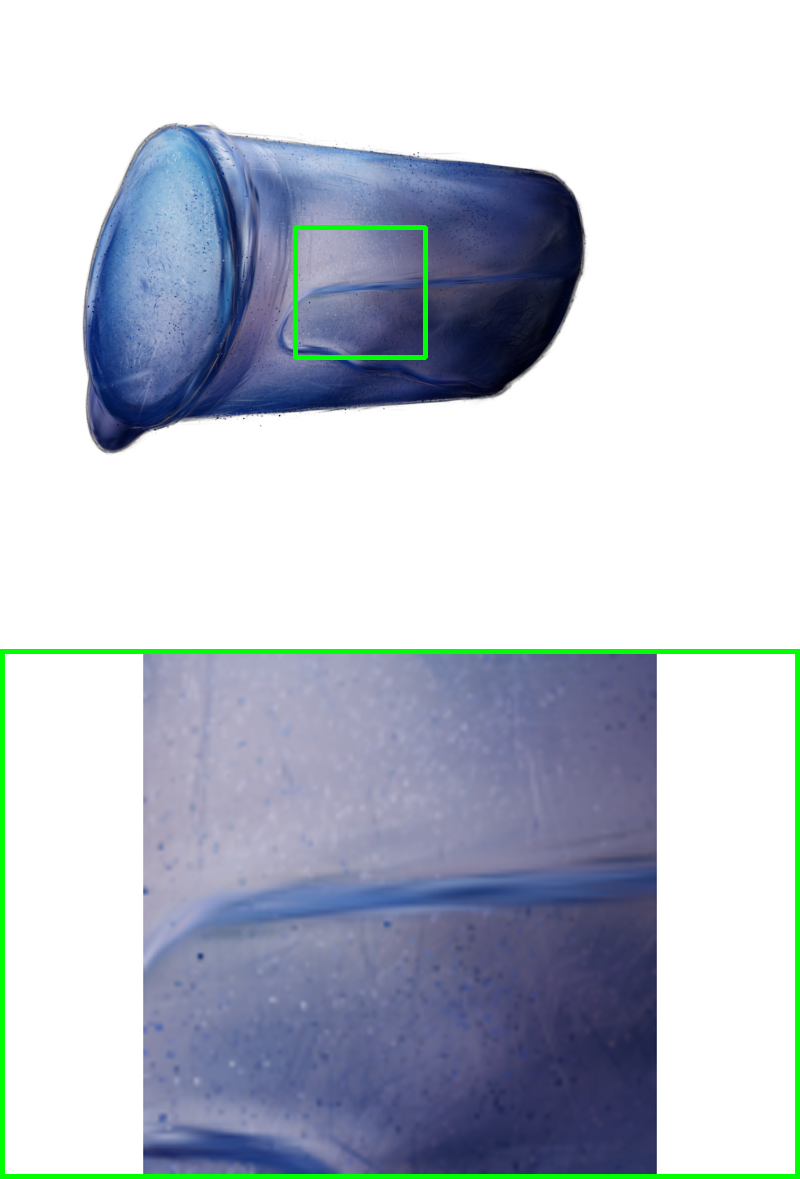} &
\includegraphics[width=0.10\linewidth]{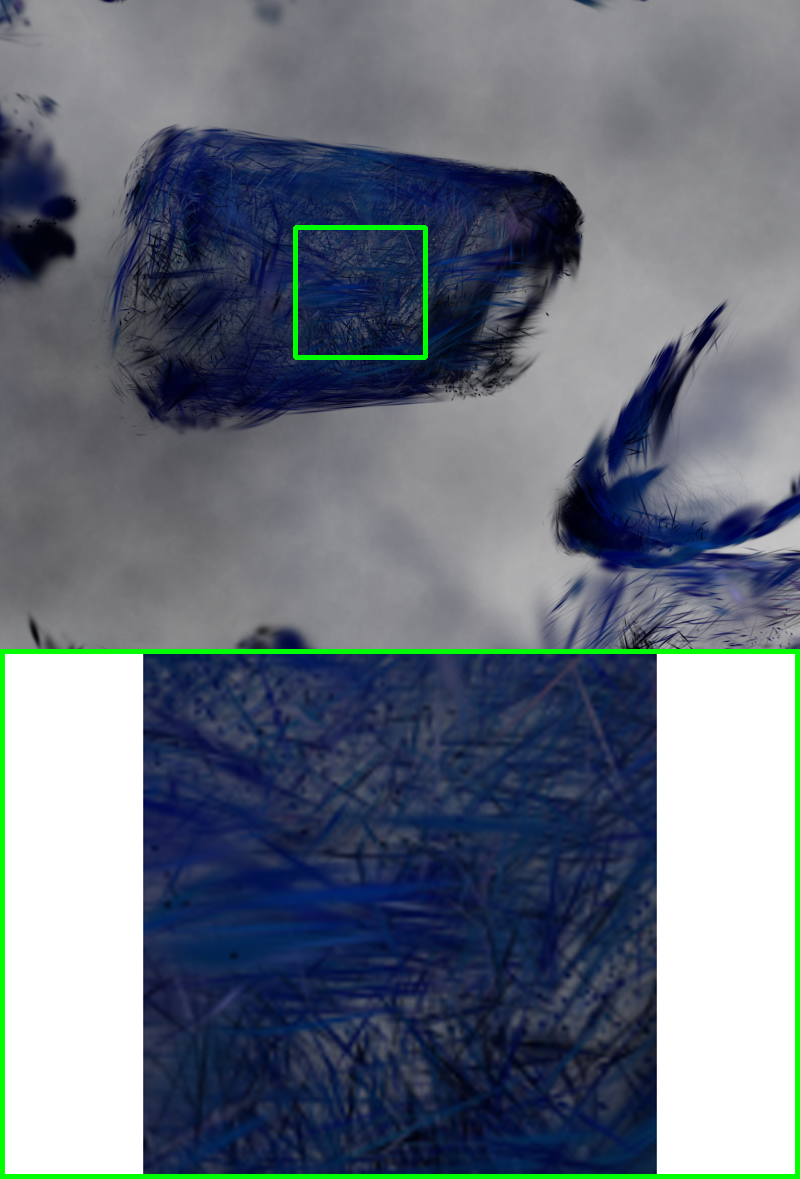} &
\scriptsize 5\% views,\;5\% lights &
\includegraphics[width=0.10\linewidth]{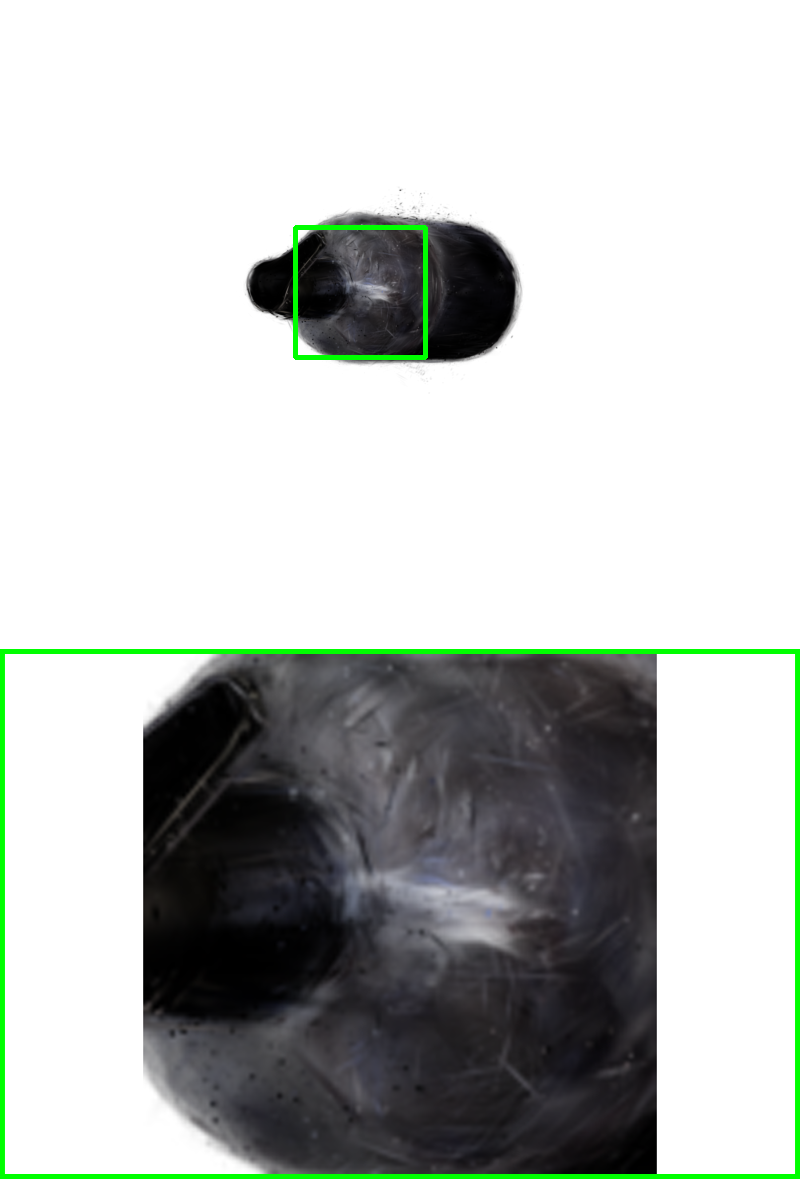} &
\includegraphics[width=0.10\linewidth]{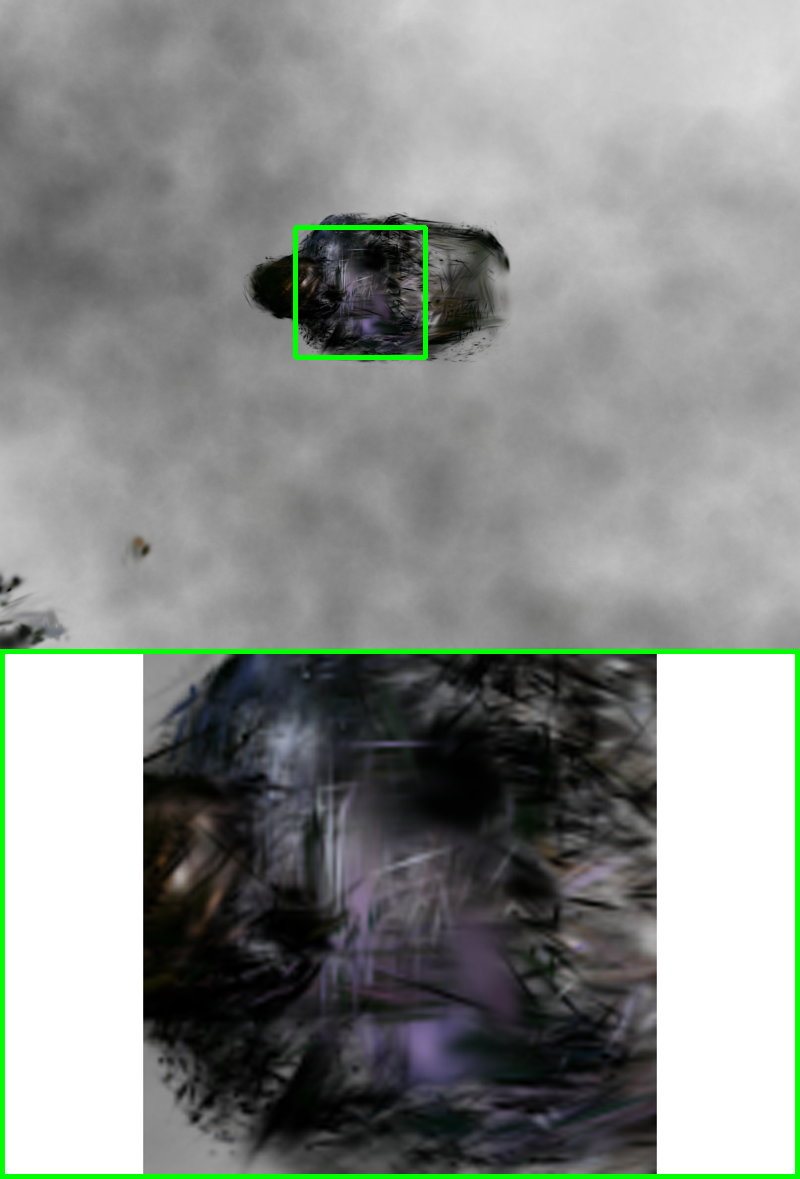} \\
\midrule

\scriptsize 3\% views,\;3\% lights &
\includegraphics[width=0.10\linewidth]{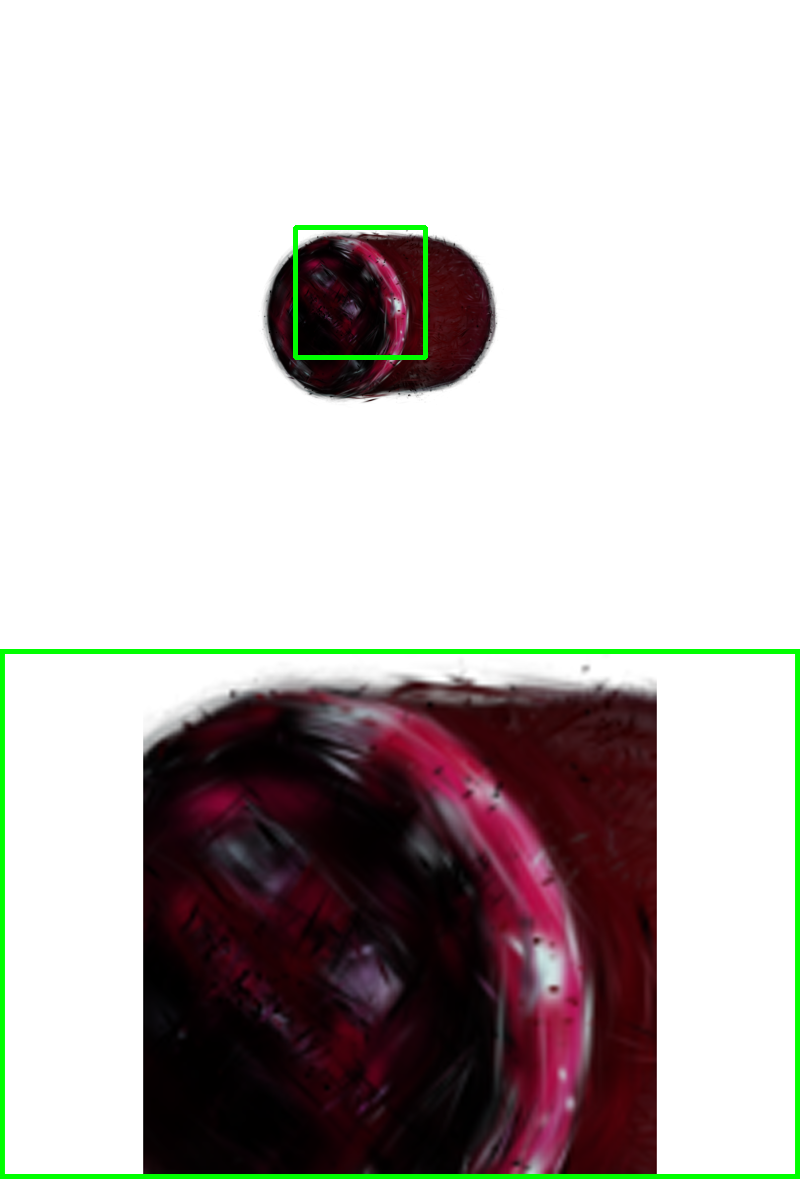} &
\includegraphics[width=0.10\linewidth]{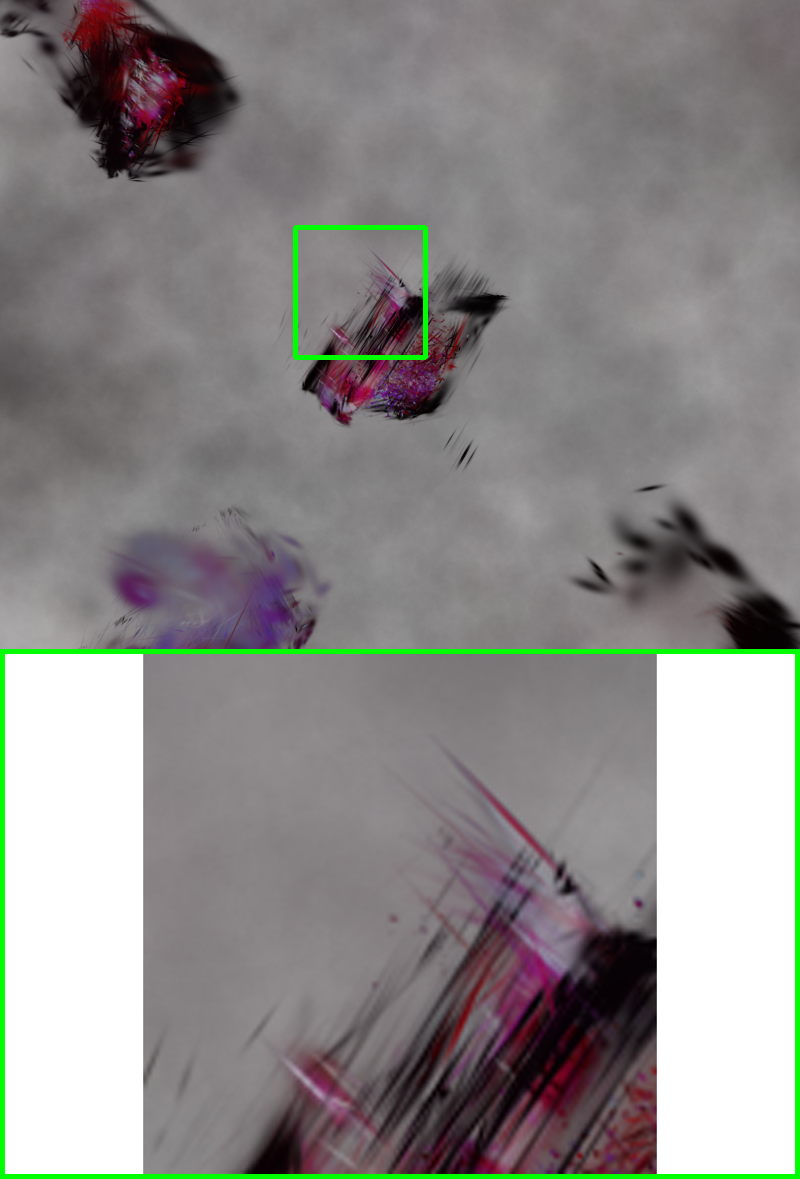} &
\scriptsize 3\% views,\;3\% lights &
\includegraphics[width=0.10\linewidth]{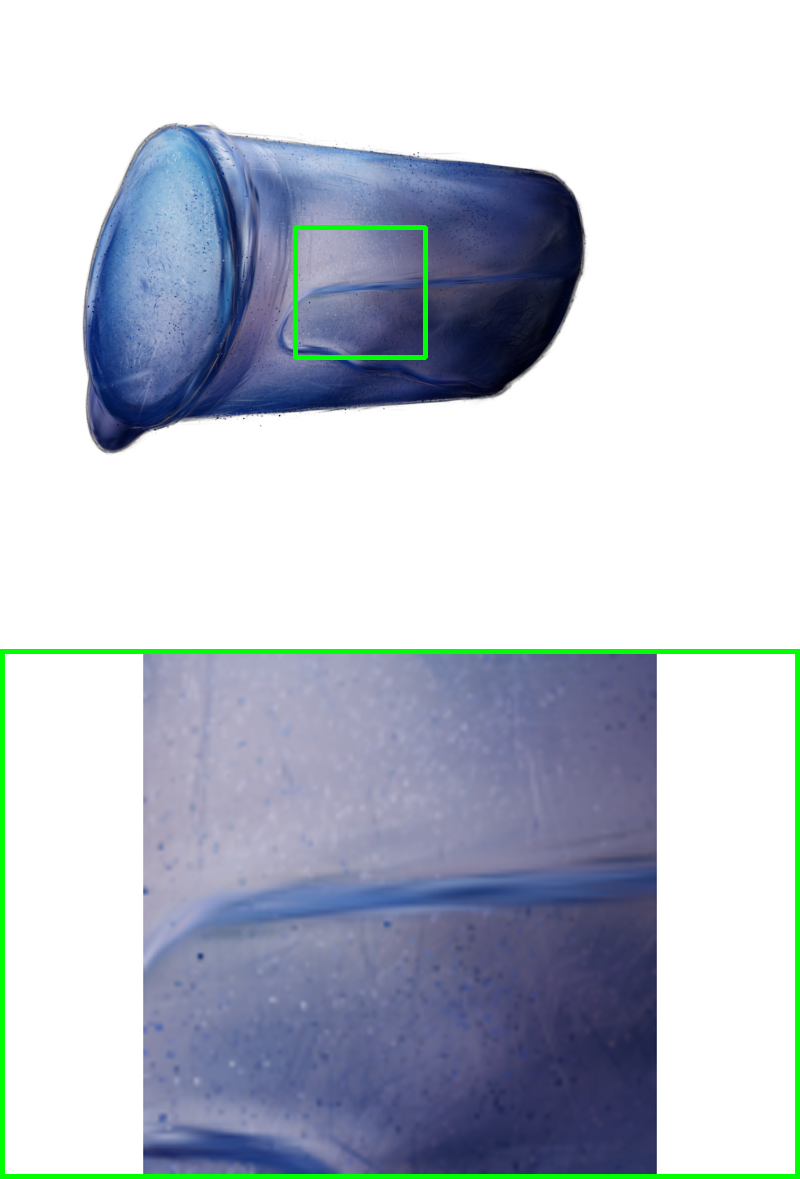} &
\includegraphics[width=0.10\linewidth]{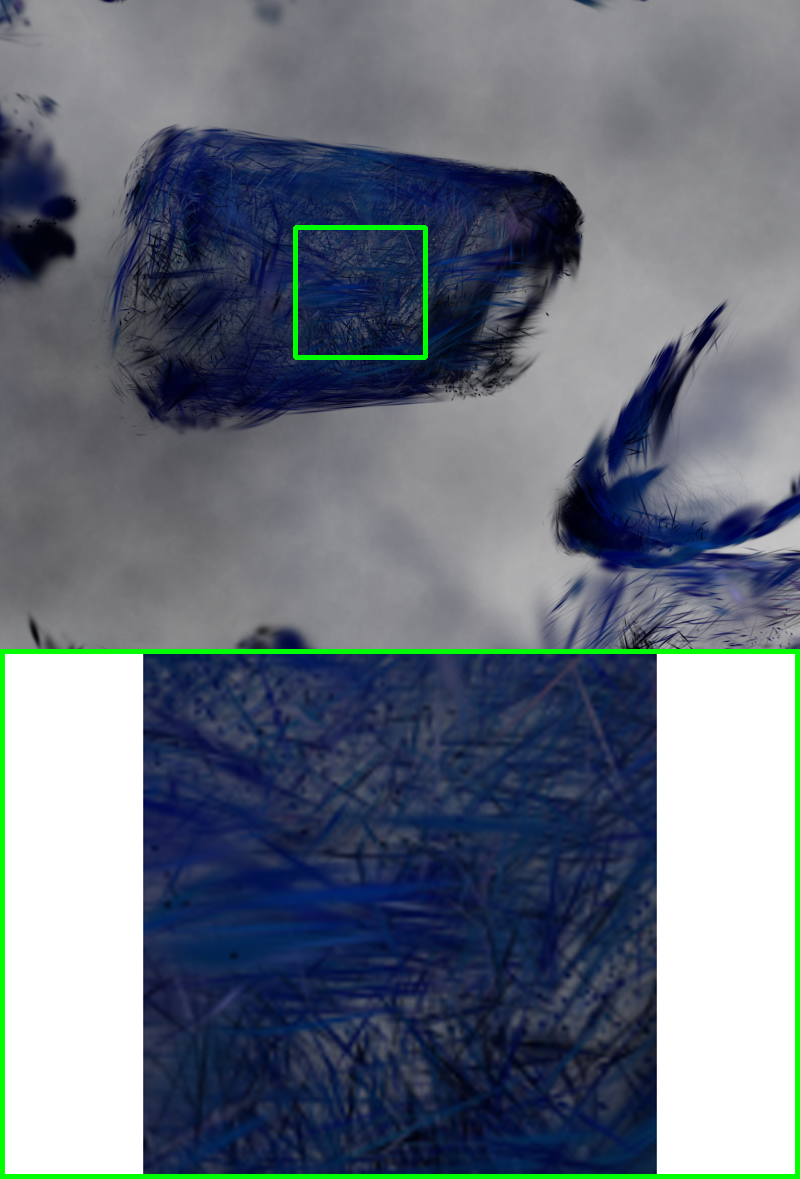} &
\scriptsize 3\% views,\;3\% lights &
\includegraphics[width=0.10\linewidth]{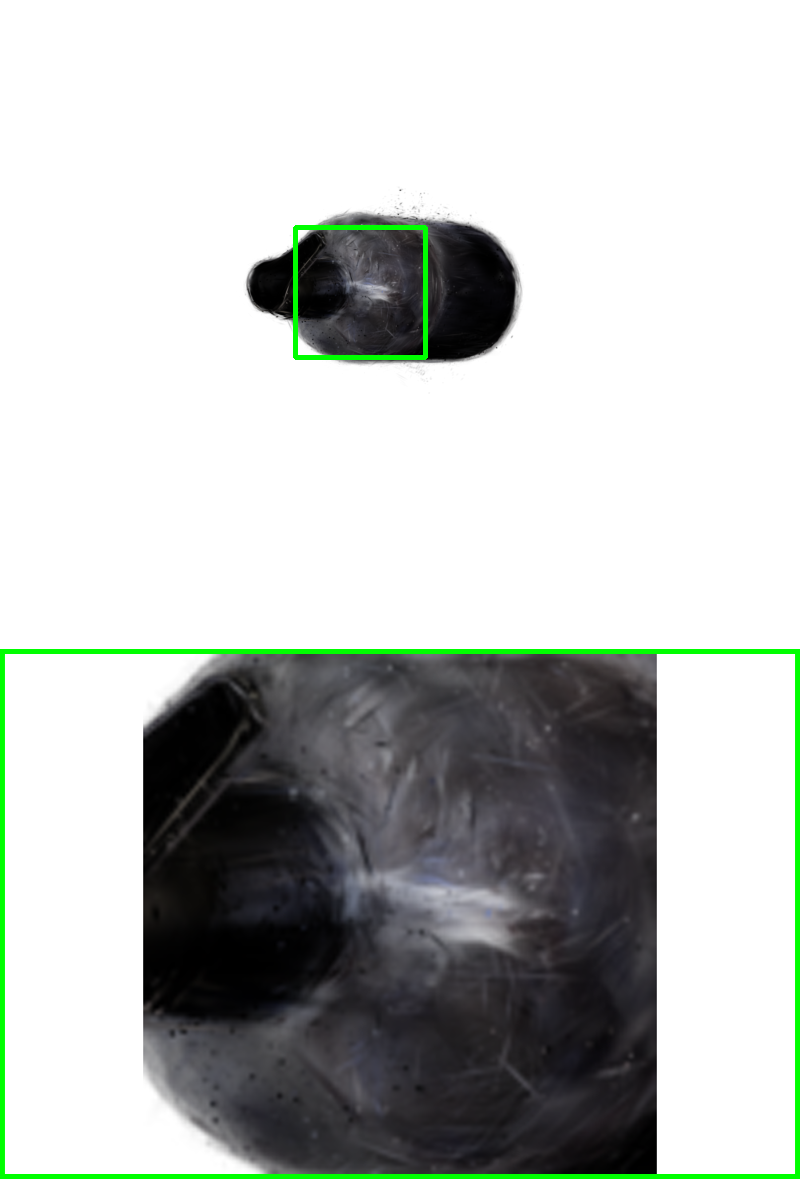} &
\includegraphics[width=0.10\linewidth]{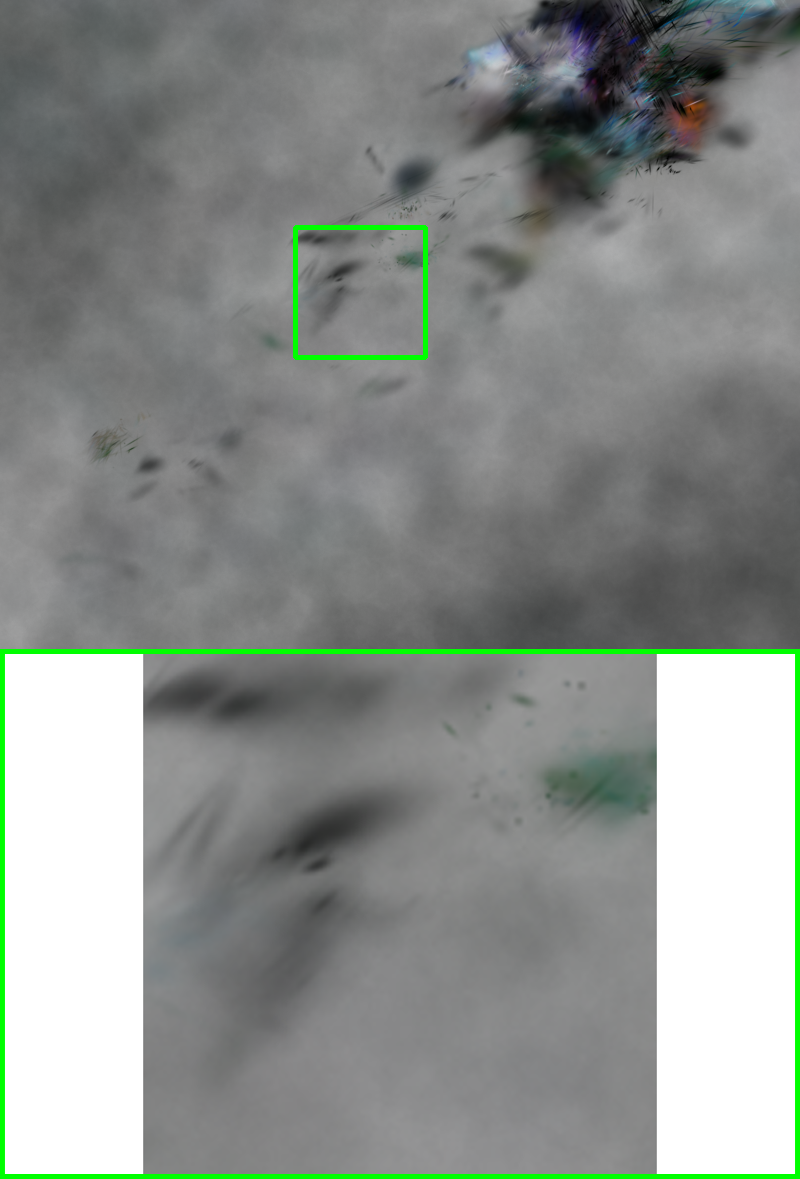} \\
\bottomrule
\end{tabular}
}
\caption{Qualitative comparison of reconstructed translucent appearance under different supervision conditions, the first row shows the setting: all views, all lights. }
\label{fig:qualitative_comparison_all_2}

\end{figure*}

\paragraph{Fine-grained appearance comparisons.}
Figure~\ref{fig:qualitative_comparison_all_2} provides additional close-up views
highlighting translucent appearance, color diffusion, and internal glow.  
Across objects, DIAMOND-SSS preserves the soft radiance falloff characteristic of
translucent materials while maintaining geometric fidelity.
In contrast, SSS-3DGS often exhibits noisy silhouettes, incorrect color
bleeding near boundaries, or inconsistent opacity when supervision becomes sparse.

These qualitative results reinforce the core claim of the paper: by integrating
diffusion-based augmentation with multi-view geometric consistency, DIAMOND-SSS
remains robust even under highly incomplete OLAT captures, providing
state-of-the-art reconstructions of translucent materials.

\section{Diffusion Fine-Tuning Details}
\label{sec:supp_diffusion}

The diffusion-based augmentation pipeline is introduced in \cref{subsec:diffusion_augmentation} and integrated into the reconstruction procedure in \cref{subsec:diamond_pipeline}.
Here we provide additional training and conditioning details for the two diffusion models used in DIAMOND-SSS: a multi-view (novel-view) diffusion model and a relighting diffusion model.
Both models are fine-tuned once on a small OLAT subset and are then reused across all test objects and capture regimes, without any per-object retraining.

\subsection{Training Data for Diffusion Models}
\label{sec:supp_diff_data}

\begin{figure*}[t]
    \centering
    \includegraphics[width=\textwidth, keepaspectratio, height=\textheight]{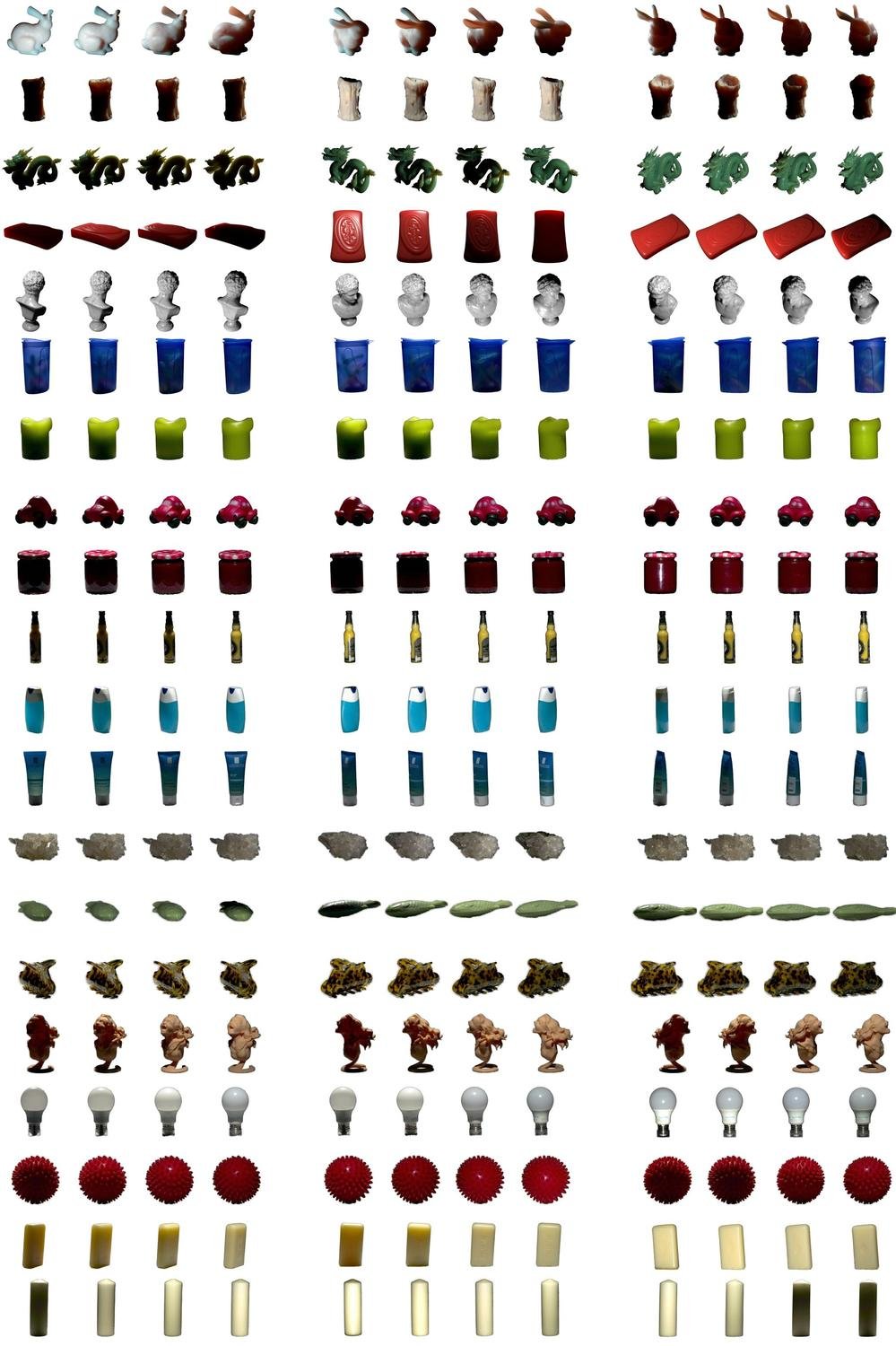}
    \caption{Dataset overview showing synthetic (top) and real (bottom) objects under different illumination and viewpoints.}
    \label{fig:sss_objects}
\end{figure*}

\begin{figure*}
    \centering
    \includegraphics[width=\linewidth]{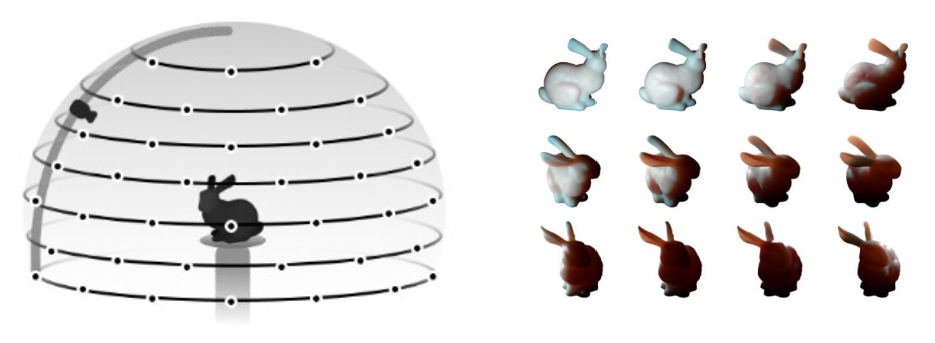}
    \caption{Overview of OLAT capture setup. A hemispherical lighting rig sequentially illuminates the object with one light at a time, capturing per-light images from fixed viewpoints.}
    \label{fig:OLAT}
\end{figure*}

\Cref{fig:sss_objects} shows an overview of the dataset and~\cref{fig:OLAT} depicts the OLAT capture setup used by \cite{dihlmann2024subsurfacescattering3dgaussian} to generate the dataset. As described in \cref{subsec:datasets,subsec:implementation}, both diffusion models are fine-tuned on a small subset ($\leq 7\%$) of the full OLAT training data from SSS-3DGS~\cite{dihlmann2024subsurfacescattering3dgaussian}.
We select four representative translucent objects---\emph{red car}, \emph{jam jar}, \emph{wax candle}, and \emph{marble head}---covering a range of scattering behaviors and specularities.

We emphasize that:
\begin{itemize}[leftmargin=*]
    \item Fine-tuning uses only training views and illuminations; no evaluation frames are seen by the diffusion models.
    \item The same fine-tuned models are reused across \emph{all} test objects and capture regimes (no per-object or per-regime retraining).
    \item Illumination is kept fixed for multi-view diffusion (to isolate geometry) and view is kept fixed for relighting diffusion (to isolate light transport).
\end{itemize}

Together, these two diffusion models are used to expand a sparse OLAT capture into a denser multi-view, multi-illumination training set for DIAMOND-SSS.

\subsection{Multi-View Diffusion (Novel-View Synthesis)}
\label{sec:supp_nvs}

\paragraph{Purpose and model selection.}
Given a reference image $I_{v_r}$ from viewpoint $v_r$ and a target pose $P_v = (K, T_v)$, the multi-view diffusion module synthesizes the corresponding novel view $\hat I_v$.
This densifies the camera trajectory under fixed illumination, enabling cross-combination with the relighting module to obtain synthetic multi-view, multi-light supervision within DIAMOND-SSS.

We first benchmarked several diffusion-based NVS models in inference mode—Zero123~\cite{liu2023zero}, Zero123++~\cite{shi2023zero123plus}, MVD-Fusion~\cite{mvdfusion}, and Free3D~\cite{zheng24}.
On SSS-3DGS OLAT data, Free3D showed the strongest geometry and silhouette consistency out of the box, and we adopt it as our baseline prior.
Representative comparisons are provided in \cref{supp:fig:nvs_statue,supp:fig:nvs_bunny,supp:fig:nvs_soap}.
See \cref{fig:approach} for an overview of the architecture.

\paragraph{Training procedure.}
We fine-tune the official Free3D checkpoint~\cite{zheng24} (latent-diffusion UNet, $32{\times}32$ latents, scaling $0.18215$) using the original denoising objective and linear noise schedule ($0.00085 \rightarrow 0.0120$ over 1000 steps).
Training sequences are sampled under fixed illumination so that the model learns geometric variation without entangling lighting.

Camera poses are injected through ray-conditioning normalization layers, and lightweight multi-view attention and noise-sharing modules promote cross-view consistency.
No architectural changes are introduced; only the weights are adapted to the OLAT domain.

\textbf{Fine-tuning hyperparameters:}
\begin{itemize}[leftmargin=*]
  \item \textbf{Batch size:} $16$ target views sampled from $4$ distinct objects,
  \item \textbf{Learning rate:} $1\times 10^{-4}$ with $100$ warm-up steps,
  \item \textbf{Training duration:} $\approx 500$ epochs on the OLAT subset,
  \item \textbf{Memory optimization:} Gradient checkpointing enabled.
\end{itemize}

\paragraph{Use within DIAMOND-SSS.}
During reconstruction, synthetic views generated by the multi-view diffusion module are rendered through the same camera poses as real observations.
Photometric losses on synthetic images are down-weighted by the factor $\alpha$ defined in \cref{subsec:diamond_pipeline}, while geometric consistency losses (e.g., silhouette or depth consistency) benefit fully from the increased viewpoint density.

\subsection{Relighting Diffusion}
\label{sec:supp_relit}

\paragraph{Purpose and model selection.}
Given a single-view RGB input $I_{v,l_s}$ under source illumination $l_s$, together with auxiliary conditioning maps $C_v$ (depth and normals), the relighting diffusion module predicts the same view under a novel target light $l_{tgt}$.
This expands the illumination coverage under fixed geometry and is especially useful in regimes such as ``all views, 1 light per view,’’ where synthetic OLAT variants provide dense lighting supervision.

We adopt a ControlNet-style conditional diffusion framework inspired by~\cite{zhang2023controlnet,poirier2024radiancefield}, starting from the public implementation of Poirier-Ginter~\etal~\cite{poirier2024radiancefield}.
An architectural overview is included in \cref{fig:approach}.

\paragraph{Training procedure.}
We augment the model original model to receive the following conditioning signals:
\begin{itemize}[leftmargin=*]
  \item \textbf{Source RGB image:} $I_{v,l_s}$,
  \item \textbf{Estimated depth map:} from Depth Anything~v2~\cite{depth_anything_v2},
  \item \textbf{Estimated surface normals:} from Marigold~\cite{marigold},
  \item \textbf{Target illumination:} 9D spherical harmonic encoding $\Phi(l_{tgt})$.
\end{itemize}

\begin{figure*}[htbp]
\label{app:sanity_samples}
\centering
\setlength{\tabcolsep}{2pt}
\begin{tabular}{*{8}{p{.11\textwidth}}}

\includegraphics[width=\linewidth]{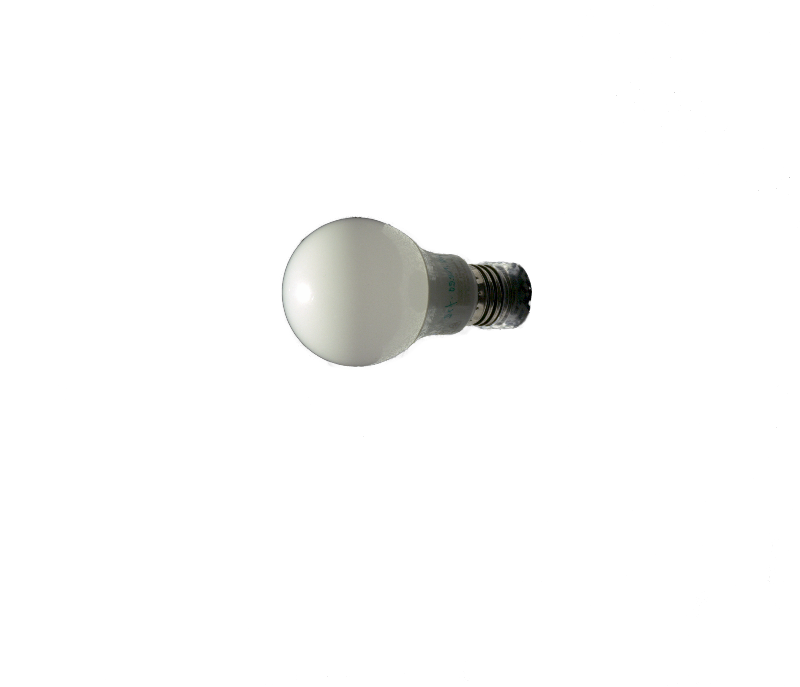} &
\includegraphics[width=\linewidth]{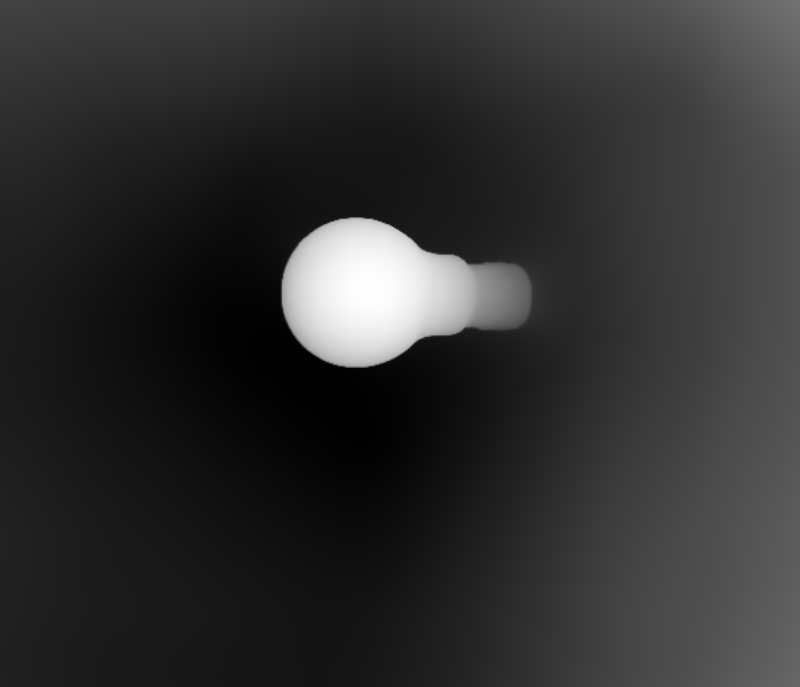} &
\includegraphics[width=\linewidth]{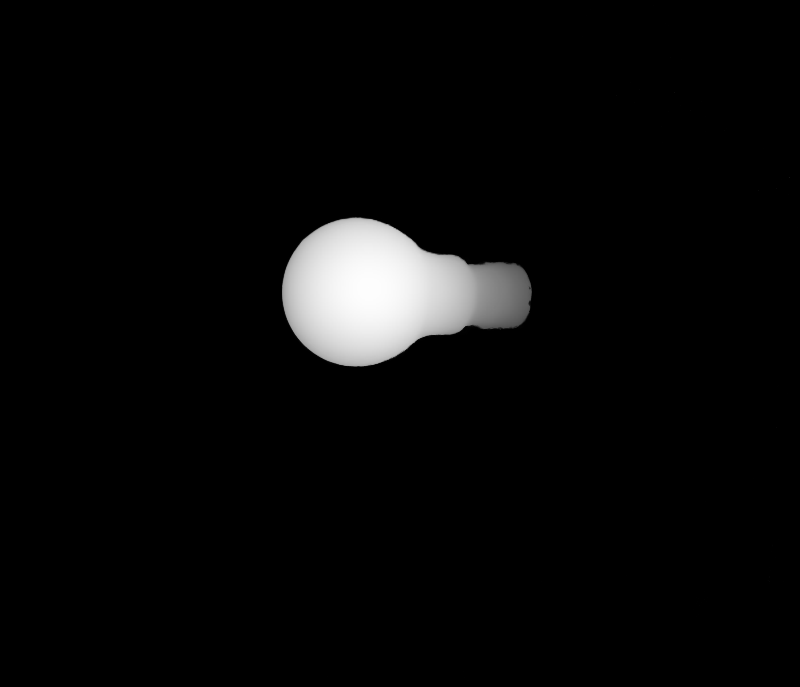} &
\includegraphics[width=\linewidth]{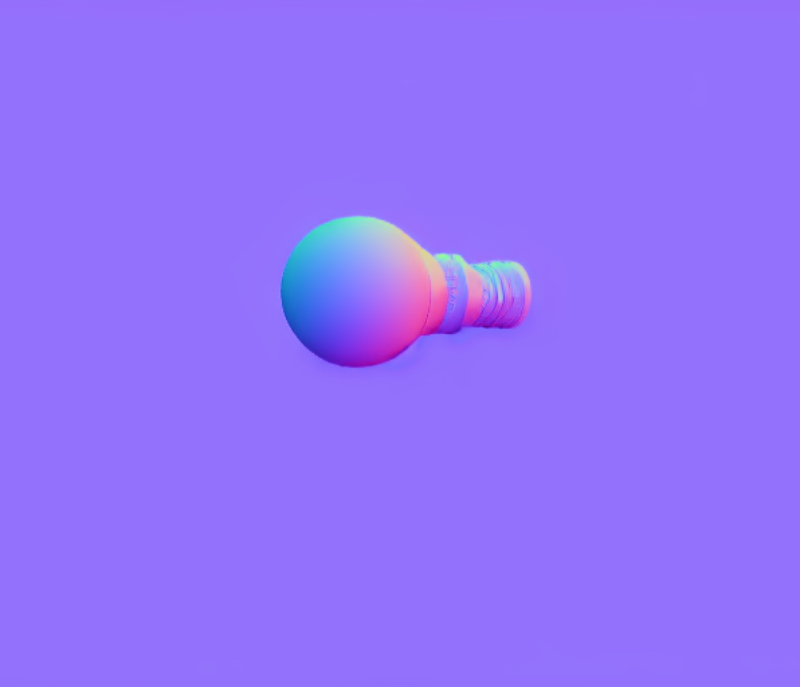} &
\includegraphics[width=\linewidth]{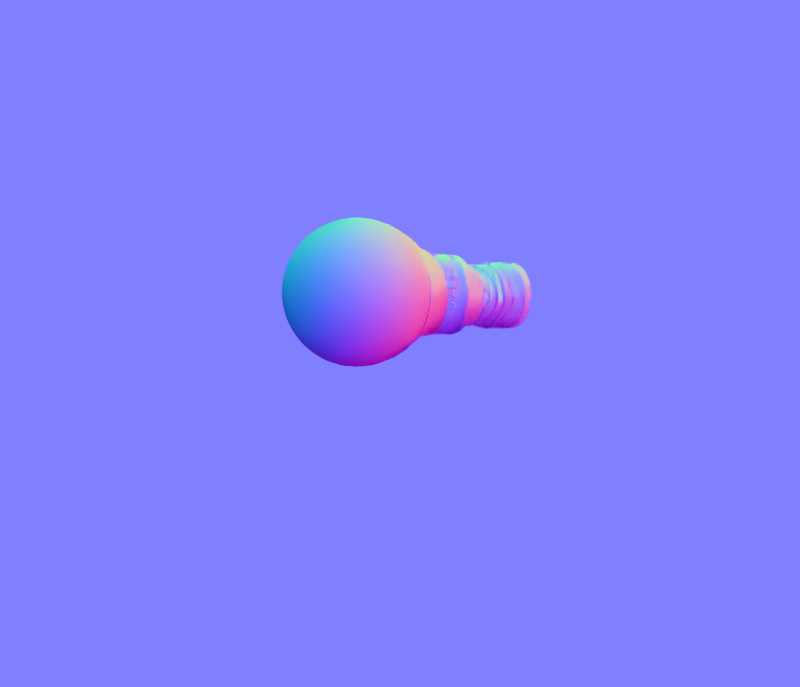} &
\includegraphics[width=\linewidth]{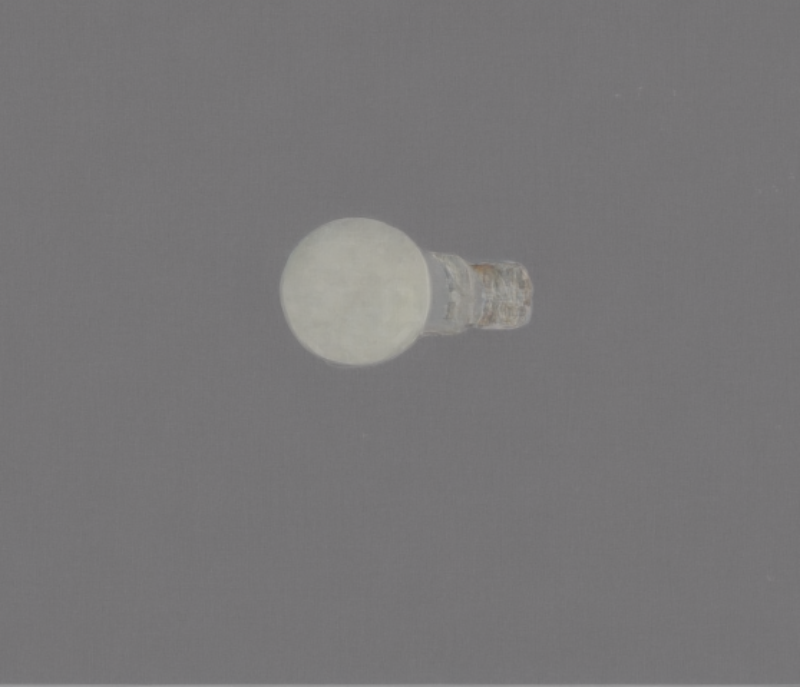} &
\includegraphics[width=\linewidth]{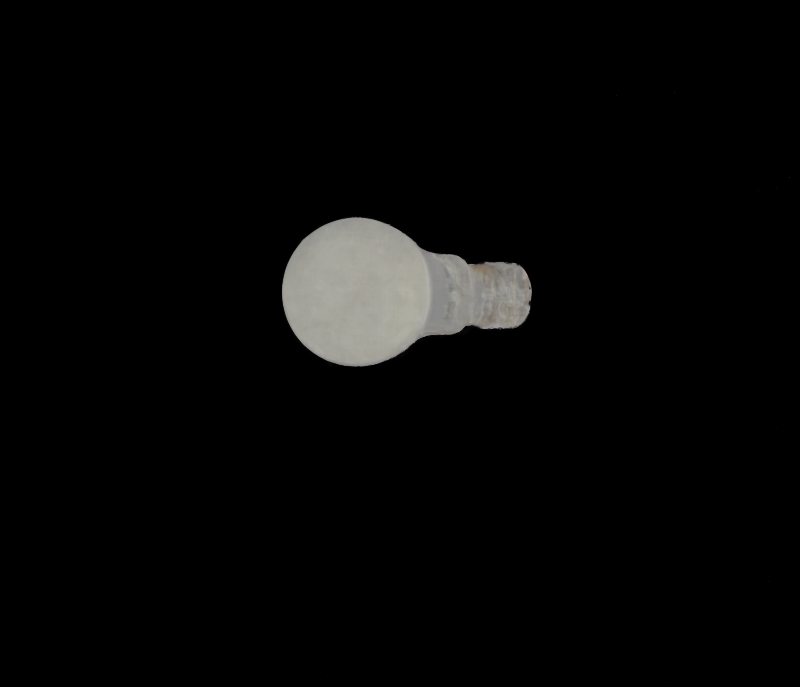} &
\includegraphics[width=\linewidth]{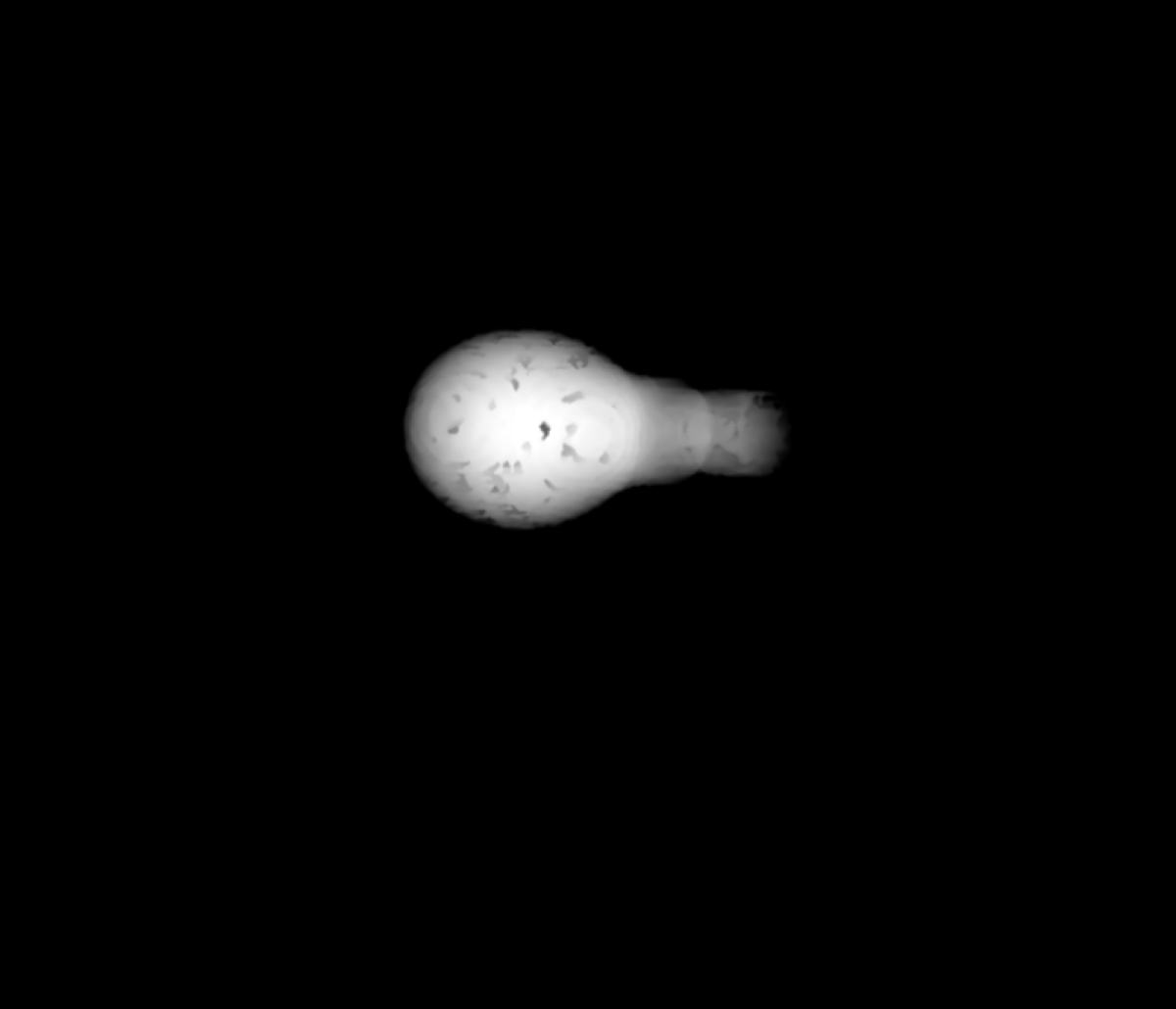}\\

\includegraphics[width=\linewidth]{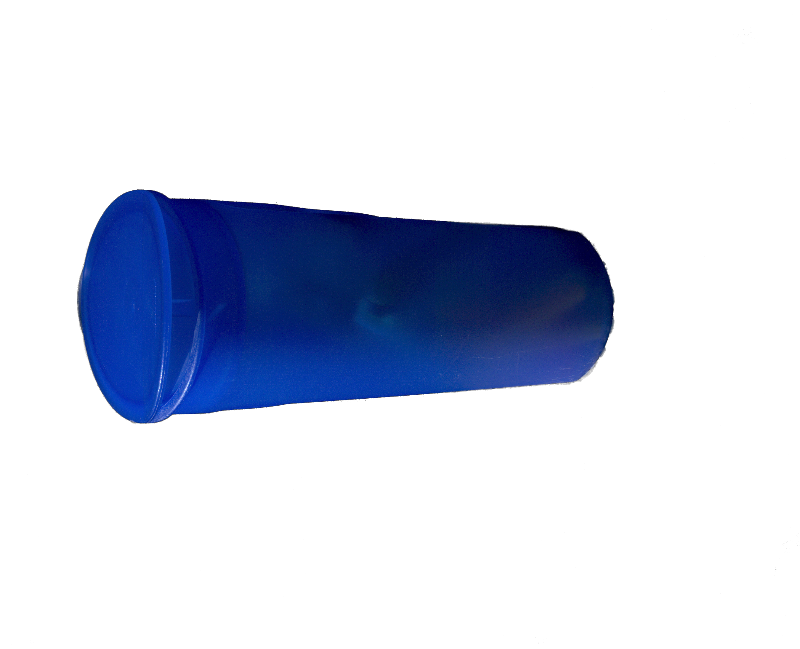} &
\includegraphics[width=\linewidth]{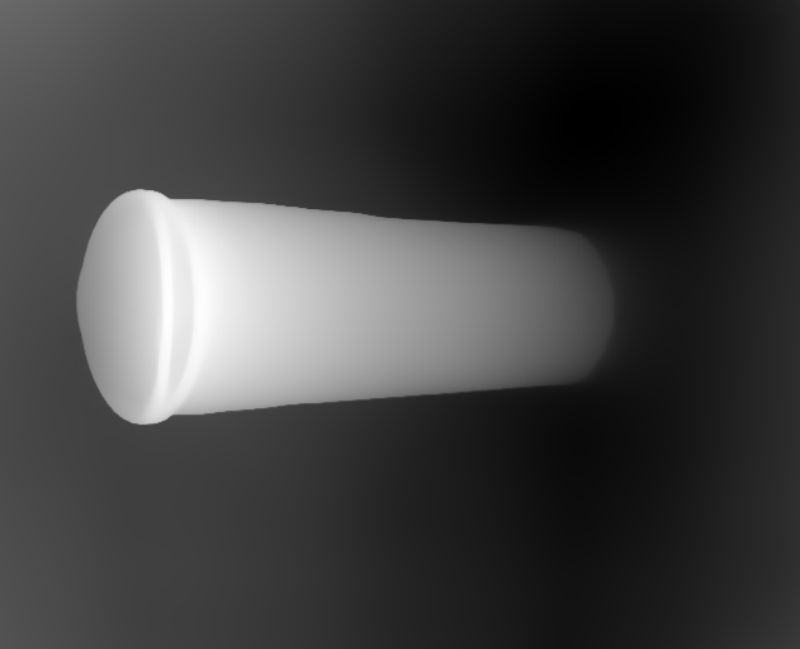} &
\includegraphics[width=\linewidth]{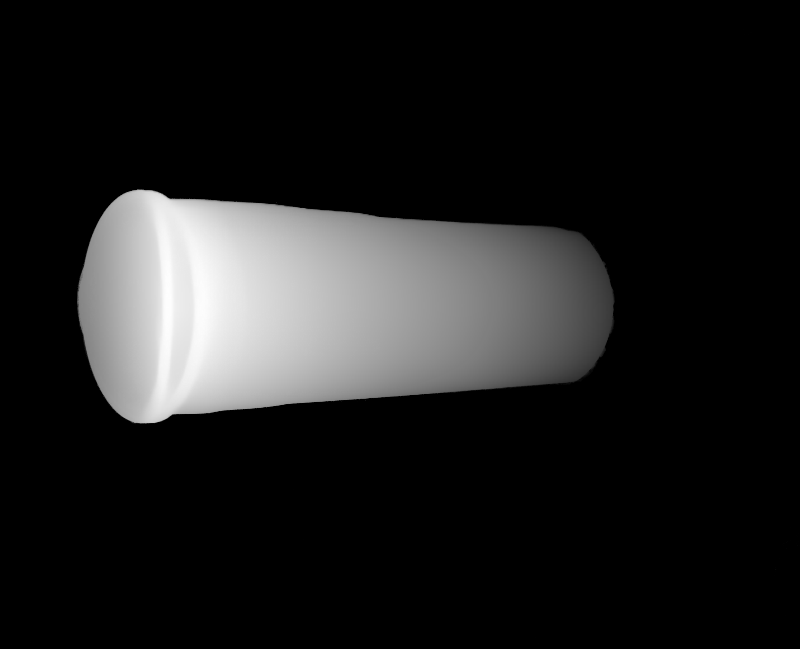} &
\includegraphics[width=\linewidth]{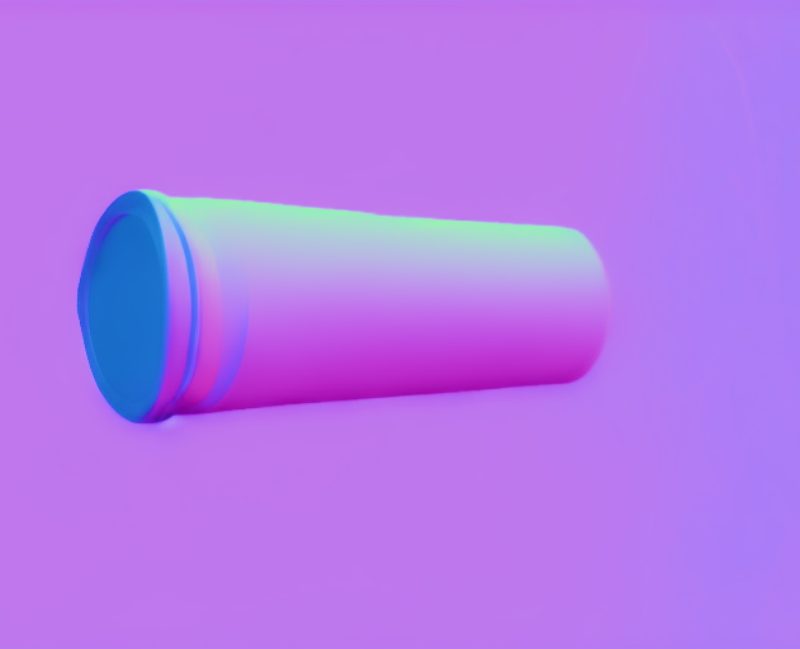} &
\includegraphics[width=\linewidth]{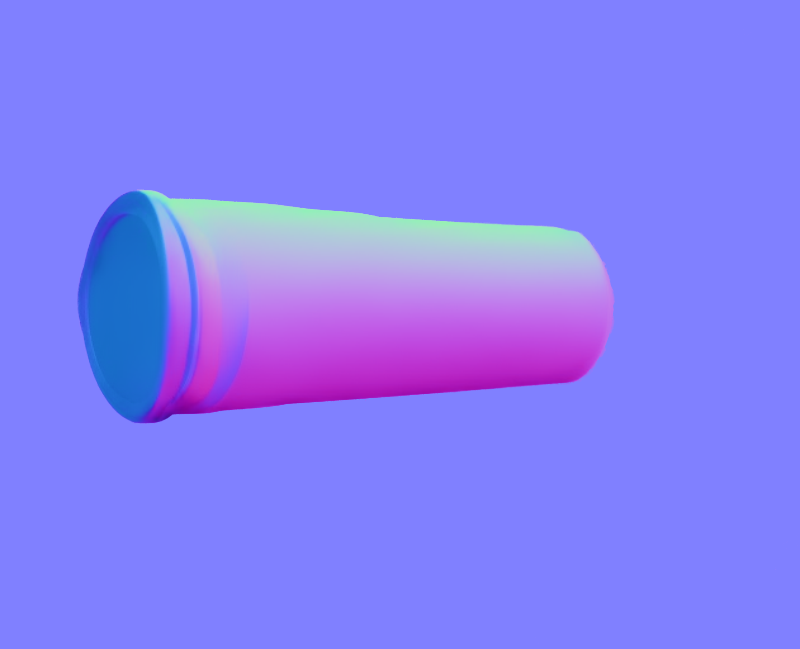} &
\includegraphics[width=\linewidth]{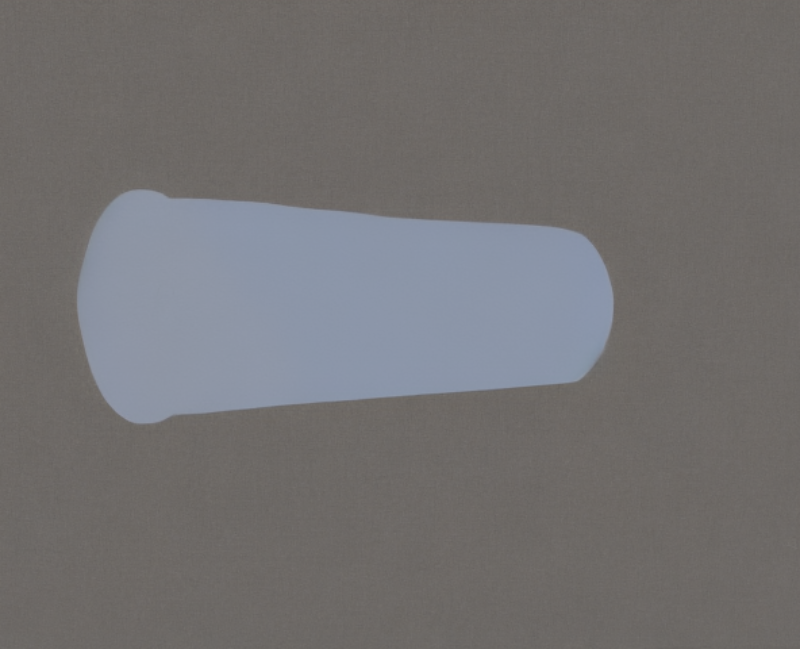} &
\includegraphics[width=\linewidth]{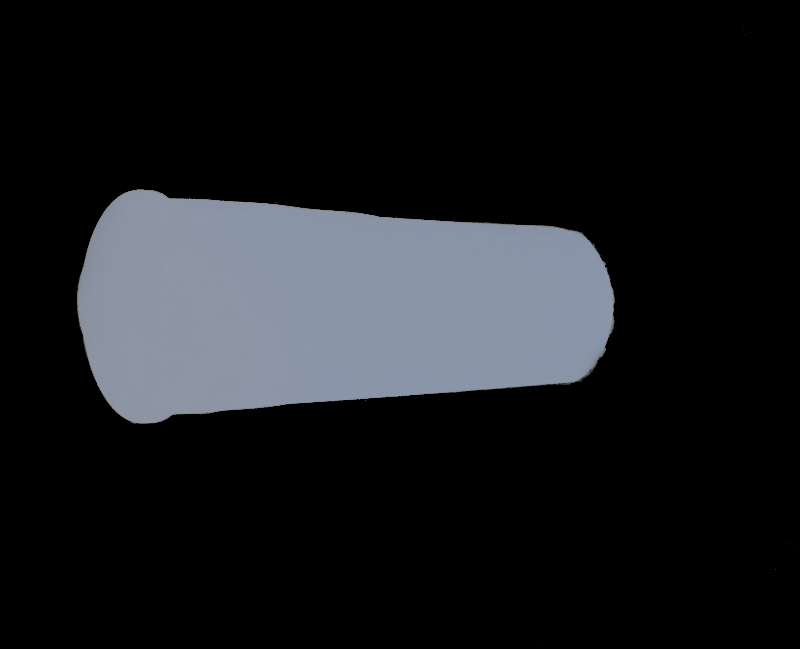} &
\includegraphics[width=\linewidth]{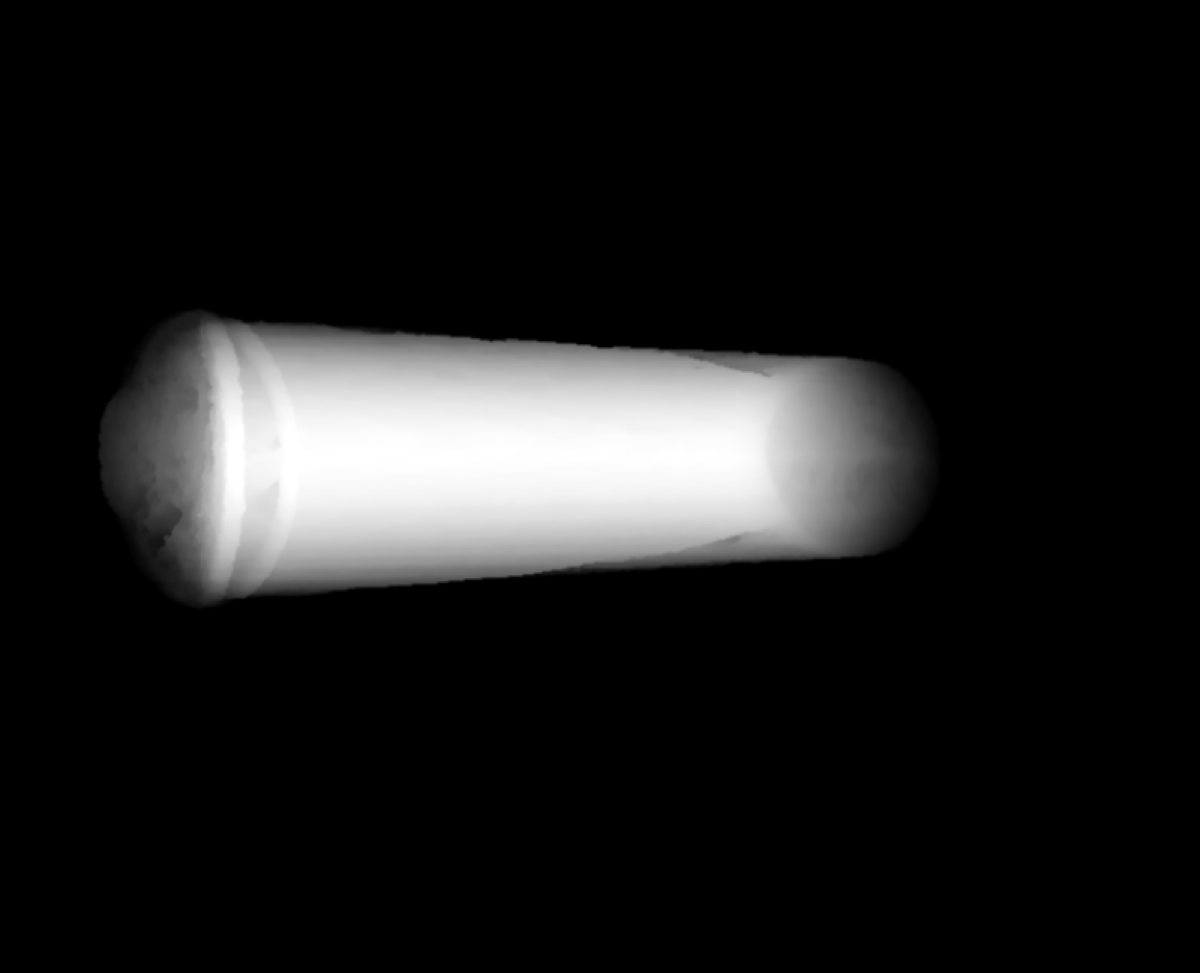}\\
 
\includegraphics[width=\linewidth]{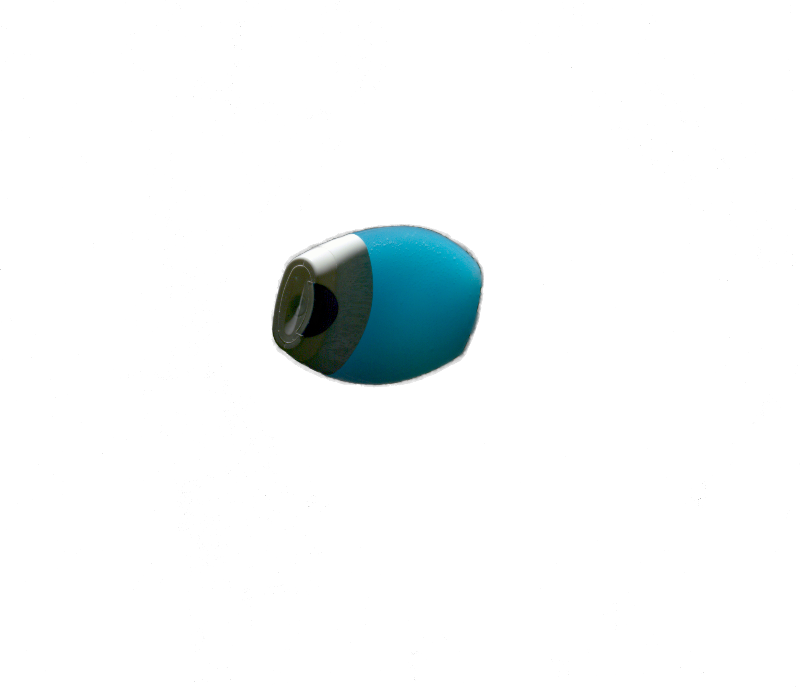} &
\includegraphics[width=\linewidth]{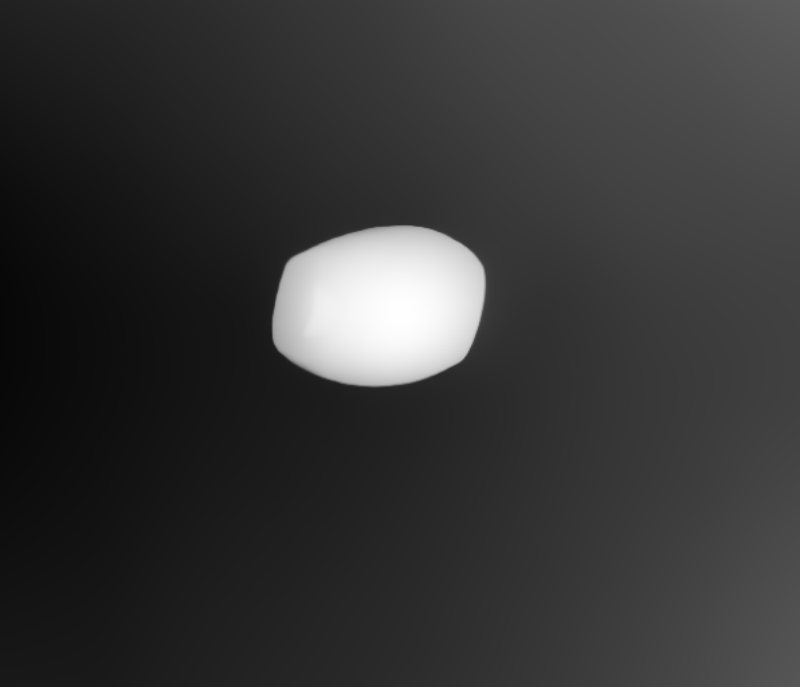} &
\includegraphics[width=\linewidth]{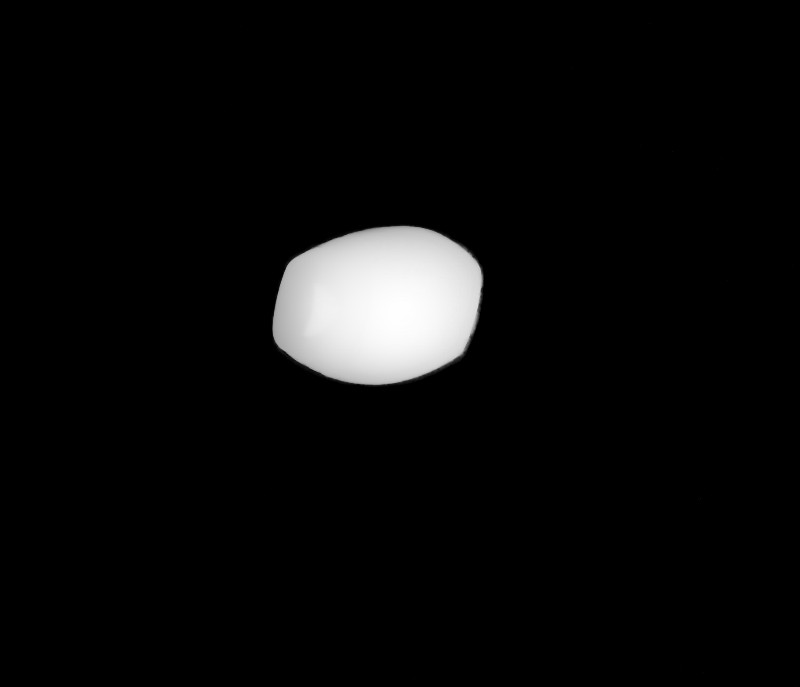} &
\includegraphics[width=\linewidth]{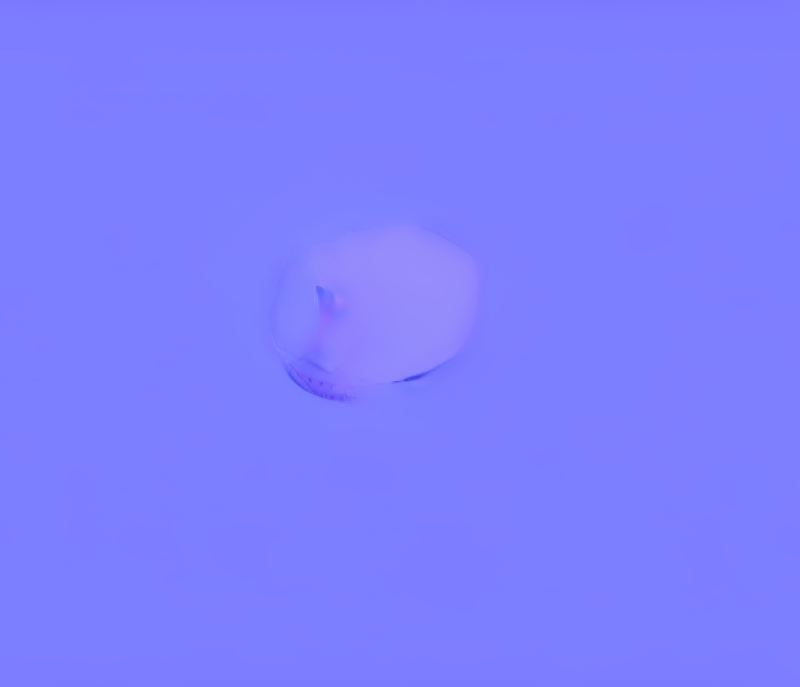} &
\includegraphics[width=\linewidth]{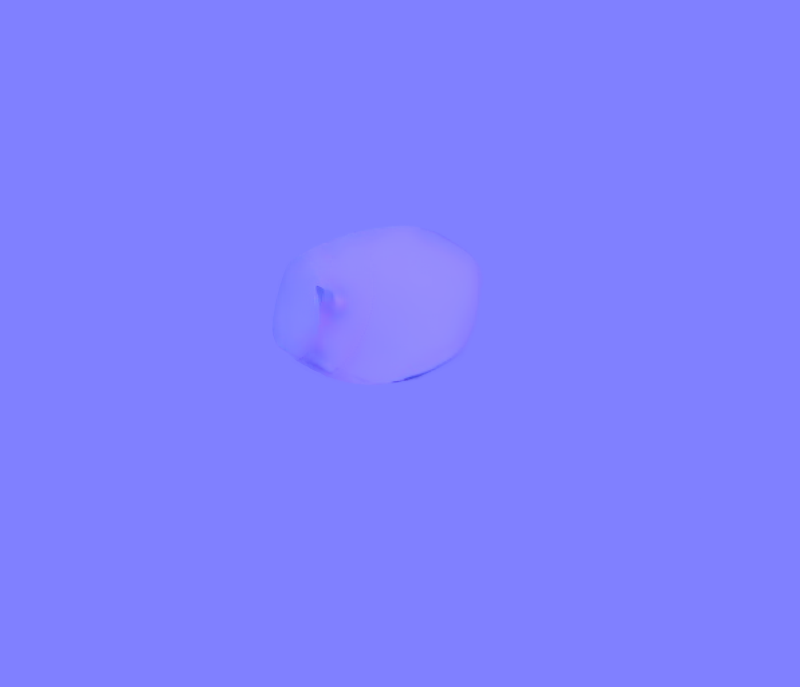} &
\includegraphics[width=\linewidth]{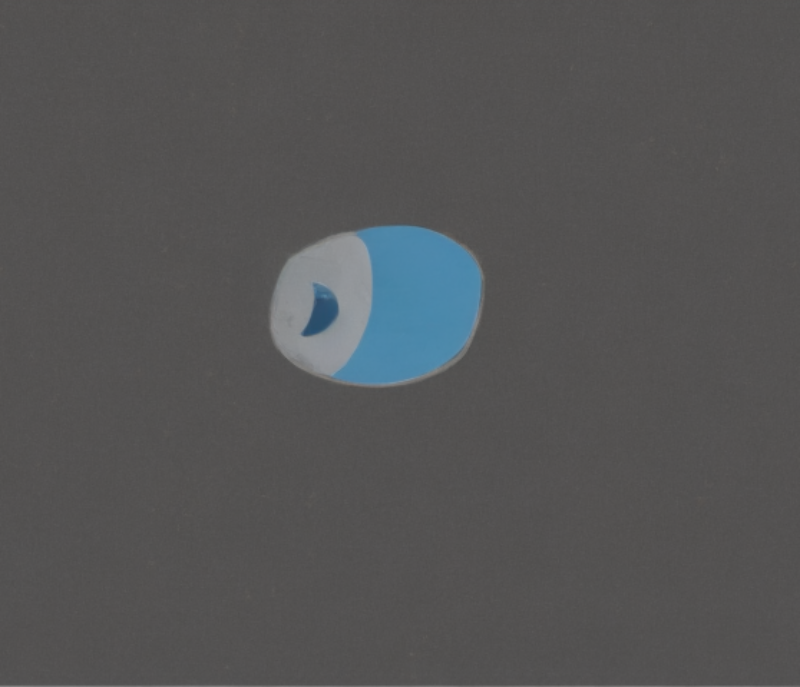} &
\includegraphics[width=\linewidth]{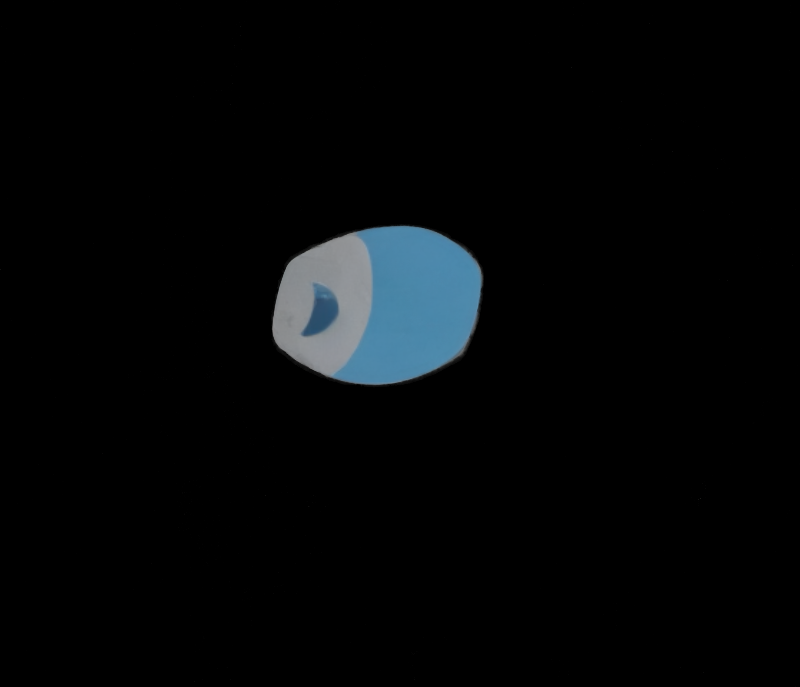} &
\includegraphics[width=\linewidth]{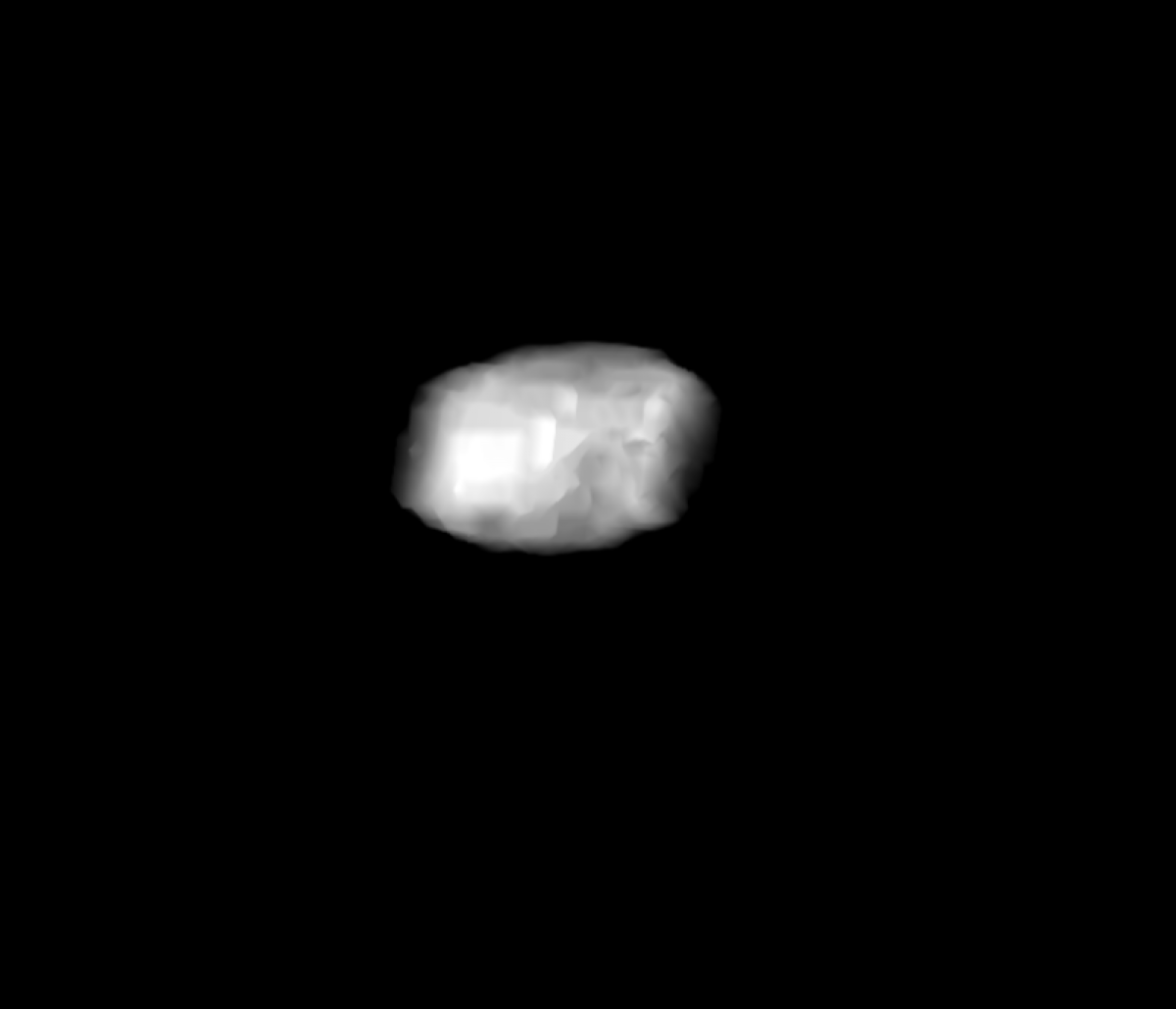}\\

\includegraphics[width=\linewidth]{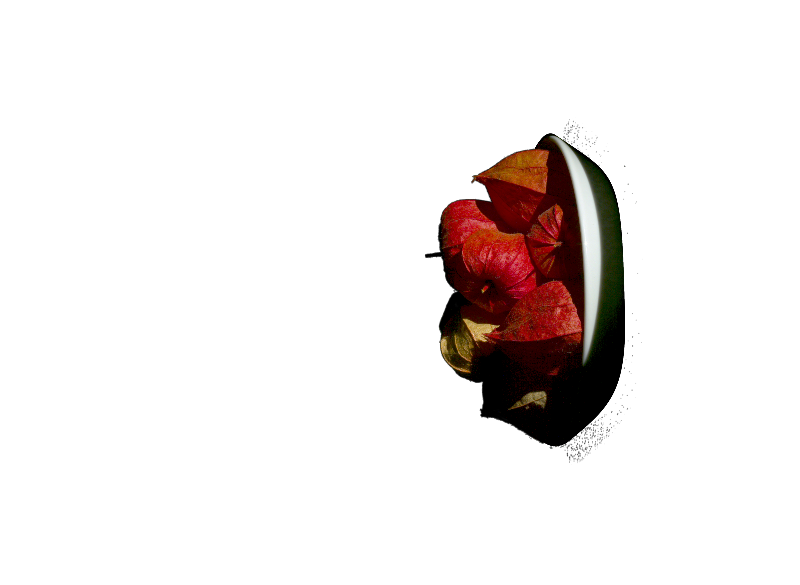} &
\includegraphics[width=\linewidth]{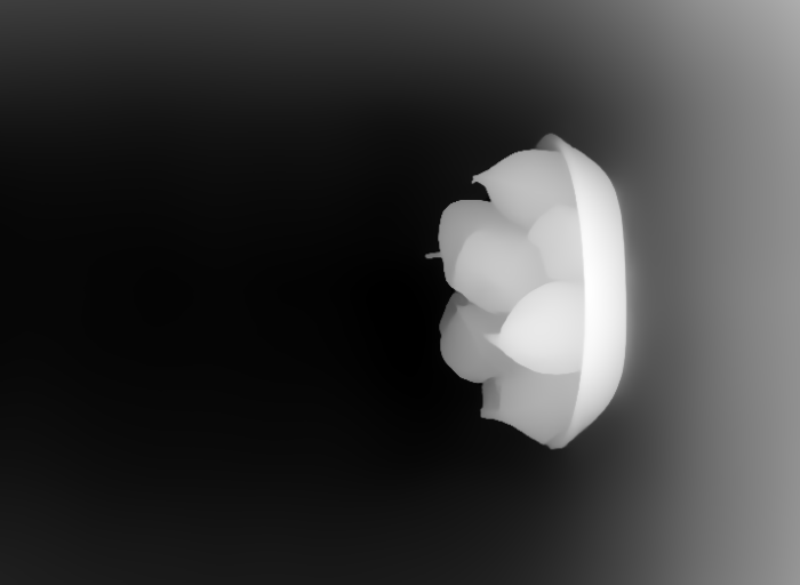} &
\includegraphics[width=\linewidth]{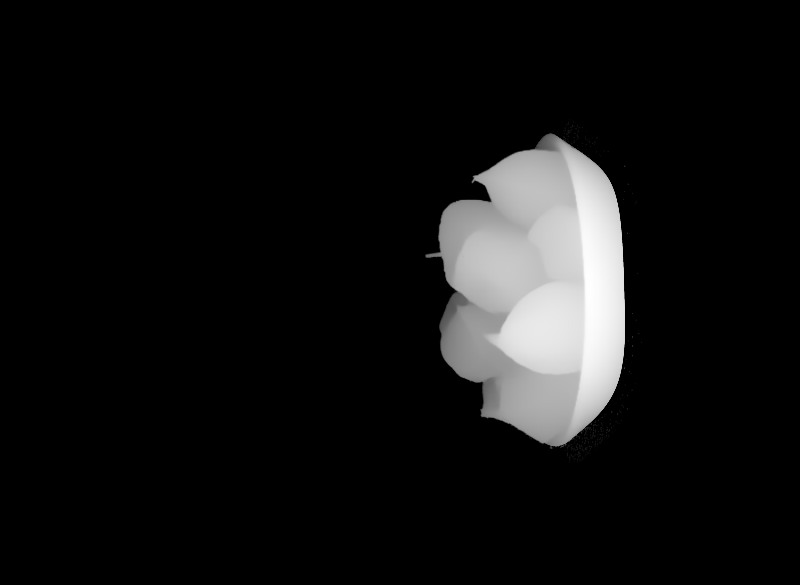} &
\includegraphics[width=\linewidth]{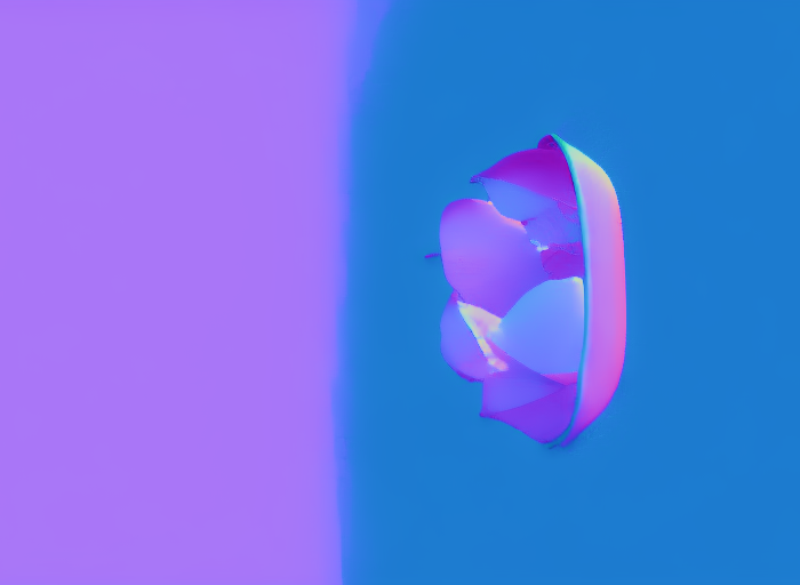} &
\includegraphics[width=\linewidth]{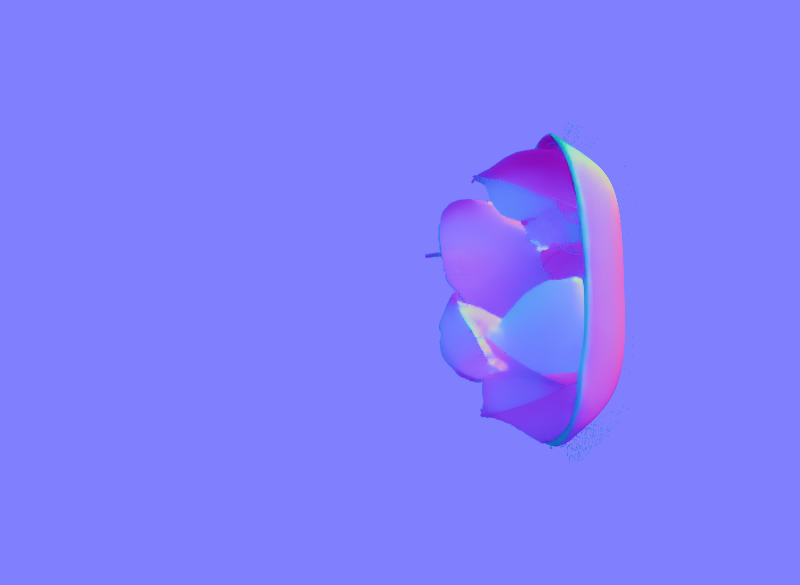} &
\includegraphics[width=\linewidth]{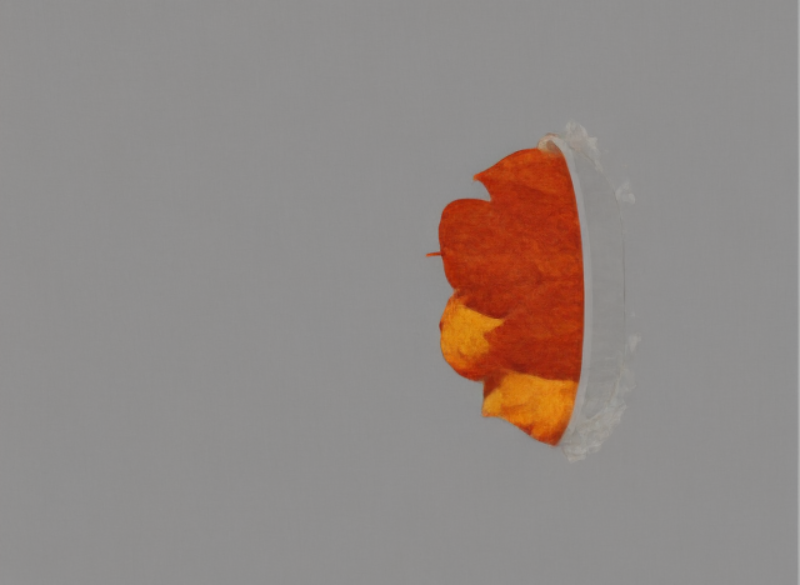} &
\includegraphics[width=\linewidth]{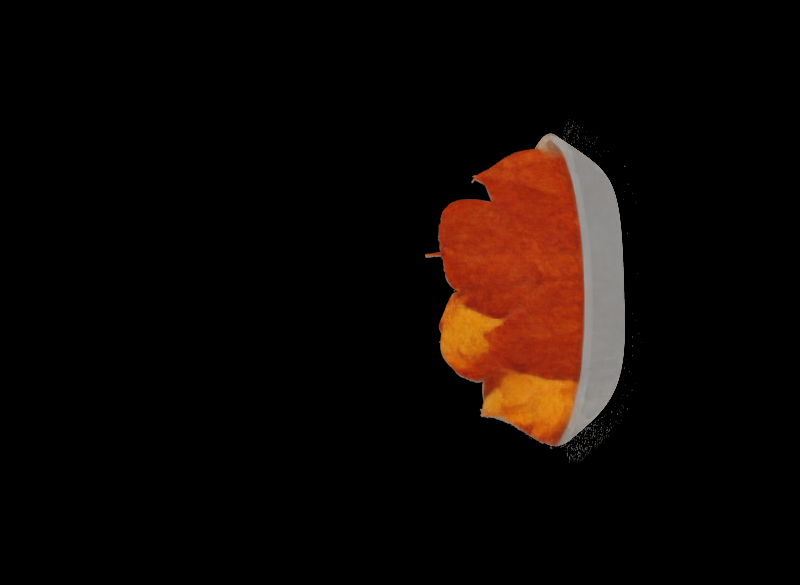} &
\includegraphics[width=\linewidth]{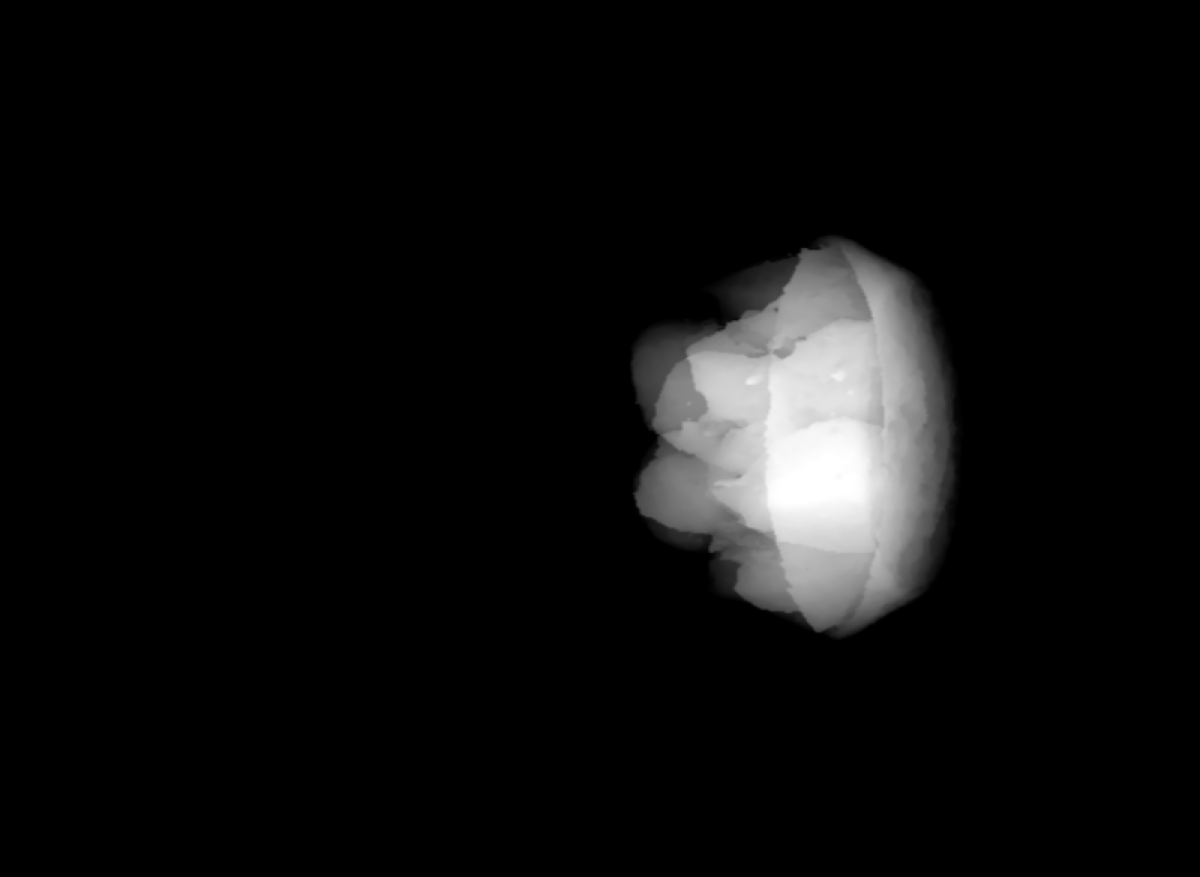}\\

\includegraphics[width=\linewidth]{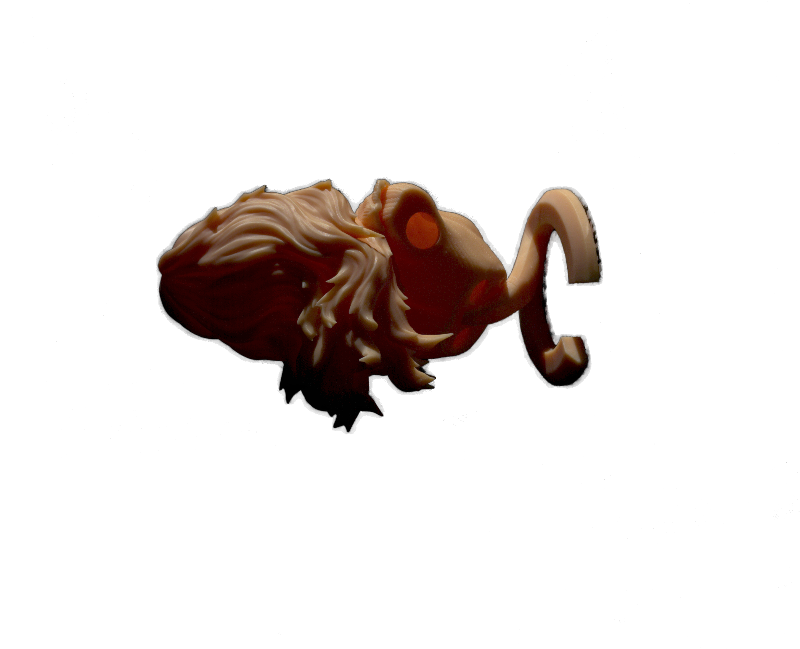} &
\includegraphics[width=\linewidth]{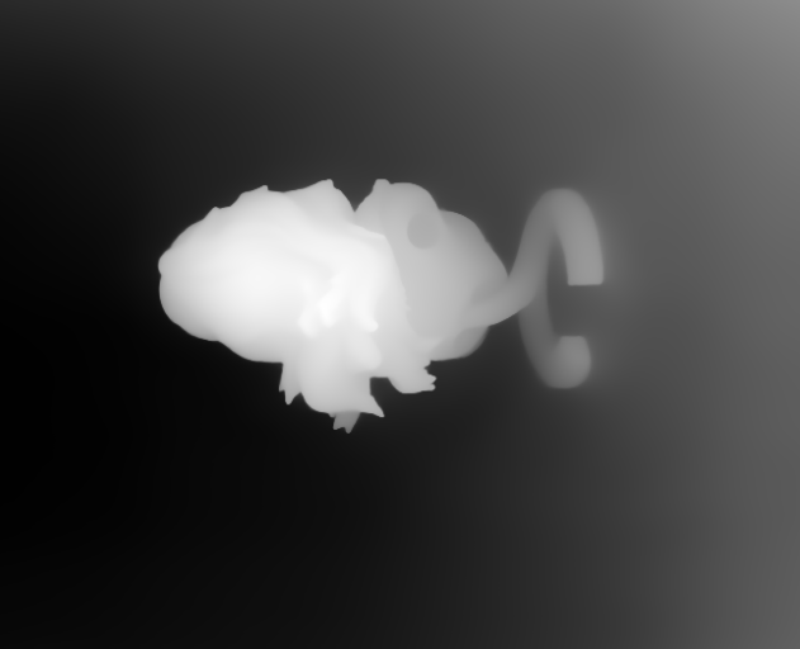} &
\includegraphics[width=\linewidth]{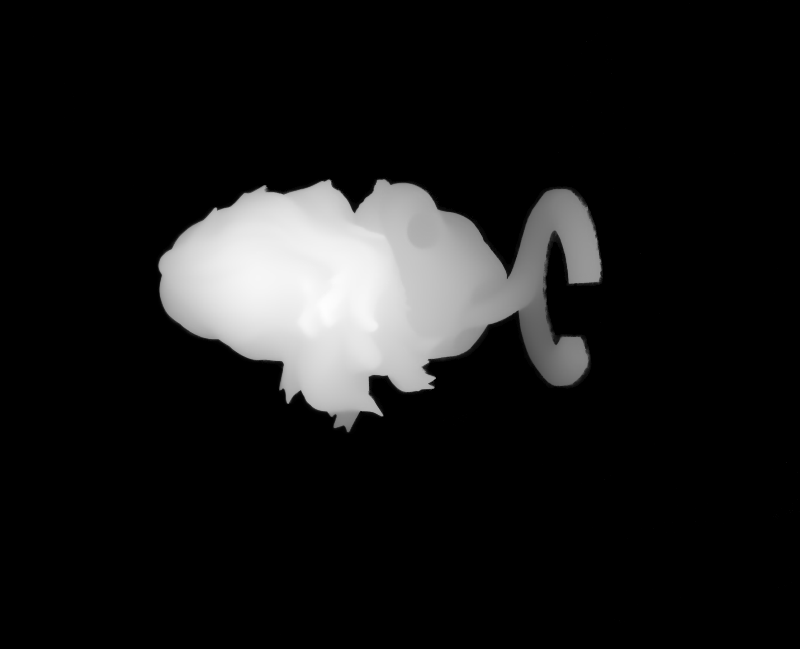} &
\includegraphics[width=\linewidth]{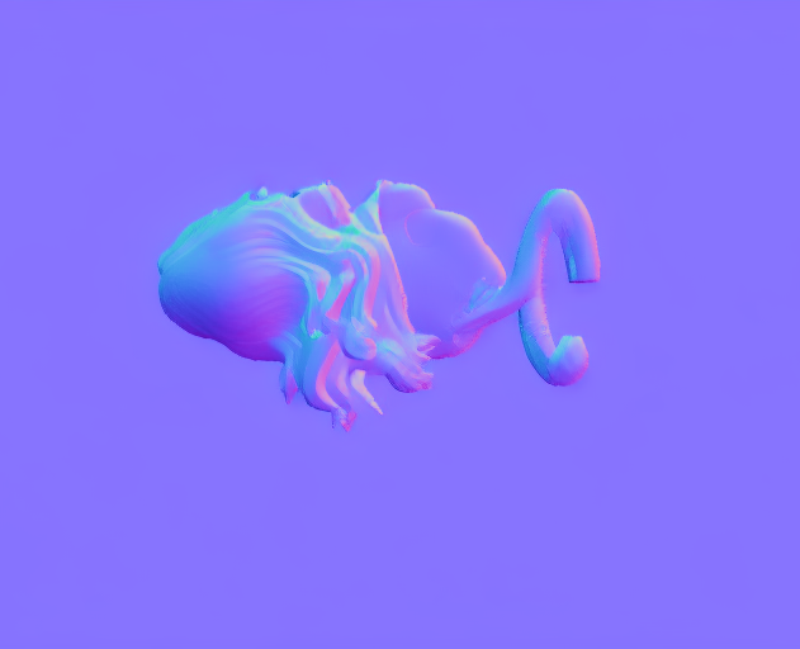} &
\includegraphics[width=\linewidth]{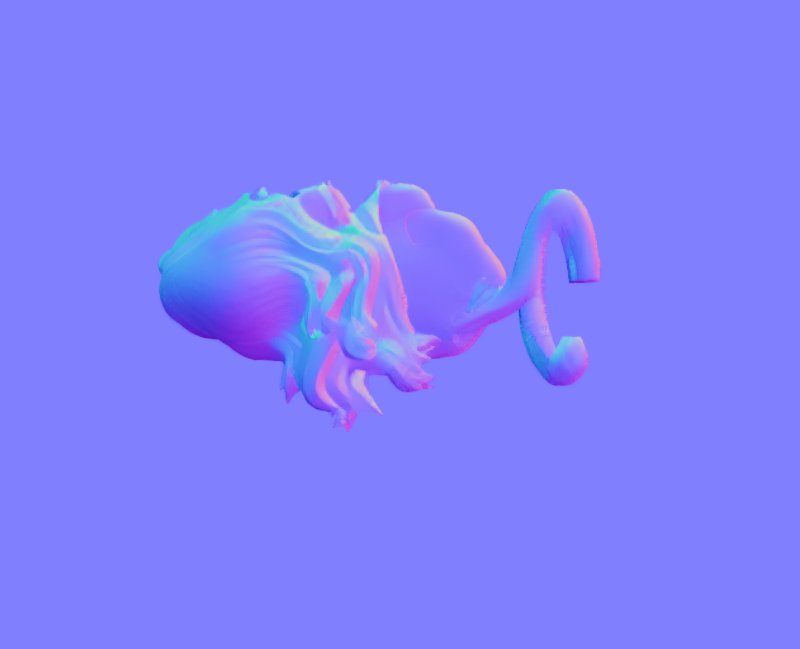} &
\includegraphics[width=\linewidth]{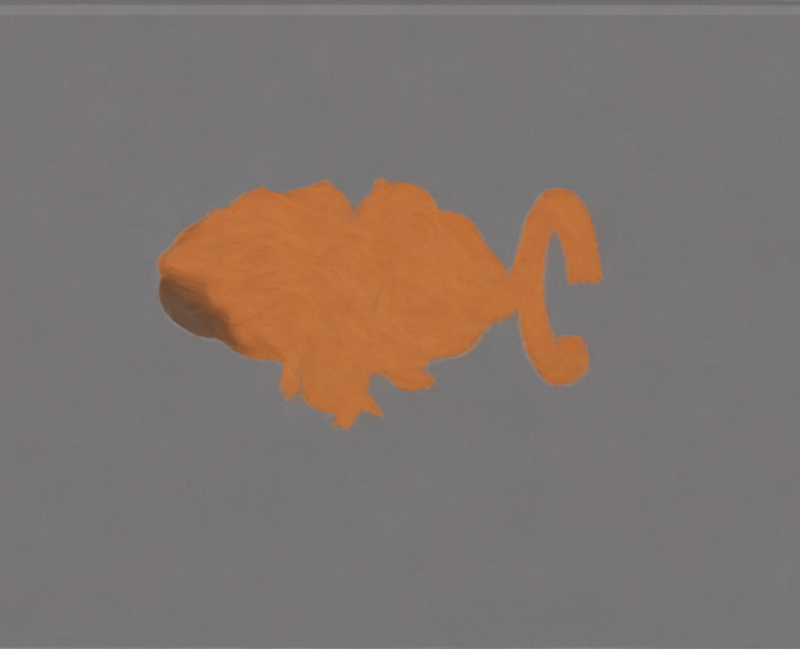} &
\includegraphics[width=\linewidth]{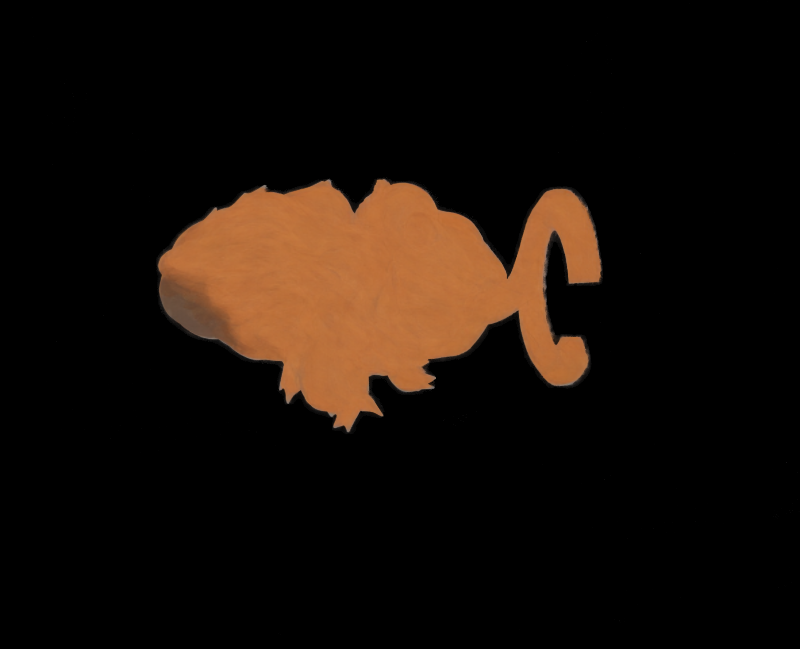} &
\includegraphics[width=\linewidth]{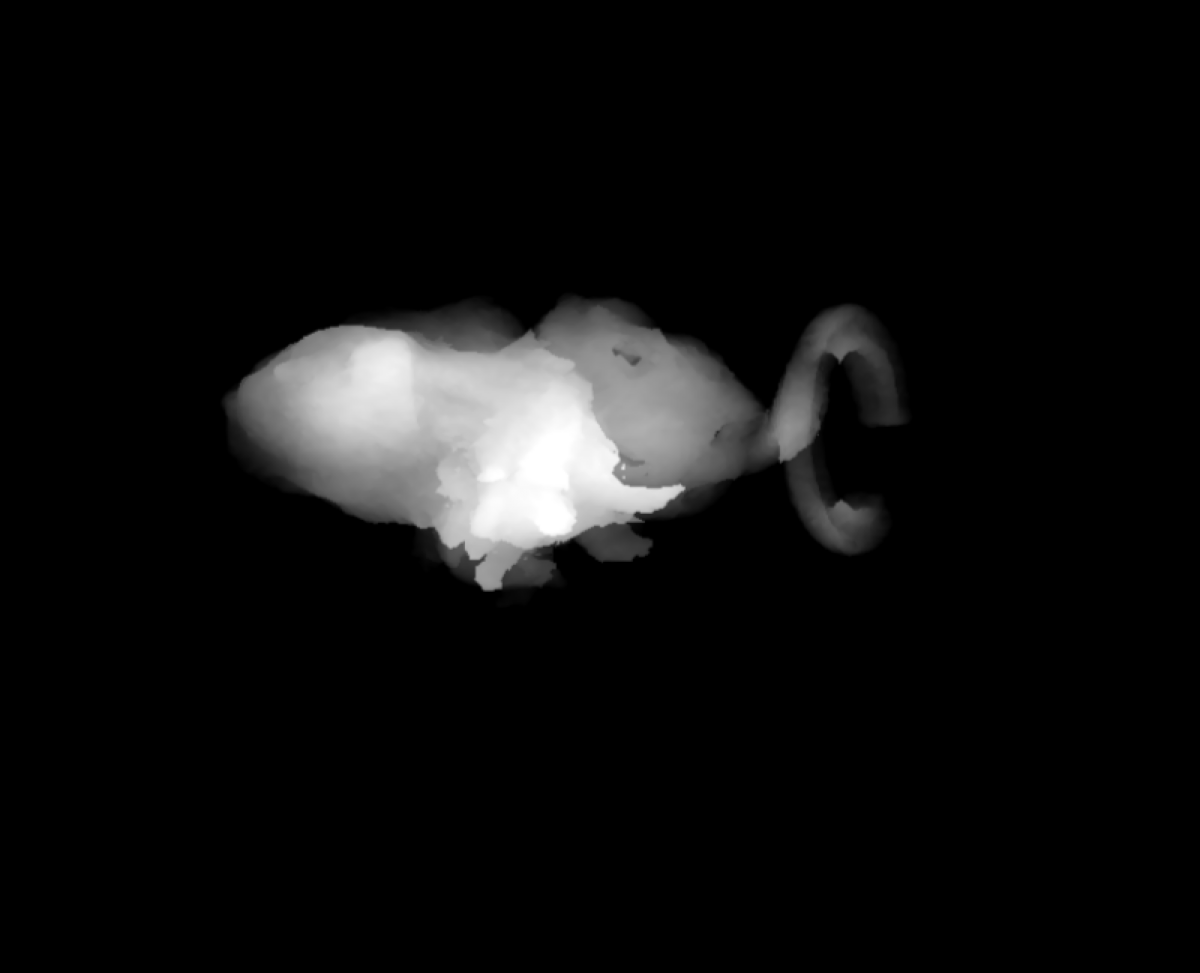} \\[4pt]

\parbox{\linewidth}{\centering\footnotesize input image} &
\parbox{\linewidth}{\centering\footnotesize depth w/o alpha} &
\parbox{\linewidth}{\centering\footnotesize depth w alpha} &
\parbox{\linewidth}{\centering\footnotesize normal w/o alpha} &
\parbox{\linewidth}{\centering\footnotesize normal w alpha} &
\parbox{\linewidth}{\centering\footnotesize albedo w/o alpha} &
\parbox{\linewidth}{\centering\footnotesize albedo w alpha} &
\parbox{\linewidth}{\centering\footnotesize thickness} \\

\end{tabular}
\caption{\textbf{Samples illustrating the compositing conventions for different conditions.} Each row shows the input RGB image and its corresponding map before and after applying the foreground mask.}
\label{fig:example_conditoins}
\end{figure*}

Depth and normals are resized and normalized to match the UNet resolution. SH features are injected into both the ControlNet branch and the denoiser’s timestep and cross-attention embeddings. The architecture itself is unchanged; only the training data and conditioning inputs are adapted to OLAT captures. In~\cref{fig:example_conditoins} depicts examples of the conditions used per object.

\paragraph{Training objective.}
We employ the $v$-parameterization objective combined with a mixture of
L1, SSIM, and LPIPS losses,
as well as cycle-consistency and blur-aware terms to encourage stable low-frequency light transport and reduce high-frequency artifacts.
Training samples follow:
\[
(I_{v,l_s},\, D_v,\, N_v,\, \Phi(l_{tgt}),\, I_{v,l_{tgt}}),
\]
where $I_{v,l_{tgt}}$ is the corresponding OLAT ground truth.  
The implementation follows \cite{poirier2024radiancefield}, with modifications only to the data and conditioning.

\paragraph{Use within DIAMOND-SSS.}
During reconstruction, the relighting diffusion module produces synthetic OLAT variants for each view.
As with multi-view diffusion outputs, synthetic photometric losses are down-weighted by $\alpha$, but these augmented illuminations still contribute meaningfully to multi-view silhouette and depth consistency, improving supervision across illumination conditions.

\subsection{Setup: Use with Real, Relighted, and Fully Synthetic Data}
\paragraph{Real OLAT captures.}
We use native OLAT images with provided $(K_v, T_v)$ without modification.

\paragraph{Relighting augmentation.}
For settings such as “all views, 1 light per view,” missing illuminations are synthesized using the relighting diffusion model (\cref{sec:supp_relit}). The camera parameters $(K_v, T_v)$ remain unchanged.

\paragraph{Fully synthetic multi-view sets.}
When using multi-view diffusion + relighting, we generate:
\begin{itemize}[leftmargin=*]
    \item novel poses $P_v$ from the Free3D-based NVS model (\cref{sec:supp_nvs}),
    \item relighted OLAT equivalents under all target directions.
\end{itemize}
Generated data is aligned with the original OLAT lattice (same lights, comparable poses).

\subsection{Benchmarking and ablation studies}
\label{sec:supp_diffusion_ablation}

\begin{figure*}[t]
    \centering
    \small
    \setlength{\tabcolsep}{2pt}
    \renewcommand{\arraystretch}{0}
    \newcolumntype{s}[1]{>{\scriptsize\raggedright\arraybackslash}p{#1}}
    \begin{tabular}{s{0.1\textwidth} p{0.17\textwidth} *{5}{p{0.17\textwidth}}}
        & \centering\textbf{Condition} &
        \multicolumn{4}{c}{\textbf{Generated Views}} \\[4pt]

        \textbf{Free3D (ours, fine-tuned)~\cite{zheng24}} &
        \centering\includegraphics[width=\linewidth]{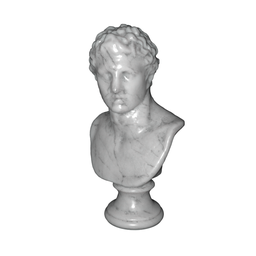} &
        \includegraphics[width=\linewidth]{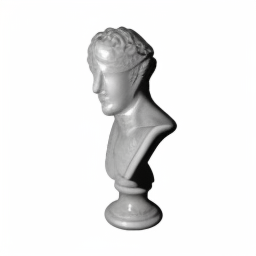} &
        \includegraphics[width=\linewidth]{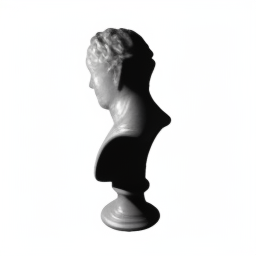} &
        \includegraphics[width=\linewidth]{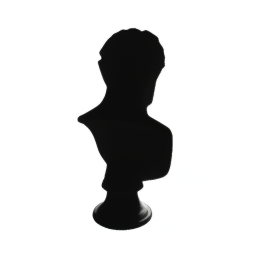} &
        \includegraphics[width=\linewidth]{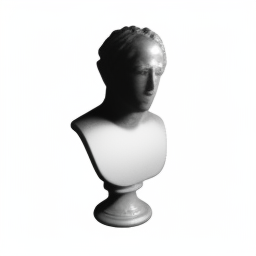} \\[6pt]
        \textbf{Free3D (off-the-shelf)~\cite{zheng24}} &
        \centering\includegraphics[width=\linewidth]{assets/experiments/nvs/ours/statue/r_0_l_45_src.png} &
        \includegraphics[width=\linewidth]{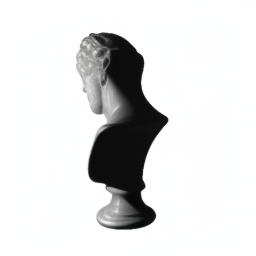} &
        \includegraphics[width=\linewidth]{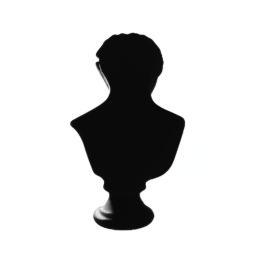} &
        \includegraphics[width=\linewidth]{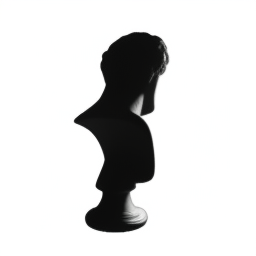} &
        \includegraphics[width=\linewidth]{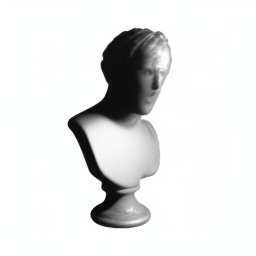} \\ [6pt]

        \textbf{Zero123~\cite{liu2023zero}} &
        \centering\includegraphics[width=\linewidth]{assets/experiments/nvs/ours/statue/r_0_l_45_src.png} &
        \includegraphics[width=\linewidth]{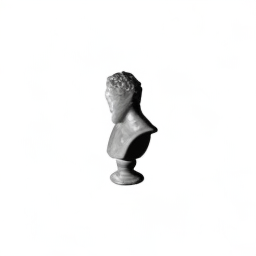} &
        \includegraphics[width=\linewidth]{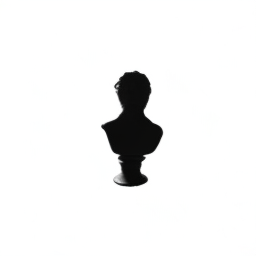} &
        \includegraphics[width=\linewidth]{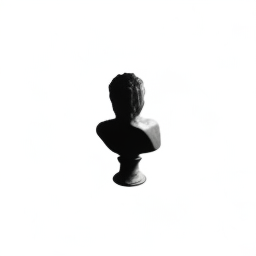} &
        \includegraphics[width=\linewidth]{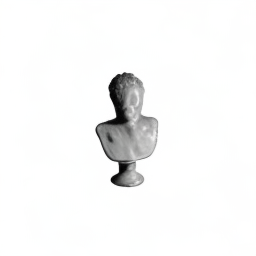} \\[6pt]

        \textbf{Zero123-XL~\cite{shi2023zero123plus}} &
        \centering\includegraphics[width=\linewidth]{assets/experiments/nvs/ours/statue/r_0_l_45_src.png} &
        \includegraphics[width=\linewidth]{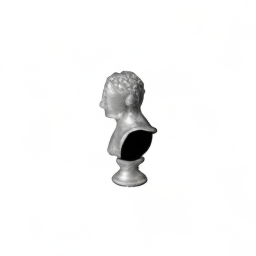} &
        \includegraphics[width=\linewidth]{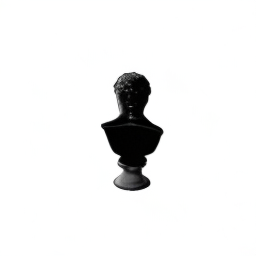} &
        \includegraphics[width=\linewidth]{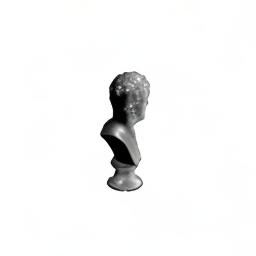} &
        \includegraphics[width=\linewidth]{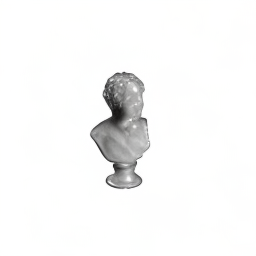} \\[6pt]

        \textbf{MVD-Fusion~\cite{mvdfusion}} &
        \centering\includegraphics[width=\linewidth]{assets/experiments/nvs/ours/statue/r_0_l_45_src.png} &
        \includegraphics[width=\linewidth]{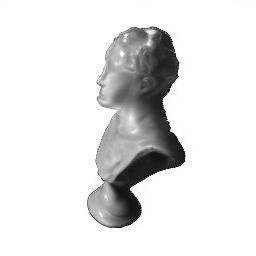} &
        \includegraphics[width=\linewidth]{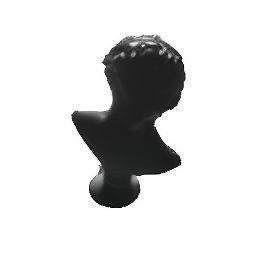} &
        \includegraphics[width=\linewidth]{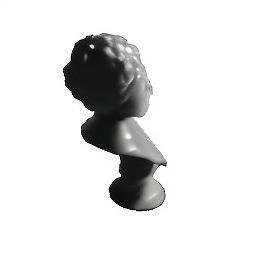} &
        \includegraphics[width=\linewidth]{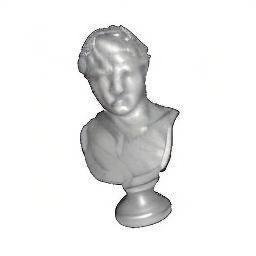} \\[6pt]

    \end{tabular}
    \caption{\textbf{Multi-view diffusion: comparison on a synthetic ``statue'' object.}
    Each row shows novel views generated from a single conditioning image (first column) by different NVS models.
    Our Free3D fine-tuning better preserves object identity and silhouette coherence than off-the-shelf Free3D and improves geometric consistency over other baselines.}
    \label{supp:fig:nvs_statue}
\end{figure*}

\begin{figure*}[t]
    \centering
    \small
    \setlength{\tabcolsep}{2pt}
    \renewcommand{\arraystretch}{0}
    \newcolumntype{s}[1]{> {\scriptsize\raggedright\arraybackslash}p{#1}}
    \begin{tabular}{s{0.1\textwidth} p{0.17\textwidth} *{5}{p{0.17\textwidth}}}
        & \centering\textbf{Condition} &
        \multicolumn{4}{c}{\textbf{Generated Views}} \\[4pt]

        \textbf{Free3D (ours, fine-tuned)~\cite{zheng24}} &
        \centering\includegraphics[width=\linewidth]{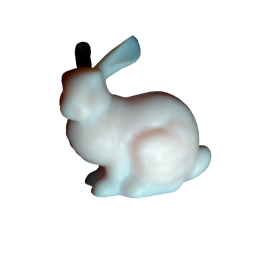} &
        \includegraphics[width=\linewidth]{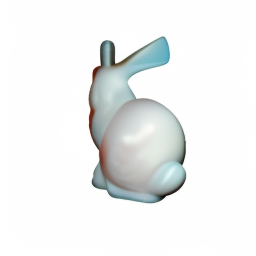} &
        \includegraphics[width=\linewidth]{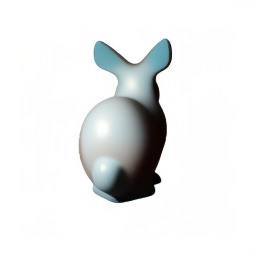} &
        \includegraphics[width=\linewidth]{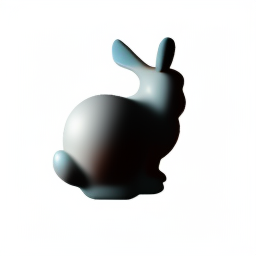} &
        \includegraphics[width=\linewidth]{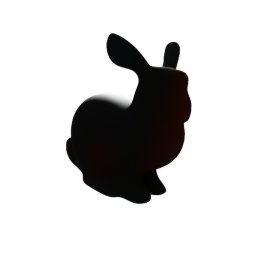} \\[6pt]

        \textbf{Free3D (off-the-shelf)~\cite{zheng24}} &
        \centering\includegraphics[width=\linewidth]{assets/experiments/nvs/condition/r_12_l_53.png} &
        \includegraphics[width=\linewidth]{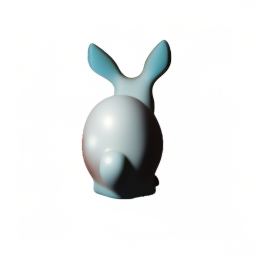} &
        \includegraphics[width=\linewidth]{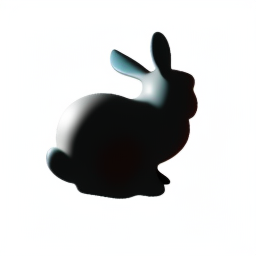} &
        \includegraphics[width=\linewidth]{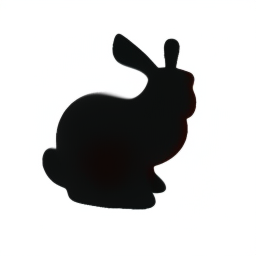} &
        \includegraphics[width=\linewidth]{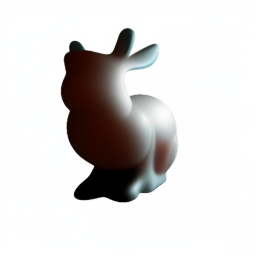} \\

        \textbf{Zero123~\cite{liu2023zero}} &
        \centering\includegraphics[width=\linewidth]{assets/experiments/nvs/condition/r_12_l_53.png} &
        \includegraphics[width=\linewidth]{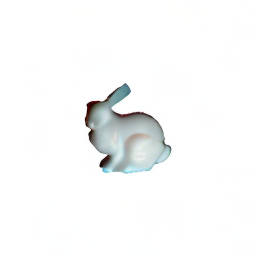} &
        \includegraphics[width=\linewidth]{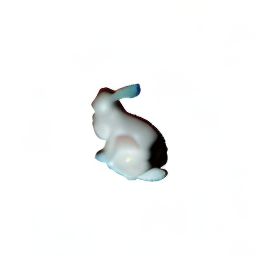} &
        \includegraphics[width=\linewidth]{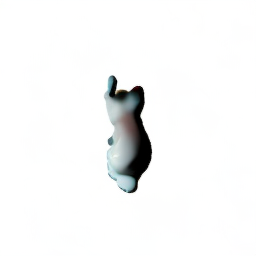} &
        \includegraphics[width=\linewidth]{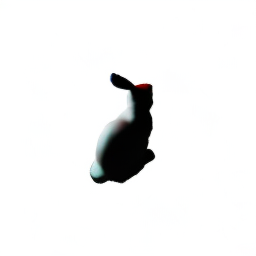} \\[6pt]

        \textbf{Zero123-XL~\cite{shi2023zero123plus}} &
        \centering\includegraphics[width=\linewidth]{assets/experiments/nvs/condition/r_12_l_53.png} &
        \includegraphics[width=\linewidth]{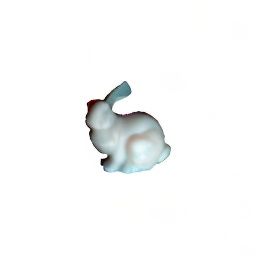} &
        \includegraphics[width=\linewidth]{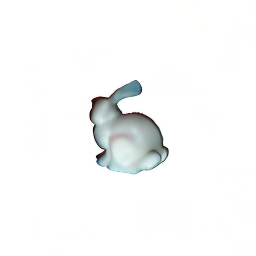} &
        \includegraphics[width=\linewidth]{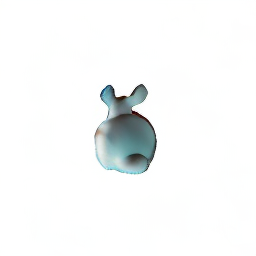} &
        \includegraphics[width=\linewidth]{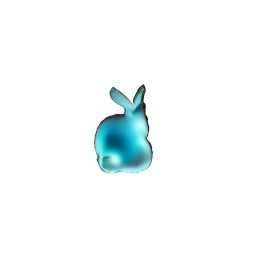} \\[6pt]
        
        \textbf{MVD-Fusion~\cite{mvdfusion}} &
        \centering\includegraphics[width=\linewidth]{assets/experiments/nvs/condition/r_12_l_53.png} &
        \includegraphics[width=\linewidth]{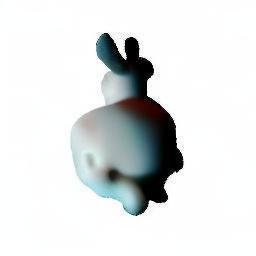} &
        \includegraphics[width=\linewidth]{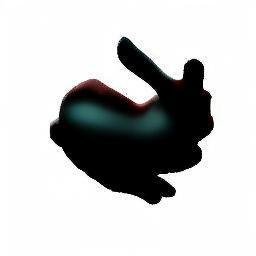} &
        \includegraphics[width=\linewidth]{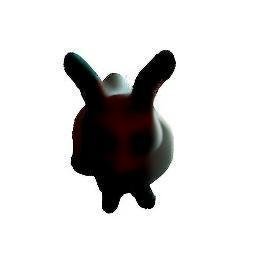} &
        \includegraphics[width=\linewidth]{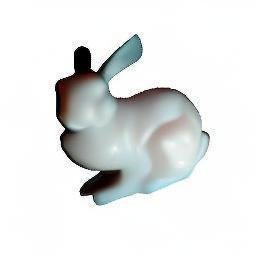} \\[6pt]

    \end{tabular}
    \caption{\textbf{Multi-view diffusion on a real ``bunny'' object.}
    Comparison between our fine-tuned Free3D model and several off-the-shelf NVS baselines.
    Fine-tuning improves viewpoint coherence and reduces geometric distortions while preserving appearance across views.}
    \label{supp:fig:nvs_bunny}
\end{figure*}

\begin{figure*}[t]
    \centering
    \small
    \setlength{\tabcolsep}{2pt}
    \renewcommand{\arraystretch}{0}
    \newcolumntype{s}[1]{> {\scriptsize\raggedright\arraybackslash}p{#1}}
    \begin{tabular}{s{0.1\textwidth} p{0.17\textwidth} *{5}{p{0.17\textwidth}}}
        & \centering\textbf{Condition} &
        \multicolumn{4}{c}{\textbf{Generated Views}} \\[4pt]

        \textbf{Free3D (ours, fine-tuned)~\cite{zheng24}} &
        \centering\includegraphics[width=\linewidth]{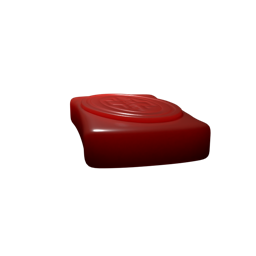} &
        \includegraphics[width=\linewidth]{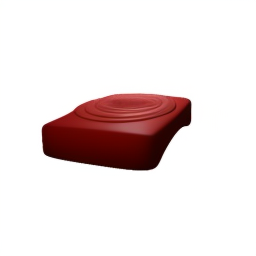} &
        \includegraphics[width=\linewidth]{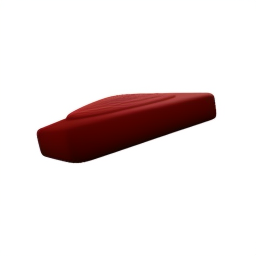} &
        \includegraphics[width=\linewidth]{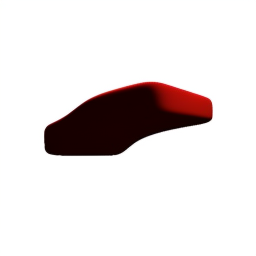} &
        \includegraphics[width=\linewidth]{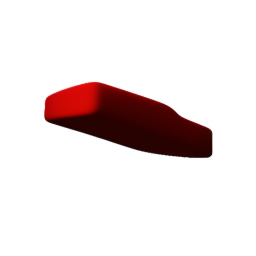} \\[6pt]

        \textbf{Free3D (off-the-shelf)~\cite{zheng24}} &
        \centering\includegraphics[width=\linewidth]{assets/experiments/nvs/condition/soap_cond.png} &
        \includegraphics[width=\linewidth]{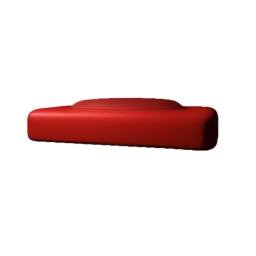} &
        \includegraphics[width=\linewidth]{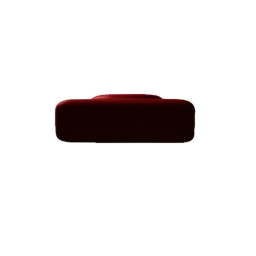} &
        \includegraphics[width=\linewidth]{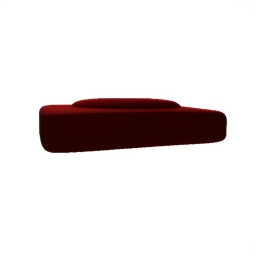} &
        \includegraphics[width=\linewidth]{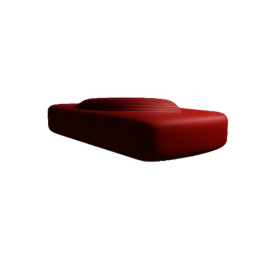} \\

        \textbf{Zero123~\cite{liu2023zero}} &
        \centering\includegraphics[width=\linewidth]{assets/experiments/nvs/condition/soap_cond.png} &
        \includegraphics[width=\linewidth]{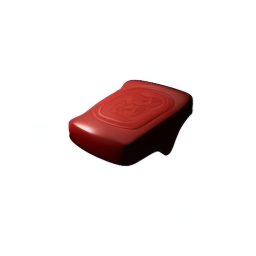} &
        \includegraphics[width=\linewidth]{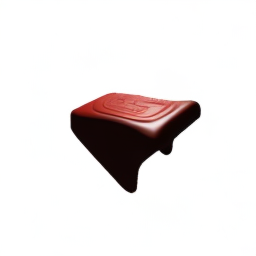} &
        \includegraphics[width=\linewidth]{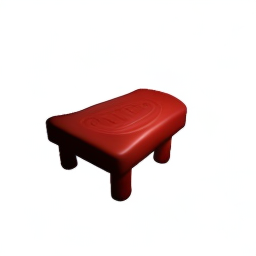} &
        \includegraphics[width=\linewidth]{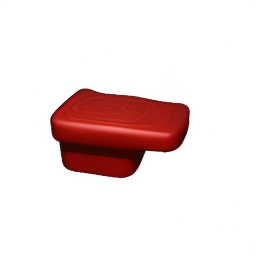} \\[6pt]

        \textbf{Zero123-XL~\cite{shi2023zero123plus}} &
        \centering\includegraphics[width=\linewidth]{assets/experiments/nvs/condition/soap_cond.png} &
        \includegraphics[width=\linewidth]{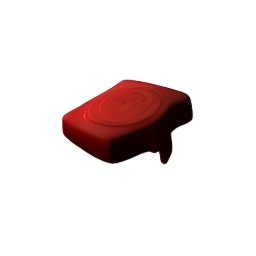} &
        \includegraphics[width=\linewidth]{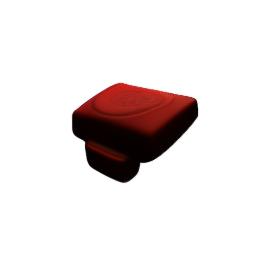} &
        \includegraphics[width=\linewidth]{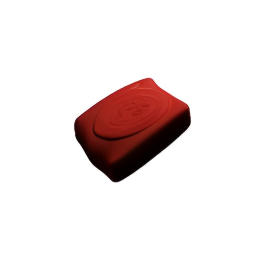} &
        \includegraphics[width=\linewidth]{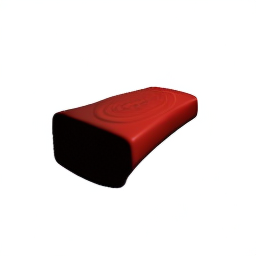} \\[6pt]

        \textbf{MVD-Fusion~\cite{mvdfusion}} &
        \centering\includegraphics[width=\linewidth]{assets/experiments/nvs/condition/soap_cond.png} &
        \includegraphics[width=\linewidth]{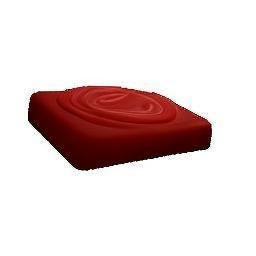} &
        \includegraphics[width=\linewidth]{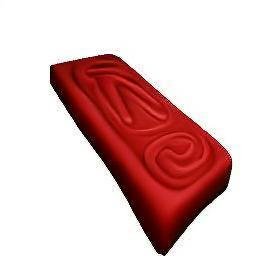} &
        \includegraphics[width=\linewidth]{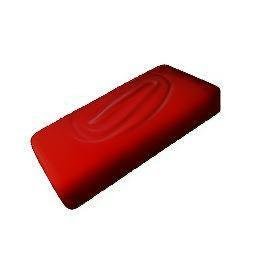} &
        \includegraphics[width=\linewidth]{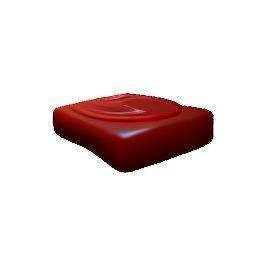} \\[6pt]

    \end{tabular}

    \caption{\textbf{Multi-view diffusion on a translucent ``soap'' object.}
    Our fine-tuned Free3D model better respects the global shape and translucency cues across viewpoints than off-the-shelf Free3D and other NVS baselines, while remaining suitable as an augmentation prior for DIAMOND-SSS.}
    \label{supp:fig:nvs_soap}
\end{figure*}

\begin{figure*}[t]
\centering
\setlength{\tabcolsep}{2pt}
\renewcommand{\arraystretch}{0}
\resizebox{\textwidth}{!}{%
\begin{tabular}{*{6}{p{.19\textwidth}}}
\includegraphics[width=\linewidth]{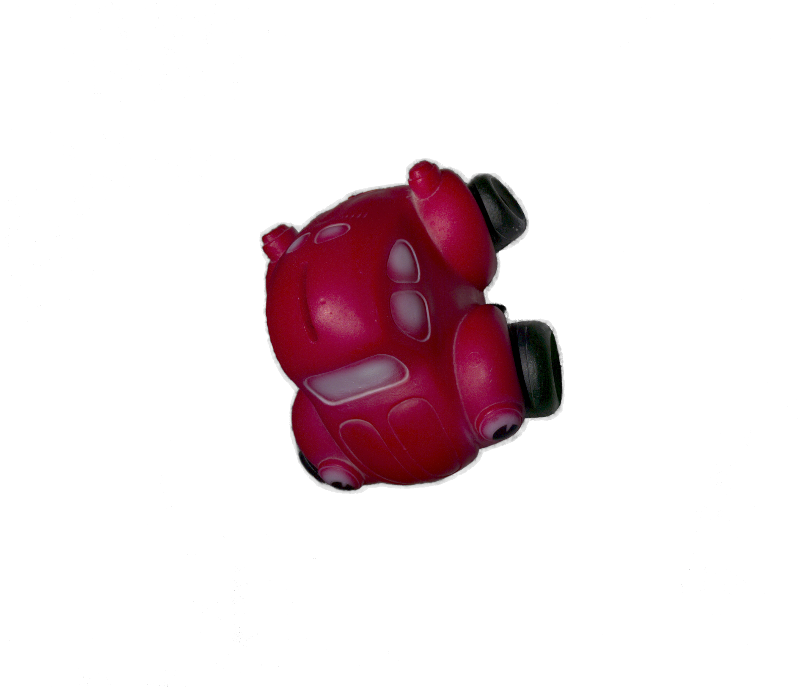} &
\includegraphics[width=\linewidth]{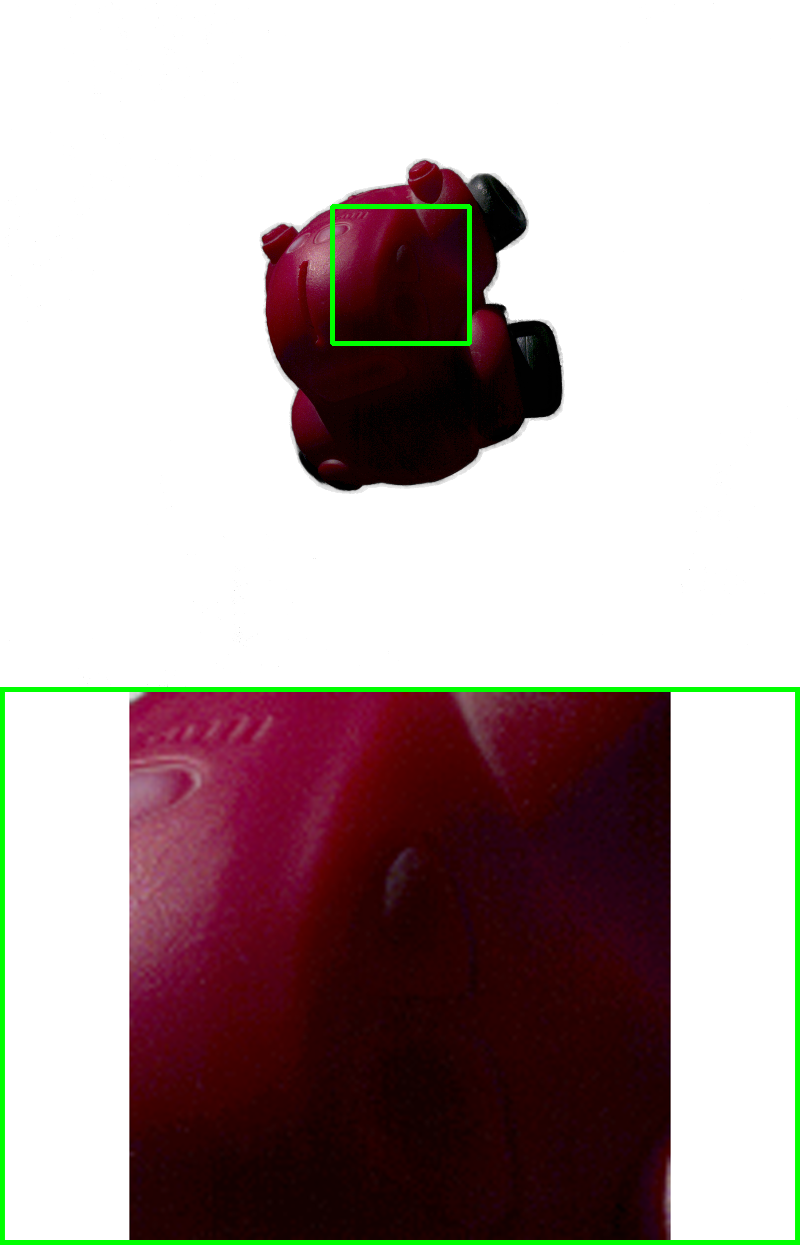} &
\includegraphics[width=\linewidth]{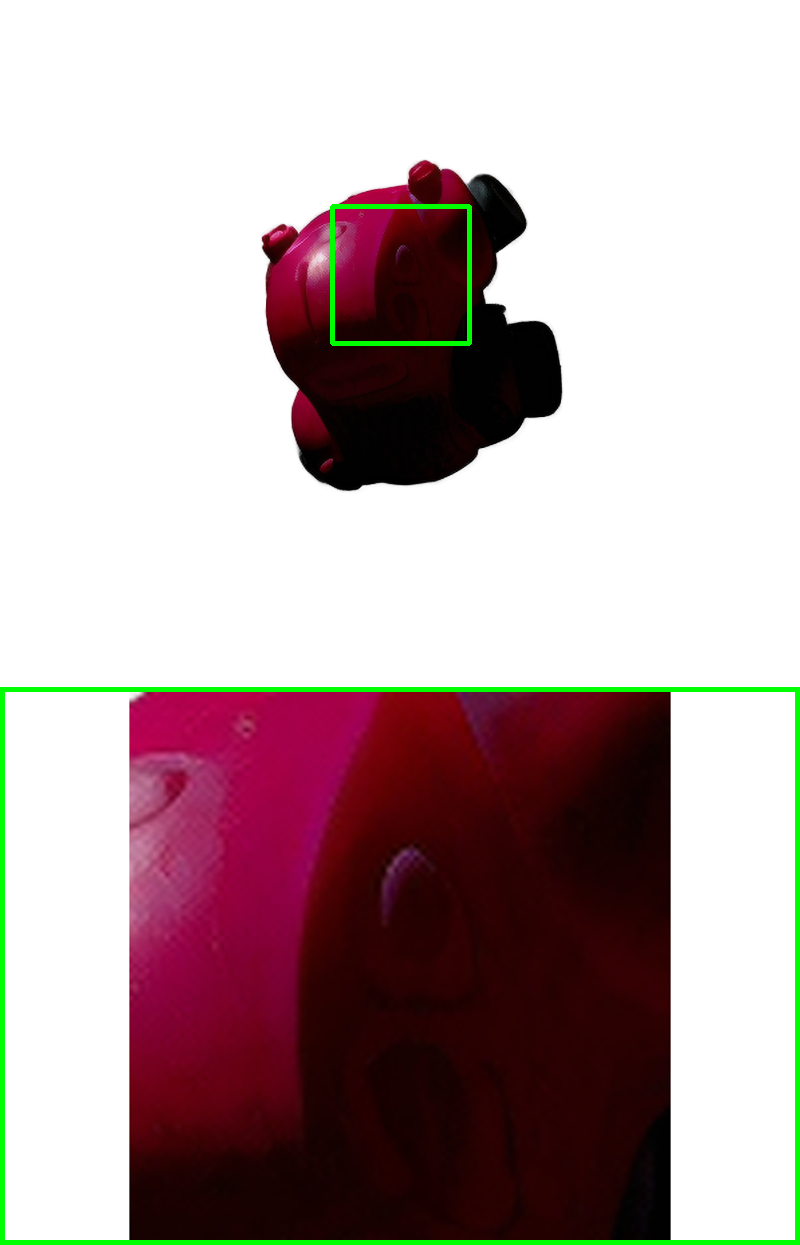} &
\includegraphics[width=0.68\linewidth]{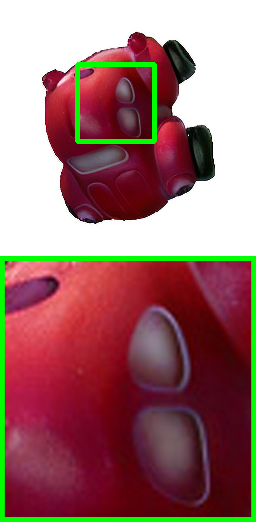} &
\includegraphics[width=.68\linewidth]{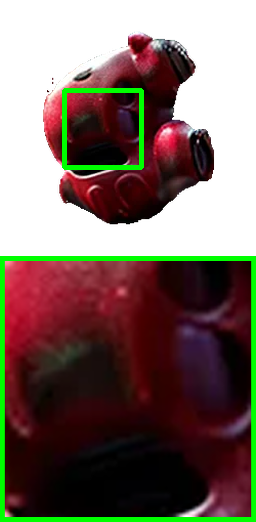} &
\includegraphics[width=.7\linewidth]{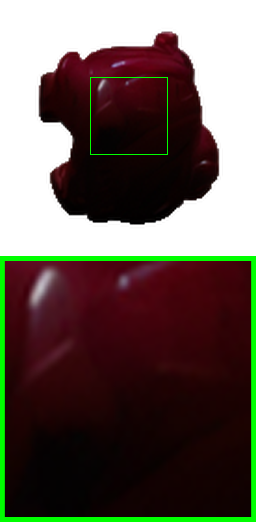} \\[4pt]

\includegraphics[width=\linewidth]{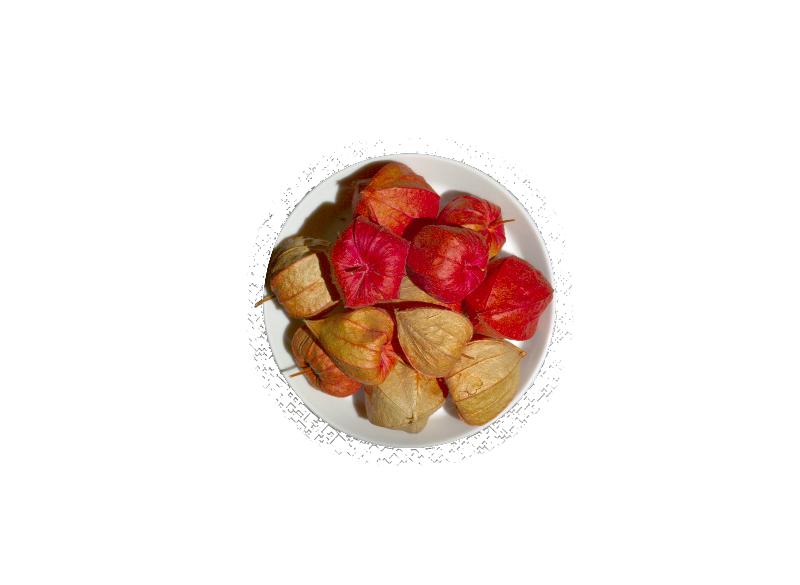} &
\includegraphics[width=\linewidth]{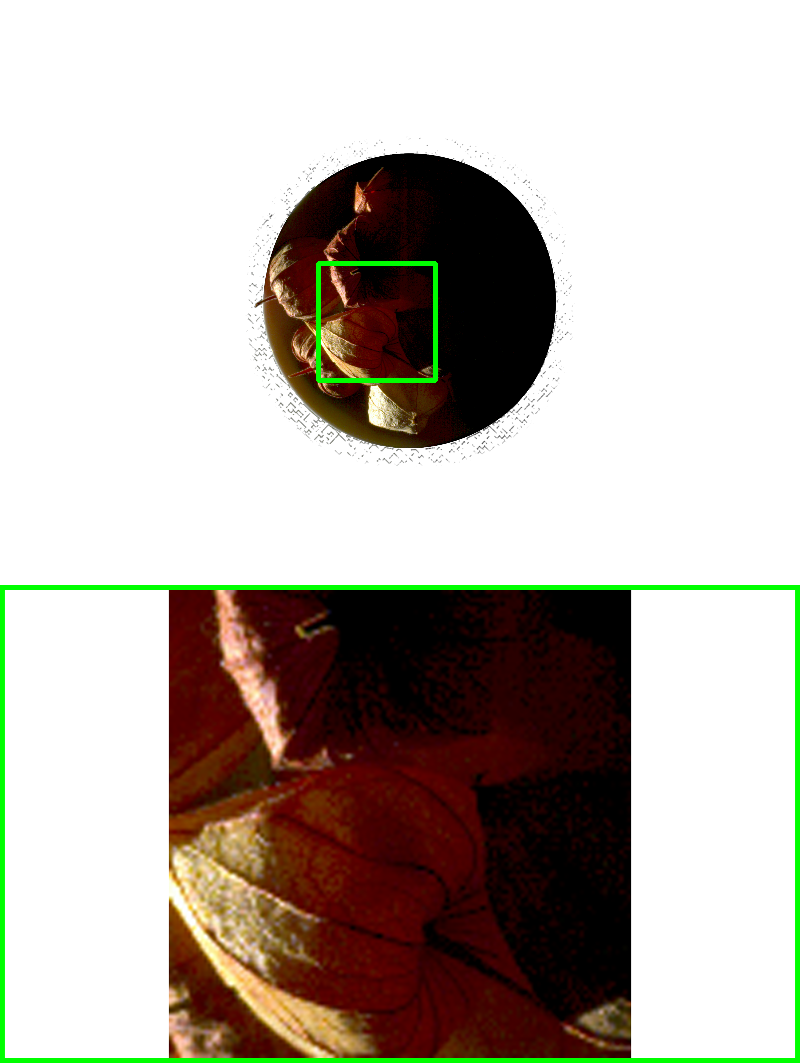} &
\includegraphics[width=\linewidth]{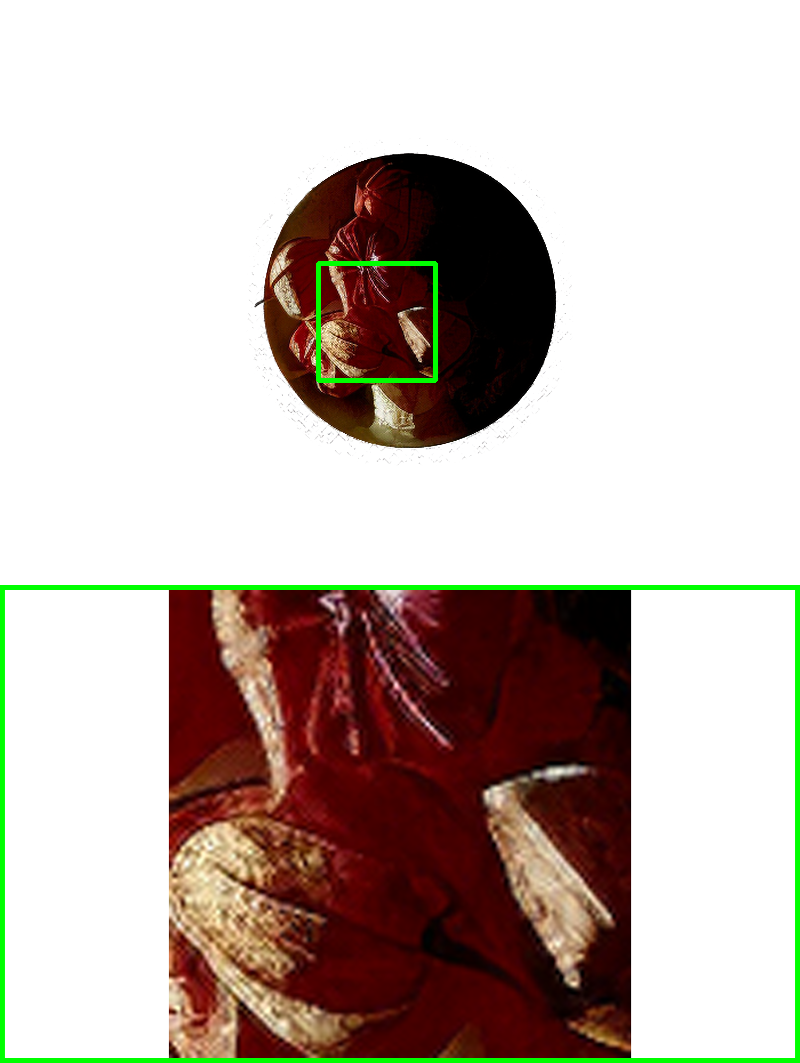} &
\hspace{1mm}\includegraphics[width=0.6\linewidth]{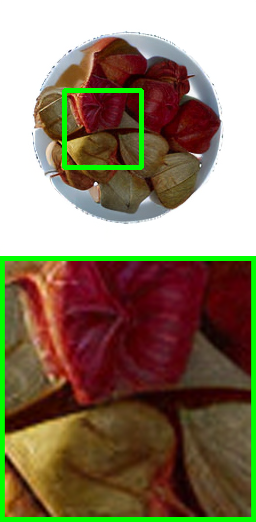} &
\includegraphics[width=.6\linewidth]{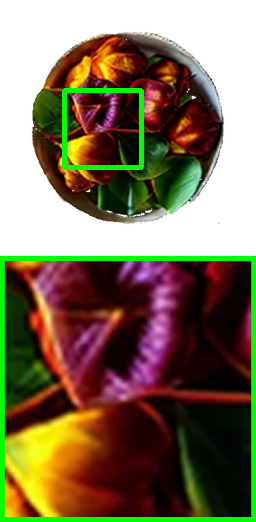} &
\includegraphics[width=.7\linewidth]{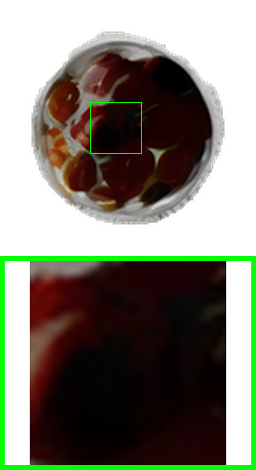} \\[4pt]

\includegraphics[width=\linewidth]{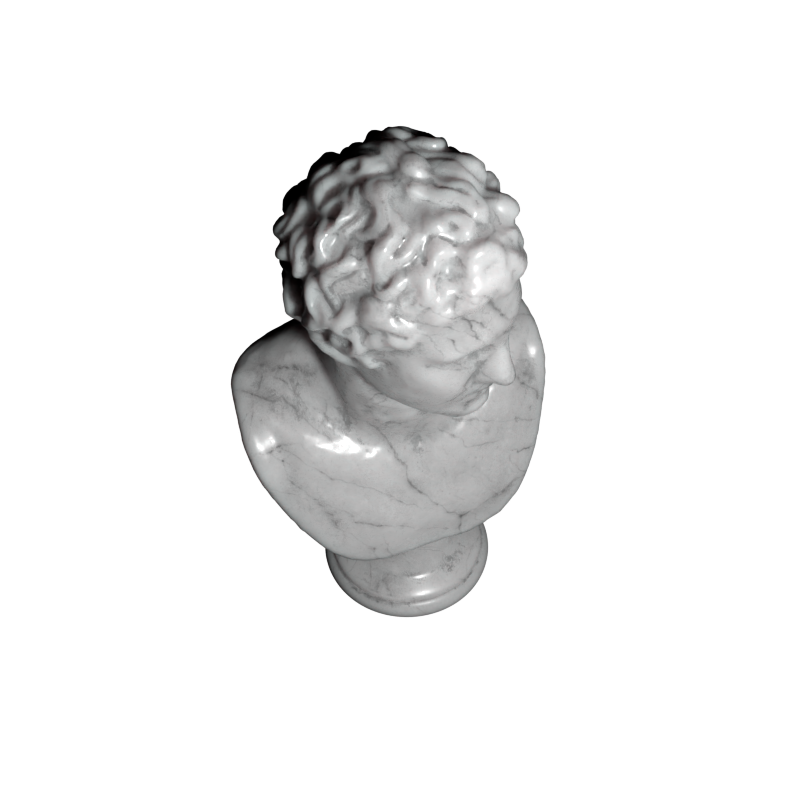} &
\includegraphics[width=\linewidth]{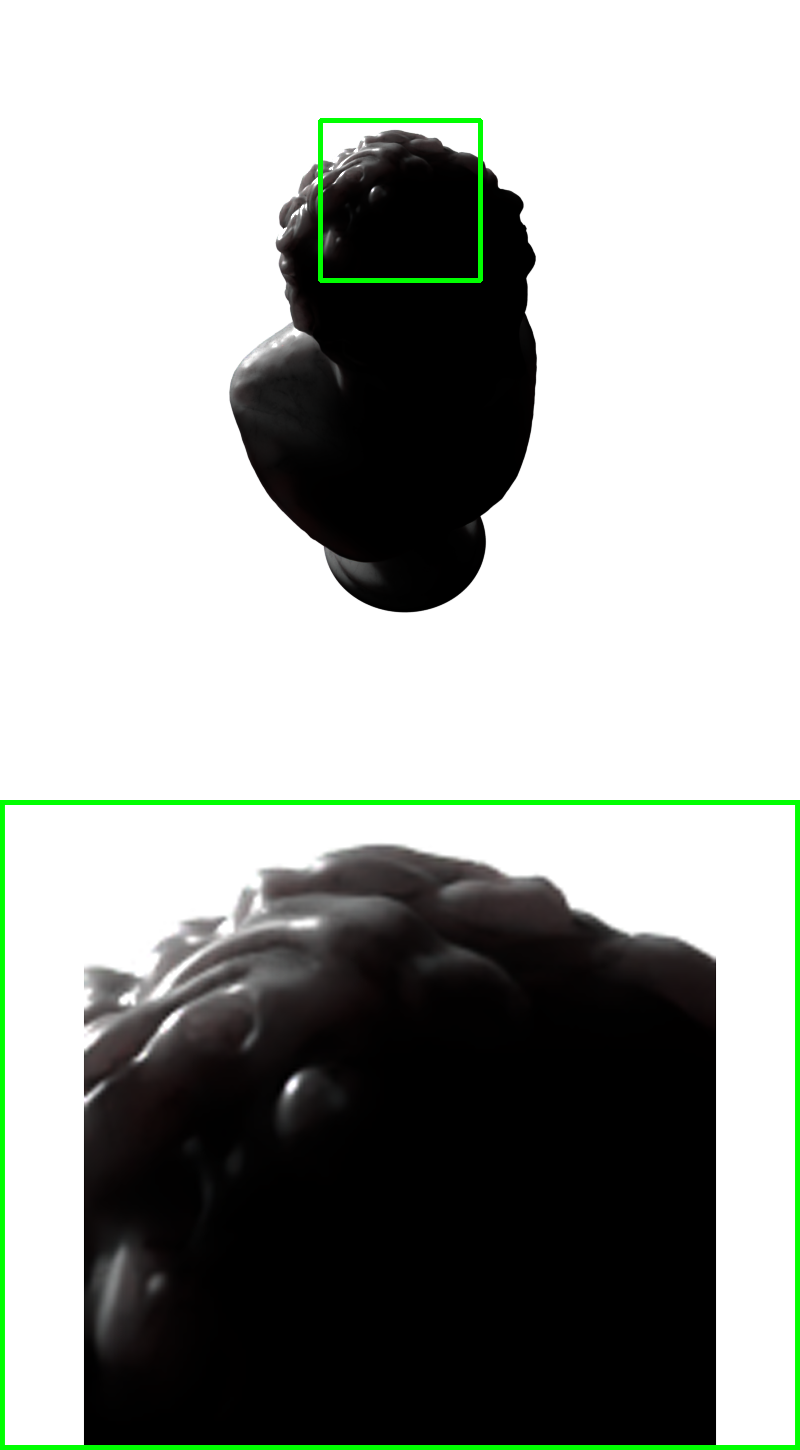} &
\includegraphics[width=\linewidth]{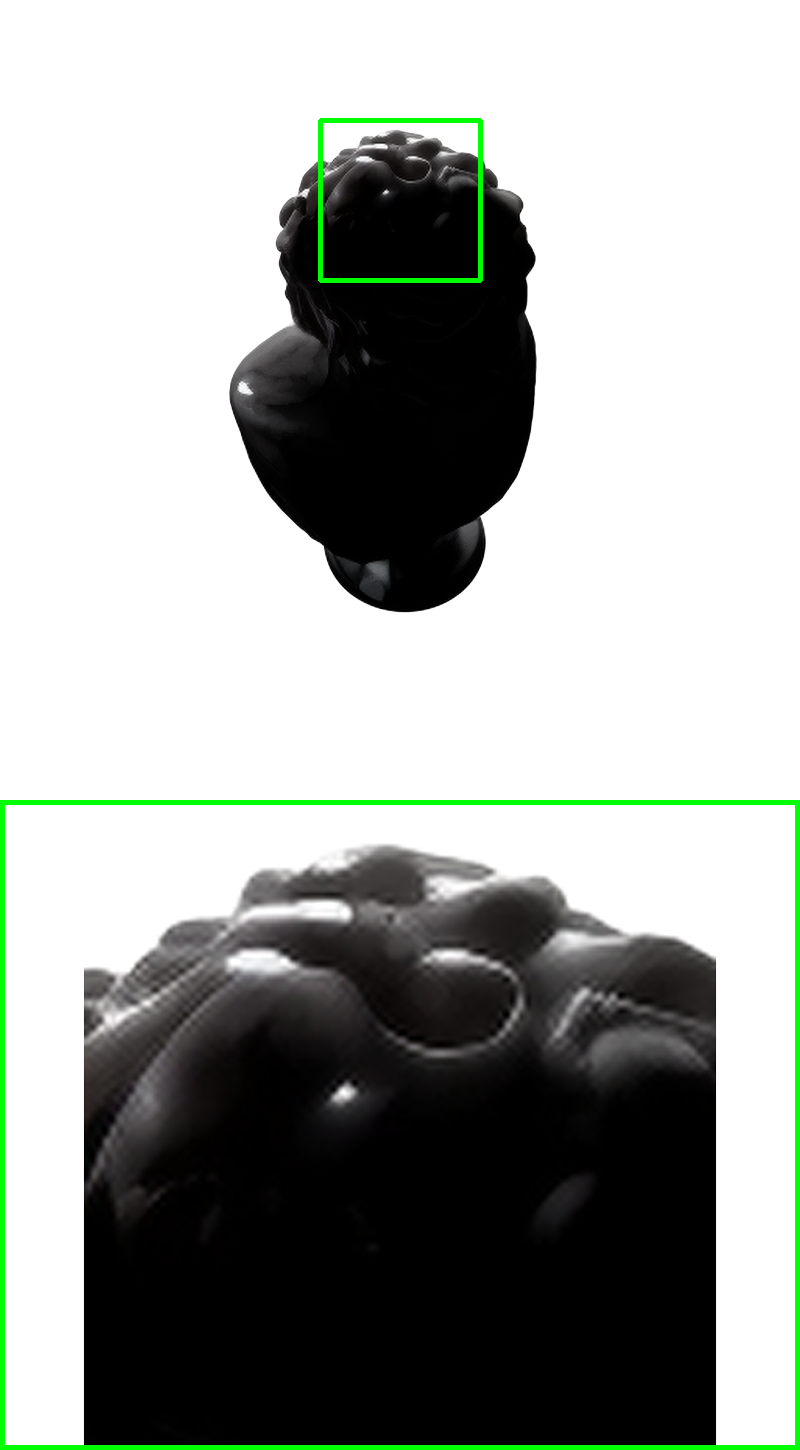} &
\includegraphics[width=\linewidth]{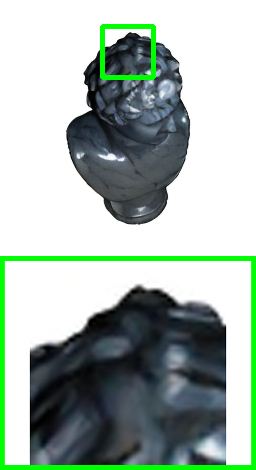} &
\includegraphics[width=\linewidth]{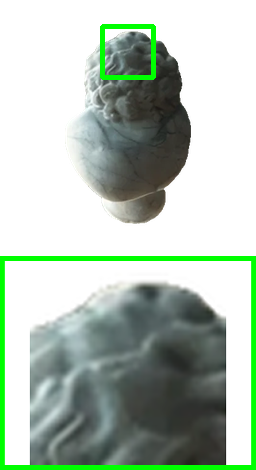} &
\includegraphics[width=\linewidth]{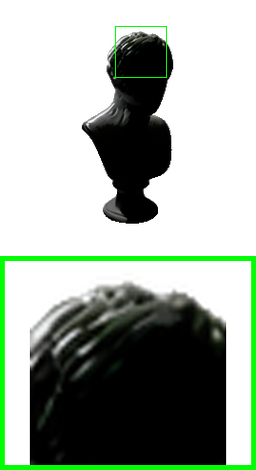} \\[4pt]

\parbox{\linewidth}{\centering\footnotesize Condition} &
\parbox{\linewidth}{\centering\footnotesize Ground Truth} &
\parbox{\linewidth}{\centering\footnotesize Relighting (ours)} &
\parbox{\linewidth}{\centering\footnotesize NeuralGaffer~\cite{neuralgaffer}} &
\parbox{\linewidth}{\centering\footnotesize IC-Light~\cite{iclight}} &
\parbox{\linewidth}{\centering\footnotesize NVS + Relighting (ours)} \\
\end{tabular}
}
\caption{\textbf{Qualitative relighting comparison.}
For three representative objects (two real, one synthetic), we compare our relighting-only model and our full NVS+relighting pipeline against NeuralGaffer~\cite{neuralgaffer} and IC-Light~\cite{iclight} under matched target illumination.
Our method better preserves geometry and material appearance, and produces more consistent subsurface scattering cues (e.g., soft shadows and glow) across diverse lighting conditions.}
\label{supp:fig:relighting_comparison}
\end{figure*}

\begin{figure*}[t]
\centering
\setlength{\tabcolsep}{2pt}
\renewcommand{\arraystretch}{0}
\begin{tabular}{*{4}{p{.19\textwidth}}}
\includegraphics[width=\linewidth]{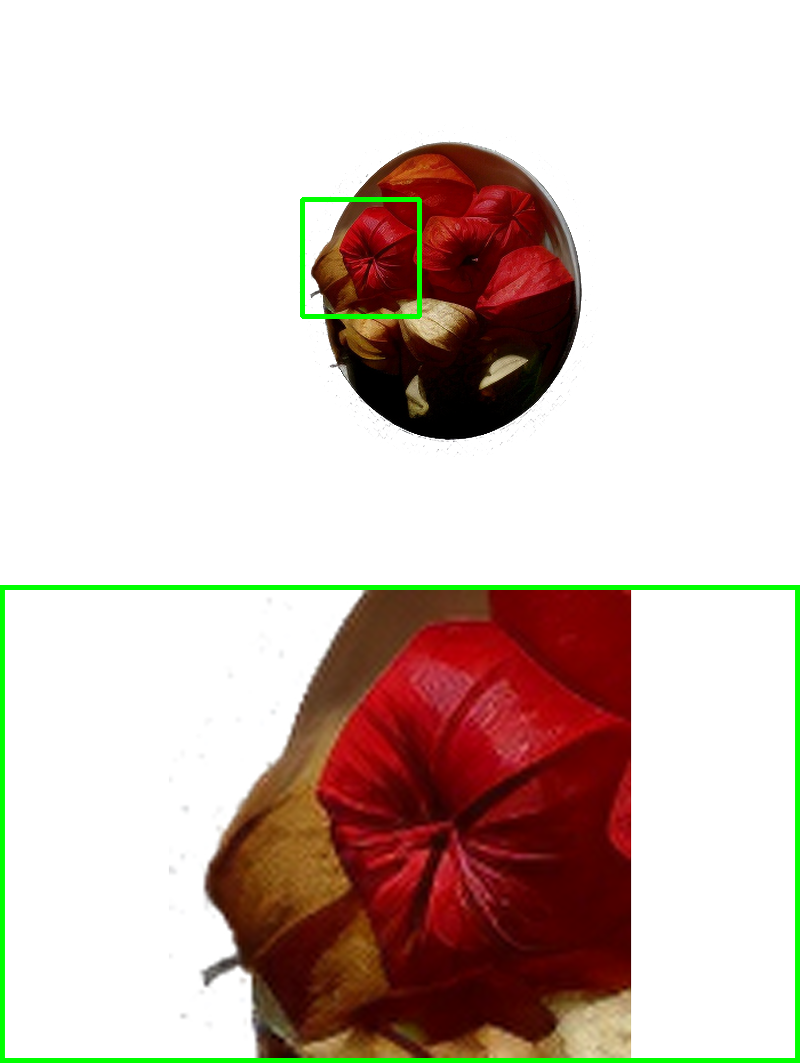} &
\includegraphics[width=\linewidth]{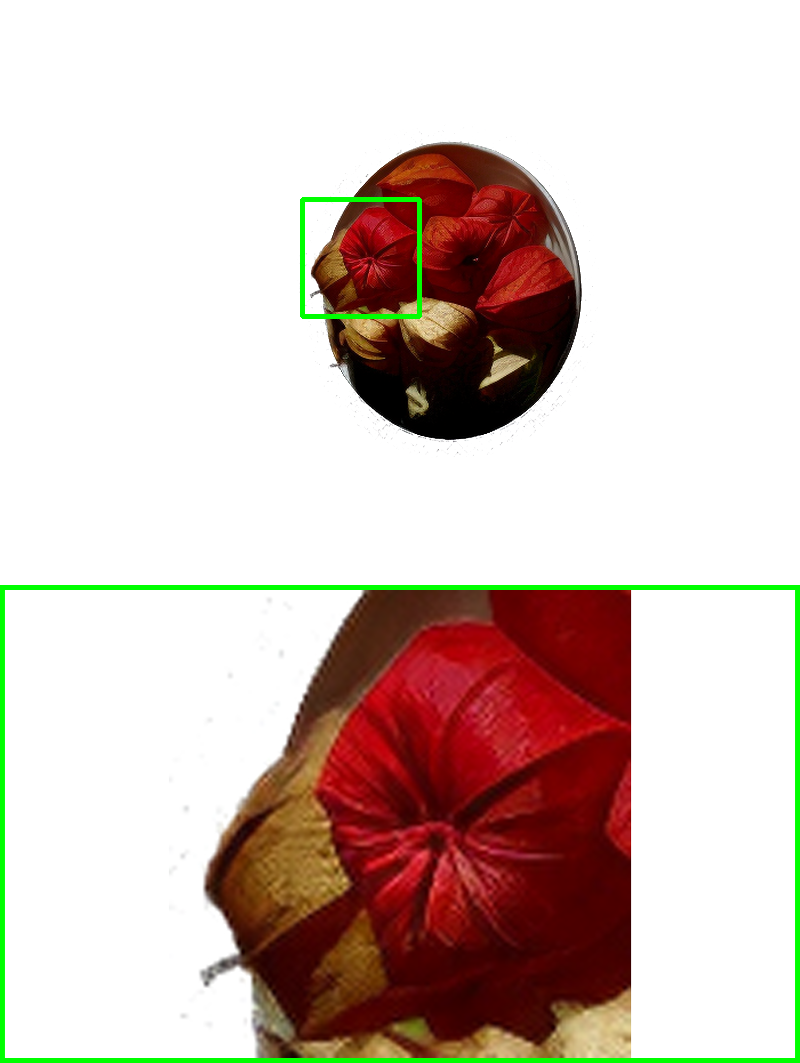} &
\includegraphics[width=\linewidth]{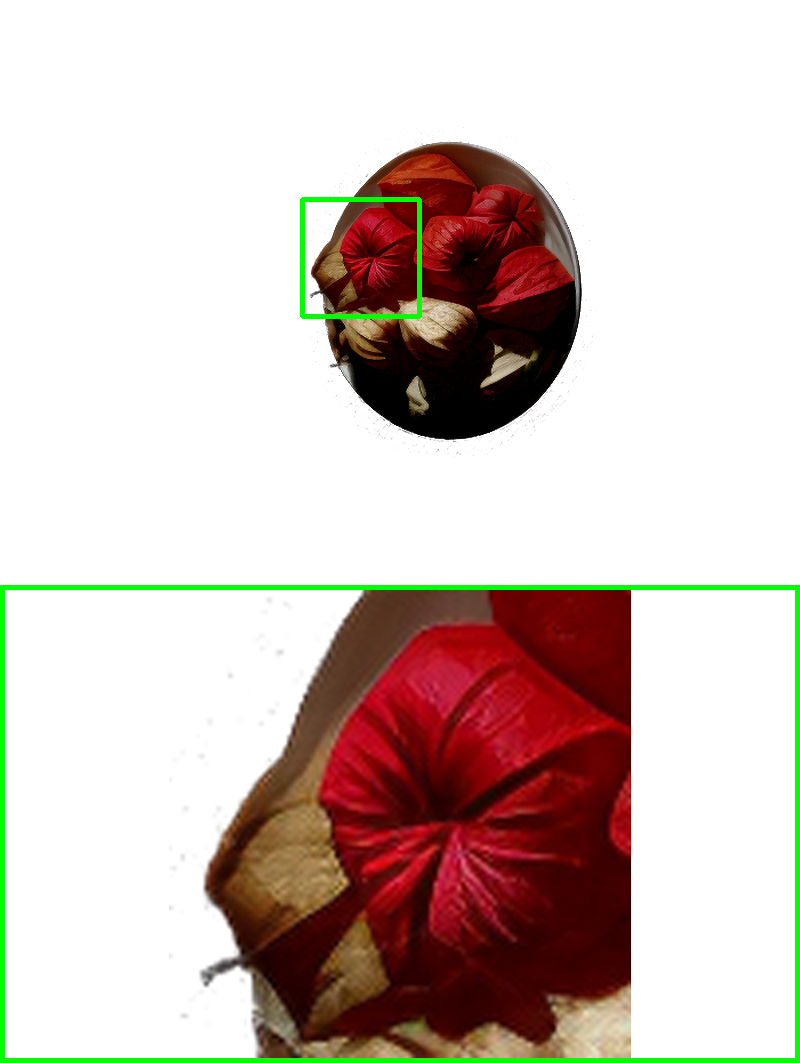} & 
\includegraphics[width=\linewidth]{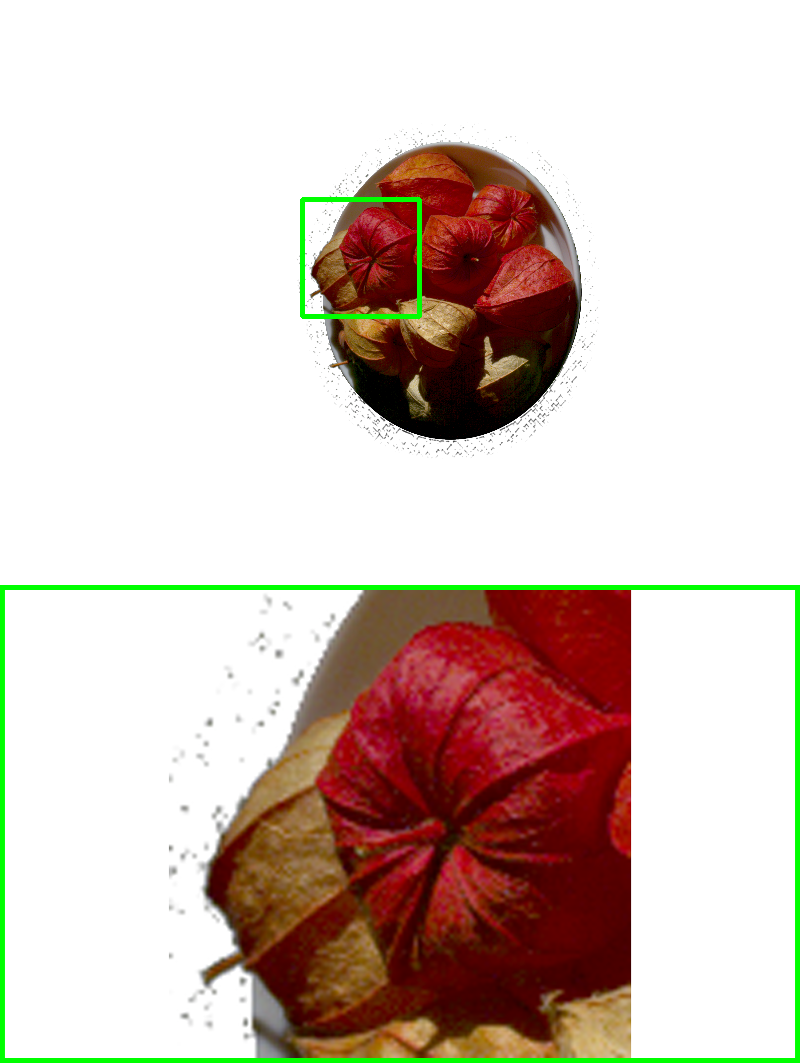} \\

\includegraphics[width=\linewidth]{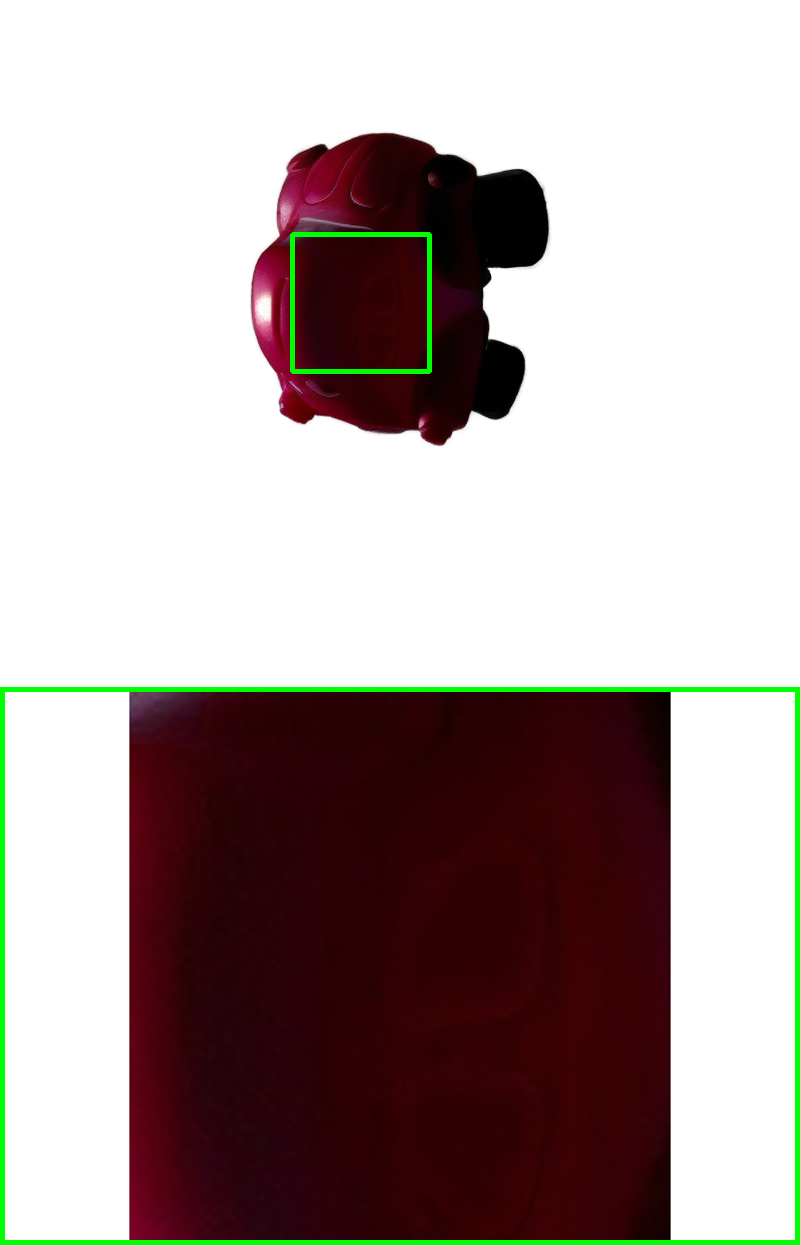} &
\includegraphics[width=\linewidth]{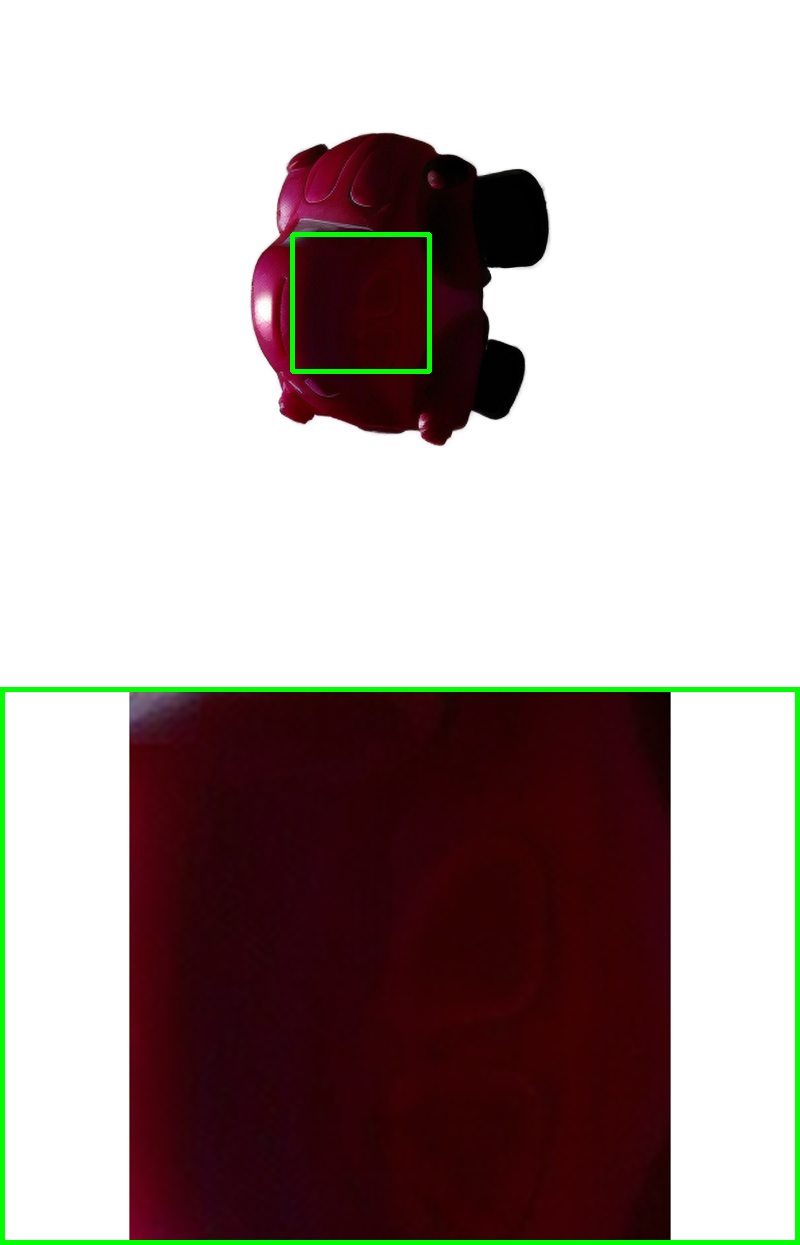} &
\includegraphics[width=\linewidth]{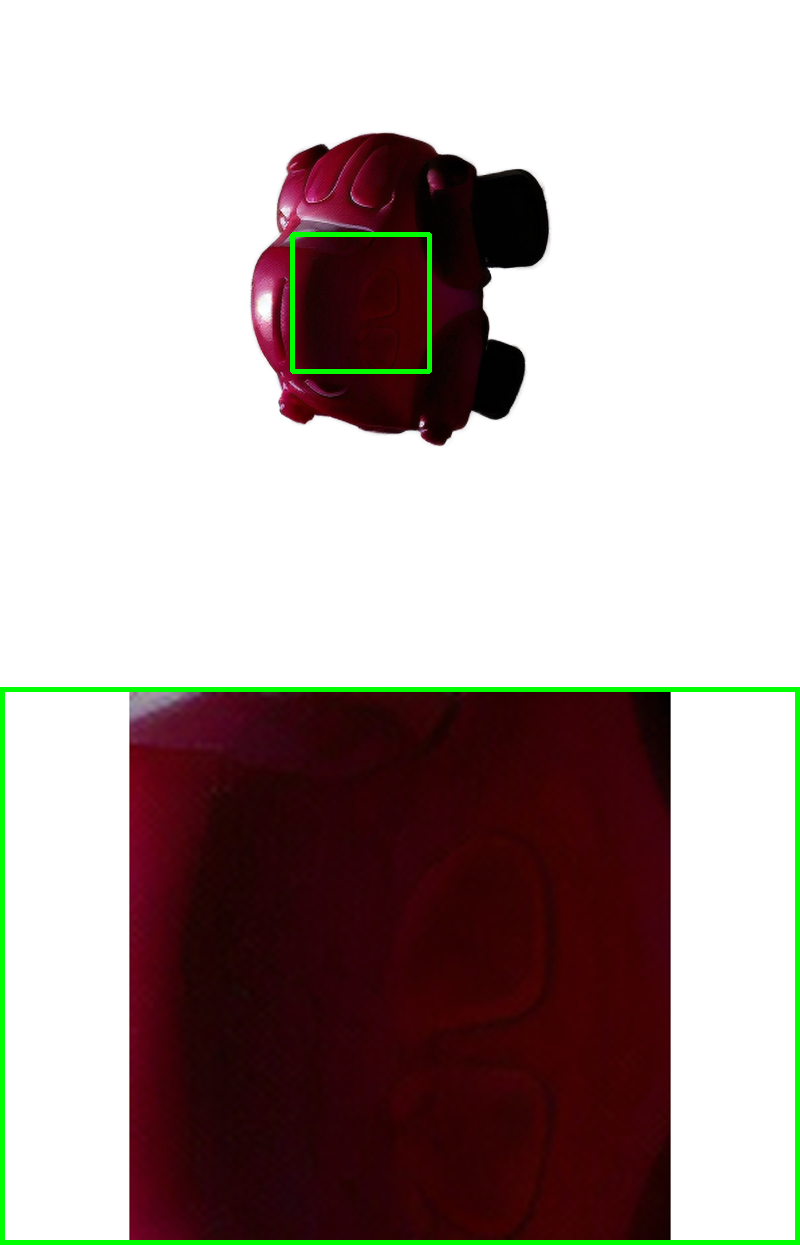} &
\includegraphics[width=\linewidth]{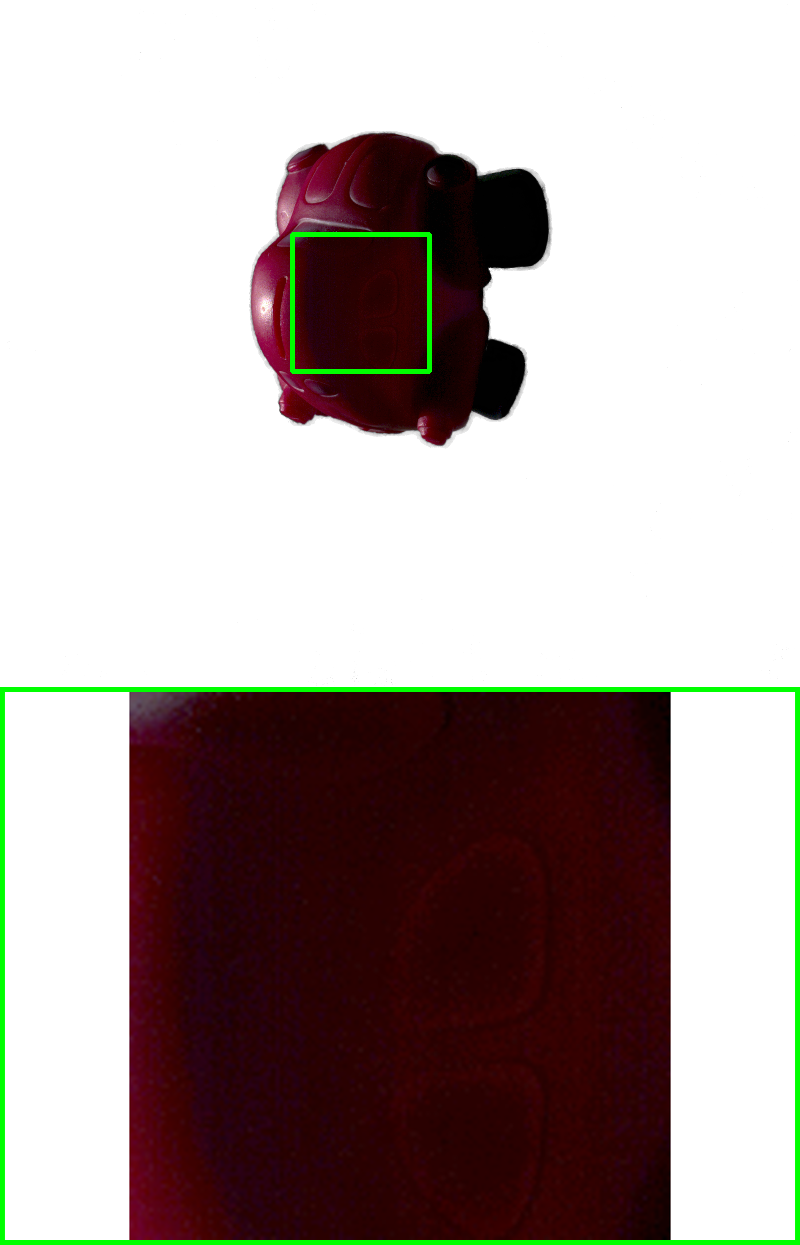}\\

\parbox{\linewidth}{\centering\footnotesize baseline} &
\parbox{\linewidth}{\centering\footnotesize baseline + depth + normals} &
\parbox{\linewidth}{\centering\footnotesize baseline + depth + normals + $\mathcal{L}_{1}$ + perceptual losses} &
\parbox{\linewidth}{\centering\footnotesize ground truth}\\
\end{tabular}
\caption{\textbf{Relighting ablation.}
Effect of progressively adding geometric conditioning (depth, normals) and pixel/perceptual losses to the relighting diffusion model.
Additional conditioning and perceptual supervision improve shadow placement, reduce halo artifacts, and better reproduce the soft scattering patterns of translucent materials.}
\label{supp:fig:relighting_ablations}
\end{figure*}

We evaluate the effect of diffusion fine-tuning for both the multi-view synthesis module and the relighting module.  
Figures~\ref{supp:fig:nvs_statue}--\ref{supp:fig:nvs_soap} summarize the multi-view synthesis behavior across several baselines, while Figures~\ref{supp:fig:relighting_comparison} and~\ref{supp:fig:relighting_ablations} focus on the relighting model. In addition, Table~\ref{tab:ablation_combined} shows our numerical results for the ablation studies. 

\paragraph{Multi-view diffusion (NVS).}
We compare our fine-tuned Free3D model with:  
(1) off-the-shelf Free3D,  
(2) Zero123,  
(3) Zero123-XL, and  
(4) MVD-Fusion.  

Across synthetic and real objects (Figures~\ref{supp:fig:nvs_statue}, \ref{supp:fig:nvs_bunny}, \ref{supp:fig:nvs_soap}), our fine-tuned model consistently produces:
\begin{itemize}[leftmargin=*]
    \item sharper and more stable silhouettes,
    \item reduced geometric drift across viewpoints,
    \item fewer distortions in thin structures,
    \item better preservation of object identity under large viewpoint changes.
\end{itemize}
This justifies using Free3D as the backbone for DIAMOND-SSS and highlights the importance of domain adaptation to OLAT capture conditions.

\paragraph{Relighting diffusion.}
We further ablate the effects of progressively increasing conditioning and supervision:
\begin{enumerate}
    \item Baseline RGB-only conditioning,
    \item + depth and normal conditioning,
    \item + L1 and perceptual losses.
\end{enumerate}

As illustrated in Figures~\ref{supp:fig:relighting_comparison} and~\ref{supp:fig:relighting_ablations}, each addition improves physical realism:
\begin{itemize}[leftmargin=*]
    \item depth improves global shadow placement,
    \item normals sharpen high-frequency surface cues,
    \item perceptual losses reduce haloing and over-sharp reflections,
    \item the full model best reproduces soft-scattering cues, achieving smooth light transitions characteristic of SSS material appearance.
\end{itemize}

\paragraph{Summary.}
The ablations show that diffusion models benefit substantially from fine-tuning under our OLAT dataset and conditioning strategy.  
For NVS, fine-tuning improves geometric coherence and silhouette stability.  
For relighting, deeper conditioning leads to more faithful subsurface-scattering behavior and reduced artifacts.  
These findings support the design choices adopted in DIAMOND-SSS and demonstrate that diffusion-based augmentation becomes reliable only after domain adaptation.

\section{Multi-View Geometric Losses: Practical Details}
\label{sec:supp_geom}

The multi-view geometric consistency losses introduced in \cref{subsec:geometric_consistency}—namely multi-view silhouette consistency and multi-view depth consistency—play a central role in stabilizing reconstructions under sparse or synthetic supervision. This section provides additional implementation details, motivation, and qualitative analyses that complement the main text.

\subsection{Motivation and Intuition}
Multi-view photometric supervision alone does not sufficiently constrain the geometry of translucent objects, in particular in our very sparse setting with synthetic data augmentation. Small inconsistencies in predicted depth or occupancy can lead to view-dependent artifacts, such as boundary bleeding or inconsistent surface placement, which propagate into both shading and appearance, especially under SSS transport.

The two geometric consistency terms complement each other:

\paragraph{Silhouette Consistency.}
The silhouette-consistency term enforces that pixels corresponding to the object’s foreground in one view remain foreground when reprojected into another view. Concretely, for a pixel $x$ in view $i$, we back-project it using its predicted depth and reproject it into view $j$ (see~\cref{eq:sil_loss}). The loss penalizes cases where this reprojected point falls outside the soft silhouette mask of view $j$, thereby stabilizing object boundaries, reducing opacity bleeding, and improving cross-view alignment of object outlines.

\paragraph{Depth Consistency.}
Silhouette alignment alone does not provide depth information. The depth consistency loss addresses this by enforcing agreement between the predicted metric depth in view $j$ and the back-projected/reprojected depth from view $i$ (see~\cref{eq:depth_loss} in the main paper). This ensures that the same physical surface point occupies a consistent 3D position across camera viewpoints, thereby improving global geometric coherence.

Together, these losses constrain both \emph{where} the surface should appear (silhouette) and \emph{how far} it should lie along each viewing ray (depth), reducing cross-view inconsistencies and improving stability under downstream relighting.

\subsection{Visibility and Correspondence Filtering}
\label{sec:supp_visibility}

Both $\mathcal{L}_{\text{sil}}^{\text{MV}}$ and $\mathcal{L}_{\text{depth}}^{\text{MV}}$ rely on correspondences obtained via back-projection from a source view $i$ to a target view $j$ using the camera models. In practice, we improve robustness by:

\begin{itemize}[leftmargin=*]
    \item \textbf{Domain filtering:} Correspondences whose reprojected coordinates $x'_{i,j}$ fall outside the target image plane are discarded.
    \item \textbf{Silhouette filtering:} Pixels reprojected outside the target soft silhouette (beyond a small tolerance) are ignored.
    \item \textbf{Depth validity:} Correspondences with invalid or missing depth (e.g., occlusions or empty space) in either view are removed.
    \item \textbf{Normalization:} Each loss term is normalized by the number of valid correspondences to avoid biasing toward dense or front-facing views.
\end{itemize}

This visibility-aware masking substantially reduces artifacts near occlusion boundaries and improves stability in sparse-view settings.

\subsection{Depth Normalization and Numerical Stability}
\label{sec:supp_depth_norm}

Depth values depend on the scene scale and camera intrinsics. To improve numerical conditioning, both the reprojection depth $\hat{z}_{i,j}$ and the rendered depth $D_j(x')$ are linearly normalized to $[0,1]$ using scene-specific near/far clipping planes. This removes the need for per-object tuning of $\lambda_{\text{depth}}$ and yields consistent gradients across scenes.

\subsection{Ablation Studies}
\label{sec:supp_geom_ablation}

\begin{figure*}[htbp]
\centering
\resizebox{\textwidth}{!}{
\setlength{\tabcolsep}{2pt}
\renewcommand{\arraystretch}{0}

\begin{tabular}{*{5}{p{.24\textwidth}}}
\includegraphics[width=\linewidth]{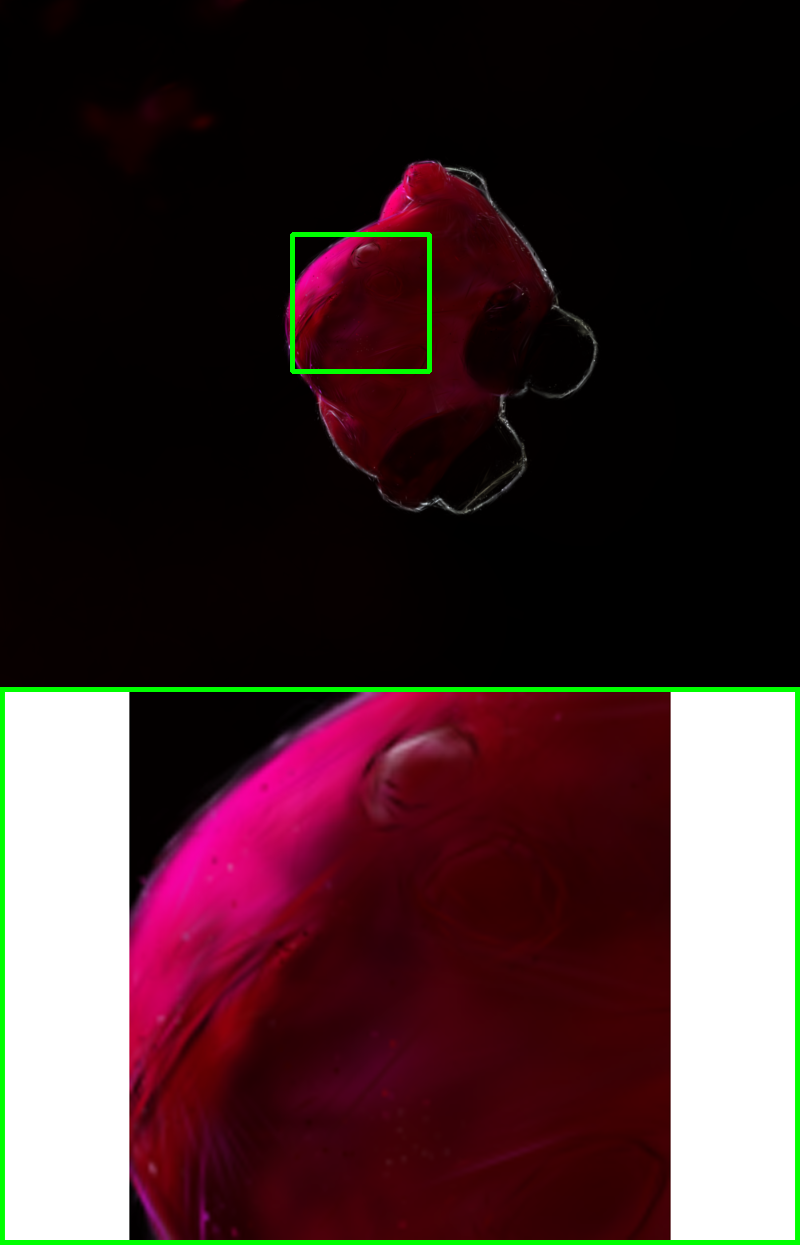} &
\includegraphics[width=\linewidth]{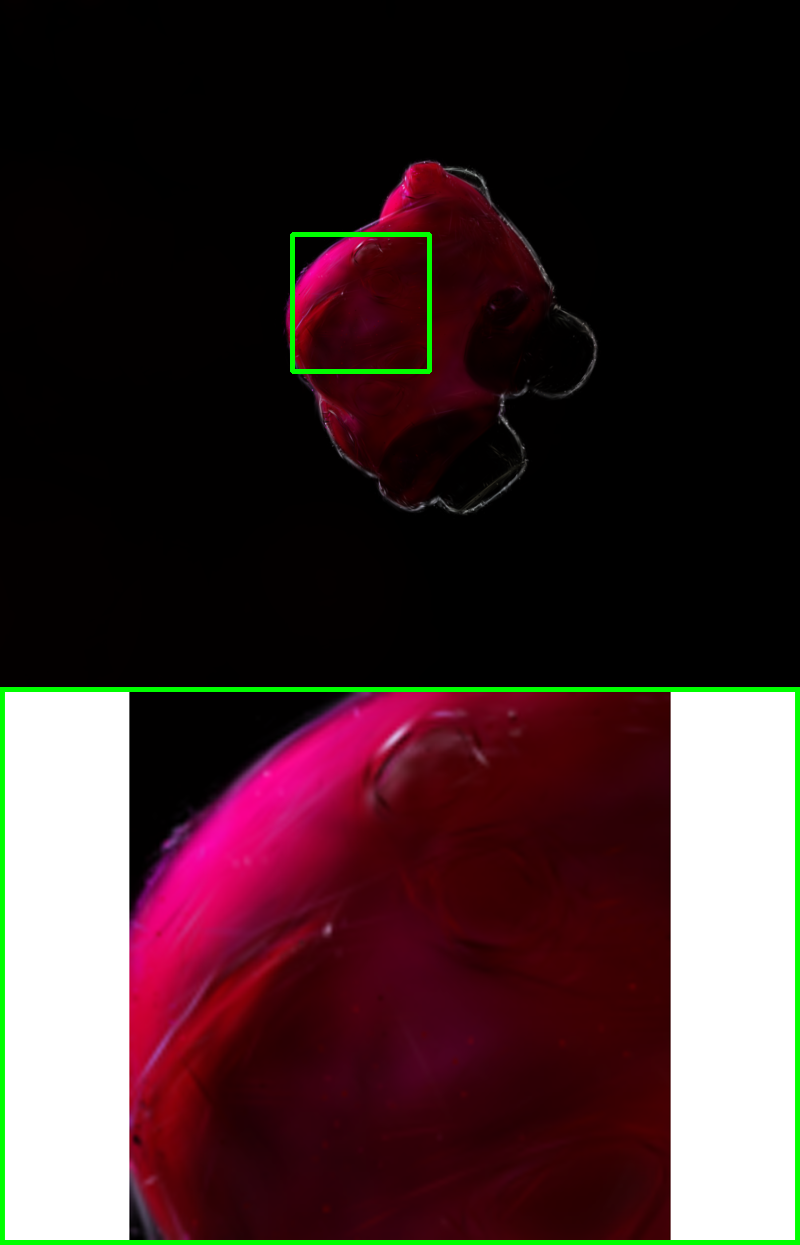} &
\includegraphics[width=\linewidth]{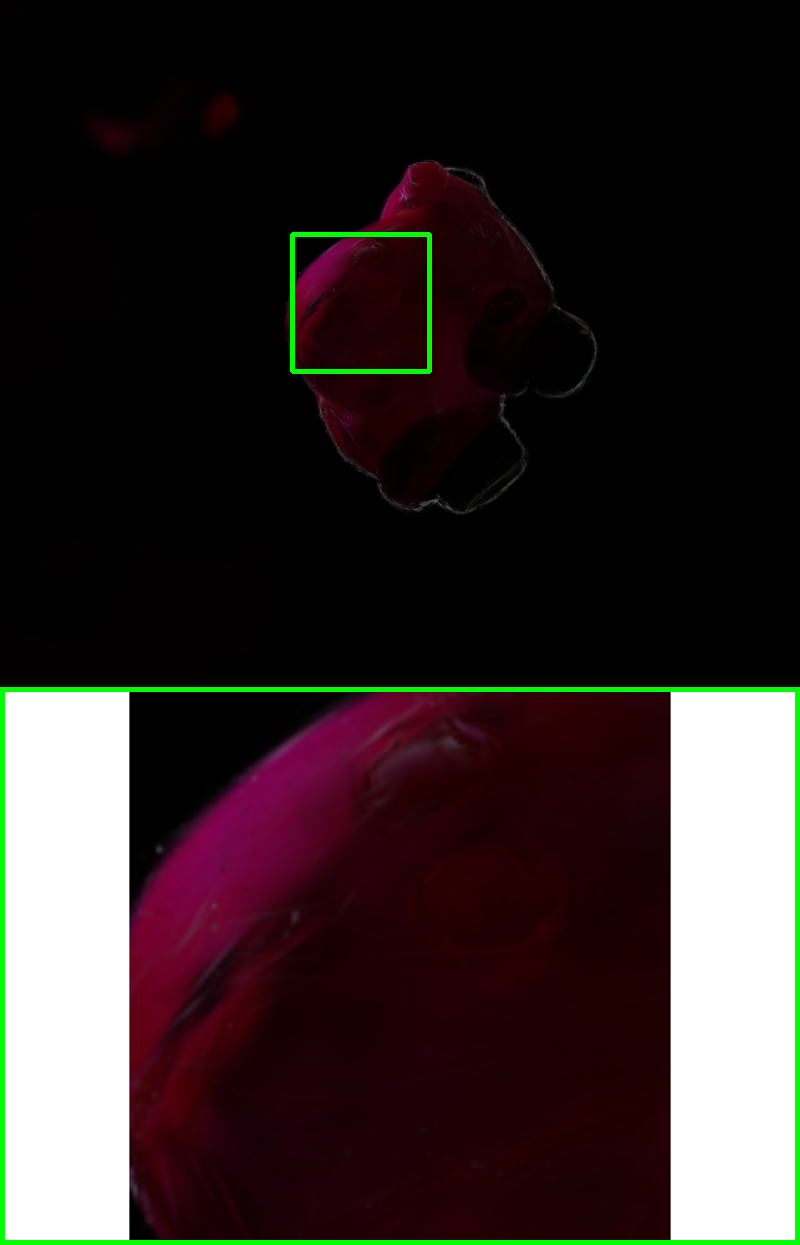} &
\includegraphics[width=\linewidth]{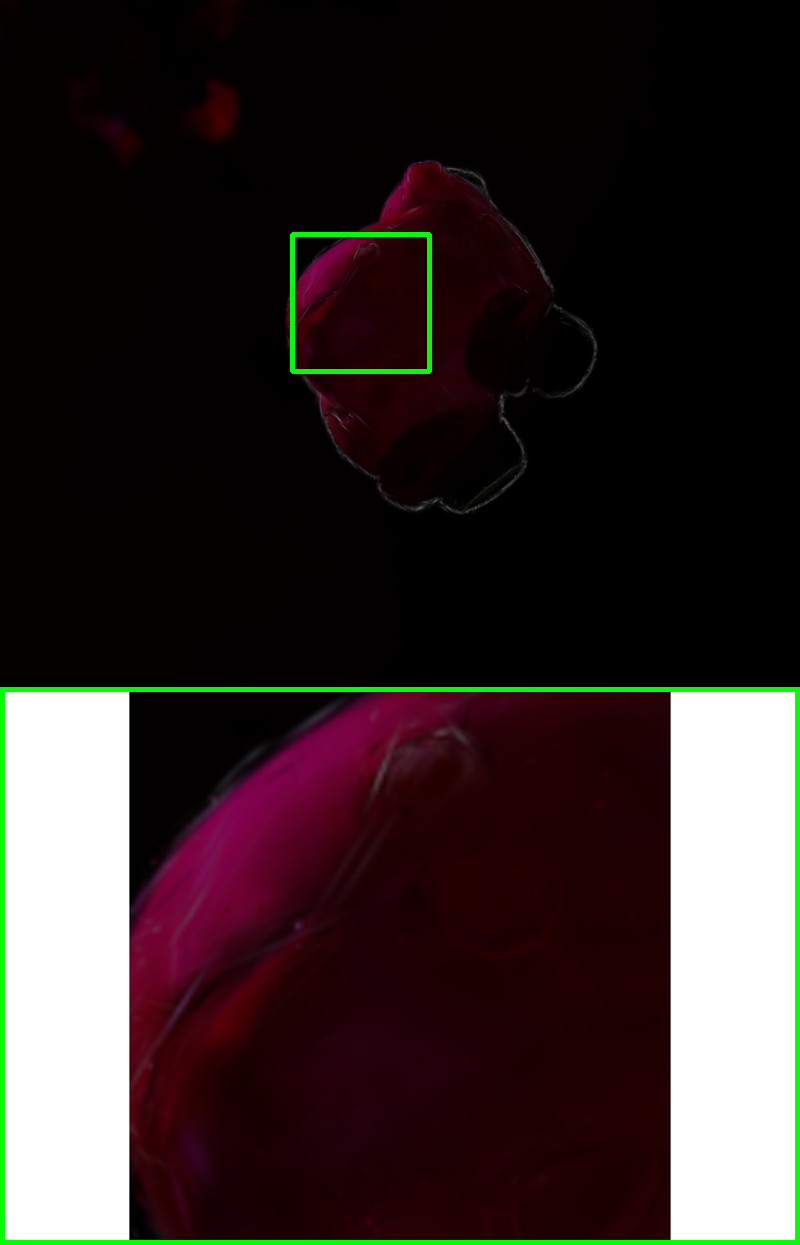} &
\includegraphics[width=\linewidth]{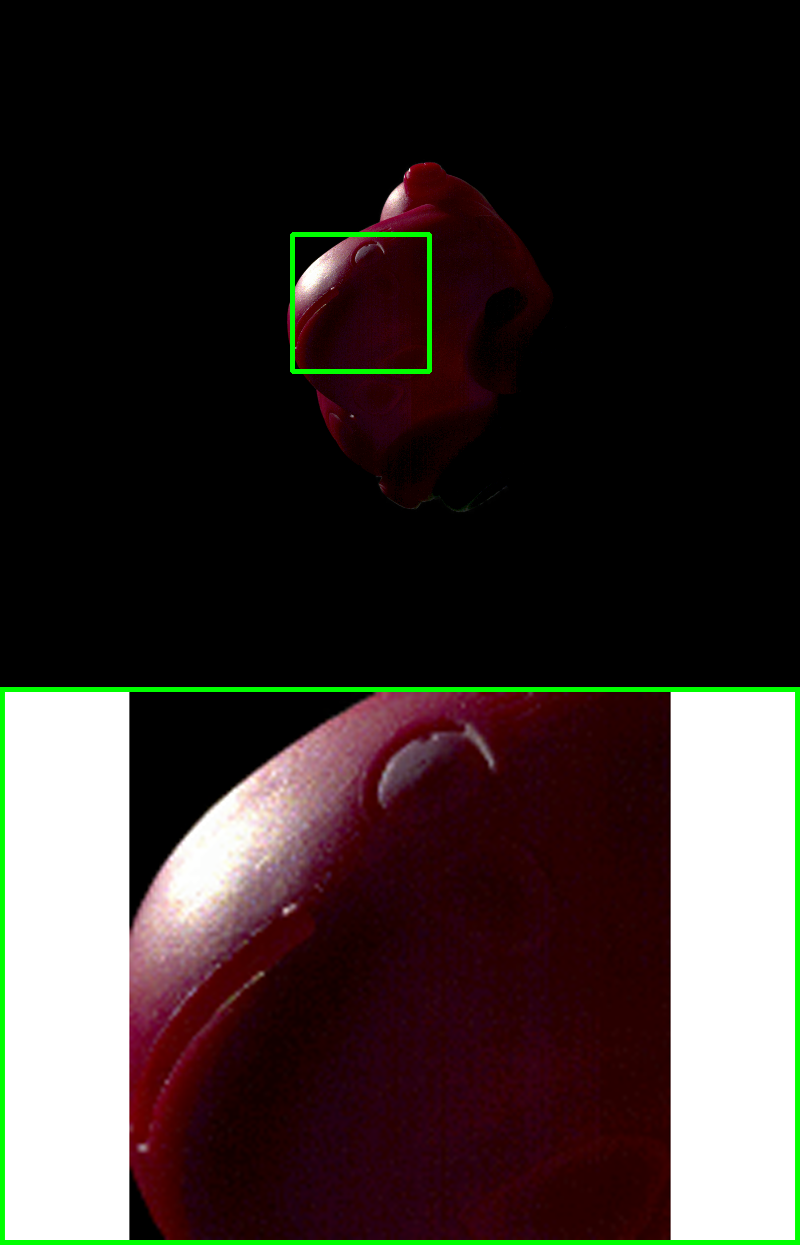} \\[4pt]
\parbox{\linewidth}{\centering\footnotesize SSS-3DGS \cite{dihlmann2024subsurfacescattering3dgaussian}} &
\parbox{\linewidth}{\centering\footnotesize + Silhouette} &
\parbox{\linewidth}{\centering\footnotesize + Depth} &
\parbox{\linewidth}{\centering\footnotesize + Depth + Silhouette} &
\parbox{\linewidth}{\centering\footnotesize Ground Truth}
\end{tabular}
}
\caption{\textbf{Synthetic-data ablation of multi-view geometric losses.} Silhouette and depth consistency each improve boundary stability and metric alignment, with the combination yielding the most coherent results.}
\label{fig:reconstruction_metrics}
\end{figure*}

\begin{figure*}[h!]
\centering
\resizebox{\textwidth}{!}{
\setlength{\tabcolsep}{2pt}
\renewcommand{\arraystretch}{0}

\begin{tabular}{*{3}{p{.24\textwidth}}}
\includegraphics[width=\linewidth]{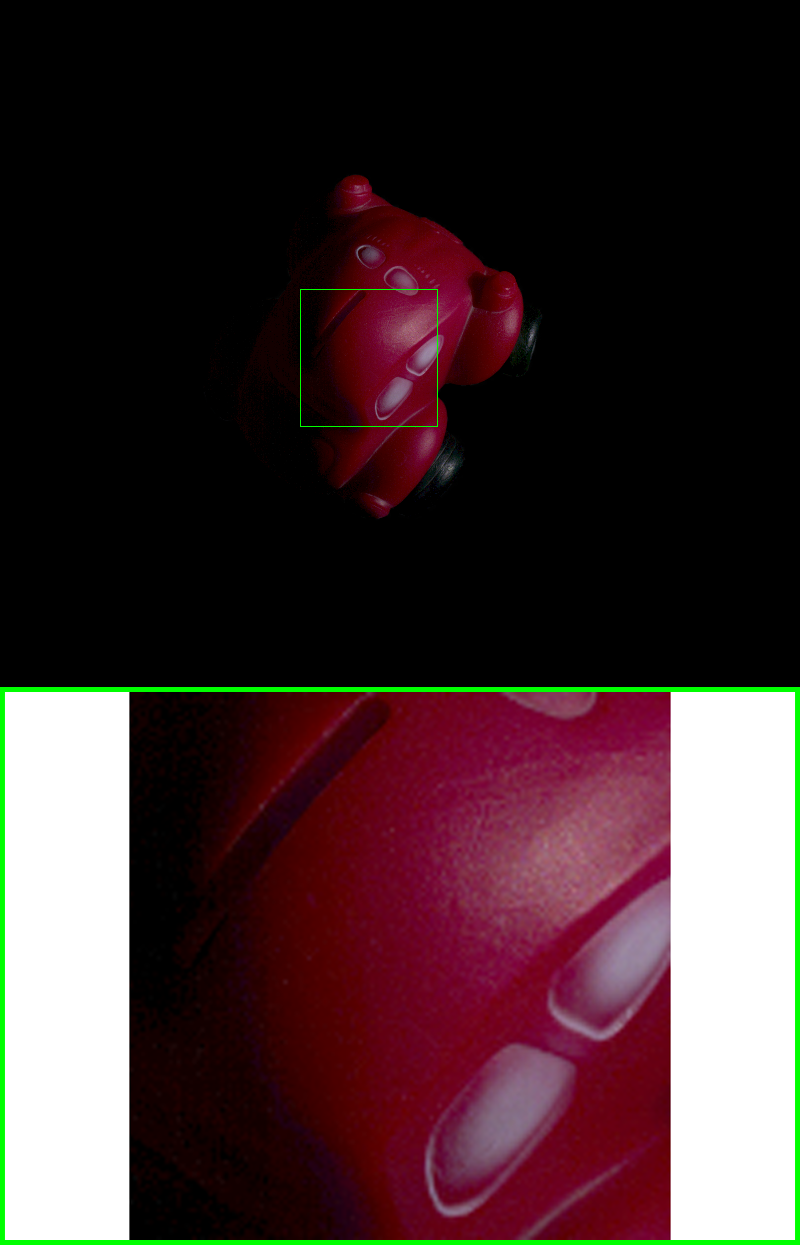} &
\includegraphics[width=\linewidth]{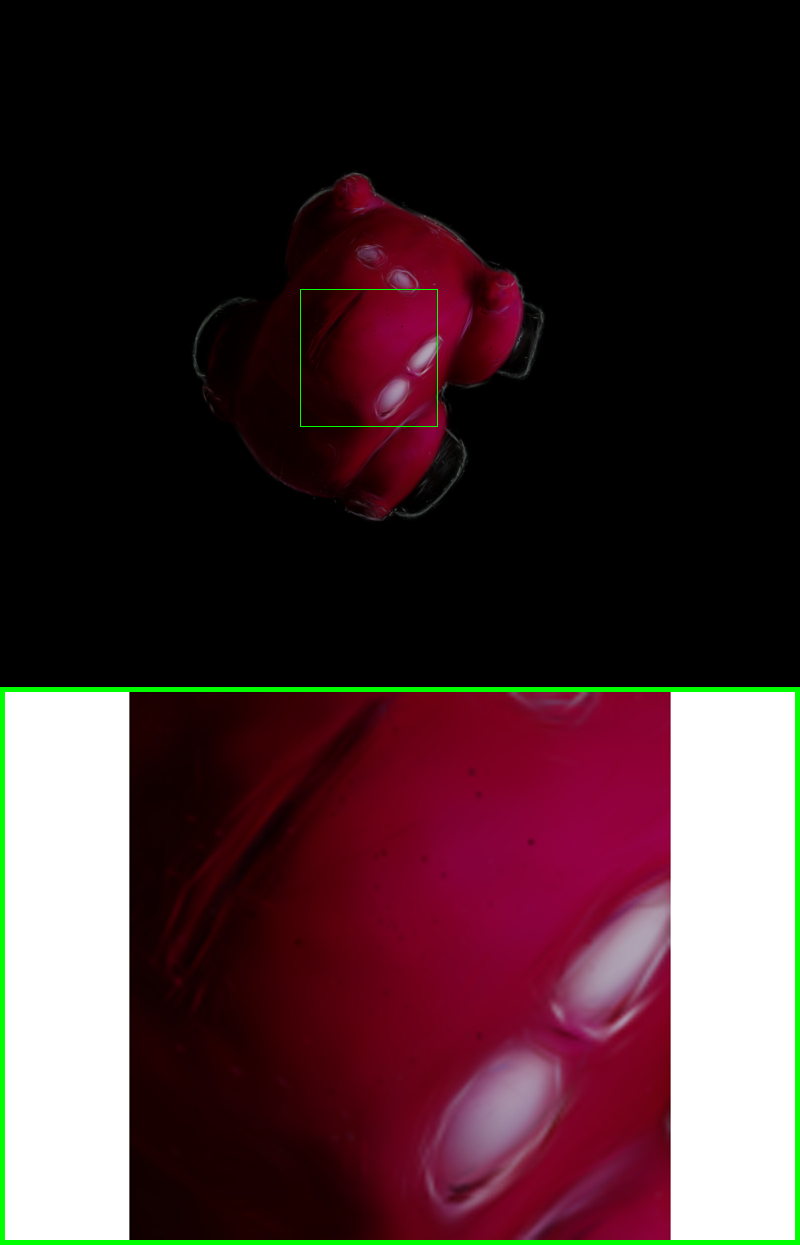} &
\includegraphics[width=\linewidth]{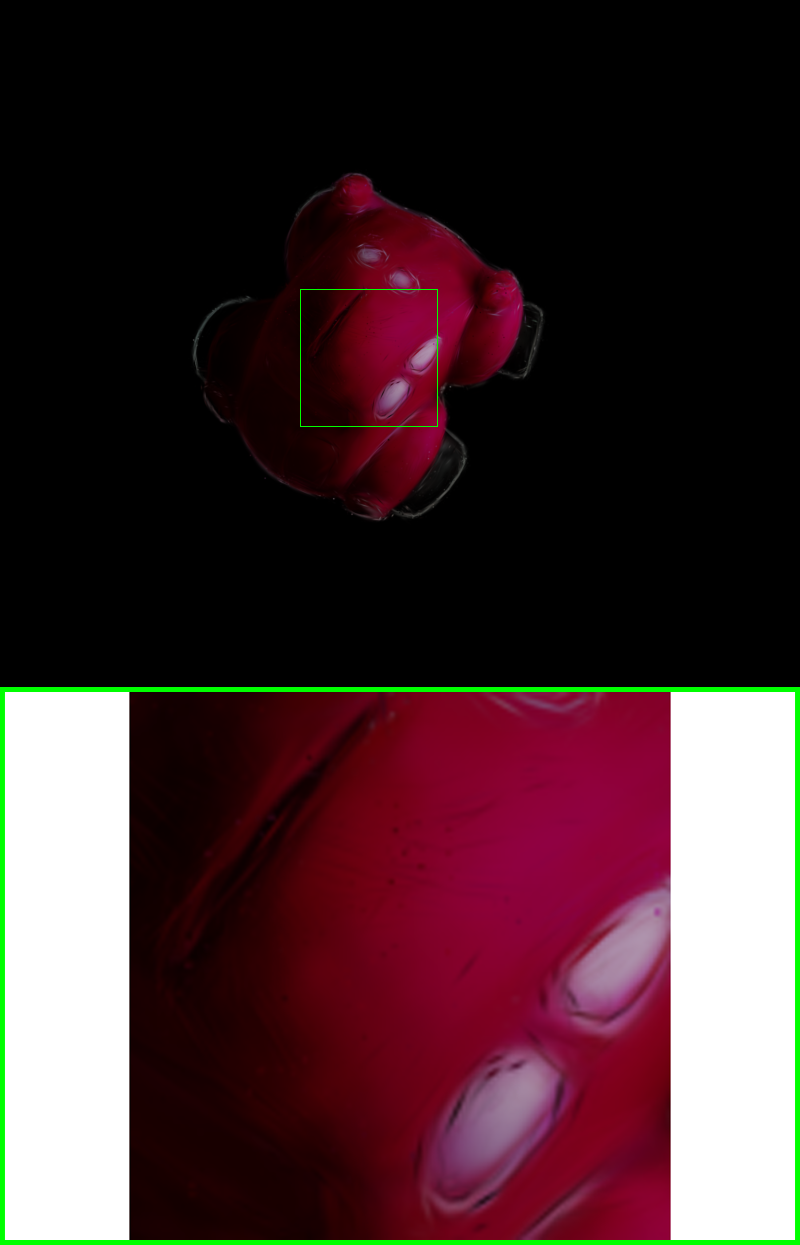} \\[4pt]
\parbox{\linewidth}{\centering\footnotesize Ground Truth} &
\parbox{\linewidth}{\centering\footnotesize  Ours} &
\parbox{\linewidth}{\centering\footnotesize SSS-3DGS \cite{dihlmann2024subsurfacescattering3dgaussian} + Synthetic augmentation}
\end{tabular}
}
\caption{\textbf{Real-data reconstruction comparison.} Multi-view geometric losses significantly improve boundary correctness and reduce view-dependent distortions. In this images we are showing the full sampling setting.}
\label{fig:reconstruction_comparison}
\end{figure*}

\begin{figure*}[htbp]
    \centering
    \includegraphics[width=0.86\linewidth]{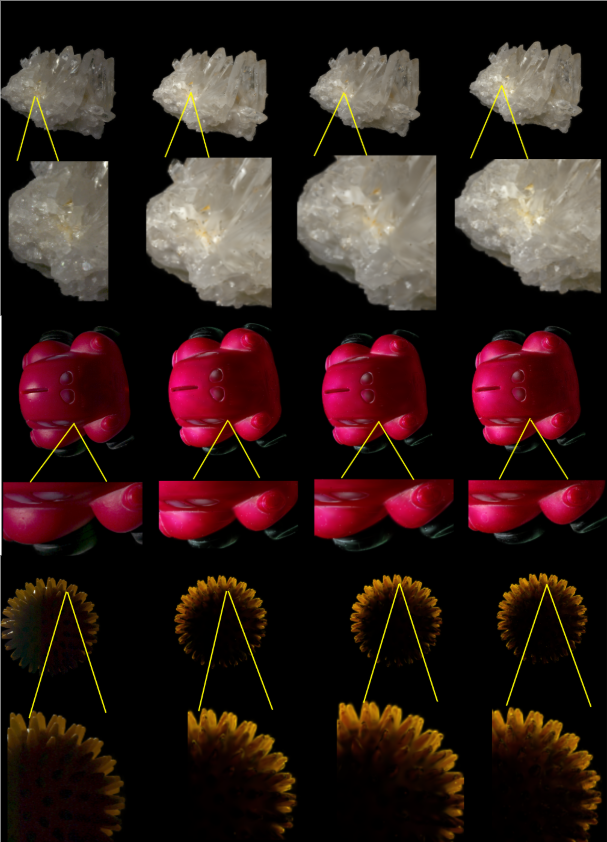}
    \begin{minipage}[t]{0.24\linewidth}
        \centering
        \textbf{(a)} Ground Truth
    \end{minipage}
    \begin{minipage}[t]{0.24\linewidth}
        \centering
        \textbf{(b)} SSS \cite{dihlmann2024subsurfacescattering3dgaussian} Tuned
    \end{minipage}
    \begin{minipage}[t]{0.24\linewidth}
        \centering
        \textbf{(c)} Ours
    \end{minipage}
    \begin{minipage}[t]{0.24\linewidth}
        \centering
        \textbf{(d)} SSS 3DGS \cite{dihlmann2024subsurfacescattering3dgaussian} + Losses
    \end{minipage}
    \caption{\textbf{Multi-view geometric losses on real objects.} The full multi-view supervision produces sharper boundaries and reduces positional drift. Here we also compare the original method with our tuned hyperparemeters.}
    \label{fig:qualitative_comparisonablation2}
\end{figure*}

\begin{figure*}[htbp]
    \centering
    \includegraphics[width=\linewidth]{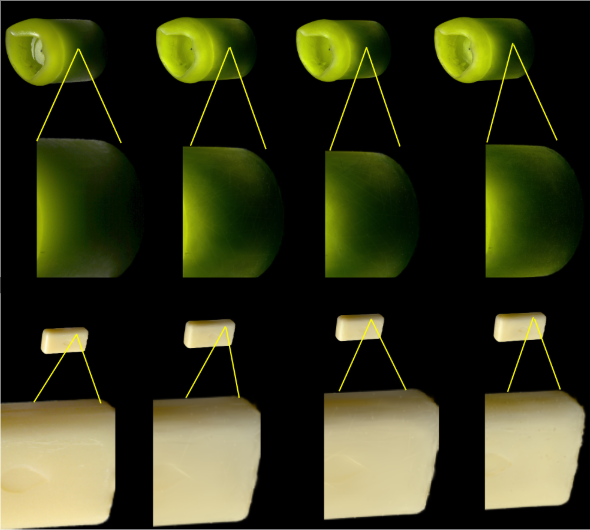}
    \vspace{0.5em}
    \begin{minipage}[t]{0.24\linewidth}
        \centering
        \textbf{(a)} Ground Truth
    \end{minipage}
    \begin{minipage}[t]{0.24\linewidth}
        \centering
        \textbf{(b)} SSS 3DGS \cite{dihlmann2024subsurfacescattering3dgaussian} Tuned
    \end{minipage}
    \begin{minipage}[t]{0.24\linewidth}
        \centering
        \textbf{(c)} Ours
    \end{minipage}
    \begin{minipage}[t]{0.24\linewidth}
        \centering
        \textbf{(d)} SSS 3DGS \cite{dihlmann2024subsurfacescattering3dgaussian} Multi-View
    \end{minipage}
    \caption{\textbf{Synthetic vs. tuned multi-view results.} Combining parameter tuning with multi-view losses yields the most stable reconstructions.  Here we also compare the original method with our tuned hyperparemeters.}
    \label{fig:qualitative_comparisonablation3}
\end{figure*}

\begin{figure*}[htbp]
    \centering
    \includegraphics[width=0.8\linewidth]{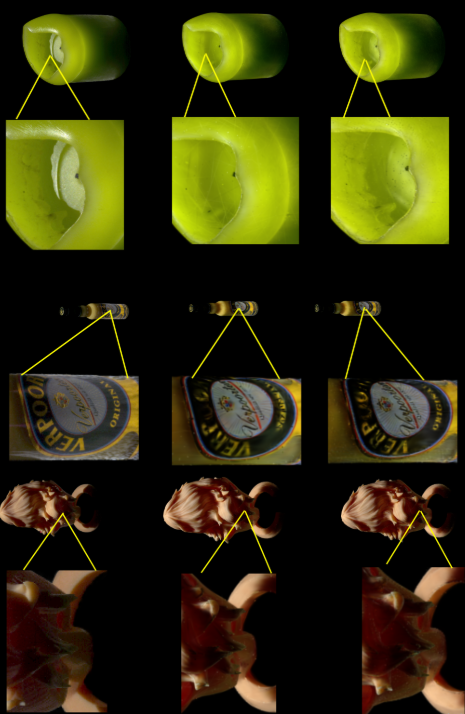}
    \vspace{0.5em}
    \begin{minipage}[t]{0.32\linewidth}
        \centering
        \textbf{(a)} Ground Truth
    \end{minipage}
    \begin{minipage}[t]{0.32\linewidth}
        \centering
        \textbf{(b)} Ours
    \end{minipage}
    \begin{minipage}[t]{0.32\linewidth}
        \centering
        \textbf{(c)} SSS 3DGS \cite{dihlmann2024subsurfacescattering3dgaussian} Tuned
    \end{minipage}
    \caption{\textbf{Silhouette loss effect on synthetic data.} The addition of silhouette regularization reduces boundary bleeding and sharpens object contours.}
    \label{fig:loss_results}
\end{figure*}

\begin{figure*}[htbp]
    \centering
    \includegraphics[width=0.8\linewidth]{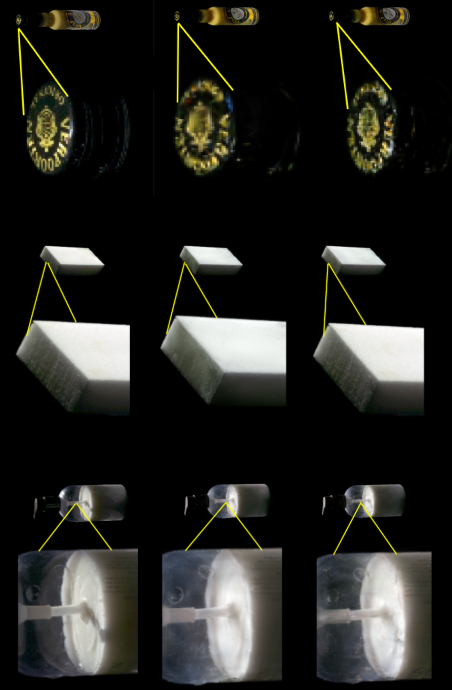}
    
    \vspace{0.5em}
    \begin{minipage}[t]{0.32\linewidth}
        \centering
        \textbf{(a)} Ground Truth
    \end{minipage}
    \begin{minipage}[t]{0.32\linewidth}
        \centering
        \textbf{(b)} Ours
    \end{minipage}
    \begin{minipage}[t]{0.32\linewidth}
        \centering
        \textbf{(c)} SSS 3DGS \cite{dihlmann2024subsurfacescattering3dgaussian}
    \end{minipage}
    
    \caption{\textbf{Silhouette loss effect on synthetic data.} The addition of silhouette regularization reduces boundary bleeding and sharpens object contours.}
    \label{fig:loss_results2}
\end{figure*}

We evaluate the contribution of each multi-view geometric term through controlled ablations on synthetic and real OLAT data. The quantitative results are summarized in Table~\ref{tab:ablation_combined}, and qualitative comparisons appear in Figures~\ref{fig:reconstruction_metrics}--\ref{fig:loss_results2}. 

\paragraph{Silhouette-only supervision.}
Adding only the silhouette consistency term improves boundary alignment by ensuring that reprojected points fall within the target-view silhouette. As shown in Figure~\ref{fig:reconstruction_metrics} (second column), this reduces edge bleeding and stabilizes contours but does not fully resolve metric inconsistencies.

\paragraph{Depth-only supervision.}
The depth consistency loss enforces agreement between the reprojected depth $\hat{z}_j(X)$ and the rendered depth $D_j(x')$ in the target view. Figure~\ref{fig:reconstruction_metrics} (third column) shows that this significantly reduces geometric drift across views, improving spatial coherence even in sparse-view settings.

\paragraph{Combined silhouette + depth supervision.}
Using both terms yields the most stable reconstructions. As seen in Figures~\ref{fig:reconstruction_metrics} (fourth column) and~\ref{fig:reconstruction_comparison}, the combination delivers accurate geometry and clean boundaries, with fewer view-dependent artifacts than either component alone.

\paragraph{Generalization across objects.}
Figures~\ref{fig:qualitative_comparisonablation2} and~\ref{fig:qualitative_comparisonablation3} demonstrate that these improvements generalize across diverse synthetic and real objects, including highly translucent materials. The combined loss consistently produces sharper silhouettes and more metrically consistent shapes.

\paragraph{Isolated silhouette effects.}
Figures~\ref{fig:loss_results} and~\ref{fig:loss_results2} highlight the specific gain from silhouette consistency alone—most notably the suppression of boundary haloing and improved contour sharpness.

\paragraph{Summary.}
Across all experiments, multi-view geometric supervision (i) sharpens boundaries, (ii) improves metric alignment, and (iii) reduces view-dependent distortions. These trends match the improvements reported quantitatively in Table~\ref{tab:ablation_combined} and play a central role in the reliability of DIAMOND-SSS under sparse or noisy view regimes.

\section{Reproducibility and Compute}
\label{sec:supp_repro}

DIAMOND-SSS is built entirely on publicly released components in 3D Gaussian Splatting and diffusion architectures.  
To facilitate reproducibility, we summarize below the practical considerations, compute settings, and implementation choices that complement the hyperparameters in
\cref{supp:tab:sss3dgs_hyperparams} and the training details provided in
\cref{subsec:implementation,sec:supp_diffusion}.

\paragraph{Hardware.}
All experiments were conducted on a single NVIDIA RTX~A6000 GPU (49\,GB VRAM) using PyTorch~2.2 with automatic mixed precision.
A local workstation with 32\,GB RAM and $\sim$100\,GB of disk storage (datasets, checkpoints, intermediate outputs) is sufficient to reproduce every experiment.
Training the Free3D fine-tuning and the relighting model does not require multi-GPU setups.

\paragraph{Code base.}
Our implementation builds on:
\begin{itemize}[leftmargin=*]
    \item \textbf{3DGS}~\cite{kerbl20233dgaussiansplattingrealtime} for the base Gaussian renderer,
    \item \textbf{Relightable 3DGS}~\cite{gao2024relightable} for material and lighting branches,
    \item \textbf{SSS-3DGS}~\cite{dihlmann2024subsurfacescattering3dgaussian} for subsurface-scattering modeling,
    \item \textbf{Free3D}~\cite{zheng24} for multi-view diffusion,
    \item \textbf{Poirier-Ginter~et~al.}~\cite{poirier2024radiancefield} for relighting diffusion (ControlNet-style conditioning).
\end{itemize}
All components required to reproduce our results are open source.

\paragraph{Training schedule and runtime.}
Reconstruction of each scene (including multi-view geometric consistency) requires approximately:
\begin{itemize}[leftmargin=*]
    \item \textbf{1--1.5 hour} for \textbf{full-view} OLAT settings,
    \item \textbf{0.5--1 hours} for \textbf{sparse-view} settings,
\end{itemize}
measured on an RTX~A6000.  
Fine-tuning the diffusion models is performed once:
\begin{itemize}[leftmargin=*]
    \item \textbf{Free3D NVS fine-tuning:} $\approx$15 hours for 500 epochs,
    \item \textbf{Relighting model:} $\approx$12--20 hours depending on object subset size.
\end{itemize}
These models are then reused for all experiments without per-object retraining.

\paragraph{Scene reuse and caching.}
Rendered depth, normals, and intermediate diffusion results are cached to reduce repeated compute.
Gaussian densification schedules and multi-view loss masks (\cref{sec:supp_geom}) are shared across experiments to ensure comparability.

\paragraph{Randomness and determinism.}
All experiments use fixed seeds for camera sampling, appearance model initialization, and diffusion noise, ensuring repeatability with no run-to-run variations.

\paragraph{Reconstruction pipeline consistency.}
Evaluation scripts, loss computation, and image-space metrics (PSNR, SSIM, LPIPS) follow the same routines across synthetic and real scenes for consistent reporting.

\medskip
Overall, DIAMOND-SSS can be reproduced on a single high-memory GPU and relies only on publicly available components. The training schedules, hyperparameters, and data augmentations are fully documented in the main paper and this supplementary material.

\end{document}